\documentclass[11pt, a4paper, logo, copyright, nonumbering]{mll_style}
\usepackage[authoryear, sort&compress, round]{natbib}
\usepackage{graphicx}
\usepackage{xspace}
\usepackage{adjustbox}
\usepackage{array}
\usepackage{float}
\usepackage{dblfloatfix}
\usepackage{enumitem}
\usepackage{setspace}
\usepackage{booktabs}
\usepackage{multirow}
\usepackage{nicefrac}
\usepackage{wrapfig}
\usepackage{placeins}
\usepackage[figuresright]{rotating}
\usepackage{url}
\usepackage{fontawesome5}
\usepackage{amsmath}
\usepackage{amssymb}
\usepackage{amsfonts}
\usepackage{mathtools}
\usepackage{bm}
\usepackage[most]{tcolorbox}
\usepackage{minitoc}

\renewcommand \partname{}

\makeatletter
\def\@BTrule[#1]{%
  \ifx\longtable\undefined
    \let\@BTswitch\@BTnormal
  \else\ifx\hline\LT@hline
    \nobreak
    \let\@BTswitch\@BLTrule
  \else
     \let\@BTswitch\@BTnormal
  \fi\fi
  \global\@thisrulewidth=#1\relax
  \ifnum\@thisruleclass=\tw@\vskip\@aboverulesep\else
  \ifnum\@lastruleclass=\z@\vskip\@aboverulesep\else
  \ifnum\@lastruleclass=\@ne\vskip\doublerulesep\fi\fi\fi
  \@BTswitch}
\makeatother

\addto\extrasenglish{
}

\newcommand{\hd}[2]{{\scriptsize\begin{tabular}[b]{@{}r@{}}#1\\#2\end{tabular}}}

\reportnumber{001}

\makeatletter\renewcommand{\titlefont}{\color{black}\normalfont\bfseries\fontsize{15}{19}\selectfont}\makeatother
\title{\centering WOVEN: Weaving Visual World Modeling into Multimodal LLMs}
\date{}

\author[*]{%
\begin{tabular}{@{}c@{}}
Zheyu Fan$^{1}$, Yue Zhang$^{3}$, Mingkai Deng$^{2}$, Kangrui Wang$^{1}$, 
Qineng Wang$^{1}$, Canyu Chen$^{1}$,\tabularnewline[5pt]
Jie Hao$^{4}$, Xing Fan$^{4}$, Chenlei Guo$^{4}$,
Eric P.~Xing$^{2}$, Mohit Bansal$^{3}$, Manling Li$^{1}$
\end{tabular}\\
\small $^1$Northwestern University~~~
$^2$Carnegie Mellon University~~~
$^3$UNC Chapel Hill~~~
$^4$Amazon\\
{\footnotesize \href{https://woven-ai.github.io}{\faGlobe~Website}\quad \href{https://github.com/mll-lab-nu/WOVEN}{\faGithub~Code}\quad \href{https://huggingface.co/datasets/MLL-Lab/WOVEN}{\faDatabase~Dataset}}\\
\vspace{-10pt}
}

\makeatletter
\DeclareRobustCommand{\_}{\ifmmode\nfss@text{\textunderscore}\else\textunderscore\allowbreak\fi}
\makeatother
\begin{abstract}
\looseness=-1 Multimodal large language models (MLLMs) struggle with spatial, embodied, physical, and temporal reasoning. We hypothesize that these failures reflect a shared deficit in \textbf{visual transition reasoning}, and test whether this capability can serve as a \textbf{shared training primitive} (one that different models can learn from different supervision sources and reuse across different tasks), and seek a systematic training recipe that benefits diverse downstream tasks. Existing benchmarks document these deficits separately but do not support controlled comparisons across scenes, actions, and reasoning operations. To enable such comparisons, we introduce \textbf{WOVEN}, a training source and benchmark for visual transition reasoning that organizes transition supervision by scene, action, and reasoning type, using diverse, realistic rollouts from video-pretrained generative models. It comprises $36{,}076$ examples across $20$ scene types (e.g., kitchen, park), $5$ action types (e.g., object manipulation, camera motion), and $8$ reasoning types (e.g., forward dynamics, counterfactual substitution). Using WOVEN, we first evaluate $38$ frontier MLLMs (e.g., GPT-5.4 and Qwen3-VL-235B-A22B) and find a substantial and systematic deficit: even the strongest models fall far below humans, and the failures recur across model families and persist with scale. We then train MLLMs at multiple scales on WOVEN and find that they learn a shared capability that transfers broadly: training subsets of only about $2{,}000$ items each collectively improve $22$ of $26$ external benchmarks by up to $27.3$ percentage points, and WOVEN data can even replace $30$--$50\%$ of a task's own training data with comparable accuracy. More importantly, controlled comparisons reveal what supervision transfers best and yield a \textbf{training recipe} for visual world modeling, validated prospectively on held-out benchmarks: supervision should be selected by the reasoning operation it teaches rather than by the actions, scenes, or domains it shows, and larger changes to the visual state during training bring greater robustness. Our work establishes visual transition reasoning as a reusable foundation for improving diverse downstream capabilities, paving the way for systematic visual world-model training in MLLMs.
\end{abstract}

\begin{document}

\doparttoc
\faketableofcontents

\maketitle

\section{Introduction}
\label{sec:intro}

\begin{figure}[!t]
\centering
\includegraphics[width=\linewidth]{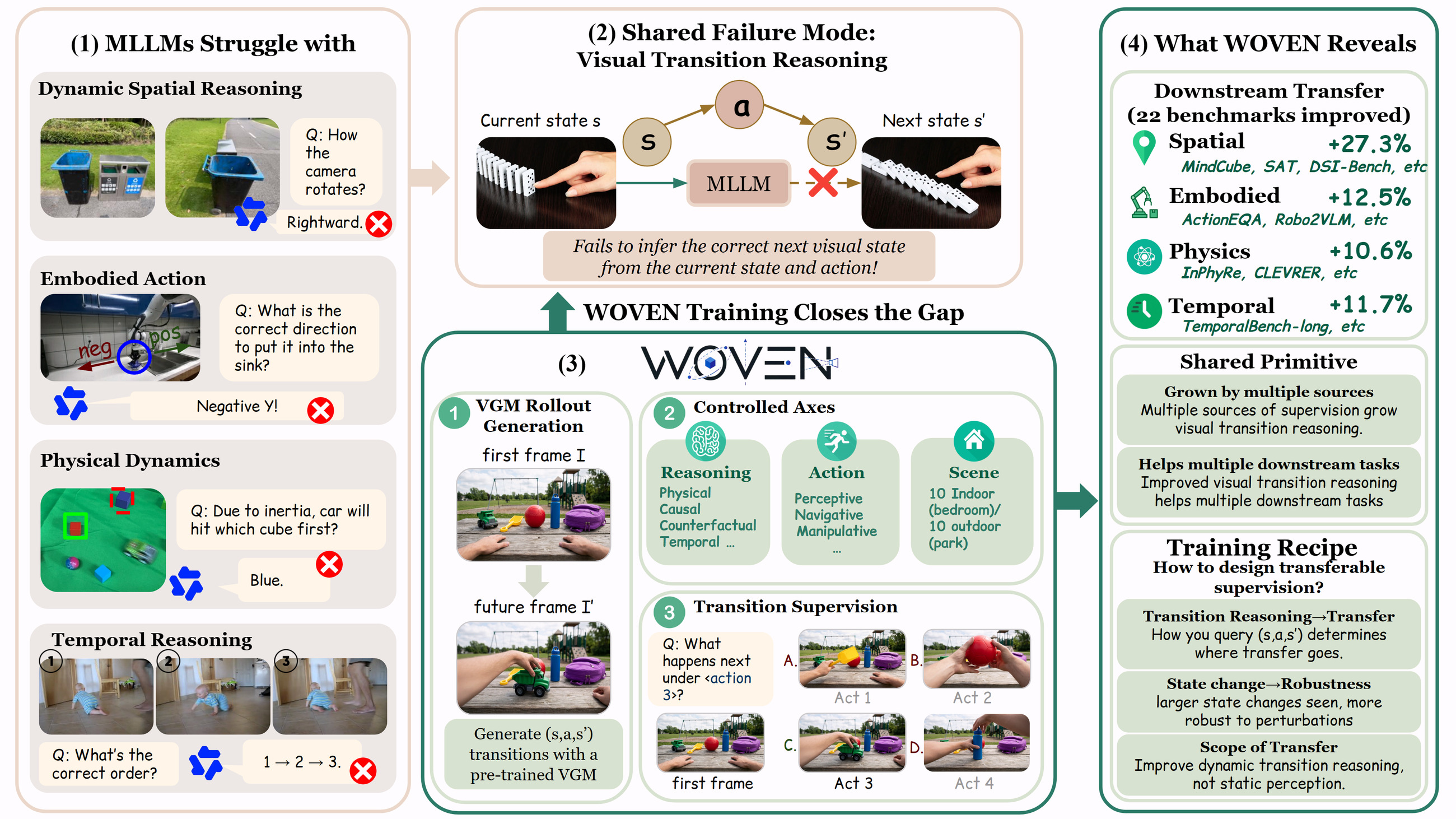}
\caption{
\textbf{Overview.} (1) MLLMs fail across spatial, embodied, physical, and temporal tasks. (2) These failures share a weakness: reasoning about how visual states, actions, and outcomes relate. (3) WOVEN generates $(s,a,s')$ transitions with a video generation model, organizes them along controlled axes of reasoning, action, and scene, and turns them into error-typed multiple-choice supervision. (4) Training on WOVEN improves $22$ of $26$ external benchmarks; the gains come from a shared primitive that different sources build and different tasks use; and controlled comparisons yield a training recipe: the reasoning operation decides where transfer goes, larger state changes give robustness, and transfer covers transition tasks rather than static perception.
}
\label{fig:teaser}
\end{figure}

Multimodal large language models (MLLMs) perform well on visual recognition and language-conditioned reasoning~\citep{bai2025qwen25vl}, yet remain unreliable on tasks involving visual change, including spatial, physical, embodied, and temporal reasoning~\citep{yin2025mindcube,chow2025physbench,ma2026causalspatial,huang2026thinkingdynamics}.
For example, they may misjudge camera rotation, predict outcomes that violate physical constraints, recommend actions that conflict with the task goal, or misorder video frames (Fig.~\ref{fig:teaser}(1)).
Despite their different goals, many of these tasks share a common structure: an action $a$ links an initial visual state $s$ to a resulting state $s'$ (Fig.~\ref{fig:teaser}(2)).
Predicting outcomes, inferring actions from observed changes, and reasoning about alternative actions are different inferences over the same $(s,a,s')$ triplet.
We hypothesize that a shared deficit in \emph{action-conditioned visual transition reasoning} underlies failures across these tasks.

\looseness=-1 This hypothesis motivates exploring a shared training primitive for otherwise distinct downstream capabilities: rather than improving each capability separately, we ask whether and how visual transition reasoning can serve as a common training target across models and benefit diverse downstream tasks. In this paper, we pursue this goal through four key questions. First, do failures in visual transition reasoning exhibit consistent patterns across model families, and how pronounced are these deficits? Second, can it be learned through transition supervision, and do the benefits extend beyond the training tasks? Third, do these gains arise from separate training components helping separate tasks, or from learning visual transition reasoning that different models can reuse across different tasks? Fourth, and more fundamentally, what governs the downstream transfer and should therefore guide a systematic training recipe: similarities in actions, scenes, or application domains, or the reasoning operation applied to the transition $(s,a,s')$?

Answering these questions calls for controlled comparisons that allow us to examine the roles of individual components of the $(s,a,s')$ triplet. Existing task-specific benchmarks document important deficits, but do not readily support such comparisons \citep{yin2025mindcube,chow2025physbench,causalvqa,temporalbench,ma2026causalspatial,huang2026thinkingdynamics}. We therefore introduce \textbf{WOVEN} (\textbf{WO}rld modeling via controllable \textbf{V}ideo g\textbf{EN}eration), a diagnostic and training source that draws on the transition priors of video-pretrained generative models (\citealp{wan2025}). Generated rollouts provide known actions and endpoints for forward and inverse supervision, outcomes under alternative actions for counterfactual supervision, and temporally coherent intermediate states for temporal supervision. Organizing this supervision along scene, action, and reasoning dimensions enables controlled studies of learning and transfer (Fig.~\ref{fig:teaser}(3)). The collection comprises $36{,}076$ examples spanning 20 scene types (e.g., kitchen, park, and living room), five action types (e.g., passive physical events, camera motion, and object manipulation), and eight reasoning types across four families (e.g., causal dynamics, temporal coherence, and counterfactual reasoning).  We also provide diagnostic splits which cover  in-distribution evaluation, held-out scenes, and state perturbations, with error-typed distractors supporting fine-grained analysis of the effects of training (\S\ref{sec:benchmark}).

To examine visual transition reasoning as a common training target, we first ask whether its deficits recur across model families and whether learning it yields gains beyond the training tasks. Through fine-grained evaluation of 38 frontier MLLMs (e.g., GPT-5.4~\citep{openai2026gpt54}, Qwen3-VL-235B-A22B~\citep{bai2025qwen3vl}, and Llama-4-Maverick~\citep{meta2025llama4}), we identify recurring failure patterns across model families. These deficits are substantial: even the strongest model achieves only 65.8\% accuracy against a 92.3\% human baseline. These recurring deficits identify visual transition reasoning as a common target for improving current MLLMs, establishing a training need that extends beyond the weaknesses of any individual model (\S\ref{sec:assess}). We then ask whether this capability can be learned and whether the resulting gains extend beyond the training tasks.
Crucially, our empirical results from post-training MLLMs at multiple scales (\S\ref{sec:enhance:transfer}) demonstrate substantial gains: training on the full WOVEN training set raises the in-distribution accuracy of Qwen2.5-VL-3B~\citep{bai2025qwen25vl} from 26.4\% to 89.3\%, with 88.2\% on held-out scenes (\S\ref{sec:enhance:sft}). The benefits extend to 12 of 26 external benchmarks across multiple domains. Controlled subsets containing approximately $2{,}000$ examples each collectively expand this coverage to 22 of 26 benchmarks, with gains of up to 27.3 percentage points (Fig.~\ref{fig:teaser}(4)).

\looseness=-1 However, broad transfer alone does not establish visual transition reasoning as a shared training primitive that different models can reuse from different sources in learning different tasks, as the downstream gains could arise from separate parts of the training source, each helping only its own tasks. To serve as a shared training primitive, visual transition reasoning must be a common capability that models acquire from different sources, and the gains on downstream tasks must result from this common capability. We therefore first test whether different sources improve the same capability, and find that every controlled subset improves WOVEN accuracy, including on the action types it does not train on (\S\ref{sec:enhance:sft}), meeting the first requirement. We then examine whether the downstream gains result from this common capability, by checking whether subsets that improve it more also transfer more. Results show that the more a subset improves WOVEN accuracy, the larger its average gain on the external benchmarks ($r{=}0.86$), and that the gains come from visual transition reasoning itself rather than from a generally stronger model or the answer format (\S\ref{sec:enhance:substitution}). We further test causality by intervening on the training data: replacing $30$--$50\%$ of task-specific examples with WOVEN under a fixed training budget preserves comparable accuracy (Fig.~\ref{fig:g2-mix}; \S\ref{sec:enhance:substitution}), showing that transition supervision can take over part of the training normally supplied by task-specific data. Together with transfer across model scales (\S\ref{sec:enhance:transfer}), these results establish visual transition reasoning as a shared training primitive.

More importantly, our controlled comparisons yield a training recipe for visual world modeling, validated prospectively on held-out benchmarks (\S\ref{sec:enhance:org}). \textbf{(1) Select supervision by reasoning operation.}  Surprisingly, supervision that teaches the same reasoning operation produces similar patterns of downstream gains even when the actions differ, whereas supervision that shares only an action type does not (Fig.~\ref{fig:teaser}(4)). The reasoning operation therefore provides a basis for selecting supervision with relevant downstream benefits. Once the operation is selected, \textbf{(2) choose supervision with larger changes to the visual state for robustness.} Supervision whose actions change more of the scene, such as object manipulation, yields larger robustness gains than supervision that only moves the camera. \textbf{(3) Supervise temporal and agent-driven tasks directly.} Transfer does not cross the boundary between the temporal family and the other three reasoning families, nor the boundary between passive physical events and agent-driven actions. \textbf{(4) Keep dedicated supervision for capabilities beyond state transitions.} Transition supervision improves diverse tasks that involve state transitions, but does not necessarily generalize to capabilities such as static perception, which therefore need their own supervision (\S\ref{sec:enhance:org}).
Taken together, our study establishes visual transition reasoning as a shared, trainable primitive and provides a systematic recipe for training it, offering a controlled framework for advancing visual world modeling in MLLMs.

\begin{table}[t]
\centering
\small
\captionsetup{font=small,skip=3pt}
\caption{Eight reasoning operations considered in this work. The same operation can be applied to different actions and scenes (\S\ref{sec:benchmark}). Notation: $a_{\mathrm{alt}}$, an alternative action from the same initial state; $a_{\mathrm{exo}}$, a passive event; $a_q$, $a_{\mathrm{exo}}$ with a cue naming the physical principle; $\mathbf{s}=(s_0,\ldots,s_L)$, sampled states in time order with $s_0=s$ and $s_L=s'$; $\widetilde{\mathbf{s}}$, the same states shuffled; $s_-,s_+$, the neighbors in time of a reference state $s_{\mathrm{ref}}$. In the physical and temporal families the action description is optional.}
\label{tab:query-operators}
\setlength{\tabcolsep}{3pt}
\renewcommand{\arraystretch}{1.05}
\begin{tabular*}{\linewidth}{@{\extracolsep{\fill}}lllll@{}}
\toprule
Family & Reasoning operation & Distribution & Reasoning operation & Distribution \\
\midrule
Causal & Forward dynamics & $P(s'\mid s,a)$ & Inverse dynamics & $P(a\mid s,s')$ \\
Counterfactual & Counterfactual removal & $P(s\mid a,s')$ & Counterfactual substitution & $P(s'_{a_{\mathrm{alt}}}\mid a,s',a_{\mathrm{alt}})$ \\
Physical & Outcome prediction & $P(s'\mid s,a_{\mathrm{exo}})$ & Cued prediction & $P(s'\mid s,a_q)$ \\
Temporal & Temporal ordering & $P(\mathbf{s}\mid\widetilde{\mathbf{s}},a)$ & Temporal adjacency & $P(s_-,s_+\mid s_{\mathrm{ref}},\widetilde{\mathbf{s}},a)$ \\
\bottomrule
\end{tabular*}
\end{table}

\section{WOVEN}
\label{sec:benchmark}

\begin{figure}[!t]
\centering
\includegraphics[width=\linewidth]{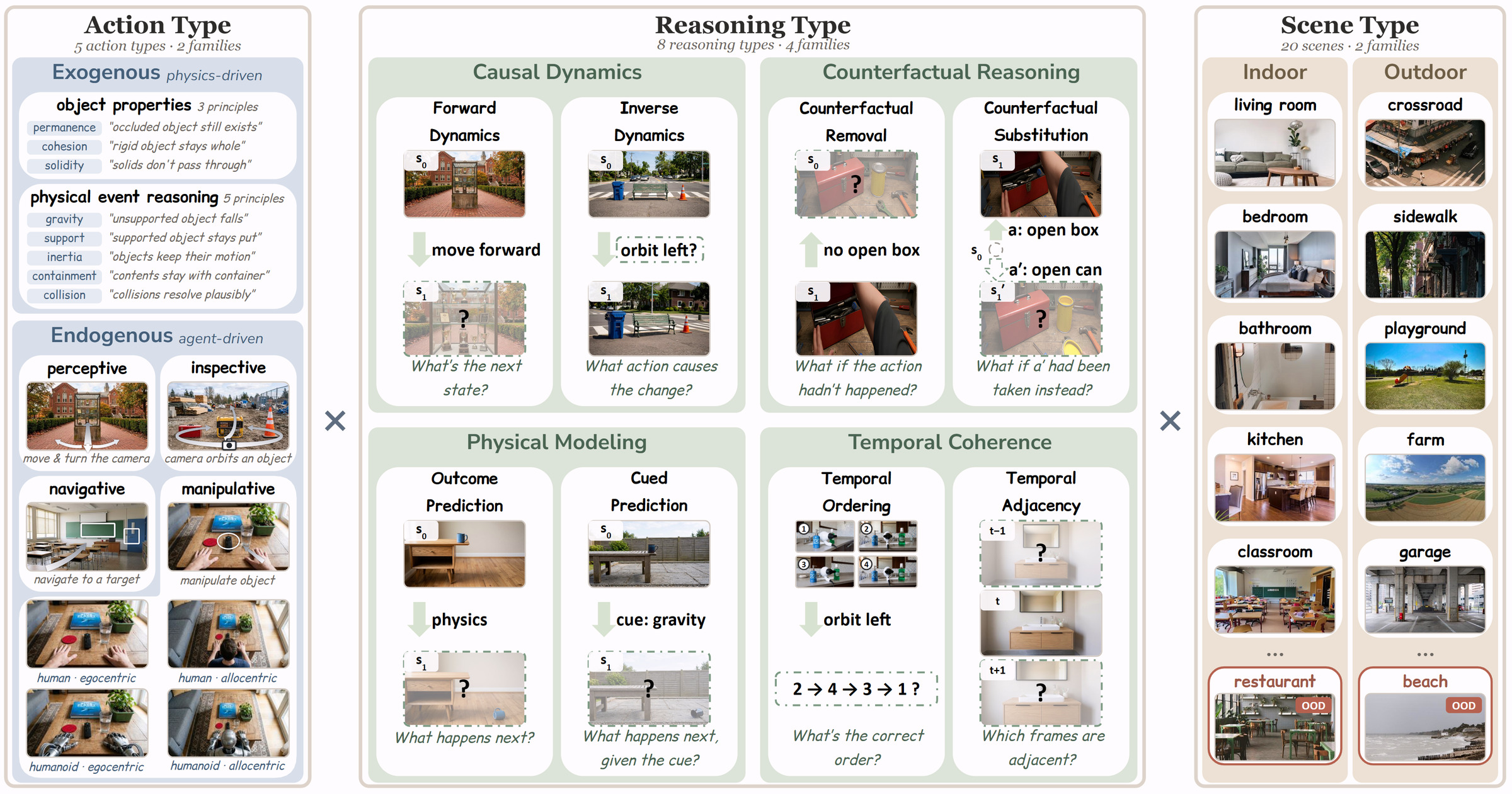}
\caption{\textbf{The WOVEN taxonomy.} Every item is labeled along the three axes shown from left to right (\S\ref{sec:bench:axes}). Left: the exogenous action type with its eight principles, and the four endogenous types, where manipulative actions combine subject (human, humanoid) and view (egocentric, allocentric). Middle: the eight reasoning types in four families; in each schematic, the dashed box marks the element of $(s,a,s')$ that the question asks for. Right: example scenes; restaurant and beach are held out for the scene-OOD test.}
\label{fig:taxonomy}
\end{figure}

\looseness=-1 We present WOVEN, a training source and benchmark for visual transition reasoning, designed for controlled comparisons of supervision. We first formalize visual transition reasoning in \S\ref{sec:formalization} as the theoretical foundation. Then, we define our taxonomy in \S\ref{sec:bench:axes}, which labels every item along three axes, scene, action, and reasoning type, so that supervision can be varied independently along each axis. As real video offers neither controlled actions nor alternative outcomes from the same initial state, and simulators are limited in realism and in scene and object diversity, we generate the transitions specified by our taxonomy with a \textit{video generation model}, which produces rollouts from specified initial states and actions, and alternative rollouts from the same state under different actions (\S\ref{sec:curation}). Finally, we turn the rollouts into multiple-choice items with error-typed distractors and split them into a training set and three test sets (\S\ref{sec:bench:supervision}).

\subsection{Visual Transition Reasoning}
\label{sec:formalization}

\looseness=-1 Given a transition $(s,a,s')$, where $s$ and $s'$ are the visual states of a scene before and after an action $a$, which may be agent-driven or passive, \emph{visual transition reasoning} is the process of inferring unobserved component(s) of the transition from its known or observed parts. Different choices of which parts are given or to be inferred define different reasoning operations, each written as a conditional distribution such as $P(s'\mid s,a)$ (shorthand for $P(S'=s'\mid S=s,A=a)$). This work considers eight operations in four families (Table~\ref{tab:query-operators}). Appendix~\ref{app:primitive} gives the semantics of each operation.

\subsection{Controlled Axes}
\label{sec:bench:axes}

\label{sec:methodology}We define the \textbf{taxonomy} along three axes and categorize each item accordingly: (1) the \emph{scene} in which its transition occurs, (2) the \emph{action} producing the change, and (3) the \emph{reasoning operation} applied to the transition (Fig.~\ref{fig:taxonomy}). \emph{Scene type} has $20$ categories, evenly divided between indoor and outdoor environments (e.g., kitchen, living room, park, and beach). \emph{Action type} has $5$ categories: one exogenous type, passive physical events (Exo), divided into $8$ principles from cognitive science ($3$ object properties and $5$ physical events), and four endogenous, agent-driven types: perceptive, that is, camera motion (Perc), inspective, that is, object inspection (Insp), navigative, that is, navigation (Navi), and manipulative, that is, object manipulation (Mani) (Fig.~\ref{fig:taxonomy}). \emph{Reasoning type} instantiates the eight operations of Table~\ref{tab:query-operators}, grouped in pairs into four families: forward dynamics (Fwd) and inverse dynamics (Inv) in causal dynamics, counterfactual removal (Rmv) and counterfactual substitution (Sub) in counterfactual reasoning, outcome prediction (Otm) and cued prediction (Cue) in physical modeling, and temporal ordering (Ord) and temporal adjacency (Adj) in temporal coherence. The tables use these abbreviations. Within compatible combinations, each axis can be varied while the other two are held fixed. Category definitions and cognitive motivation are in Appendices~\ref{app:why-agency}--\ref{app:manip}.

\subsection{Controlled Rollout Generation}
\label{sec:curation}

We use Wan2.2-I2V-A14B~\mbox{\citep{wan2025}} to generate rollouts from initial frames and action descriptions, following two construction principles. \textbf{(P1) World-model isolation.} Resulting and intermediate states are produced by the rollout rather than prescribed in the conditioning prompt, so supervision draws on learned transition dynamics. \textbf{(P2) Counterfactual rollouts.} From the same initial state, we generate several independent rollouts under different actions, so that alternative outcomes of the same scene are available. Generation, frame extraction, and quality-control details are provided in Appendix~\ref{app:pipeline}.

\subsection{Supervision Construction and Dataset Splits}
\label{sec:bench:supervision}

\label{sec:meth:stat} We build multiple-choice questions from the rollouts, each referred to as an item and is defined on all three axes: its scene and action types are those of the transition it is built from, and its reasoning type (Table~\ref{tab:query-operators}) determines which parts of the transition are given and which part is asked for. Each question is phrased with one of $30$ templates drawn at random to ensure linguistic diversity. The three distractor options are usually drawn from other rollouts sharing the same initial state, each wrong in a specific way (e.g., the outcome of a different action) and labeled with its error type (e.g., ``reversed direction'' or ``violating gravity''), enabling fine-grained error analysis and statistical attribution beyond overall accuracy. For physical-modeling items, the distractors are produced by an image editor, so the correct option is also passed through the same editor with an instruction to change nothing; a control experiment shows that editing artifacts do not reveal the answer. See Appendices~\ref{app:distractors} and~\ref{app:stat-methods} for details.

\label{sec:bench:overview}\label{sec:meth:diag}\label{sec:bench:perturb} WOVEN has one training set ($22{,}728$), one validation set ($2{,}508$), and three test sets, totaling $36{,}076$ four-option multiple-choice items (Appendix~\ref{app:splits}). Each test set differs from the training set in one respect. The \emph{in-distribution} test set ($6{,}228$) contains new transitions from the same $18$ scenes as the training set. The \emph{scene-OOD} test set ($3{,}496$) contains transitions from two held-out scenes, restaurant and beach, following held-out-environment evaluation in embodied AI~\citep{openeqa2024,cheng2025embodiedeval}. The \emph{state-perturbation} test set ($1{,}116$) measures robustness to controlled changes in the initial state: following the logic of contrast sets~\citep{gardner2020contrast}, each item perturbs one condition of an in-distribution item's initial state, either its spatial configuration or its appearance, while keeping the action and the multiple-choice question fixed, testing whether a model actually models the state rather than the typical effect of the action (Appendix~\ref{app:perturb-design}).

\begin{table}[t]
\centering
\scriptsize
\setlength{\tabcolsep}{3pt}
\caption{\textbf{Visual transition reasoning in frontier MLLMs against the human baseline.} Accuracy (\%) on the in-distribution test set, overall and by action type and by reasoning type, and on the state-perturbation test set under geometric and appearance changes. Human accuracy is the mean of three annotators. Results for all $38$ models are in Appendix~\ref{app:eval-tables}, with the corresponding analyses in Appendix~\ref{sec:assess:findings}.}
\label{tab:assess-main}
\begin{tabular*}{\linewidth}{@{\extracolsep{\fill}}l r rrrrr rrrrrrrr rr@{}}
\toprule
 & & \multicolumn{5}{c}{Action type} & \multicolumn{8}{c}{Reasoning type} & \multicolumn{2}{c}{Perturbation} \\
\cmidrule(lr){3-7}\cmidrule(lr){8-15}\cmidrule(l){16-17}
 & Overall & Exo & Perc & Insp & Navi & Mani & Fwd & Inv & Rmv & Sub & Otm & Cue & Ord & Adj & Geo & App \\
\midrule
Human & 92.3 & 92.8 & 92.8 & 92.6 & 90.6 & 92.8 & 92.8 & 92.8 & 91.1 & 92.2 & 93.3 & 91.1 & 92.8 & 91.7 & 97.3 & 98.7 \\
\midrule
GPT-5.4 & 65.8 & 62.4 & 61.2 & 58.3 & 68.8 & 81.2 & 83.8 & 81.7 & 64.5 & 76.0 & 81.9 & 85.8 & 57.1 & 27.0 & 62.7 & 95.5 \\
Intern-S1-Pro & 63.2 & 56.2 & 52.8 & 53.7 & 70.1 & 87.1 & 77.6 & 77.5 & 66.7 & 73.3 & 61.5 & 69.1 & 63.4 & 26.5 & 33.3 & 79.2 \\
Qwen2.5-VL-72B & 58.9 & 53.7 & 47.7 & 47.8 & 66.4 & 83.4 & 74.7 & 74.7 & 54.8 & 72.0 & 59.4 & 76.0 & 49.4 & 28.0 & 38.5 & 83.0 \\
Gemma-4-31B & 57.8 & 50.8 & 49.4 & 53.1 & 61.2 & 74.3 & 76.3 & 76.6 & 63.6 & 64.5 & 64.9 & 70.5 & 30.8 & 25.3 & 59.0 & 95.5 \\
\bottomrule
\end{tabular*}
\end{table}

\section{Is Visual Transition Reasoning Learnable and Transferable?}
\label{sec:enhance}

Using WOVEN as an error-typed and shortcut-resistant diagnostic and training source, we ask how far current MLLMs fall short in visual transition reasoning (\S\ref{sec:assess}), whether it can be learned from transition supervision (\S\ref{sec:enhance:sft}), and whether the learned reasoning transfers beyond the training tasks (\S\ref{sec:enhance:transfer}). Throughout, training is carried out in two complementary ways: on the full WOVEN training set, and on controlled subsets, each containing only items of a specified action type and reasoning family and named after the two (perc\_causal, for example, holds camera-motion items of the causal family, and insp\_cf object-inspection items of the counterfactual family); WOVEN's training set consists of $11$ such subsets, each of about $2{,}000$ items.
\phantomsection\label{sec:enhance:setup} The full training set is trained with supervised fine-tuning, and each subset with supervised fine-tuning followed by GRPO~\citep{shao2024deepseekmath} (Appendix~\ref{app:E-recipes}), on MLLMs of multiple scales (Appendix~\ref{app:E-F13}).
Evaluation uses the three WOVEN test sets and, zero-shot, $26$ external benchmarks: $22$ that involve visual transitions, spanning spatial, embodied, physical, temporal reasoning, etc., and $4$ benchmarks of static perception and general video QA. A benchmark counts as improved when the gain over the base model exceeds $2$ percentage points.

\subsection{Current MLLMs Show a Systematic Deficit}
\phantomsection\label{sec:assess}\label{sec:assess:eval}

\begin{table}[t]
\centering
\footnotesize
\setlength{\tabcolsep}{3.6pt}
\caption{\textbf{WOVEN test accuracy (\%) of Qwen2.5-VL-3B-Instruct after training on the full training set, and gains (percentage points over the base model) after training on each controlled subset.} In-distribution results are reported overall, by action type, and on temporal adjacency; bold marks the action type a subset trains on. Results after SFT alone and by every reasoning type are in Appendix~\ref{app:E-F15}.}
\label{tab:woven-train}
\begin{tabular}{@{}l rrrrrrr r rrr@{}}
\toprule
 & \multicolumn{7}{c}{ID test} & \multicolumn{1}{c}{OOD scenes test} & \multicolumn{3}{c}{Perturbation test} \\
\cmidrule(lr){2-8}\cmidrule(lr){9-9}\cmidrule(l){10-12}
Training & Overall & Exo & Perc & Insp & Navi & Mani & Adj & Overall & Overall & Geo & App \\
\midrule
Base & 26.4 & 23.8 & 27.0 & 23.5 & 30.0 & 28.8 & 25.7 & 27.8 & 21.1 & 22.6 & 19.7 \\
Full training set & 89.3 & 79.4 & 90.7 & 90.5 & 90.0 & 95.3 & 69.7 & 88.2 & 38.3 & 35.0 & 41.6 \\
\midrule
\multicolumn{12}{@{}l}{\textit{Gain over the base model after training on one subset}} \\
perc\_causal & +22.9 & +5.1 & \textbf{+33.0} & +13.4 & +26.9 & +39.8 & +0.2 & +20.0 & +1.7 & +0.9 & +2.5 \\
perc\_temporal & +6.6 & +8.4 & \textbf{+10.9} & +4.9 & +8.0 & +1.3 & +29.9 & +2.9 & +4.1 & +5.6 & +2.7 \\
insp\_causal & +25.3 & +6.8 & +8.9 & \textbf{+39.9} & +27.2 & +37.7 & +0.9 & +20.8 & +3.0 & +1.1 & +4.8 \\
insp\_cf & +25.1 & +6.6 & +7.3 & \textbf{+38.5} & +24.8 & +42.7 & +1.3 & +24.7 & +3.2 & +3.2 & +3.2 \\
insp\_temporal & +6.1 & +9.2 & +6.4 & \textbf{+3.4} & +11.2 & +1.2 & +33.1 & $-$2.0 & +5.2 & +2.5 & +7.9 \\
navi\_causal & +21.6 & +7.6 & +10.1 & +12.7 & \textbf{+30.0} & +50.9 & +4.9 & +20.7 & +13.6 & +16.0 & +11.3 \\
navi\_temporal & +6.9 & +8.1 & +12.2 & +4.0 & \textbf{+8.6} & +2.9 & +32.0 & +5.7 & +6.8 & +6.8 & +6.8 \\
mani\_causal & +19.9 & +8.2 & +3.9 & +5.7 & +26.2 & \textbf{+61.4} & +2.3 & +25.2 & +23.2 & +13.1 & +33.3 \\
mani\_cf & +21.3 & +10.6 & +5.2 & +5.5 & +27.2 & \textbf{+64.2} & +2.9 & +23.6 & +20.2 & +14.0 & +26.4 \\
exo\_physical & +9.4 & \textbf{+27.4} & $-$1.2 & +5.7 & +6.7 & +9.6 & $-$1.5 & +11.1 & +29.9 & +15.4 & +44.4 \\
exo\_temporal & +6.6 & \textbf{+8.4} & +7.1 & +5.5 & +10.1 & +2.3 & +30.6 & +5.5 & +6.9 & +5.6 & +8.2 \\
\bottomrule
\end{tabular}
\end{table}

To establish the need for training, we first quantify the size and distribution of the deficit in visual transition reasoning by evaluating $38$ frontier MLLMs on the in-distribution and state-perturbation test sets; Table~\ref{tab:assess-main} reports the four strongest against the human baseline. The best model, GPT-5.4, reaches $65.8\%$ in-distribution accuracy against $92.3\%$ for humans (the mean of three annotators), and the next three fall between $57.8\%$ and $63.2\%$, a gap that persists across model scales and cannot be closed by scaling alone (Appendix~\ref{sec:assess:scaling}). The gap also has a consistent structure across model families: for every model, temporal adjacency is harder than ordering ($25.3\%$--$28.0\%$ versus $30.8\%$--$63.4\%$), counterfactual removal is harder than substitution ($54.8\%$--$66.7\%$ versus $64.5\%$--$76.0\%$), and accuracy drops more under geometric than under appearance changes ($33.3\%$--$62.7\%$ versus $79.2\%$--$95.5\%$), suggesting a systematic deficit in visual transition reasoning rather than an idiosyncrasy of any particular family or training recipe. These results provide a fine-grained baseline for the training study that follows. The model inventory, human protocol, and detailed results are in Table~\ref{tab:T01} and Appendices~\ref{app:human-baseline}, \ref{app:eval-tables}, and~\ref{sec:assess:findings}; complementary evaluations of a VGM, V-JEPA~2~\citep{assran2025vjepa2}, and an IGM are in Appendix~\ref{app:video-priors}.

\begin{table}[t]
\centering
\footnotesize
\setlength{\tabcolsep}{1.2pt}
\caption{\textbf{Zero-shot transfer after training on the complete WOVEN training set.} Accuracy (\%) on the $12$ improved benchmarks; $\Delta$ is the larger of the SFT and GRPO gains. AEQA: ActionEQA; TB-L: TemporalBench-L; WM-AB: WM-ABench; WPred: WorldPrediction; BSwan: BlackSwan; SpViz: SpatialViz. References, results for all $26$ benchmarks, and category breakdowns are in Appendix~\ref{app:E-F11}.}
\label{tab:enhance-transfer}
\begin{tabular}{@{}l*{12}{r}@{}}
\toprule
 & SAT & InPhyRe & MindCube & AEQA & TB-L & WM-AB & WPred & TOMATO & BSwan & ERQA & SpViz & CV-Bench \\
\midrule
Base & 37.3 & 22.0 & 37.2 & 43.4 & 16.6 & 31.0 & 34.8 & 31.6 & 61.8 & 33.8 & 27.7 & 69.3 \\
WOVEN SFT & 55.3 & 24.6 & 37.8 & 49.5 & 11.5 & 37.1 & 39.6 & 31.5 & 55.9 & \textbf{37.5} & 28.8 & 72.0 \\
WOVEN GRPO & \textbf{61.3} & \textbf{34.9} & \textbf{48.6} & \textbf{54.4} & \textbf{26.8} & \textbf{39.8} & \textbf{41.5} & \textbf{36.1} & \textbf{65.6} & 34.5 & \textbf{31.1} & \textbf{72.4} \\
\rowcolor{gray!12} $\Delta$ & +24.0 & +13.0 & +11.3 & +11.0 & +10.3 & +8.9 & +6.7 & +4.5 & +3.8 & +3.8 & +3.4 & +3.0 \\
\bottomrule
\end{tabular}
\vspace{10pt}

\centering
\footnotesize
\setlength{\tabcolsep}{2pt}
\caption{\textbf{Broad transfer from controlled training subsets.} Results on $8$ of the $22$ improved benchmarks and, in the last four columns, on the $4$ benchmarks of static perception and general video QA; each column uses the best-performing subset for that benchmark, named in the last row (for the last four, its selected checkpoint). Abbreviations as in Table~\ref{tab:enhance-transfer}; R2VLM: Robo2VLM; DSI: DSI-Bench; 3DSR: 3DSRBench; CoreCog: CoreCognition; PercTest: PerceptionTest; EgoTQA: EgoTaskQA. Results for all $22$ benchmarks are in Appendix~\ref{app:E-F11} (Table~\ref{tab:enhance-bestcell-full}).}
\label{tab:enhance-bestcell}
\begin{tabular}{@{}l*{12}{r}@{}}
\toprule
 & SAT & AEQA & TB-L & InPhyRe & R2VLM & WPred & DSI & MindCube & 3DSR & CoreCog & PercTest & EgoTQA \\
\midrule
Base & 37.3 & 43.4 & 16.6 & 22.0 & 33.5 & 34.8 & 34.9 & 37.2 & 51.2 & 59.8 & 62.9 & 81.5 \\
Subset SFT & 63.3 & 47.5 & 22.0 & 32.3 & 40.5 & 40.1 & 40.3 & 40.0 & \multirow{2}{*}{52.6} & \multirow{2}{*}{58.8} & \multirow{2}{*}{63.7} & \multirow{2}{*}{82.6} \\
Subset GRPO & \textbf{64.7} & \textbf{55.9} & \textbf{28.2} & \textbf{32.6} & \textbf{42.4} & \textbf{42.9} & \textbf{40.9} & \textbf{43.1} & & & & \\
\rowcolor{gray!12} $\Delta$ & +27.3 & +12.5 & +11.7 & +10.6 & +8.9 & +8.1 & +6.0 & +5.9 & +1.4 & $-1.0$ & +0.8 & +1.1 \\
\addlinespace[2pt]
Subset & \hd{perc}{causal} & \hd{perc}{causal} & \hd{mani}{cf} & \hd{insp}{cf} & \hd{insp}{cf} & \hd{mani}{causal} & \hd{perc}{causal} & \hd{perc}{temporal} & \hd{navi}{causal} & \hd{exo}{physical} & \hd{perc}{causal} & \hd{exo}{physical} \\
\bottomrule
\end{tabular}
\end{table}

\subsection{Visual Transition Reasoning Is Learnable}
\label{sec:enhance:sft}

\looseness=-1 As shown in Table~\ref{tab:woven-train}, supervised fine-tuning of Qwen2.5-VL-3B-Instruct on the full training set raises in-distribution accuracy from $26.4\%$ to $89.3\%$, with $88.2\%$ on held-out scenes. The gains also hold on models of larger scales and different families (Appendices~\ref{app:E-F13} and~\ref{app:gemma}). Each controlled subset also raises WOVEN accuracy on its own, by $6.1$ to $25.3$ percentage points in distribution, and also on action types it does not train on (Table~\ref{tab:woven-train}), so different kinds of supervision improve the same capability rather than only the action they were specifically trained on. The deficit is therefore learnable as a whole from transition supervision, and the learning generalizes across scenes. The learning is also data-efficient: with about $7$ hours of generated video, WOVEN post-training yields a larger gain on a spatial-reasoning benchmark than Orca~\citep{wang2026orca}, which mid-trains the same backbone on $125{,}000$ hours of video (Appendix~\ref{app:E-F14}).

\subsection{Learned Transition Reasoning Transfers Broadly}
\label{sec:enhance:transfer}

\looseness=-1 We then examine whether the learned reasoning transfers to other tasks. The full training set improves as many as $12$ of the $26$ external benchmarks, with gains of up to $24.0$ percentage points on SAT (Table~\ref{tab:enhance-transfer}). The controlled subsets transfer effectively as well, despite holding only about $2{,}000$ items each: with the best subset chosen for each benchmark, the coverage rises to a remarkable $22$ of $26$, that is, every benchmark that involves visual transitions, and gains reach up to $27.3$ percentage points (Table~\ref{tab:enhance-bestcell}; selection and seed checks in Appendix~\ref{app:best-subset-robustness}). The transfer also holds at scale (Appendix~\ref{app:E-F13}) and across model families (Appendix~\ref{app:gemma}). The $4$ benchmarks of static perception and general video QA do not improve under any training (Table~\ref{tab:enhance-bestcell}), so the gains are not due to an overall stronger model or to better multiple-choice answering (Appendix~\ref{app:letter-control}). Full scores and evaluation procedures are in Appendices~\ref{app:E-F11} and~\ref{app:stat-methods}.

\section{Does Visual Transition Reasoning Serve as a Shared Primitive?}
\label{sec:enhance:substitution}

\begin{figure}[!t]
\centering
\includegraphics[width=\linewidth]{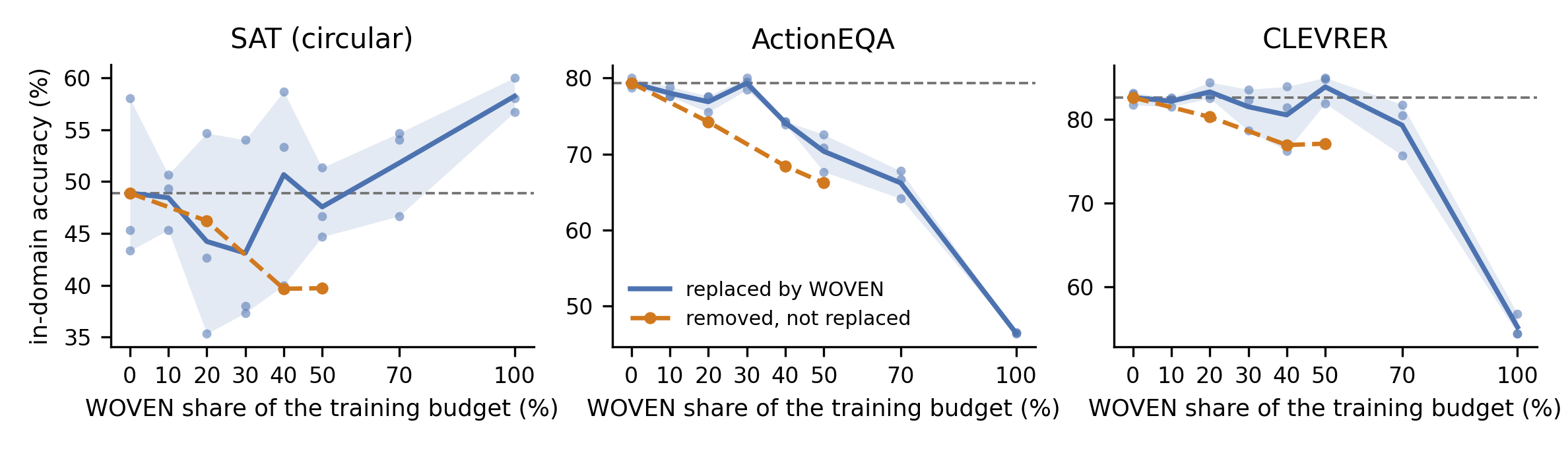}
\caption{\textbf{WOVEN supervision substitutes for in-domain training data.} Each panel retrains one external benchmark on a fixed $2{,}000$-item budget in which a share $p$ of the benchmark's own training data is replaced by items from a selected WOVEN subset (blue dots: $3$ seeds; blue line: seed mean; band: seed range; gray dashed: $p{=}0$ mean; orange dashed: deletion control). Replacing $30$--$50\%$ of task-specific supervision preserves comparable accuracy; on SAT, training exclusively on WOVEN yields the highest accuracy in the sweep. The protocol, full per-seed tables, and trend statistics for each benchmark are in Appendix~\ref{app:E-F12}.}
\label{fig:g2-mix}
\end{figure}

\looseness=-1 The gains so far could still be a sum of separate effects, with each subset helping only the downstream tasks closest to its own content. For visual transition reasoning to serve as a shared training primitive, two requirements must hold: 

\noindent\textbf{(a) Different sources improve the same capability.} Every subset improves WOVEN accuracy on the action types it does not train on (\S\ref{sec:enhance:sft}, Table~\ref{tab:woven-train}); the capability also spans reasoning operations, as training on forward prediction alone raises counterfactual substitution and removal by $48$ and $25$ percentage points (Appendices~\ref{app:E-F6} and~\ref{app:fwd-inv}), and it is not a shortcut (Table~\ref{tab:static-pull}; Appendix~\ref{app:static-option}).

\noindent\textbf{(b) The downstream gains result from that capability.} If so, all subsets should improve largely the same benchmarks, and subsets that improve the capability more should transfer more. Results show that the transfer profiles of the $11$ subsets are all positively correlated (Fig.~\ref{fig:fingerprint-grouping}, left), so the subsets improve largely the same benchmarks (detailed analysis in Appendix~\ref{app:E-F11-common}). Moreover, the more a subset improves WOVEN accuracy, the larger its average gain on the external benchmarks ($r{=}0.86$; Appendix~\ref{app:E-F11-common}), as expected, and the same WOVEN format with permuted or mirrored transitions does not reproduce the downstream gains (Appendix~\ref{app:content-control}). Together with the specificity shown in \S\ref{sec:enhance:transfer}, the downstream gains therefore come from the learned capability rather than from the individual subsets. We further test causality by intervention on the training data. With the training set fixed at $2{,}000$ items, we replace a growing share of task-specific items with a WOVEN subset on three benchmarks (SAT~\citep{ray2024sat}, ActionEQA~\citep{actioneqa2025}, and CLEVRER~\citep{yi2020clevrer}) from three different domains, and average results over three seeds. Results show that substituting up to $30$--$50\%$ (depending on the benchmark) of the task-specific items leaves accuracy comparable to training on these items alone, and above a control that simply deletes the replaced items (Fig.~\ref{fig:g2-mix}; Appendix~\ref{app:E-F12}). Transition supervision thus supplies part of what these tasks would otherwise learn from their own data, supporting the conclusion that the downstream gains result from the learned capability.

Together, these results establish visual transition reasoning as a shared training primitive; how its transfer is organized beyond this common component is the subject of \S\ref{sec:enhance:org}.

\section{Is There a Systematic Recipe for Training Visual World Modeling?}
\label{sec:enhance:org}

Sections~\ref{sec:enhance} and~\ref{sec:enhance:substitution} establish visual transition reasoning as a shared, trainable primitive. From the transfer profiles, error-typed analyses, and perturbation robustness of the controlled subsets, we distill a training recipe for visual world modeling in MLLMs (Fig.~\ref{fig:teaser}(4)): (1) to improve a downstream task, select supervision that teaches the reasoning operation the task requires, not supervision that matches its actions, scenes, or domains; (2) for robustness, prefer supervision with large state changes, such as object manipulation over camera motion; (3) supervise temporal and agent-driven tasks directly: transfer does not cross the boundary between the temporal family and the other three, nor the boundary between passive physical events and agent-driven actions; and (4) for downstream tasks that involve no transition, such as static perception, transition supervision does not help. The following subsections give the evidence behind each item; \S\ref{sec:recipe:prospective} tests the recipe prospectively on held-out benchmarks.

\subsection{What Determines Where Supervision Transfers?}
\label{sec:recipe:transfer}

\begin{figure}[!t]
\centering
\includegraphics[width=\linewidth]{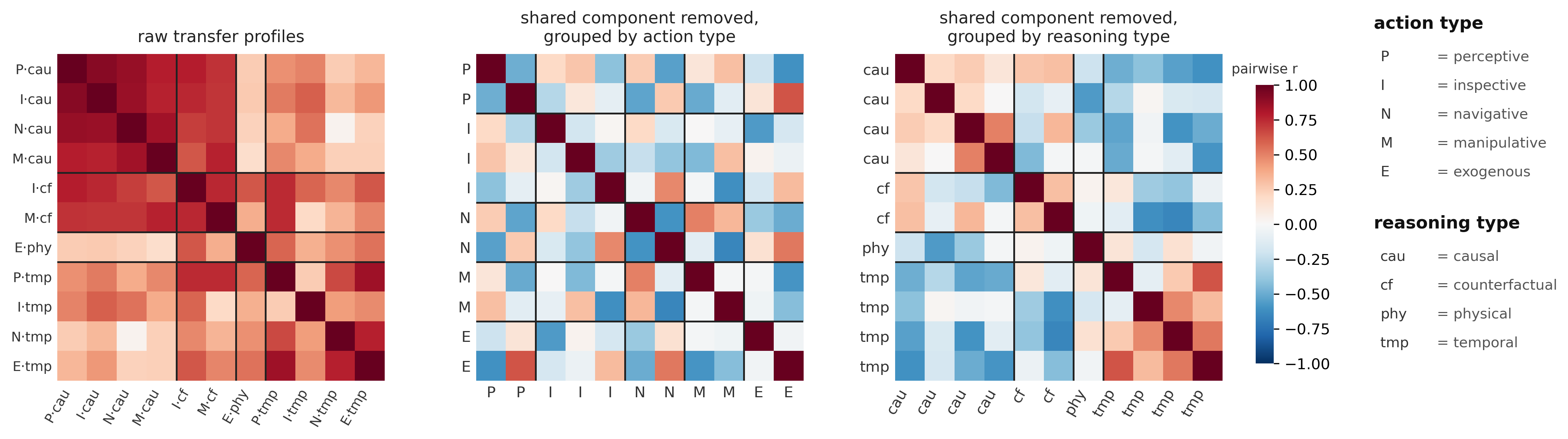}
\caption{\textbf{Correlation of transfer profiles under the two candidate groupings.} Pairwise Pearson correlation of the $11$ subsets' $26$-benchmark transfer profiles. Left: raw profiles, all pairs positive (the common component of \S\ref{sec:enhance:substitution}). Middle and right: the same residual matrix after the per-benchmark $z$-score, ordered by action type and by reasoning family; only the latter produces diagonal blocks.}
\label{fig:fingerprint-grouping}
\end{figure}

\looseness=-1 Similarity in scenes, actions, or application domains offers an intuitive basis for selecting transferable supervision. To test whether such similarity predicts downstream benefit, we compare subsets that share only an action type with those that share only a reasoning family. All transfer profiles contain the common component of \S\ref{sec:enhance:substitution}, so we remove it by standardizing the gains within each benchmark, which leaves each subset's relative pattern of benefit, and correlate these residual profiles pairwise (Fig.~\ref{fig:fingerprint-grouping}).

\looseness=-1 Surprisingly, results show that subsets teaching the same reasoning family have similar residual profiles, with a mean pairwise correlation of $+0.32$, whereas subsets sharing only an action type are no more similar than subsets sharing neither ($-0.22$ versus $-0.23$), so the action shown in a transition does not predict where its supervision transfers. Moreover, the same ordering holds under three other normalizations of the gains (Fig.~\ref{fig:fingerprint-grouping}; Appendix~\ref{app:E-F11}). Together with the transfer to held-out scenes (\S\ref{sec:enhance:sft}) and to external domains (\S\ref{sec:enhance:transfer}), the reasoning operation a source of supervision teaches, not the actions or scenes it shows, decides which downstream tasks it helps. \textbf{Recipe (1).} To improve a downstream task, select supervision that teaches the reasoning operation the task requires, not supervision that matches its actions, scenes, or domains; Table~\ref{tab:enhance-bestcell} and Appendix~\ref{app:E-F11} list the best-transferring subset for each benchmark.

\subsection{What Makes the Learned Capability Robust?}
\label{sec:recipe:robustness}

\looseness=-1 Recipe (1) leaves the extent of state change open. We examine whether larger state changes improve robustness by comparing subsets that differ in the actions that transform the scene.
Among the causal subsets, robustness gains increase from camera motion ($+1.7$) and object inspection ($+3.0$) to navigation ($+13.6$) and object manipulation ($+23.2$), and passive physical events give the largest gain of all subsets ($+29.9$; Table~\ref{tab:woven-train}; Appendix~\ref{app:E-F15}). \textbf{Recipe (2).} For robustness, prefer supervision with large state changes, such as object manipulation over camera motion.
\subsection{Where Does Transfer Break?}
\label{sec:recipe:composition}

\begin{wraptable}{r}{0.40\linewidth}
\vspace{-10pt}
\centering
\footnotesize
\setlength{\tabcolsep}{5pt}
\caption{\textbf{Gains on the two agent-action benchmarks by non-temporal subset} (percentage points over the base model; base accuracy: WorldPrediction $34.8$, ActionEQA $43.4$). Temporal subsets are covered by (a).}
\label{tab:exo-agent}
\begin{tabular}{@{}lrr@{}}
\toprule
Subset & \hd{World}{Prediction} & \hd{Action}{EQA} \\
\midrule
perc\_causal & $+2.4$ & $+12.5$ \\
insp\_causal & $+5.1$ & $+10.9$ \\
navi\_causal & $+4.1$ & $+4.4$ \\
mani\_causal & $+8.1$ & $+11.6$ \\
insp\_cf & $-5.1$ & $+12.5$ \\
mani\_cf & $+7.7$ & $+9.8$ \\
exo\_physical & $\mathbf{-10.2}$ & $\mathbf{-1.7}$ \\
\bottomrule
\end{tabular}
\vspace{-6pt}
\end{wraptable}
Although supervision on one reasoning operation often teaches others, as forward-dynamics supervision alone raises counterfactual substitution and removal by $48$ and $25$ percentage points (\S\ref{sec:enhance:substitution}), two kinds of transition reasoning are not learned this way, nor are capabilities beyond state transitions.
\looseness=-1 (a) Temporal localization is learned only from temporal supervision: the four temporal subsets raise temporal-adjacency accuracy from $25.7\%$ to $55.6$--$58.7\%$, whereas the seven other subsets leave it near the baseline, and supervision on temporal adjacency and on temporal ordering do not transfer to each other (Table~\ref{tab:woven-train}; Appendix~\ref{app:E-F7}). (b) Endogenous (agent-driven) transitions are not learned from exogenous supervision: the subset of passive physical events contains no agent action to learn from, and it lowers accuracy on the two benchmarks built on an agent's action and its outcome, WorldPrediction ($-10.2$) and ActionEQA ($-1.7$; Table~\ref{tab:exo-agent}; Appendix~\ref{app:E-F11}). \textbf{Recipe (3).} Supervise temporal and agent-driven tasks directly: transfer does not cross the boundary between the temporal family and the other three reasoning families, nor the boundary between passive physical events and agent-driven actions. (c) Capabilities beyond state transitions are not learned from transition supervision: the $4$ benchmarks of static perception and general video QA do not improve under any training (Table~\ref{tab:enhance-bestcell}). \textbf{Recipe (4).} For downstream tasks that involve no transition, transition supervision does not help.

\subsection{Can the Recipe Predict Transfer on Held-Out Benchmarks?}
\label{sec:recipe:prospective}

\begin{table}[!t]
\centering
\footnotesize
\setlength{\tabcolsep}{4pt}
\caption{\textbf{Prospective test of the recipe on held-out benchmarks.} Accuracy (\%), mean over $8$ models; $\Delta$ = Recipe $-$ comparison mixture; Item = recipe item tested; Base = untrained model on the same questions; Set A: agent-action questions, Set B: passive-physics questions, Set C: camera-motion questions.}
\label{tab:prosp-results}
\begin{tabular}{@{}lclrrrrrr@{}}
\toprule
Comparison mixture & Item & Questions & Base & Recipe & Comp. & $\Delta$ & $95\%$ CI & $p$ \\
\midrule
Temporal-heavy & (1) & Set A & 31.1 & 37.9 & 34.4 & $+3.5$ & $[+2.03, +4.92]$ & 0.0002 \\
Uniform & (1) & Set A & 31.1 & 37.9 & 36.4 & $+1.5$ & $[+0.18, +2.81]$ & 0.03 \\
Camera-heavy & (1) & Set C & 54.1 & 58.4 & 57.8 & $+0.6$ & $[+0.02, +1.14]$ & 0.04 \\
Exogenous-heavy & (3) & Set A & 31.1 & 37.9 & 33.2 & $+4.7$ & $[+3.19, +6.17]$ & 0.0002 \\
Exogenous-heavy & (3) & Set B & 70.3 & 71.0 & 71.5 & $-0.5$ & $[-1.05, -0.02]$ & 0.04 \\
\bottomrule
\end{tabular}
\end{table}

To test the generality of our recipe, we use it to select training supervision before any training and evaluate the resulting models on external benchmarks that played no part in its derivation. For questions about agent actions and their outcomes, we compose a training mixture of WOVEN items by the recipe (Recipe) and compare it with mixtures of the same size composed in other ways: by matching the action (Camera-heavy), by a different reasoning operation (Temporal-heavy), by passive physical events (Exogenous-heavy), or with no selection (Uniform). We train Qwen2.5-VL-3B-Instruct on each mixture with $8$ seeds and evaluate the models on questions from MMSI-Bench~\citep{yang2025mmsi}, VLM4D~\citep{vlm4d2025}, PhysBench~\citep{chow2025physbench}, SPAR-Bench~\citep{zhang2025spar}, MVBench~\citep{mvbench2023}, IntPhys~2~\citep{intphys2}, and CameraBench~\citep{lin2025camerabench} (Appendix~\ref{app:recipe-prospective}).
Results show that all five comparisons come out as the recipe predicts (Table~\ref{tab:prosp-results}). On agent-action questions, the Recipe mixture exceeds the others by $1.5$ to $4.7$ percentage points ($p\le0.03$), and on camera-motion questions it also outperforms the mixture matched to the camera actions ($+0.6$, $p=0.04$), so selecting supervision by the reasoning operation matters more than matching the action. On passive-physics questions, the Exogenous-heavy mixture is ahead instead ($-0.5$, $p=0.04$), as item (3) predicts. Our recipe therefore generalizes to new benchmarks.

\section{Related Work}
\label{sec:related}

\looseness=-1 \textbf{World models and video priors.} World models predict how a scene evolves, either in a latent space~\citep{hafner2023dreamerv3,assran2025vjepa2} or by generating video~\citep{openai2024sora,genie3,wan2025,cosmospredict2}, and video models increasingly serve as simulators for planning~\citep{du2023vlp,zhou2024robodreamer} and as sources of training data for downstream learning~\citep{jiang2025dreamgen}. Transition-level supervision has also been used directly: state transition pretraining teaches GUI agents the forward and inverse dynamics of interface states~\citep{wang2026stp}, and intervention-based world-state data support controlled generation, action prediction, and policy transfer~\citep{cai2026cgworld}. WOVEN uses video-pretrained transition priors for a different purpose, as a source of controlled supervision for studying which transition reasoning MLLMs learn and reuse, which complements evaluations of rollout quality~\citep{huang2024vbench,pbench} and of world models on planning and interaction tasks~\citep{worldprediction2025,warrier2025worldtest}.

\noindent\textbf{Benchmarks and supervision for multimodal reasoning.} A large body of benchmarks tests MLLMs on one facet of visual change at a time: spatial reasoning under viewpoint change~\citep{yin2025mindcube,yang2025mmsi,li2025viewspatial,wang2025spatialviz}, physical and intuitive-physics reasoning~\citep{chow2025physbench,sreekumar2025inphyre,intphys2,yi2020clevrer}, counterfactual reasoning about videos~\citep{causalvqa,mucr2024,cripp2022,chinchure2025blackswan}, temporal ordering and localization~\citep{temporalbench,shangguan2025tomato,liu2024tempcompass}, and embodied and robotic question answering~\citep{openeqa2024,actioneqa2025,chen2025robo2vlm}, with recent suites targeting world modeling as a whole~\citep{huang2025wmabench,worldprediction2025}. On the supervision side, synthetic data has been used to teach MLLMs spatial reasoning~\citep{ray2024sat} and transferable reasoning through rule-based games~\citep{xie2026vigal}, and fully controlled scenes have been used to study what fine-tuning transfers to real images~\citep{rizzoli2026synthetic}. WOVEN relates these facets through a common $(s,a,s')$ representation and controlled training comparisons, separating the reasoning operation of a task from its action and scene content, so that supervision built for one facet can be tested on every other facet and the resulting gains compared on equal terms (Appendices~\ref{app:related-work} and~\ref{app:cell-coverage}).

\section{Conclusion}
\label{sec:conclusion}

\looseness=-1 We used a factorized experimental design to study how the content and reasoning demands of supervision shape the acquisition and transfer of visual transition reasoning. The rollout priors and conditional generation of video generation models supplied the controlled transitions for these comparisons. To the best of our knowledge, we are the first to establish visual transition reasoning as a shared training primitive in MLLMs across diverse task domains and derive a systematic training recipe from controlled comparisons of its downstream transfer. More broadly, our work paves the way for developing visual world modeling in MLLMs through reusable reasoning capabilities whose training supervision is designed by controlled experiments.

\section*{Acknowledgments}
We thank Amazon for their support of this work.

\bibliographystyle{abbrvnat}
\bibliography{references}

\appendix
\clearpage
\renewcommand{\partname}{}
\part{Appendix}
\parttoc
\clearpage

\section{Background and motivation}

\subsection{World modeling as a cognitive primitive}
\label{app:primitive}

This appendix unfolds the developmental, neuroscientific, and AI-program evidence that world modeling is a cognitive primitive with an intrinsic structure, established independently of any benchmark, summarized in \S\ref{sec:intro}. The point of assembling this evidence is not decorative: it is what licenses treating the two axes of \S\ref{sec:methodology} as the primitive's own structure rather than a decomposition we impose, and it is why the scattered external failures noted in \S\ref{sec:intro} can be read as manifestations of that structure.

\paragraph{Interpretation of the query formulation.}
The formulation in \S\ref{sec:formalization} connects forward and inverse internal models~\citep{wolpert1995internalmodels} with physical simulation~\citep{battaglia2013simulation} and counterfactual reasoning~\citep{pearl2009causality}. A shared trajectory mechanism makes the episode conditions explicit:
\begin{equation}
\tau_a=\mathcal G(s,a,u),\qquad
\operatorname{start}(\tau_a)=s,\qquad
\operatorname{end}(\tau_a)=s',
\label{eq:transition-structure}
\end{equation}
Here $u$ collects the episode's remaining background conditions; evaluating an alternative action in $\mathcal G$ preserves the same initial state and background conditions. This notation describes the transition structure underlying the query distributions.

Forward, inverse, and removal recover the resulting state, action, and initial state, respectively; removal asks what the scene would look like had the action not occurred, so its target is the initial state. Substitution retains the factual action and observed endpoint as evidence for the alternative outcome; its input does not include the initial state. Outcome and cued prediction share a physical prediction target but differ in the supplied principle information, such as containment. Ordering and adjacency recover global and local temporal relations within the same transition, with adjacency defined among the sampled frames. In the physical and temporal families, the action description enters the condition only when it is given.

\paragraph{Developmental psychology.}
Predicting the visual consequence of an action and inferring the action behind a state change is documented in infants from the first months of life. Spelke's core-knowledge programme identifies object permanence, cohesion, and solidity as innate primitives~\citep{spelke2007coreknowledge,spelke2022whatbabiesknow}, with violation-of-expectation paradigms (e.g., \citealt{baillargeon1987objectpermanence}) demonstrating sensitivity to physical events long before language acquisition. Counterfactual thought is a classical topic in cognitive science~\citep{lewis1973counterfactuals,roese1997counterfactualthinking} formalized in causal inference~\citep{pearl2009causality}. The temporal structure of a single action is represented within the forward model itself: forward models predict the time course of a movement, not only its endpoint, and this trajectory representation underlies online prediction and correction~\citep{wolpert1995internalmodels}. Mental simulation of physical outcomes admits computational analogues in noisy-simulation accounts~\citep{battaglia2013simulation}.

\paragraph{Cognitive neuroscience.}
Action-conditioned prediction maps onto the predictive-coding and forward-model accounts of motor and perception~\citep{friston2010freeenergy,wolpert1995internalmodels}. Action-from-observation maps onto the mirror-neuron and action-understanding circuitry~\citep{rizzolatti2001neurophysiological}. The exo/endo dissociation distinguishes externally-triggered from self-initiated actions with distinct prefrontal signatures and behavioral dissociations~\citep{haggard2008humanvolition,passingham2010specialization,khalighinejad2018dissociating}. Within the visual system, the dorsal/ventral two-stream dissociation~\citep{ungerleider1982twostreams,goodale1992separate,milner1995visualbrain} is the structural basis for the geometric/appearance partition of the perturbation probe (\S\ref{sec:bench:perturb}).

\paragraph{Recurring target across AI research programs.}
The primitive recurs under several labels: embodied planning requires forward rollout from current observation and a proposed action~\citep{openeqa2024,luo2025robobench,embodiedbench2025,du2023vlp}; counterfactual and causal video reasoning requires reasoning over $(s,a,s')$ triples under intervention on $a$~\citep{causalvqa,mucr2024,cripp2022}; mental simulation of ``what-if'' movement is the primary axis of recent spatial-reasoning benchmarks~\citep{yin2025mindcube,vlm4d2025}.

\subsection{The two axes as the intrinsic structure of the primitive}
\label{app:two-axes}

The action and reasoning axes are not a decomposition we impose on the data; each enumerates distinctions that cognitive science documents as real and dissociable, so their cross-product is the primitive's own structure made measurable rather than a set of principal components fit to failures.

\paragraph{Reasoning axis.}
The operations one can perform over $(s,a,s')$ are organized by three orthogonal structural properties of a predictive model. The first is causal depth: Pearl's hierarchy separates prediction from intervention ($\mathrm{do}(a)$) from counterfactual reasoning, a strictly nested ladder in which each rung expresses what the one below cannot~\citep{pearl2009causality}. Physical modeling occupies the predictive base, the passive dynamics of objects under intuitive physics~\citep{spelke2007coreknowledge,battaglia2013simulation}; causal dynamics occupies the intervention rung, the visual effect of a deliberate action; counterfactual reasoning occupies the top rung, the visual scene under an action that did not occur. The second property is direction of inference, the paired internal models of sensorimotor control: a forward model predicts the consequences of an action and an inverse model recovers the action from an observed change~\citep{wolpert1995internalmodels,friston2010freeenergy}, and within the counterfactual rung the analogous split is removal of an action versus its substitution by an alternative. The third property is temporal granularity: because a forward model represents the time course of a single action rather than its endpoints alone~\citep{wolpert1995internalmodels}, the temporal family probes that trajectory at two grains within one transition, coarse ordering of the intermediate states and fine adjacency of consecutive ones. The eight reasoning types are the cells of these three properties, and the causal-depth nesting further predicts a difficulty ordering, physical below causal below counterfactual, that the evaluation tests directly (Appendix~\ref{sec:assess:findings}).

\paragraph{Action axis.}
The kinds of transition the primitive must represent track the core systems through which human cognition represents the physical world. The coarsest cut is agency: cognitive neuroscience separates externally triggered from self-initiated action with distinct prefrontal signatures~\citep{haggard2008humanvolition,passingham2010specialization,khalighinejad2018dissociating}, giving one exogenous type, the passive physics of objects, against agentic types. The exogenous type is organized by object core knowledge, with permanence, cohesion, and solidity as object primitives and gravity, support, inertia, containment, and collision as physical-event classes~\citep{spelke2007coreknowledge,battaglia2013simulation}. The agentic types are separated by the reference frame in which the action transforms the visual state: egocentric self-motion and object-centered transformation rely on different reference frames~\citep{burgess2006spatialrefs,epstein2017cognitivemap}, and object-based mental rotation is behaviorally and neurally dissociable from egocentric perspective-taking, giving perceptive (egocentric viewpoint change) and inspective (object-centered examination) as distinct cells, with navigative (goal-directed locomotion) and manipulative (goal-directed action on an object) grounded in the agent core system. The five action types thus partition how a visual state can be caused to change, each a documented system rather than a label.

\paragraph{Why this makes the external failures legible.}
Because the two axes are the primitive's own degrees of freedom, any benchmark that exercises world modeling necessarily lands somewhere on their grid. Existing benchmarks each fix one axis and leave the other uncontrolled: intuitive-physics suites fix the exogenous, passive-prediction corner; spatial suites fix the perceptive and inspective actions while mixing operations; counterfactual video QA fixes the counterfactual operation while mixing actions; temporal suites fix the temporal operation. Their failures are therefore scattered marginals of one grid, which is why they look unrelated until the grid is drawn. WOVEN draws it by instantiating both axes jointly; the empirical correspondence between each external marginal and the WOVEN cell it falls in is quantified in \S\ref{app:correspondence}.

\subsection{Failure-mode triangulation across external benchmarks}
\label{app:gap-triangulation}

This appendix unfolds the three-sub-domain triangulation summarized in \S\ref{sec:intro}.

\paragraph{Spatial sub-domain (dorsal-stream correspondence).}
Cognitive neuroscience distinguishes object identity, processed in the ventral ``what'' pathway, from spatial relations and viewpoint, processed in the dorsal ``where'' pathway~\citep{ungerleider1982twostreams,goodale1992separate,milner1995visualbrain}. The AI analogues of dorsal-stream competence are precisely the tasks on which MLLMs fail hardest: MindCube reports near-random MLLM performance on perspective-taking and ``what-if'' camera movement~\citep{yin2025mindcube}; VLM4D identifies systematic failure modes on translational and rotational motion~\citep{vlm4d2025}; 3DSRBench documents degraded accuracy under uncommon viewpoints~\citep{sdrbench2025}; COMFORT finds VLMs fail to track reference frames across perspective change~\citep{comfort2024}. Each is, in world-modeling terms, a forward-prediction query under a self-initiated viewpoint or navigation action.

\paragraph{Physical sub-domain (Spelke ``objects'' core system).}
Developmental psychology documents infants' intuitive physics, object permanence, solidity, continuity, inertia, as core knowledge emerging in the first months of life~\citep{spelke2007coreknowledge,spelke2022whatbabiesknow,baillargeon1987objectpermanence}. AI counterparts consistently report MLLMs at or near chance: IntPhys~$2$ shows that state-of-the-art vision models approach $50\%$ on violation-of-expectation items~\citep{intphys2}; PhysBench reports persistent MLLM shortfalls against human baselines across four physical-reasoning domains~\citep{chow2025physbench}; CRIPP-VQA probes counterfactual reasoning about implicit physical properties and finds comparable weakness~\citep{cripp2022}.

\paragraph{Causal/counterfactual sub-domain.}
Direct video-reasoning analogues, CausalVQA for physically grounded counterfactual video QA~\citep{causalvqa} and MuCR for cross-modal causal-link identification~\citep{mucr2024}, together with the counterfactual-simulation dimension of WM-ABench~\citep{huang2025wmabench}, report systematic MLLM weakness on intervention-on-action queries.

\paragraph{Parallel positive evidence: VGMs pass where MLLMs fail.}
Benchmarks evaluating VGMs directly on action-conditioned rollout report frontier models~\citep{wan2025,cogvideox2024,hunyuanvideo2024,cosmospredict2,genie3} passing well above chance: PBench evaluates world-model rollout across physics, robotics, and commonsense dimensions~\citep{pbench}; VBench decomposes generation quality into $16$ hierarchical dimensions with human-aligned metrics~\citep{huang2024vbench}; ChronoMagic-Bench targets metamorphic temporal evolution~\citep{chronomagic2024}. Downstream work assumes the world-modeling capability and exploits it as a black-box simulator: MindJourney couples a frozen VLM to a video-diffusion world model and reports an $8\%$ boost on SAT~\citep{yang2025mindjourney}; DreamGen distills a video world model into robot policies that generalize to $22$ new behaviors~\citep{jiang2025dreamgen}; InterPose uses VGM-hallucinated intermediate frames to outperform DUSt3R~\citep{cai2025interpose}; RoboDreamer~\citep{zhou2024robodreamer} and VLP~\citep{du2023vlp} exploit the same rollout capability.

\paragraph{Existing world-model benchmarks do not close the gap.}
WM-ABench~\citep{huang2025wmabench}, WorldPrediction~\citep{worldprediction2025}, WorldTest~\citep{warrier2025worldtest}, and PBench~\citep{pbench} target world-modeling evaluation but leave the world modeling in MLLMs question open: they aggregate across capabilities without controlling interventions on $a$, treat VGMs rather than MLLMs as the subject, confine themselves to grid-world or text-only environments, or probe only a single sub-domain.

\subsection{Cell-level prior-work coverage and downstream surfaces}
\label{app:cell-coverage}

\begin{figure}[!htbp]
\centering
\includegraphics[width=\linewidth]{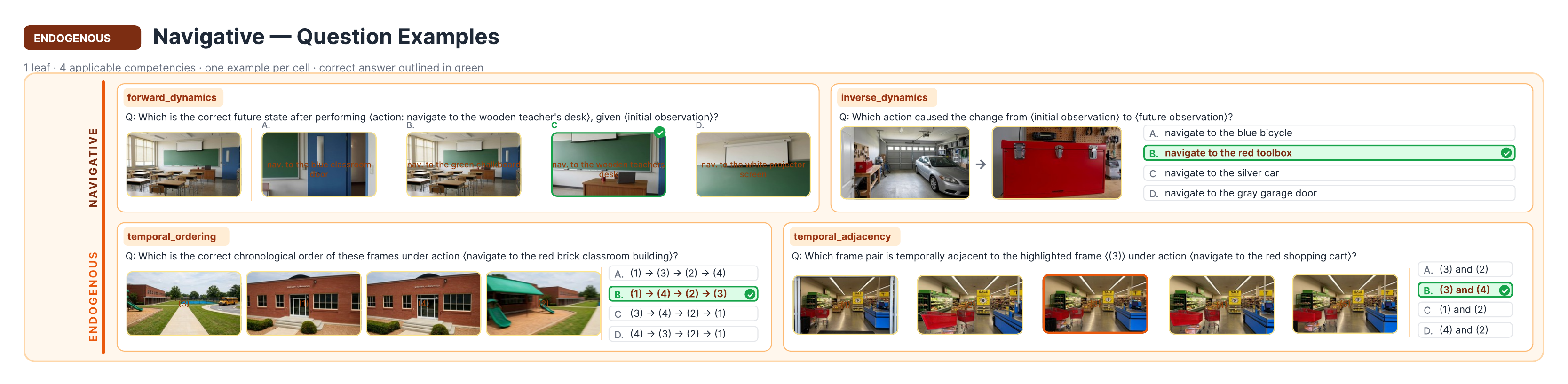}
\caption{Navigative action type: representative QA items showing in-scene navigation targets across all four navigative reasoning types.}
\label{fig:nav-examples}
\end{figure}

\begin{figure}[!htbp]
\centering
\includegraphics[width=\linewidth]{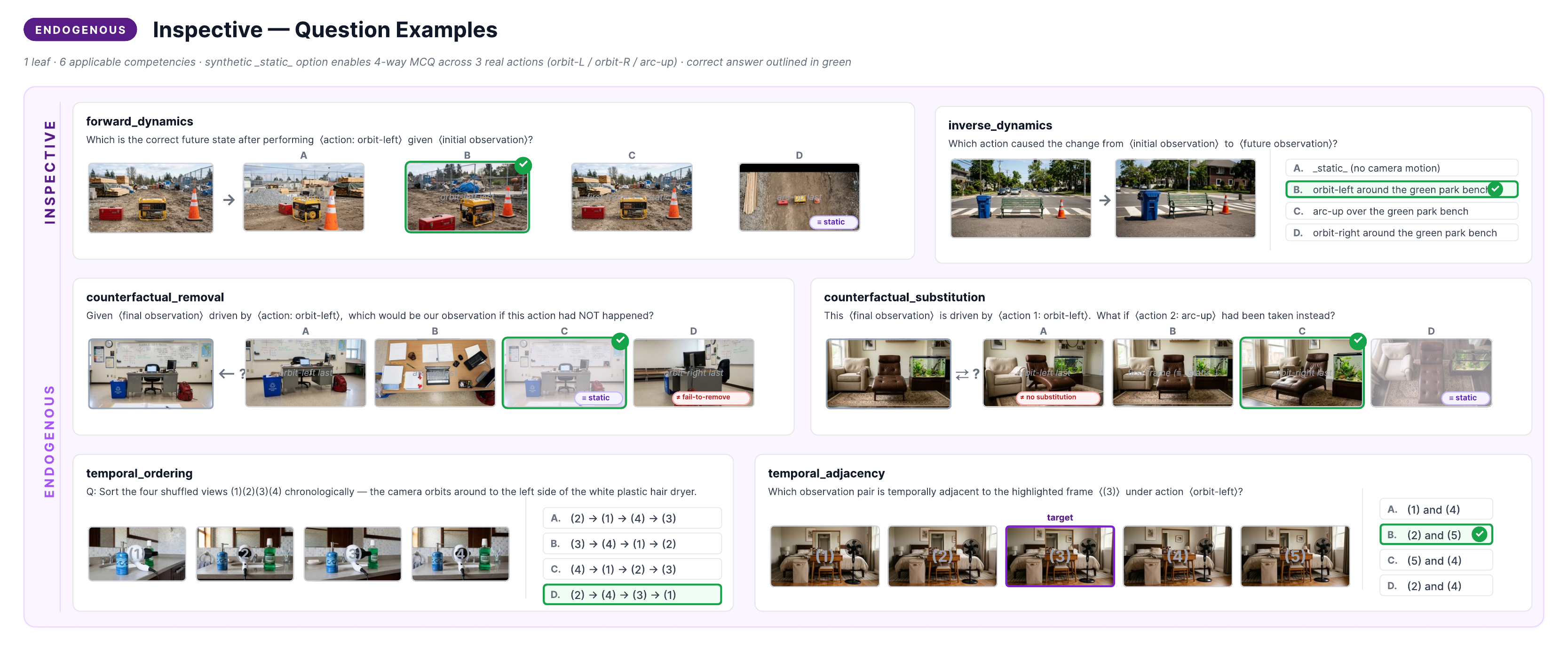}
\caption{Inspective action type: representative QA items showing orbit-L / orbit-R / arc-up actions paired with all six inspective reasoning types.}
\label{fig:insp-examples}
\end{figure}

\begin{figure}[!htbp]
\centering
\includegraphics[width=\linewidth]{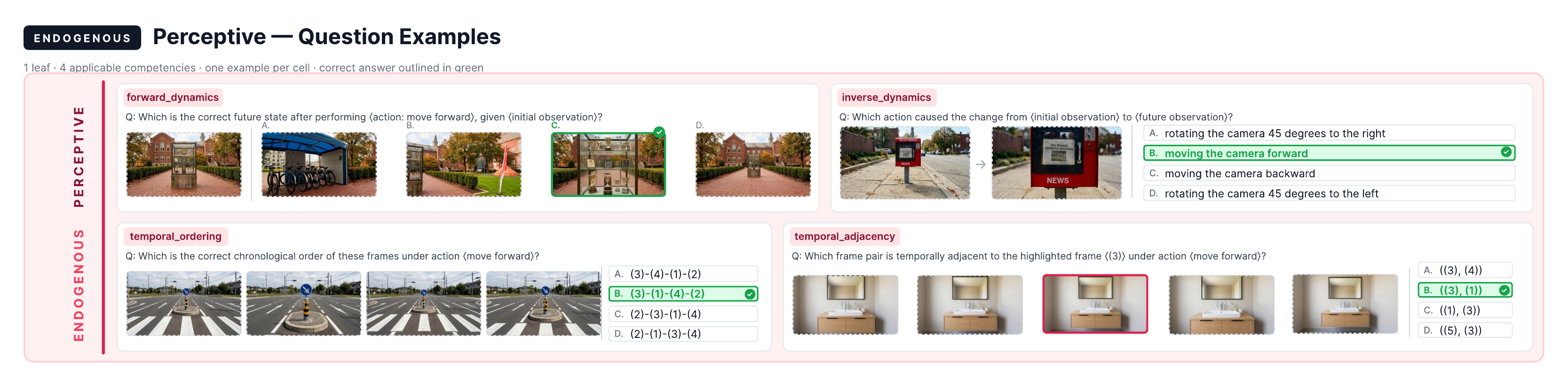}
\caption{Perceptive action type: representative QA items across all four perceptive reasoning types (forward dynamics, inverse dynamics, temporal ordering, temporal adjacency).}
\label{fig:perc-examples}
\end{figure}

This appendix collects the per-axis prior-work coverage discussion and the per-cell downstream-relevance argument deferred from \S\ref{sec:benchmark}. Together they substantiate the claim, made in \S\ref{sec:intro}, that prior benchmarks each touch a slice of world modeling but no single benchmark commits to the full action type $\times$ reasoning type cross-factoring.

\paragraph{Action axis: prior-work cell coverage.}
Existing benchmarks at the action-axis cell touch only sub-fragments of WOVEN's taxonomy. Embodied-AI evaluation suites are dominated by the \emph{manipulative} and \emph{navigative} cells~\citep{cheng2025embodiedeval,embodiedbench2025,openeqa2024,ecbench2025,luo2025robobench}; spatial-reasoning benchmarks center on \emph{perceptive} and \emph{inspective}~\citep{yin2025mindcube,yang2025mindjourney,chen2025spa,threedr1_2025,vlm4d2025,open3dvqa2025}; physical-reasoning benchmarks operate primarily on \emph{exogenous}~\citep{chow2025physbench,intphys2,qi2025bear,pbench,huang2025wmabench}. Each line implicitly probes a slice of world modeling but stops short of formalizing the action axis as such, and no single benchmark commits to a full cross-section. WOVEN's contribution at this axis is the explicit, exhaustive partition; failure attribution then becomes a question of which cell rather than whether the model is broken in some unspecified way.

\paragraph{Reasoning axis: prior-work cell coverage.}
Single reasoning types appear scattered across prior benchmarks: forward prediction is the standard probe in physical-modeling work~\citep{intphys2,chow2025physbench,pbench,huang2025wmabench}; counterfactual reasoning is the dedicated focus of \citep{causalvqa,mucr2024,comfort2024} and the concept-mixing variant of \citep{wu2024conceptmix}; temporal coherence is unpacked in video-generation evaluation~\citep{huang2024vbench,chronomagic2024,ling2025vmbench} and in video-language understanding suites~\citep{mvbench2023,yang2024videolanguage}. None co-instantiate all four families on a shared, controlled action axis, leaving the cross-family interaction structure unmeasured. WOVEN holds the four families jointly factored against the action axis of \S\ref{sec:benchmark}; the resulting (action type, reasoning type) cells are the elementary units of fine-grained diagnosis.

\paragraph{Action axis: downstream surfaces.}
Each action type tracks a distinct downstream surface: \emph{exogenous} feeds physical-AI safety~\citep{chow2025physbench,pbench}; \emph{perceptive} and \emph{inspective} feed visual spatial-mental-modeling~\citep{yin2025mindcube,yang2025mindjourney,chen2025spa,threedr1_2025}; \emph{navigative} and \emph{manipulative} feed embodied agents and visual planners~\citep{cheng2025embodiedeval,embodiedbench2025,openeqa2024,du2023vlp,jiang2025dreamgen}. A failure localized to one action type identifies which downstream surface the model is most unfit for.

\paragraph{Reasoning axis: downstream surfaces.}
Forward dynamics is the central capability behind embodied planning and visual world-model rollouts~\citep{du2023vlp,jiang2025dreamgen,zhou2024robodreamer,hafner2023dreamerv3}; inverse dynamics underlies action recognition and learning-from-observation~\citep{rizzolatti2001neurophysiological}; counterfactual templates underlie causal video reasoning~\citep{causalvqa,comfort2024}; physical-modeling templates underlie physical-AI safety~\citep{chow2025physbench,pbench}; temporal-coherence templates underlie video-understanding pipelines~\citep{mvbench2023}. As with the action type axis, failure localized to a reasoning type family identifies the downstream surface most affected.

\paragraph{Scene axis: scale and breadth.}
A central premise of the methodology is that scene diversity is a \emph{controlled} dimension to be held fixed while the action and reasoning axes are perturbed. Real-world video corpora cannot guarantee both the action-axis controllability of \S\ref{sec:benchmark} and full scene-axis breadth simultaneously. Simulators, prevalent in embodied-AI benchmarks~\citep{cheng2025embodiedeval,embodiedbench2025,openeqa2024,ecbench2025,luo2025robobench} and physical-AI benchmarks~\citep{huang2025wmabench,intphys2}, deliver controllability but commit to a small number of pre-built environments with a documented visual-domain gap. A VGM stimulus engine delivers both: under fixed scene specification, the rollout produces visually realistic stimuli at every (action type, scene type) cell, and across scenes, lexical scene specification swaps cleanly without further pipeline change. This is the most direct realization of the VGM-as-stimulus-engine argument of \S\ref{sec:methodology}.

\subsection{Empirical correspondence between external failures and WOVEN cells}
\label{app:correspondence}

This appendix makes the claim of \S\ref{app:two-axes} quantitative: the failures external benchmarks report in isolation are the marginals of WOVEN's grid, so the cell where each external failure falls should be hard in WOVEN exactly when the external task is hard. We establish the correspondence in two layers, a conceptual mapping of each benchmark to its WOVEN cell and three statistical tests of that mapping, all computed from published baselines and our existing $38$-model evaluation without additional experiments.

\paragraph{Conceptual mapping.}
Table~\ref{tab:correspondence} assigns each external benchmark, at the sub-task level where one is provided, to the WOVEN (action type, reasoning type) cell it instantiates. The mapping is many-to-one by construction: several external benchmarks fall in the same WOVEN cell because each isolates one axis while WOVEN crosses both, which is the precise sense in which WOVEN unifies rather than duplicates them.

\begin{table}[!htbp]
\centering
\small
\caption{Conceptual mapping of external benchmarks to the WOVEN cells they instantiate. Several benchmarks fall in one cell (many-to-one), the sense in which WOVEN unifies scattered marginals rather than duplicating any one of them. \textbf{Stat} marks which correspondence test the pair primarily supports: \textbf{M} model-level rank correlation, \textbf{D} cell-level difficulty correlation.}
\label{tab:correspondence}
\begin{adjustbox}{max width=\linewidth}
\begin{tabular}{lll}
\toprule
External benchmark (sub-task) & WOVEN cell / axis & Stat \\
\midrule
VSI-Bench~\citep{yang2025thinkinginspace} (relative direction) & perceptive/inspective $\times$ forward & M, D \\
VSI-Bench (appearance order) & perceptive $\times$ temporal & M, D \\
VSI-Bench (route planning) & navigative $\times$ forward & M, D \\
ViewSpatial-Bench (camera vs.\ human) & manipulative view-axis (ego/allo) & M, D \\
3DSRBench (orientation, uncommon view) & inspective $\times$ forward; orientation perturbation & M, D \\
VLM4D (rotational/directional motion) & inspective/perceptive rotation & D \\
MindCube (perspective-taking, what-if) & inspective $\times$ \{forward, counterfactual\} & M, D \\
IntPhys~2 (object principles) & exogenous $\times$ physical & D \\
PhysBench (physical reasoning) & exogenous $\times$ \{physical, forward\} & M, D \\
CLEVRER (predictive, counterfactual) & exogenous $\times$ \{physical, counterfactual\} & D \\
CausalVQA (counterfactual, hypothetical) & manipulative $\times$ counterfactual & D \\
TemporalBench (event order/adjacency) & all actions $\times$ temporal & M, D \\
\bottomrule
\end{tabular}
\end{adjustbox}
\end{table}

\paragraph{Statistical tests.}
Three complementary analyses, all from published numbers and our evaluation, test whether the mapping holds empirically. (i) \emph{Model-level rank correlation.} Our cohort and the external benchmarks share a backbone of commonly evaluated models, principally the Qwen2.5-VL ladder ($3/7/32/72$B) and Gemma-3 ($4/12/27$B), which appear in most recent spatial and physical benchmarks, often with the sub-task breakdowns that map to WOVEN cells. For each mapped pair we compute the Spearman correlation between the shared models' WOVEN-cell accuracy and their external sub-task accuracy; the Qwen2.5-VL ladder is the strongest anchor because its four scale points trace a controlled difficulty gradient, so a monotone co-trend is itself evidence. As the shared set per benchmark is small (about $4$ to $8$ models), this layer is corroborating. (ii) \emph{Cell-level difficulty correlation.} For each mapped pair we take the WOVEN-cell difficulty (frontier accuracy or human-frontier gap, from our evaluation) and the external sub-task difficulty (published frontier-vs-human gap), and correlate the difficulty profile across all pairs. This needs no model overlap and tests the structural claim directly, whether the WOVEN cells that are hard are the external tasks that are hard, and is the statistical backbone of the correspondence. (iii) \emph{Common-reference normalization.} For benchmarks reporting few of our models, we normalize both sides to a shared anchor, performance relative to Qwen2.5-VL-$72$B or to the human ceiling, so that thin-overlap benchmarks still enter the cell-level correlation on a common scale.

\paragraph{Reading the result.}
A positive correspondence supports the central claim that WOVEN's taxonomy is the primitive's intrinsic structure and reflects the same failure modes external work reports, while the many-to-one mapping preserves WOVEN's distinct contribution: the correspondence is convergence onto one grid rather than redundancy with any benchmark on it.

\subsection{Extended related work}
\label{app:related-work}

WOVEN sits at the intersection of three programs; cell-level prior-work coverage and downstream-surface mapping are in Appendix~\ref{app:cell-coverage}.

\paragraph{World modeling.}
The term \emph{world model} spans heterogeneous formal commitments: latent predictive architectures (DreamerV3~\citep{hafner2023dreamerv3}, V-JEPA~2~\citep{assran2025vjepa2}); generative pixel-rollout simulators (Sora~\citep{openai2024sora}, Genie~3~\citep{genie3}, Cosmos-Predict2~\citep{cosmospredict2}, HunyuanVideo~\citep{hunyuanvideo2024}; quality benchmarks~\citep{huang2024vbench,chronomagic2024,ling2025vmbench}); symbolic-environment learners~\citep{warrier2025worldtest,chen2025spa}; and procedural-planning formulations~\citep{worldprediction2025}. Recent surveys note that no single operationalization dominates~\citep{li2025critiquesworldmodels}, and existing world-model evaluation benchmarks inherit this fragmentation~\citep{huang2025wmabench,pbench,zhang2025worldinworld,wang2025diswm}. We focus on action-conditioned visual transition reasoning over $(s,a,s')$, using a common transition representation to study the shared and distinct supervision requirements of visual reasoning tasks.

\paragraph{VGMs as world simulators.}
The Sora technical report~\citep{openai2024sora} introduced the ``video generation as world simulator'' framing, since adopted by CogVideoX~\citep{cogvideox2024}, HunyuanVideo~\citep{hunyuanvideo2024}, the Wan family~\citep{wan2025}, Cosmos-Predict2~\citep{cosmospredict2}, and Genie~3~\citep{genie3}, with \citet{yang2024videolanguage} articulating the strongest version of the claim. Empirical support comes from direct rollout measurement~\citep{pbench,huang2024vbench,chronomagic2024,ling2025vmbench} and downstream exploitation as black-box simulators~\citep{yang2025mindjourney,jiang2025dreamgen,cai2025interpose,zhou2024robodreamer,du2023vlp,wang2025diswm}; closed-loop evaluation~\citep{zhang2025worldinworld} additionally validates that VGM-rollout \emph{controllability} drives embodied utility. These findings motivate our use of Wan2.2-I2V-A14B~\citep{wan2025} to generate diverse, controllable transitions for structured supervision.

\paragraph{Multimodal reasoning benchmarks.}
Independent literatures document MLLM failures along axes that each touch a slice of action-conditioned visual world modeling without naming it: \emph{counterfactual video reasoning} (CausalVQA~\citep{causalvqa}, MuCR~\citep{mucr2024}, CRIPP-VQA~\citep{cripp2022}); \emph{post-event physical prediction} (IntPhys~2~\citep{intphys2}, PhysBench~\citep{chow2025physbench}, BEAR~\citep{qi2025bear}); \emph{viewpoint-conditioned spatial inference} (MindCube~\citep{yin2025mindcube}, VLM4D~\citep{vlm4d2025}, 3DSRBench~\citep{sdrbench2025}, COMFORT~\citep{comfort2024}, Open3D-VQA~\citep{open3dvqa2025}, 3D-R1~\citep{threedr1_2025}); and \emph{video understanding plus embodied evaluation} (MVBench~\citep{mvbench2023}, ECBench~\citep{ecbench2025}, OpenEQA~\citep{openeqa2024}, EmbodiedBench~\citep{embodiedbench2025}, EmbodiedEval~\citep{cheng2025embodiedeval}, RoboBench~\citep{luo2025robobench}; on the training side, EMCompress~\citep{fan2025emcompress}). These four families have been pursued in isolation with disjoint distractor designs, yet their failure modes factor through a shared structural deficit (Appendix~\ref{app:gap-triangulation} makes the triangulation explicit). WOVEN builds on these diagnostic insights by expressing reasoning tasks over a common $(s,a,s')$ representation and varying the semantic components of supervision, enabling controlled study of shared training requirements and the organization of transfer.

\section{Benchmark construction details}

\subsection{Off-the-shelf video priors on WOVEN}
\label{app:video-priors}

Before training, we ask whether engines pretrained at scale on video reconstruction and future-frame prediction already solve WOVEN, and whether the transition knowledge they hold is directly queryable. We evaluate three frozen off-the-shelf engines on \texttt{test\_id} under one shared read-out recipe: the engine scores every option of the $4$-way MCQ and the argmax is taken (chance $=25\%$).

\textbf{Video encoder.} V-JEPA2 (ViT-L)~\citep{assran2025vjepa2}. Non-action reasoning types use its native energy read-out zero-shot; action-conditioned types are read out by a probe trained on the frozen representation at two capacities, \emph{linear} and \emph{attentive}, with an open-vocabulary action space reached through a CLIP text bridge.

\textbf{Video generator.} I2VGen-XL~\citep{zhang2023i2vgenxl} ($2.84$B denoiser). Options are scored by conditional denoising likelihood over a fixed grid of diffusion steps with shared noise. Its forward-only rollouts cover $7$ of the $8$ reasoning types (counterfactual substitution has no forward likelihood read-out).

\textbf{Image generator.} OmniGen~\citep{xiao2024omnigen} ($3.8$B, matched to the video generator's denoiser scale), an image-editing generative model conditioned on the initial frame and the action instruction, scored with the identical likelihood recipe. This engine tests whether an image prior could stand in for a video prior as the WOVEN rollout generator.

\begin{table}[h]
\centering
\caption{\textbf{Off-the-shelf video priors on WOVEN} (\texttt{test\_id}, $4$-way MCQ accuracy in \%, chance $=25$). Columns are reasoning types, abbreviated as in \S\ref{sec:bench:axes}. For the encoder, the event cue of cued prediction is text and outside its input space, so outcome and cued prediction share one read-out. Overall for the generator engines spans the seven reasoning types their forward-only read-out covers.}
\label{tab:video-priors}
\begin{adjustbox}{max width=\linewidth}
\begin{tabular}{lccccccccc}
\toprule
Model & Fwd & Inv & Rmv & Sub & Otm & Cue & Ord & Adj & Overall \\
\midrule
V-JEPA2 linear & 51.3 & 51.9 & \textbf{66.5} & 20.0 & \textbf{29.2} & \textbf{29.2} & 37.5 & 16.9 & 39.3 \\
V-JEPA2 attentive & \textbf{70.7} & \textbf{75.4} & 57.7 & \textbf{35.9} & \textbf{29.2} & \textbf{29.2} & 37.5 & 16.9 & \textbf{47.7} \\
Video generator & 19.8 & 27.7 & 37.6 & --- & 13.2 & 13.9 & \textbf{53.1} & \textbf{56.2} & 36.4 \\
Image generator & 20.0 & 22.5 & 24.5 & --- & 13.5 & 12.5 & 22.0 & 21.5 & 19.5 \\
\bottomrule
\end{tabular}
\end{adjustbox}
\end{table}

\textbf{WOVEN is not trivially solvable from off-the-shelf priors.} The strongest read-out reaches $47.7\%$ overall against the $92.3\%$ human baseline, and each engine's profile is partial: the encoder is strong on action-conditioned types (forward $70.7$, inverse $75.4$) but weak on temporal ones ($37.5/16.9$), while the video generator shows the complementary profile (temporal $53.1/56.2$, forward $19.8$).

\textbf{The transition prior lives in the video prior.} At matched denoiser scale and under the identical likelihood recipe, the image-editing generator scores $19.5\%$ overall with no reasoning type above chance. An image prior cannot stand in for a video prior as the rollout generator, which supports building WOVEN on VGM rollouts.

\textbf{The knowledge is present but not directly accessible.} On identical frozen V-JEPA2 features, the attentive probe recovers ${+}19$ to ${+}34$pp over the linear probe on action-conditioned types, and a random-initialization control (same probe, same training data, randomly initialized encoder) trails the pretrained encoder by $27.6$pp on average, positive on all $12$ action cells: the gain is attributable to the pretrained representation, and the transition knowledge it holds is recoverable but not linearly accessible. This is the premise behind \S\ref{sec:enhance}: video-pretrained engines hold the $(s,a,s')$ prior (which is what makes them strong rollout generators for WOVEN) but do not expose it to direct querying, and post-training on $(s,a,s')$ supervision is what installs the access.

\subsection{Construction pipeline details}
\label{app:pipeline}

\begin{figure}[!htbp]
\centering
\includegraphics[width=\linewidth]{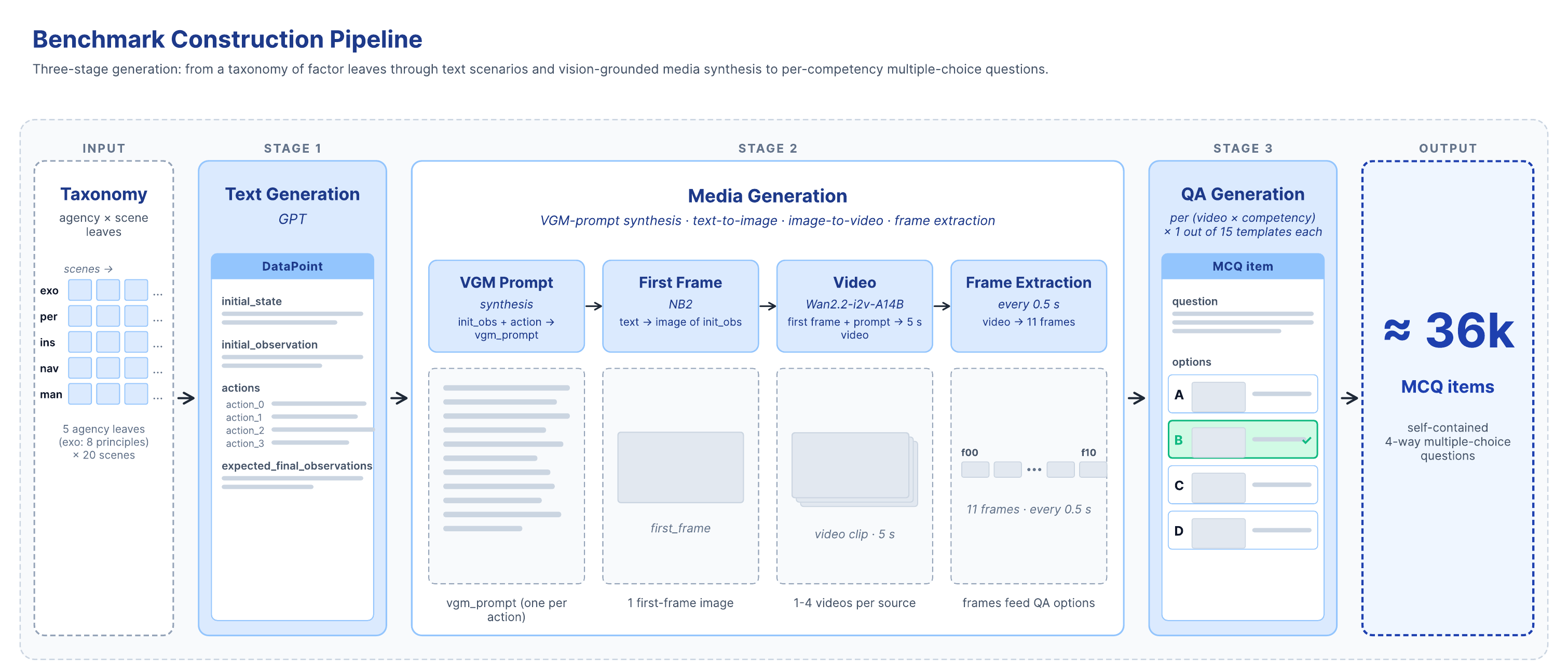}
\caption{Construction pipeline: stage-1 textual specification, stage-2 VGM rollout (Wan2.2-I2V-A14B), stage-3 typed-distractor MCQ assembly, with quality-control filters at each stage.}
\label{fig:pipeline}
\end{figure}

This appendix documents the technical realization of the three-stage construction pipeline summarized in \S\ref{sec:bench:overview}.

\paragraph{Stage 1 (text specification).}
For each (action type, scene) combination we generate a textual specification consisting of: (i) the initial environment state and the first-person observation of $s_t$; (ii) one or more action descriptions with deterministic identifiers; and (iii) the expected post-action state. Action vocabularies are action-type-specific: \emph{exogenous} draws one principle per item from the closed inventory of $8$ principles; \emph{perceptive} draws all $4$ fixed camera-translation/rotation actions; \emph{inspective} draws all $3$ fixed camera-orbit actions; \emph{navigative} draws $4$ in-scene targets; \emph{manipulative} draws $4$ object interactions consistent with the scenario-group definition (\S\ref{sec:benchmark}).

\paragraph{Stage 2 (media synthesis).}
The first frame is synthesized from the initial-observation text by an image generative model. The post-action $5$-second video at $24$\,fps is then synthesized by Wan2.2-I2V-A14B~\citep{wan2025} conditioned on $(s_t, a)$. We extract $11$ uniformly spaced frames per video, indexed $f_{00},\ldots,f_{10}$, with $f_{00}=s_t$ and $f_{10}=s_{t+1}$. Intermediate frames are reserved for temporal-coherence reasoning types.

\paragraph{Stage 3 (MCQ assembly).}
Four-way MCQs are assembled from the resulting frames per the reasoning-type-specific distractor strategies of \S\ref{sec:benchmark}. Each item additionally exposes structured metadata, taxonomy path, action-type-specific auxiliary fields (e.g., manipulated object, paired-variant indices, exogenous principle), per-frame paths, and the deterministic distractor type label where applicable, to support fine-grained downstream analysis.

\paragraph{Quality control.}
Media-level rule-based and model-based filters are applied to first frames and videos to reject outputs that fail per-action-type physical plausibility checks; layout and orientation perturbations carry an additional automated edit-faithfulness, scene-consistency, and artifact-absence check. Quality control operates only on media; it does not gate question generation.

\subsection{Why action type over motor command}
\label{app:why-agency}

Existing world-model action taxonomies organize $a$ by \emph{embodiment}: discrete vs.\ continuous in latent-policy world models~\citep{hafner2023dreamerv3}, joint/wheel/keyboard by application domain in driving and robotics WMs, or DOF-counted action vectors in egocentric WMs. These conventions are appropriate for embodiment-bound WM training, where the action space is fixed by the agent's effectors and the state space by the simulator. WOVEN instead evaluates a cognitive primitive across heterogeneous embodiments, bodiless cameras, simulated humans, simulated humanoids, externally-caused physical events. An embodiment-centric taxonomy would force inter-action-type alignment via low-level motor parameters that have no shared semantics across these settings (a wheel-radian and a translation-meter are not commensurable diagnostics for world modeling). Organizing by \emph{action type}, ``who initiates the change and how it changes the visual state'', operationalizing the cognitive-science notion of \emph{agency}~\citep{haggard2008humanvolition,passingham2010specialization,khalighinejad2018dissociating}, preserves cross-embodiment comparability while remaining cognitively grounded. We retain the cognitive-science terms \emph{agency} (for the action axis) and \emph{competency} (for the reasoning axis) only here, as the interface that motivates the taxonomy; the benchmark itself is named by action type, reasoning type, and scene type.

\subsection{Scene set}
\label{app:scenes}

The indoor set is \{\emph{living\_room, bedroom, bathroom, restaurant, kitchen, classroom, office, supermarket, gym, hospital}\}; the outdoor set is \{\emph{crossroad, sidewalk, playground, farm, garage, park, construction\_site, yard, school\_campus, beach}\}. The held-out OOD pair is (\emph{restaurant}, \emph{beach}).

\subsection{Cognitive Science support of Exogenous principles}
\label{app:exo-principles}

\begin{figure}[!htbp]
\centering
\includegraphics[width=\linewidth]{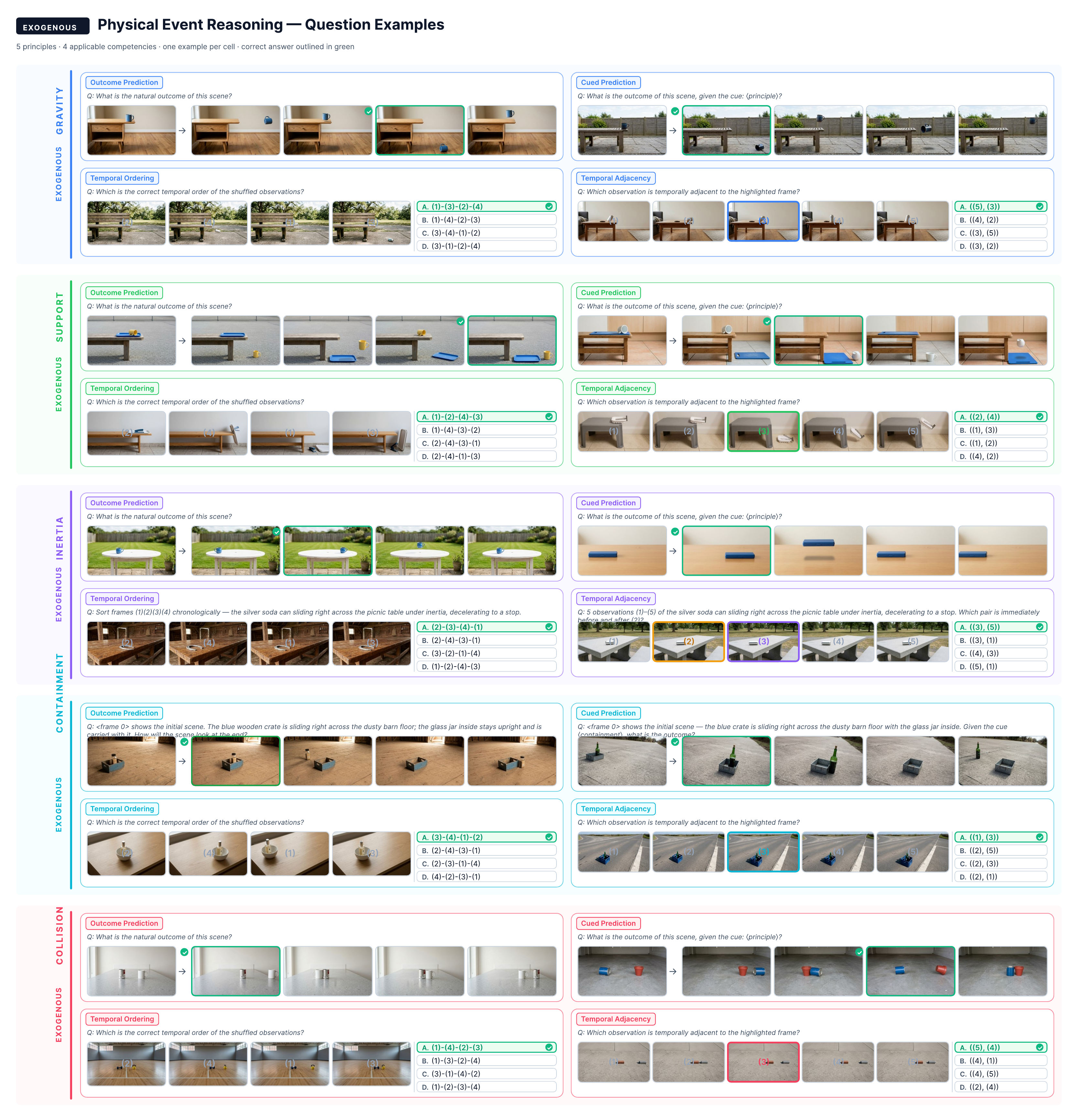}
\caption{Exogenous \emph{physical events} (gravity, support, inertia, containment, collision): per-principle illustration showing the action-conditioned violation-of-expectation distractor design.}
\label{fig:exo-events}
\end{figure}

\begin{figure}[!htbp]
\centering
\rotatebox{90}{\includegraphics[height=0.95\linewidth]{figs/WOVEN_Figure_3_Exogenous_Object_Properties.pdf}}
\caption{Exogenous \emph{object properties} (permanence, cohesion, solidity): per-principle illustration with prompt template, video, and four-way distractor structure.}
\label{fig:exo-objprops}
\end{figure}

The eight exogenous principles in WOVEN are organized in two cog-sci-grounded groups, each instantiating a separate strand of intuitive-physics evidence.

\paragraph{Group 1: object properties (\emph{permanence}, \emph{cohesion}, \emph{solidity}).}
These are the three foundational primitives of Spelke's core-knowledge inventory~\citep{spelke2007coreknowledge,spelke2022whatbabiesknow}. \emph{Permanence}: an object continues to exist when occluded; the diagnostic violation is dis-appearance during occlusion~\citep{baillargeon1987objectpermanence}. \emph{Cohesion}: an object's parts move together as a connected whole; the diagnostic violation is mid-event part-splitting. \emph{Solidity}: distinct objects do not interpenetrate; the diagnostic violation is one object passing through another. Each principle has its own Stage~$1$ prompt template in the construction pipeline, ensuring coverage of the principle-specific scene type and event type.

\paragraph{Group 2: physical events (\emph{gravity}, \emph{support}, \emph{inertia}, \emph{containment}, \emph{collision}).}
These are five principal physical-event classes routinely catalogued in computational intuitive-physics work~\citep{battaglia2013simulation} and in violation-of-expectation studies~\citep{baillargeon1987objectpermanence}. \emph{Gravity}: unsupported objects fall in expected directions and trajectories. \emph{Support}: an object resting on a surface remains supported absent additional cause. \emph{Inertia}: motion persists or stops in cause-consistent ways. \emph{Containment}: contained objects move with their containers and constrain motion. \emph{Collision}: contact events transfer motion in expected ways.

The two groups jointly span the principles of object-state-as-property (Group~$1$) and object-state-as-event (Group~$2$), giving the exogenous action type a clean cross-section. Each principle is tied to its own IEM violation-of-expectation distractor inventory in outcome prediction and cued prediction (\S\ref{sec:benchmark}).

\subsection{Manipulative cross-factoring}
\label{app:manip}

\begin{figure}[!htbp]
\centering
\includegraphics[width=\linewidth]{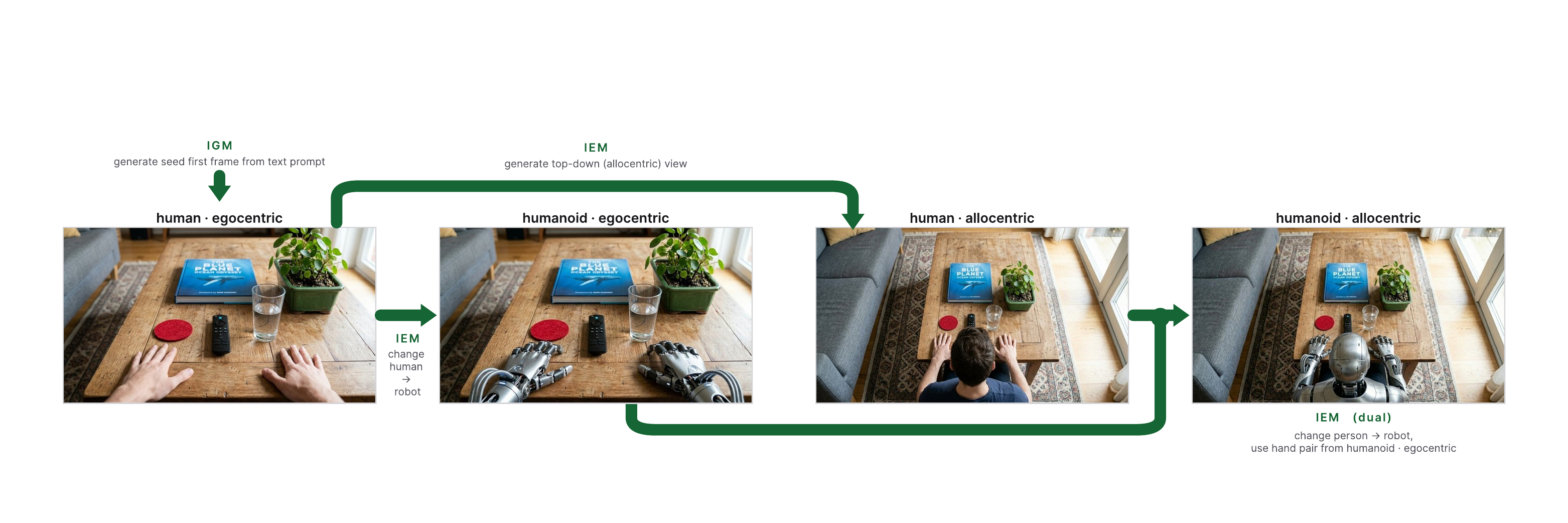}
\caption{Manipulative cross-factoring: $\{$human, humanoid$\}\times\{$egocentric, allocentric$\}$ derivation showing the four pair-mate cells used by the pair-mate independence analysis (Appendix~\ref{app:F-H}) and the cross-action-type transfer analysis (Appendix~\ref{app:E-F5}).}
\label{fig:manip-2x2}
\end{figure}

\begin{figure}[!htbp]
\centering
\includegraphics[width=\linewidth]{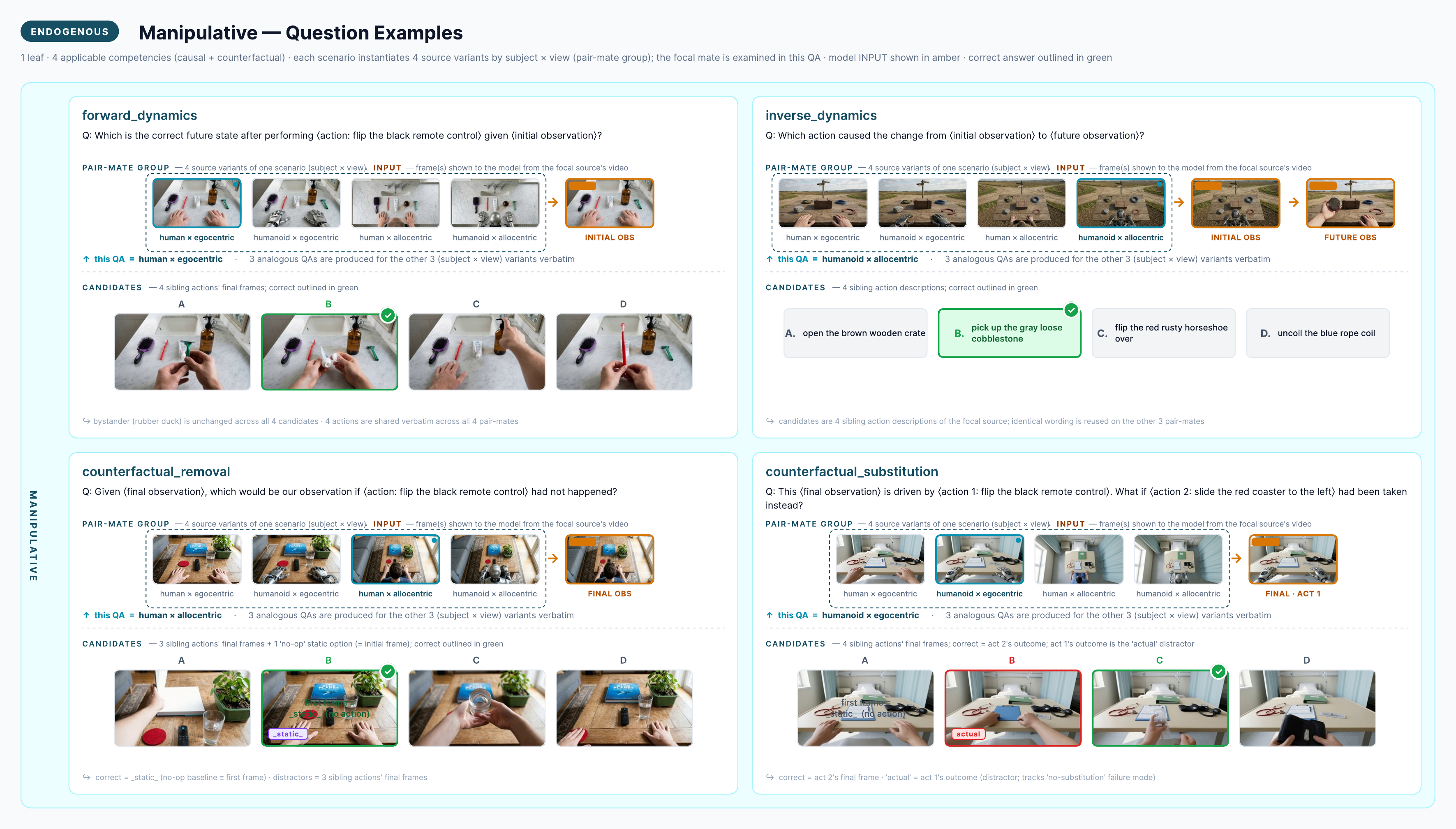}
\caption{Manipulative action type: representative QA items showing object-interaction A/B/C/D action structure with all four manipulative reasoning types (forward dynamics, inverse dynamics, counterfactual substitution, counterfactual removal).}
\label{fig:manip-examples}
\end{figure}

\paragraph{Why $\boldsymbol{2\times 2}$.}
The cross-factoring along $\textbf{subject}\in\{\text{human},\text{humanoid}\}$ ${\times}$ $\textbf{view}\in\{\text{egocentric},\text{allocentric}\}$ is motivated by two complementary cog-sci traditions. The subject axis distinguishes biologically natural actors from kinematically constrained robotic actors, isolating an MLLM's robustness to embodiment-shape variation in action-recognition pathways~\citep{rizzolatti2001neurophysiological,gibson1979ecological}. The view axis instantiates the egocentric/allocentric reference-frame distinction, a foundational dichotomy in spatial cognition~\citep{burgess2006spatialrefs,kessler2010embodiedperspective,epstein2017cognitivemap}: egocentric views ground the action in the actor's effector frame, while allocentric views ground it in a third-person scene frame. Cross-factoring isolates whether world-modeling competence is invariant to actor identity at fixed view, invariant to view at fixed actor, or jointly entangled across the two.

\paragraph{Generation pipeline.}
Each manipulative scenario specifies five non-rotationally-symmetric, color-distinct everyday objects within a fixed scene. Four of them are designated as targets of four \emph{independent} (``parallel-universe'') actions; the fifth, the bystander, is held visible and unmoved across all four actions, anchoring the geometric/appearance perturbations of \S\ref{sec:bench:perturb}. The four paired (subject, view) variants share the same scenario specification, the same five-object configuration, and the same four action verbs, differing only in actor identity and camera frame. Per scenario, this yields four sources that share underlying content but differ in embodiment and frame-of-reference; within-scenario consistency tests are the basis for the embodiment-vs-frame-of-reference dissociation analysis in Appendix~\ref{app:F-H}. The construction pipeline is multi-stage and inter-stage-dependent, subject-and-view selection $\rightarrow$ five-object selection (with diversity and rotational-asymmetry constraints) $\rightarrow$ action-verb selection (four mutually-independent verbs targeting four different objects) $\rightarrow$ bystander designation $\rightarrow$ per-(subject, view)-variant Stage~$2$ media synthesis with view-dependent prompt instructions.

\paragraph{Split-time invariant.}
The four-way scenario-group is the atomic unit of partitioning at split time, ensuring all four paired variants land in the same split. Validation is carved at the scenario-group level ($10\%$ of groups, seed $42$); training/test partition is at the source level within remaining groups. This prevents the most direct visual-leakage failure mode, a humanoid-allocentric variant in train and the matching human-egocentric variant in val/test sharing the underlying scene, object configuration, and action verbs.

\FloatBarrier
\subsection{Splits: stratification and pool sizes}
\label{app:splits}

\begin{figure}[!htbp]
\centering
\includegraphics[width=\linewidth]{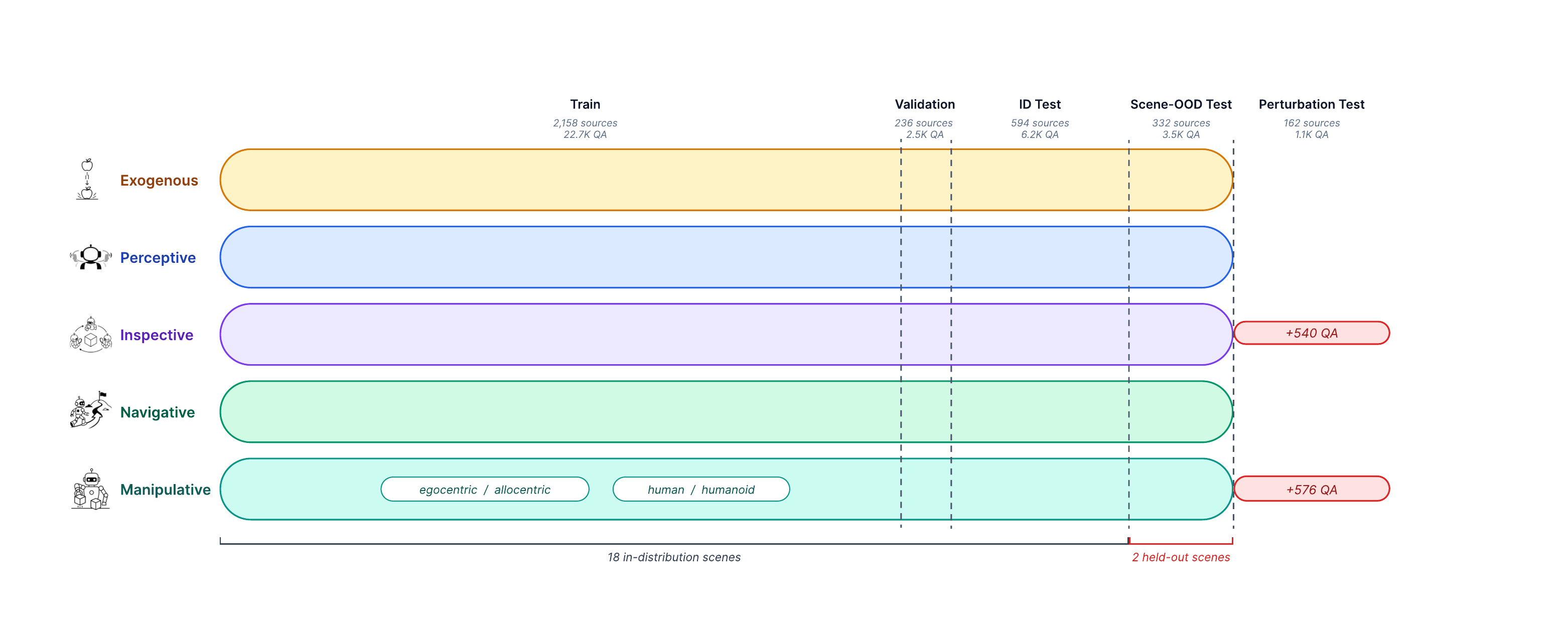}
\caption{WOVEN splits: \texttt{train\_pool} (used only for training), \texttt{test\_id} ($n{=}6{,}228$), \texttt{test\_perturbation\_id} ($n{=}1{,}116$), and held-out OOD scenes (\texttt{beach}, \texttt{restaurant}). The OOD set is reserved for future fine-tuning evaluations.}
\label{fig:splits}
\end{figure}

Partitioning is performed at the pre-MCQ-assembly stage. For each action type $\times$ scene stratum, the underlying generation pool is partitioned $80\%$/$20\%$ into a train candidate pool and a test candidate pool. Manipulative is partitioned at the four-way scenario-group level (Appendix~\ref{app:manip}). The validation set is then carved as $\sim 10\%$ of the train pool, stratified by action type and exogenous principle (random seed $42$). The scene-OOD test set is generated by an identical Stage~$1$--$3$ pipeline restricted to the two held-out scenes (\emph{restaurant}, \emph{beach}). The perturbation test is generated by the same pipeline from ID-test sources whose initial frames are edited with a geometric or appearance change before rollout (\S\ref{sec:bench:perturb}, Appendix~\ref{app:perturb-design}). The resulting per-split item counts are reported in Fig.~\ref{fig:splits} and Table~\ref{tab:T02}.

\subsection{Action type--reasoning type adjacency}
\label{app:adjacency}

Not every (action type, reasoning type) cell is instantiated in the benchmark: counterfactual reasoning applies where the action meaningfully alters scene state, and physical modeling is exogenous-specific. The full instantiation matrix is given in Table~\ref{tab:adjacency}.

\begin{table}[!htbp]
\centering
\small
\caption{Action type--reasoning type adjacency: which reasoning types instantiate items for each action type. \checkmark: instantiated; ---: not instantiated. Reasoning types are abbreviated as in \S\ref{sec:bench:axes}.}
\label{tab:adjacency}
\begin{adjustbox}{max width=\linewidth}
\begin{tabular}{l ccccccccc}
\toprule
Action type & Otm & Cue & Ord & Adj & Fwd & Inv & Rmv & Sub \\
\midrule
Exogenous     & \checkmark & \checkmark & \checkmark & \checkmark & --- & --- & --- & --- \\
Perceptive    & --- & --- & \checkmark & \checkmark & \checkmark & \checkmark & --- & --- \\
Inspective    & --- & --- & \checkmark & \checkmark & \checkmark & \checkmark & \checkmark & \checkmark \\
Navigative    & --- & --- & \checkmark & \checkmark & \checkmark & \checkmark & --- & --- \\
Manipulative  & --- & --- & --- & --- & \checkmark & \checkmark & \checkmark & \checkmark \\
\bottomrule
\end{tabular}
\end{adjustbox}
\end{table}

\paragraph{Why each cell is or is not instantiated.}
\emph{Physical Modeling} (Otm, Cue) is exogenous-specific: only externally-driven physical events admit a violation-of-expectation outcome judgement that does not depend on agent choice.
\emph{Causal Dynamics} (Fwd, Inv) is endogenous-specific: the source datapoint must contain multiple sibling actions for the cross-action MCQ to admit non-trivial distractors, which exogenous's single-action structure does not afford.
\emph{Temporal Coherence} (Ord, Adj) applies wherever visual change unfolds over time within a single rollout, which excludes manipulative whose inter-frame dynamics are dominated by hand-occlusion artifacts unrelated to scene-state evolution.
\emph{Counterfactual Reasoning} (\texttt{counterfactual\_substitution}, \texttt{counterfactual\_removal}) requires the action to meaningfully alter scene state, producing an $s_{t+1}$ informatively different from $s_t$, which holds for inspective (camera viewpoint changes the visible object configuration) and manipulative (object configuration itself changes), but not perceptive (camera-only, scene unchanged) or navigative (scene-positional change without state change).

\subsection{Distractor design: scope of typed labeling}
\label{app:distractors}

The scope of the typed-label property of distractors (\S\ref{sec:benchmark}) varies by reasoning type family.

\paragraph{Fully-typed distractors.}
For reasoning types that draw distractors from sibling actions in action types with a small fixed action vocabulary, \emph{perceptive} (4 fixed actions: translation and rotation) and \emph{inspective} (3 fixed actions: orbit-L, orbit-R, arc-up; plus a static-camera padding option to bring the option set to four), each distractor carries a deterministic, globally trackable label such as \texttt{rotate\_left}. Item-level error attribution to specific failure modes is therefore directly cross-comparable across items within an action type.

\paragraph{Family-level-typed distractors.}
For reasoning types whose action vocabulary varies per item, \emph{navigative} (the four targets are scene-specific) and \emph{manipulative} (the four object interactions are scenario-specific), distractors are still drawn from the closed sibling-action set, but their labels are not globally trackable across items. Failure-mode attribution is informative at the family level (e.g., ``selected a sibling action that shares object identity but differs in motion direction'') but does not align item-by-item.

\paragraph{Closed-class distractors.}
For temporal reasoning types (temporal ordering and temporal adjacency), distractors are drawn from a closed inventory of permutation classes (swap, off-by-one, endpoint-swap, adjacent-swap variants). Each class corresponds to a specific temporal-cognition error mode and is family-level trackable.

\paragraph{IEM-edit distractors.}
For physical-modeling reasoning types (outcome prediction and cued prediction), distractors are constructed by editing $s_{t+1}$ via Gemini 3.1 Flash Image~\citep{deepmind2026nanobanana2} from a closed inventory of physically-violating edit types (object-property and physical-event violations). Each edit type is globally trackable and aligned with a specific violation-of-expectation hypothesis. The correct outcome frame is also passed once through the same editor under a null instruction that asks for the same image with nothing changed, so every image option of a physical-modeling item has been through the editor exactly once (all $1{,}488$ exogenous sources). All four options of an item are released at the same size and in the same format.

\paragraph{Editing-artifact control.}
The null pass gives the correct option and the distractors the same editing history only if the editor re-renders the whole image rather than inpainting the edited region. We test this on all $864$ edited distractor frames of the test set. Each edited frame is resized to the size of its source frame and compared with the source in the half of the image farthest from the edit; the control is the source frame resized to the editor's output size and back, which is what an untouched region looks like after resampling alone. Table~\ref{tab:iem-rerender} reports the result. The far-region residual is about $20\times$ the resampling floor, present in every frame and every violation type, spatially structured (lag-1 autocorrelation $0.58$), and not reduced by any global shift, so it is neither a resampling nor an alignment effect. The null-edited correct options show the same footprint (far-region residual median $2.0$, whole-image median $3.4$). The editor therefore re-renders the full image regardless of the instruction, and the correct option and the distractors share the same editing process everywhere in the image.

We further test whether any residual difference between a null pass and a violation edit is detectable. The test is designed to see editing artifacts only and no task-relevant content. We randomly sample $166$ test sources and use their $278$ distractors that are edits of the outcome frame, since only these share unedited content with the correct option. A probe sees only $28{\times}28$ patches from the half of each image farthest from the edited region, where the content is identical across options, taken at the same positions for the correct option and its distractors. The probe is a six-layer convolutional network ($3{\times}3$ convolutions with $32$--$128$ channels, batch normalization, ReLU, global average pooling, and a linear output), trained with AdamW (learning rate $2{\times}10^{-3}$, weight decay $10^{-4}$, one-cycle schedule) for $1{,}500$ steps on batches of $1{,}024$ patches balanced across labels, with random horizontal flips. Evaluation uses five-fold cross-validation over sources with three training seeds, so no source appears in both training and test folds. We report patch-level balanced accuracy (chance $50\%$) and image-level AUC from each image's mean patch logit (chance $0.50$), against a permutation null that relabels one randomly chosen option per source as correct and reruns the same pipeline ($10$ seeds). Table~\ref{tab:iem-probe} shows no detectable signal above the null. As a positive control, we run the same probe, cross-validation, and permutation null on the same $166$ option sets with the correct option replaced by the original outcome frame that has not been through the editor (Table~\ref{tab:iem-probe-raw}). There the correct option is clearly separable ($55.9\%$ balanced accuracy, AUC $0.718$; $z{=}11.3$ and $14.2$), so the probe has the power to detect editing artifacts when they exist. Without the null edit the correct option is separable from the distractors; with it, it is not. The null pass therefore removes the editing signature that would otherwise mark the correct option, and the only systematic difference between the correct option and a distractor is the edited content itself.

\begin{table}[!htbp]
\centering
\small
\caption{The editor re-renders the whole image. Far-region residual between edited distractor frames and their source frames (test set, $864$ edited frames; gray levels on a $0$--$255$ scale).}
\label{tab:iem-rerender}
\begin{tabular}{@{}p{8.6cm}r@{}}
\toprule
Measurement & Value \\
\midrule
Mean absolute residual, far half of the image, edited vs.\ source & $3.8$ (median $1.9$, p90 $8.0$) \\
Same measurement for the resize round-trip of the source frame (control) & $0.17$ \\
Ratio edited\,/\,control, per frame & median $20.6\times$; $0$ of $864$ below $1.5\times$ \\
Lag-1 spatial autocorrelation of the far-region residual & $0.58$ \\
Reduction of the far-region residual by the best global shift ($\pm3$ px, $0.5$ px steps, $80$ frames) & $0\%$ \\
Ratio edited\,/\,control by violation type ($23$ types) & $10\times$--$36\times$, every type \\
\bottomrule
\end{tabular}
\end{table}

\begin{table}[!htbp]
\centering
\small
\caption{Editing-artifact probe on the released data (test set, $166$ sources, $166$ correct options, $278$ distractors).}
\label{tab:iem-probe}
\begin{tabular}{@{}lccc@{}}
\toprule
Metric & Observed & Permutation null ($10$ seeds) & $z$ \\
\midrule
Patch-level balanced accuracy (\%) & $51.1$ & $50.9\pm0.4$ & $0.5$ \\
Image-level AUC & $0.532$ & $0.520\pm0.008$ & $1.4$ \\
\bottomrule
\end{tabular}
\end{table}

\begin{table}[!htbp]
\centering
\small
\caption{Positive control: the same probe on a constructed subset without the null edit, with the correct option replaced by the original, unedited outcome frame (test set, $166$ sources, $166$ correct options, $278$ distractors).}
\label{tab:iem-probe-raw}
\begin{tabular}{@{}lccc@{}}
\toprule
Metric & Observed & Permutation null ($10$ seeds) & $z$ \\
\midrule
Patch-level balanced accuracy (\%) & $55.9$ & $50.5\pm0.5$ & $11.3$ \\
Image-level AUC & $0.718$ & $0.508\pm0.015$ & $14.2$ \\
\bottomrule
\end{tabular}
\end{table}

\paragraph{Per-cell traceability coverage.}
Table~\ref{tab:distractor-coverage} summarizes which (action type, reasoning type) cells receive fully-typed (item-level globally trackable) versus family-level (closed-class but item-dependent) distractors.

\begin{table}[!htbp]
\centering
\small
\caption{Distractor-type coverage by (action type, reasoning type) cell. \textbf{F} = fully-typed sibling-action distractors, globally trackable item-by-item. \textbf{C} = closed-class sibling-action distractors, family-level trackable. \textbf{I} = IEM-edit distractors, globally trackable. \textbf{S} = sequence-permutation distractors, family-level trackable; ---: cell not instantiated. Reasoning types are abbreviated as in \S\ref{sec:bench:axes}.}
\label{tab:distractor-coverage}
\begin{adjustbox}{max width=\linewidth}
\begin{tabular}{l c c c c c c c c}
\toprule
& Otm & Cue & Ord & Adj & Fwd & Inv & Rmv & Sub \\
\midrule
Exogenous     & I & I & S & S & --- & --- & --- & --- \\
Perceptive    & --- & --- & S & S & F & F & --- & --- \\
Inspective    & --- & --- & S & S & F & F & F & F \\
Navigative    & --- & --- & S & S & C & C & --- & --- \\
Manipulative  & --- & --- & --- & --- & C & C & C & C \\
\bottomrule
\end{tabular}
\end{adjustbox}
\end{table}

\noindent
A cell labeled \textbf{F} or \textbf{I} contributes per-item failure-mode counts that license the per-cell binomial, $\chi^2$, and two-proportion-$z$ tests of \S\ref{sec:meth:stat}. A cell labeled \textbf{C} or \textbf{S} contributes to family-level distributional tests but not to per-item attribution. The fully-typed cells cover all of perceptive's and inspective's causal-dynamics + counterfactual structure, i.e., the action types on which the perturbation probe (\S\ref{sec:bench:perturb}) also concentrates, ensuring that the most diagnostically loaded subset of the benchmark is also the subset with the strongest statistical instrumentation.

\subsection{Perturbation probe: construction and scope}
\label{app:perturb-design}

\paragraph{Construction.}
The perturbation test follows the same Stage~$1$--$3$ pipeline as the other splits. For each source, the designated object in the initial frame is edited with a geometric (layout or orientation) or an appearance change, the edited frame passes the edit-faithfulness, scene-consistency, and artifact-absence checks of Appendix~\ref{app:pipeline}, and the VGM is rolled out again from the edited frame under the same action. Questions are assembled from the new rollout as for the in-distribution test, so each perturbed item shares its action and query with an unperturbed item and differs only in the initial state.

\paragraph{Scope: inspective + manipulative only.}
Geometric and appearance perturbations are well-defined when the action operates on a discrete object whose perturbation can be cleanly localized. \emph{Inspective} actions orbit/arc around a designated main object, and the perturbation targets that main object. \emph{Manipulative} actions act on one of four target objects per scenario; the perturbation targets the bystander object that is held visible and unmoved across all four actions (\S\ref{sec:benchmark}, Appendix~\ref{app:manip}), so the perturbation does not disturb the action's primary target. For \emph{perceptive} and \emph{navigative}, no such designated object exists and the action changes the whole frame, so a localized edit of the initial frame has no stable target whose change the prediction should track. \emph{Exogenous} similarly lacks a clean target: the relevant ``state'' is the multi-object configuration jointly, and the perturbation would re-define the principle being probed. Inspective and manipulative thus offer the cleanest diagnostic signal for the dorsal/ventral state-axis dualization the probe targets.

\section{Evaluation cohort and human studies}

\subsection{Cohort: model inventory}
\label{app:cohort}

We evaluate $38$ frontier MLLMs spanning closed-source frontier, open-source dense ladders, and MoE families (Table~\ref{tab:T01}). Where vendors do not publish parameter counts (e.g., GPT-5.x, Qwen-VL-Max, Intern-S1, MiniMax, Seed, GLM-4.6V), the column is marked N/A; where they publish active vs.\ total separately (e.g., 30B-A3B = 30B total / 3B active), we list both.

\begin{table}[!htbp]
\centering
\caption{Model inventory.}
\label{tab:T01}
\small
\setlength{\tabcolsep}{4pt}
\begin{tabular}{@{}lll@{}}
\toprule
Model & Family & Params \\
\midrule
GPT-5.4 & GPT & N/A \\
GPT-5.2 & GPT & N/A \\
Qwen-VL-Max & Qwen-max & N/A \\
Qwen2.5-VL-3B & Qwen2.5 & 3B \\
Qwen2.5-VL-7B & Qwen2.5 & 7B \\
Qwen2.5-VL-32B & Qwen2.5 & 32B \\
Qwen2.5-VL-72B & Qwen2.5 & 72B \\
Qwen3-VL-8B-Instruct & Qwen3 & 8B \\
Qwen3-VL-30B-A3B & Qwen3 & 30B-A3B (MoE) \\
Qwen3-VL-32B-Instruct & Qwen3 & 32B \\
Qwen3-VL-235B-A22B & Qwen3 & 235B-A22B (MoE) \\
Intern-S1 & Intern & N/A \\
Intern-S1-Mini & Intern & N/A \\
Intern-S1-Pro & Intern & N/A \\
InternVL3.5-241B-A28B & Intern & 241B-A28B (MoE) \\
InternVL3.5-Latest & Intern & N/A \\
Gemma-3-4B & Gemma3 & 4B \\
Gemma-3-12B & Gemma3 & 12B \\
Gemma-3-27B & Gemma3 & 27B \\
Gemma-4-26B-A4B & Gemma4 & 26B-A4B (MoE) \\
Gemma-4-31B & Gemma4 & 31B \\
Llama-4-Scout & Llama4 & Scout (109B-MoE) \\
Llama-4-Maverick & Llama4 & Maverick (400B-MoE) \\
Mistral-Small-3.2 & Mistral & 24B \\
Mistral-Large-2512 & Mistral & N/A \\
Pixtral-Large-2411 & Mistral & 123B \\
MiniMax-01 & MiniMax & N/A \\
Qwen3.5-9B & Qwen3.5 & 9B \\
Qwen3.5-27B & Qwen3.5 & 27B \\
Qwen3.5-35B-A3B & Qwen3.5 & 35B-A3B (MoE) \\
Qwen3.5-122B-A10B & Qwen3.5 & 122B-A10B (MoE) \\
Qwen3.5-397B-A17B & Qwen3.5 & 397B-A17B (MoE) \\
GLM-4.6V & GLM & N/A \\
Seed-1.6-Flash & Seed & N/A \\
Seed-2.0-Mini & Seed & N/A \\
Nemotron-Nano-12B-V2-VL & Nemotron & 12B \\
ERNIE-4.5-VL-28B-A3B & ERNIE & 28B-A3B (MoE) \\
ERNIE-4.5-VL-424B-A47B & ERNIE & 424B-A47B (MoE) \\
\bottomrule
\end{tabular}
\end{table}

\subsection{Human baseline}
\label{app:human-baseline}

\begin{figure}[!htbp]
\centering
\includegraphics[width=0.62\linewidth]{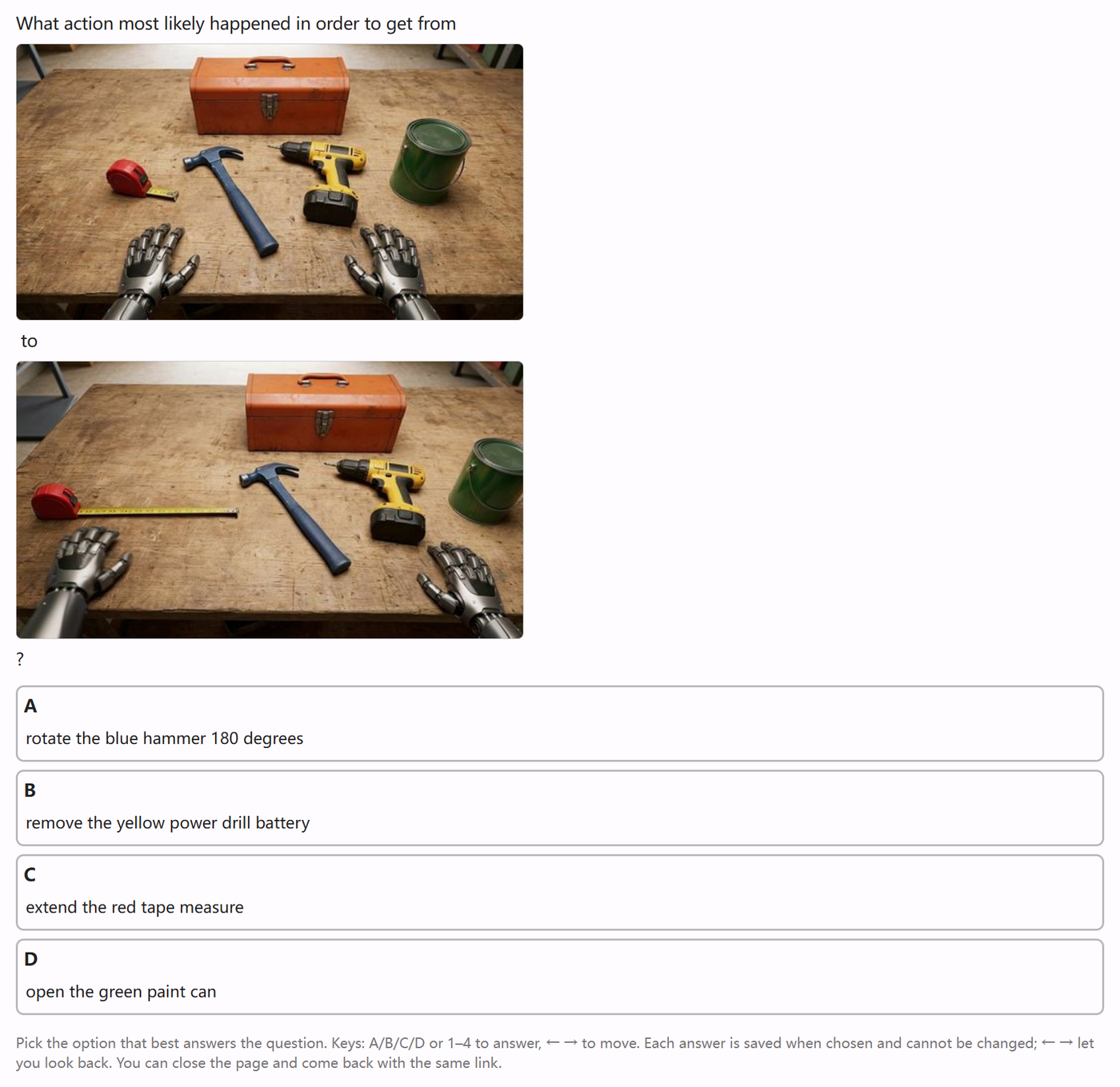}
\caption{\textbf{Human study interface.} Annotators see the same frames, question, and four options as the models. Each answer is saved when chosen and cannot be changed.}
\label{fig:human-interface}
\end{figure}

Three annotators, none of whom are authors, answered $990$ in-distribution items, sampled uniformly over the combinations of action type and reasoning type, and $1{,}000$ perturbation items ($500$ appearance, $500$ geometric) through the interface in Fig.~\ref{fig:human-interface}, which presents the same frames, question, and four options as the model input. Each answer is saved when chosen and cannot be changed. We compute each annotator's accuracy per category and report the mean over the three annotators: $92.3\%$ on the in-distribution items (Table~\ref{tab:T37}) and $98.0\%$ on the perturbation items (Table~\ref{tab:T38}). Annotators were paid in accordance with local wage standards.

\begin{table}[!htbp]
\centering
\caption{Human accuracy (\%) on the ID test, averaged over three annotators.}
\label{tab:T37}
\scriptsize
\setlength{\tabcolsep}{3pt}
\begin{tabular}{@{}lccccc|cccccccc|c@{}}
\toprule
& \multicolumn{5}{c}{Per-action type} & \multicolumn{8}{c}{Per-reasoning type} & \\
\cmidrule(lr){2-6} \cmidrule(lr){7-14}
& Exo & Perc & Insp & Navi & Mani & Cue & Otm & Adj & Ord & Fwd & Inv & Rmv & Sub & Overall \\
\midrule
human & 92.8 & 92.8 & 92.6 & 90.6 & 92.8 & 91.1 & 93.3 & 91.7 & 92.8 & 92.8 & 92.8 & 91.1 & 92.2 & 92.3 \\
\bottomrule
\end{tabular}
\end{table}

\begin{table}[!htbp]
\centering
\caption{Human accuracy (\%) on the perturbation probe, averaged over three annotators.}
\label{tab:T38}
\small
\begin{tabular}{@{}lccc@{}}
\toprule
& appearance & geometric & overall \\
\midrule
human & 98.7 & 97.3 & 98.0 \\
\bottomrule
\end{tabular}
\end{table}

\subsection{Human audit of the test splits}
\label{app:audit}

The labelled answers of WOVEN's test questions are valid for at least $98\%$ of the questions, $92.5\%$ of the rollouts carry out the scripted action fully and $98.5\%$ at least partly, and $98.7\%$ of the perturbation edits follow their instruction without other changes. WOVEN's questions and answers are produced automatically from generated rollouts, so three things could be wrong without being visible in model accuracy: a labelled answer may not be the correct one, a rollout may not carry out the scripted action or may show an implausible or discontinuous process, and an edited option in the perturbation split may not follow its edit instruction or may change more than instructed. Two annotators, who are not authors, independently audited a stratified sample of the test splits in three parts through a web page with one item per screen, both seeing the same items in the same fixed, shuffled order; every choice is stored server-side in a hash-chained log per annotator. The sample holds $250$ questions from $228$ sources of the in-distribution and held-out-scene test splits: $200$ main questions stratified by action type (exogenous $43$, inspective $43$, manipulative $43$, perceptive $36$, navigative $35$; exogenous questions are physical-modeling questions, and for the other action types the questions cycle over every reasoning type the action type has) and $50$ perturbation questions (inspective $25$, manipulative $25$; geometric and appearance edits split evenly), each with three edited options.

In Part 1 ($250$ questions), the annotator answers each question as a model does, without seeing the labelled answer, states a confidence (guessing, fairly sure, sure), and says whether another option could also be correct; answers can be changed until the page is confirmed, after which it is locked. In Part 2 ($100$ rollout videos, $20$ per action type), the annotator sees the video the question was generated from, with its start frame, frames, and scripted action (for exogenous items, the described event), and judges whether the scripted action is carried out (yes, partly, no), whether the final state is a plausible result of it, whether the video is continuous (nothing appears, vanishes, jumps, or changes shape without cause), and, for exogenous items, whether the process follows the stated physical principle. In Part 3 ($50$ perturbation questions, $150$ edits), the annotator sees the unedited image beside the three edited options with their edit instructions and judges whether all three edits were carried out as instructed with nothing else changed, and if not, which edits have a problem. Parts 2 and 3 follow Part 1 because the videos and the unedited images reveal the answers.

\begin{table}[!htbp]
\centering
\footnotesize
\caption{Human audit, Part 1: blind answering (accuracy \%, $250$ questions).}
\label{tab:audit-p1}
\begin{tabular}{@{}lrrr@{}}
\toprule
 & Annotator 1 & Annotator 2 & Pooled $[95\%$ CI$]$ \\
\midrule
All questions & 94.0 & 93.6 & 93.8 $[91.3, 95.6]$ \\
Main questions (200) & 94.0 & 93.5 & 93.8 \\
Perturbation questions (50) & 94.0 & 94.0 & 94.0 \\
Exogenous (43) & 95.3 & 93.0 & 94.2 \\
Inspective (43) & 95.3 & 93.0 & 94.2 \\
Manipulative (43) & 95.3 & 93.0 & 94.2 \\
Perceptive (36) & 91.7 & 91.7 & 91.7 \\
Navigative (35) & 91.4 & 97.1 & 94.3 \\
Answered with ``sure'' & 97.5 (204) & 96.2 (212) & 96.9 \\
\bottomrule
\end{tabular}
\end{table}

\begin{table}[!htbp]
\centering
\footnotesize
\caption{Human audit, Part 2: rollout videos (\% of $100$ videos).}
\label{tab:audit-p2}
\scriptsize
\setlength{\tabcolsep}{3pt}
\begin{tabular}{@{}lrrrr@{}}
\toprule
Question & Annotator 1 & Annotator 2 & Pooled $[95\%$ CI$]$ & Agreement \\
\midrule
Scripted action carried out (yes) & 94 & 91 & 92.5 $[88.0, 95.4]$ & 92\% \\
Scripted action carried out (yes or partly) & 99 & 98 & 98.5 $[95.7, 99.5]$ & --- \\
Final state plausible & 99 & 98 & 98.5 $[95.7, 99.5]$ & 97\% \\
Video continuous & 98 & 97 & 97.5 $[94.3, 98.9]$ & 97\% \\
Follows the stated physical principle (exogenous, 20) & 100 & 100 & 100 $[91.2, 100]$ & 100\% \\
\bottomrule
\end{tabular}
\end{table}

\begin{table}[!htbp]
\centering
\footnotesize
\caption{Human audit, Part 3: edited images.}
\label{tab:audit-p3}
\begin{tabular}{@{}lrrr@{}}
\toprule
 & Annotator 1 & Annotator 2 & Pooled $[95\%$ CI$]$ \\
\midrule
Edits carried out as instructed with nothing else changed & 149/150 & 147/150 & 98.7\% $[96.6, 99.5]$ \\
Questions with all three edits correct & 49/50 & 47/50 & 96.0\% \\
\bottomrule
\end{tabular}
\end{table}

In Part 1 the annotators answered blind with $93.8\%$ accuracy (Table~\ref{tab:audit-p1}); they chose the same option on $92.0\%$ of the questions (Cohen's $\kappa = 0.89$) and both answered correctly on $225$ questions. On $5$ questions ($2.0\%$ $[0.9, 4.6]$) both chose the same option different from the labelled answer, which bounds the rate of invalid labelled answers from above. In Part 2, $92.5\%$ of the rollouts carry out the scripted action fully and $98.5\%$ at least partly, $98.5\%$ end in a plausible state, $97.5\%$ are continuous, and all exogenous events follow the stated physical principle (Table~\ref{tab:audit-p2}); both annotators judged the scripted action as not carried out on $1$ of the $100$ videos. In Part 3, $98.7\%$ of the edits follow their instruction without other changes and the annotators agreed on whether all three edits of a question were correct for $96\%$ of the questions (Table~\ref{tab:audit-p3}), so the perturbation split tests what it is designed to test.

\section{Statistical methodology}
\label{app:stat-methods}

This appendix has two parts. \textbf{\ref{app:stat-methods:O1}} enumerates the non-default methodological choices made across the paper, each with a one-line rationale. \textbf{\ref{app:stat-methods:per-finding}} gives per-finding statistical details (test, $n$, statistic, raw and Bonferroni-adjusted $p$, effect size) for every quantitative claim that enters the main text.

\subsection{Methodological choices}
\label{app:stat-methods:O1}

\textbf{Above-chance threshold.} A model is treated as ``above chance'' on a cell when accuracy $> 25\%$ \emph{and} one-sided binomial $p{<}10^{-4}$. Twenty-five percent is the four-way-MCQ chance floor; the $10^{-4}$ threshold gives Bonferroni-style headroom for the cohort-level screen of $38$ models $\times$ up to $8$ reasoning types ($\approx\!300$ simultaneous binomials), preserving family-wise $\alpha < 0.05$ before any downstream test is applied.

\textbf{OLS on $\log_{10}N$.} Within-family scaling fits use $y = a + b \log_{10} N$ with $b$ in pts/decade. The log axis is the standard scaling-law parameterisation; pts/decade is directly comparable across axes (geometric vs.\ appearance, ID vs.\ perturbation).

\textbf{Cram\'er's $V$.} Effect-size companion to per-model $\chi^2$ goodness-of-fit on distractor-class concentrations. Bounded in $[0,1]$ and normalized by sample size, so it is comparable across models with different total error counts; conventional thresholds (small $0.10$, medium $0.30$, large $0.50$) apply at $\mathrm{df}{=}2$.

\textbf{Centred cosine similarity.} Cosine similarity on the $8$-dimensional per-principle accuracy vector after subtracting each model's own mean. The centring step isolates representational \emph{shape} (which principles a model is relatively strong on) from \emph{level} (overall accuracy), so two strong models with the same competence profile look similar even if one is uniformly higher.

\textbf{Two-proportion $z$ for within-model gap.} Used for the within-model appearance-vs-geometric perturbation gap (e.g., Qwen2.5-VL-7B at $50.5\%$ vs.\ $23.5\%$). Standard form, no continuity correction at the $n$ values reported here ($n_{\text{geo}}{=}n_{\text{app}}{=}558$).

\textbf{Paired Wilcoxon signed-rank.} Per-model paired test for the counterfactual-removal cost asymmetry (\S\ref{app:stat-methods:per-finding}; Appendix~\ref{app:F-G}), pairing $\mathrm{cost}(\mathrm{Fwd}{\to}\mathrm{Rmv})$ vs.\ $\mathrm{cost}(\mathrm{Fwd}{\to}\mathrm{Sub})$ within each model. Two-sided; ties handled by mid-rank (Pratt method); restricted to the $24$ above-chance models so that ``cost'' is measured against a non-degenerate forward-dynamics baseline.

\textbf{Bootstrap, $1000\times$ on residuals.} OLS scaling-fit confidence intervals come from parametric bootstrap on regression residuals, $1000$ resamples, slope CIs as $2.5$/$97.5$ quantiles. The slope estimates and CIs are stable to two decimal places at this resample count given the four-rung ladders in question.

\textbf{Bonferroni for family-wise control.} The seven findings of Appendix~\ref{sec:assess:findings} tested in \S\ref{app:stat-methods:per-finding} form the relevant family. We apply Bonferroni at $k{=}7$ on top of each finding's raw $p$ value. Bonferroni rather than Benjamini--Hochberg is chosen for conservativeness: at our cohort size, controlling family-wise error is the stronger guarantee a benchmark paper can offer reviewers. All seven findings remain significant ($p_{\text{adj}} < 0.001$) under this correction.

\textbf{Scene-CV per model.} Per-scene accuracy variability is summarized by $\mathrm{CV}{=}\sigma/\mu$ within each model across the $15$ ID scenes used for the noise-floor estimate. The dimensionless form makes the noise floor (median CV $5.6\%$) interpretable across models at different overall accuracy levels.

\textbf{Pair-mate Pearson aggregation.} Pair-mate independence (Appendix~\ref{app:F-H}) reports Pearson $r$ between the four (subject $\times$ view) cells of each manipulative scenario, averaged over the $\sim\!152$ available (model $\times$ reasoning type) combinations. Aggregation is by Fisher-$z$ transform, mean over combinations, inverse-transform back; reported $r$ is the inverse-transformed mean.

\subsection{Per-finding statistical details}
\label{app:stat-methods:per-finding}

Every finding of Appendix~\ref{sec:assess:findings} is summarized in one row. Raw and Bonferroni-adjusted $p$ are reported with $k{=}7$ across this family of findings. Effect sizes use Cram\'er's $V$ for $\chi^2$, Cohen's $d$ for $t$-tests, slope-$R^2$ for OLS, $r$ for Spearman/Pearson, and the matched within-model gap for paired Wilcoxon.

\begin{table}[!htbp]
\centering
\caption{Per-finding statistical details for the findings of Appendix~\ref{sec:assess:findings}. Family-wise correction applies Bonferroni at $k{=}7$. All seven findings remain significant under correction.}
\label{tab:stat-per-finding}
\setlength{\tabcolsep}{3pt}
\resizebox{\linewidth}{!}{
\begin{tabular}{@{}llrlllll@{}}
\toprule
ID & Claim & $n$ & Test & Statistic (df) & Raw $p$ & Bonf.\ adj.\ $p$ & Effect size \\
\midrule
\multirow{4}{*}{\ref{app:F-A}} & Inspective below manipulative, frontier & 11 models  & sign test                  & $11/11$ positive    & $9.8{\times}10^{-4}$ & $6.9{\times}10^{-3}$ & gap range $16.2$--$35.6$pp \\
                     & Mirror confusion, frontier         & 7 families & 1-sided binomial vs.\ $1/3$ & $\hat{p}_{\min}{=}0.55$ & $\ll\!10^{-10}$ each & $\ll\!10^{-9}$ each & $\hat{p}{-}1/3 \in [0.22, 0.56]$ \\
                     & Orientation concentration, GPT-5.4 & $231$ err  & $\chi^2$ vs.\ uniform $1/3$ & $65.22$ ($\mathrm{df}{=}2$) & $6.9{\times}10^{-15}$ & $4.8{\times}10^{-14}$ & $V{=}0.376$ (med--large) \\
                     & Orientation concentration, Qwen2.5-VL-7B & $477$ err & $\chi^2$ vs.\ uniform $1/3$ & $31.74$ ($\mathrm{df}{=}2$) & $1.3{\times}10^{-7}$ & $9.1{\times}10^{-7}$ & $V{=}0.182$ (small) \\
\midrule
\multirow{2}{*}{\ref{app:F-B}} & Appearance vs.\ geometric scaling slope         & 4 dense ladders, 11 models & OLS log-linear, per-ladder intercepts, slope $b$ & app $b{=}39.8$, geo $b{=}11.1$ pts/dec ($t{=}4.43$, $4.83$; $\mathrm{df}{=}6$) & $4.4{\times}10^{-3}$, $2.9{\times}10^{-3}$ & $3.1{\times}10^{-2}$, $2.0{\times}10^{-2}$ & slope difference $28.7$ pts/dec, $95\%$ CI $[8.1, 49.4]$ \\
                     & Within-model app--geo gap, Qwen2.5-VL-7B        & $558{+}558$ & two-proportion $z$ & $z{=}9.34$ & $<10^{-20}$ & $<10^{-19}$ & $\Delta{=}27$pp \\
\midrule
\multirow{2}{*}{\ref{app:F-C}} & ID vs.\ Appearance correlation & $38$ & Spearman & $\rho{=}{+}0.90$ ($\mathrm{df}{=}36$) & $1.5{\times}10^{-14}$ & $1.1{\times}10^{-13}$ & $R^2{=}0.86$ \\
                     & ID vs.\ Geometric correlation  & $38$ & Spearman & $\rho{=}{+}0.76$ ($\mathrm{df}{=}36$) & $3.1{\times}10^{-8}$ & $2.2{\times}10^{-7}$ & $R^2{=}0.52$ \\
\midrule
\ref{app:F-D} & Ord$-$Adj within-model gap & 11 frontier & paired Wilcoxon, two-sided & $W{=}66$ ($n_{\text{nonzero}}{=}11$) & $9.8{\times}10^{-4}$ & $6.9{\times}10^{-3}$ & median gap ${+}20.7$pp \\
\midrule
\ref{app:F-E} & Dense premium at matched total params & 3 within-family pairs & one-sample $t$ vs.\ $0$ & $t{=}14.3$ ($\mathrm{df}{=}2$) & $4.8{\times}10^{-3}$ & $3.4{\times}10^{-2}$ & $d{=}8.3$, mean ${+}12.4$pp \\
\midrule
\multirow{2}{*}{\ref{app:F-F}} & Cross-ladder emergence ordering replication & 3 ladders & Spearman rank concordance & Kendall's $W{=}0.93$ & $<10^{-3}$ & $<7{\times}10^{-3}$ & $W{>}0.9$ (strong) \\
                     & Gravity \texttt{rise} suppressed below $1/3$  & 12 models & one-sample $t$ vs.\ $1/3$ & $t{=}{-}3.85$ ($\mathrm{df}{=}11$) & $2.7{\times}10^{-3}$ & $1.9{\times}10^{-2}$ & $d{=}{-}1.11$, mean $0.205$ \\
\midrule
\ref{app:F-G} & Rmv vs.\ Sub cost asymmetry & 24 above-chance & paired Wilcoxon, two-sided & $W{=}293$ ($n{=}24$, Pratt zero handling) & $1.4{\times}10^{-6}$ & $9.8{\times}10^{-6}$ & median gap ${+}11.0$pp \\
\midrule
\ref{app:F-H} & Pair-mate independence            & $\sim\!152$ groups & Pearson $r$, Fisher-$z$ aggregated & $\bar{r} \in [0.02, 0.16]$ & per-pair $>0.05$ (n.s.) & --- & all $|\bar r| < 0.20$ \\
\bottomrule
\end{tabular}}
\end{table}

\textbf{Notes on aggregation.} For the mirror-confusion test of Appendix~\ref{app:F-A}, we report the worst-of-seven raw $p$; the seven families are individually corrected (i.e.\ effective $k{=}49$ if treated as separate hypotheses), and all seven remain $p{<}10^{-9}$ adjusted. For Appendices~\ref{app:F-D} and~\ref{app:F-G}, paired Wilcoxon is computed on the per-model paired difference $\Delta_i$; ties handled by Pratt's method; only above-chance models contribute to F-G to keep ``cost'' well-defined. For F-H, the claim is the absence of a strong correlation, so we report the magnitude bound rather than a $p$ value: all six pair-mate correlations satisfy $|\bar r| < 0.20$, well below conventional moderate-correlation thresholds. For F-B's slope CIs, parametric bootstrap on residuals ($1000$ resamples) gives appearance slope $95\%$ CI $[24.0, 38.4]$ and geometric slope $95\%$ CI $[6.5, 11.3]$; the two CIs do not overlap, reproducing the $3.5{\times}$ ratio finding under resampling.

\section{Per-cohort evaluation tables}
\label{app:eval-tables}

\begin{table}[!htbp]
\centering
\caption{Benchmark scale.}
\label{tab:T02}
\small
\begin{tabular}{@{}lrr@{}}
\toprule
Set & QA count & Sources \\
\midrule
ID reasoning type items & 6,228 & 594 \\
ID perturbation items & 1,116 & 320 \\
Scene-OOD reasoning type & 3,496 & 332 \\
\bottomrule
\end{tabular}
\end{table}

\begin{table}[!htbp]
\centering
\caption{Action type $\times$ reasoning type adjacency.}
\label{tab:T03}
\small
\begin{tabular}{@{}lcl@{}}
\toprule
Action type & \#Comp & Reasoning type list \\
\midrule
exogenous & 4 & Otm, Cue, Ord, Adj \\
perceptive & 4 & Fwd, Inv, Ord, Adj \\
inspective & 6 & Fwd, Inv, Rmv, Sub, Ord, Adj \\
navigative & 4 & Fwd, Inv, Ord, Adj \\
manipulative & 4 & Fwd, Inv, Rmv, Sub \\
\bottomrule
\end{tabular}
\end{table}

\begin{table}[t]
\centering
\caption{ID overall accuracy ($n{=}6{,}228$, sorted).}
\label{tab:T04}
\small
\setlength{\tabcolsep}{4pt}
\begin{tabular}{@{}lllr@{}}
\toprule
Model & Family & Params & ID acc \\
\midrule
GPT-5.4 & GPT & N/A & $65.8\%$ \\
Intern-S1-Pro & Intern & N/A & $63.2\%$ \\
Qwen2.5-VL-72B & Qwen2.5 & 72B & $58.9\%$ \\
Qwen3-VL-235B-A22B & Qwen3 & 235B-A22B (MoE) & $58.5\%$ \\
GPT-5.2 & GPT & N/A & $58.0\%$ \\
Gemma-4-31B & Gemma4 & 31B & $57.8\%$ \\
Qwen-VL-Max & Qwen-max & N/A & $55.6\%$ \\
Qwen3.5-397B-A17B & Qwen3.5 & 397B-A17B (MoE) & $55.4\%$ \\
Qwen3-VL-32B-Instruct & Qwen3 & 32B & $55.2\%$ \\
Seed-2.0-Mini & Seed & N/A & $54.3\%$ \\
ERNIE-4.5-VL-424B-A47B & ERNIE & 424B-A47B (MoE) & $51.7\%$ \\
Pixtral-Large-2411 & Mistral & 123B & $48.4\%$ \\
Qwen2.5-VL-32B & Qwen2.5 & 32B & $47.7\%$ \\
Gemma-3-27B & Gemma3 & 27B & $47.0\%$ \\
Seed-1.6-Flash & Seed & N/A & $45.3\%$ \\
ERNIE-4.5-VL-28B-A3B & ERNIE & 28B-A3B (MoE) & $44.9\%$ \\
Gemma-4-26B-A4B & Gemma4 & 26B-A4B (MoE) & $43.9\%$ \\
Qwen3-VL-30B-A3B & Qwen3 & 30B-A3B (MoE) & $42.8\%$ \\
Mistral-Small-3.2 & Mistral & 24B & $42.5\%$ \\
Qwen2.5-VL-7B & Qwen2.5 & 7B & $41.6\%$ \\
Gemma-3-12B & Gemma3 & 12B & $41.5\%$ \\
Mistral-Large-2512 & Mistral & N/A & $40.4\%$ \\
Qwen3.5-27B & Qwen3.5 & 27B & $33.3\%$ \\
Qwen3-VL-8B-Instruct & Qwen3 & 8B & $33.1\%$ \\
Llama-4-Scout & Llama4 & Scout (109B-MoE) & $32.6\%$ \\
MiniMax-01 & MiniMax & N/A & $31.9\%$ \\
Nemotron-Nano-12B-V2-VL & Nemotron & 12B & $31.5\%$ \\
Qwen3.5-9B & Qwen3.5 & 9B & $30.1\%$ \\
Gemma-3-4B & Gemma3 & 4B & $28.9\%$ \\
Llama-4-Maverick & Llama4 & Maverick (400B-MoE) & $28.7\%$ \\
Qwen3.5-122B-A10B & Qwen3.5 & 122B-A10B (MoE) & $26.9\%$ \\
Qwen2.5-VL-3B & Qwen2.5 & 3B & $26.3\%$ \\
InternVL3.5-Latest & Intern & N/A & $25.1\%$ \\
InternVL3.5-241B-A28B & Intern & 241B-A28B (MoE) & $25.1\%$ \\
Intern-S1-Mini & Intern & N/A & $24.4\%$ \\
Intern-S1 & Intern & N/A & $24.2\%$ \\
GLM-4.6V & GLM & N/A & $22.6\%$ \\
Qwen3.5-35B-A3B & Qwen3.5 & 35B-A3B (MoE) & $22.5\%$ \\
\bottomrule
\end{tabular}
\end{table}

\begin{table}[t]
\centering
\caption{ID accuracy by reasoning type (\%, all 38 models).}
\label{tab:T05}
\scriptsize
\setlength{\tabcolsep}{3pt}
\begin{tabular}{@{}lrrrrrrrr@{}}
\toprule
Model & Fwd & Inv & Rmv & Sub & Otm & Cue & Ord & Adj \\
\midrule
GPT-5.4 & 83.8 & 81.7 & 64.5 & 76.0 & 81.9 & 85.8 & 57.1 & 27.0 \\
GPT-5.2 & 70.7 & 78.6 & 53.2 & 56.3 & 78.8 & 85.4 & 46.4 & 27.1 \\
Qwen-VL-Max & 69.6 & 79.4 & 63.3 & 74.6 & 48.6 & 59.4 & 34.7 & 26.3 \\
Qwen2.5-VL-3B & 19.7 & 42.2 & 21.0 & 17.6 & 22.6 & 23.3 & 26.9 & 25.1 \\
Qwen2.5-VL-7B & 53.4 & 53.8 & 37.1 & 54.8 & 36.5 & 43.8 & 31.2 & 24.3 \\
Qwen2.5-VL-32B & 49.4 & 70.5 & 40.3 & 46.1 & 36.1 & 49.0 & 49.3 & 28.6 \\
Qwen2.5-VL-72B & 74.7 & 74.7 & 54.8 & 72.0 & 59.4 & 76.0 & 49.4 & 28.0 \\
Qwen3-VL-8B-Instruct & 24.8 & 69.4 & 24.9 & 19.2 & 24.0 & 25.3 & 28.8 & 24.8 \\
Qwen3-VL-30B-A3B & 40.2 & 76.2 & 36.7 & 36.2 & 32.3 & 41.0 & 38.5 & 24.6 \\
Qwen3-VL-32B-Instruct & 71.4 & 81.2 & 45.9 & 71.9 & 51.4 & 62.8 & 37.1 & 26.3 \\
Qwen3-VL-235B-A22B & 62.7 & 83.4 & 55.7 & 57.7 & 69.4 & 73.6 & 57.2 & 25.7 \\
Intern-S1 & 24.8 & 24.1 & 23.3 & 25.8 & 23.6 & 22.2 & 23.8 & 24.6 \\
Intern-S1-Mini & 25.7 & 23.9 & 23.1 & 27.6 & 22.6 & 23.3 & 23.0 & 24.5 \\
Intern-S1-Pro & 77.6 & 77.5 & 66.7 & 73.3 & 61.5 & 69.1 & 63.4 & 26.5 \\
InternVL3.5-241B-A28B & 23.5 & 26.5 & 25.4 & 24.9 & 25.3 & 28.5 & 24.7 & 25.0 \\
InternVL3.5-Latest & 26.3 & 24.8 & 25.3 & 24.2 & 24.3 & 24.7 & 25.6 & 24.7 \\
Gemma-3-4B & 24.5 & 53.8 & 21.5 & 18.5 & 21.5 & 24.0 & 25.6 & 23.7 \\
Gemma-3-12B & 49.5 & 56.5 & 41.6 & 46.2 & 40.6 & 49.7 & 29.5 & 26.1 \\
Gemma-3-27B & 62.2 & 62.6 & 47.8 & 62.7 & 39.2 & 47.9 & 30.6 & 26.2 \\
Gemma-4-26B-A4B & 51.1 & 59.4 & 39.2 & 55.6 & 49.7 & 56.2 & 29.2 & 26.1 \\
Gemma-4-31B & 76.3 & 76.6 & 63.6 & 64.5 & 64.9 & 70.5 & 30.8 & 25.3 \\
Llama-4-Scout & 19.8 & 67.7 & 18.8 & 18.1 & 18.4 & 15.3 & 32.6 & 31.9 \\
Llama-4-Maverick & 14.9 & 59.2 & 12.5 & 13.6 & 15.6 & 17.4 & 49.1 & 23.9 \\
Mistral-Small-3.2 & 49.8 & 53.0 & 39.4 & 44.8 & 38.9 & 45.1 & 27.1 & 24.6 \\
Mistral-Large-2512 & 52.0 & 54.7 & 29.2 & 50.9 & 40.3 & 54.5 & 26.9 & 22.7 \\
Pixtral-Large-2411 & 56.4 & 59.9 & 45.2 & 50.5 & 35.8 & 47.9 & 21.8 & 27.1 \\
MiniMax-01 & 23.6 & 59.8 & 22.6 & 26.5 & 29.5 & 24.3 & 28.0 & 26.0 \\
Qwen3.5-9B & 16.5 & 67.5 & 19.5 & 15.2 & 22.6 & 13.5 & 29.8 & 24.5 \\
Qwen3.5-27B & 28.2 & 67.8 & 25.4 & 21.7 & 25.3 & 32.2 & 35.6 & 24.6 \\
Qwen3.5-35B-A3B & 12.2 & 67.0 & 14.6 & 12.9 & 4.2 & 5.9 & 26.2 & 25.4 \\
Qwen3.5-122B-A10B & 19.6 & 18.6 & 26.5 & 19.5 & 50.0 & 56.9 & 17.6 & 23.9 \\
Qwen3.5-397B-A17B & 75.0 & 67.7 & 47.7 & 53.4 & 53.5 & 68.4 & 25.5 & 24.6 \\
GLM-4.6V & 10.5 & 25.3 & 9.0 & 18.0 & 39.6 & 60.4 & 42.2 & 26.1 \\
Seed-1.6-Flash & 50.7 & 65.0 & 27.9 & 60.4 & 44.4 & 56.2 & 25.7 & 25.9 \\
Seed-2.0-Mini & 76.0 & 67.4 & 50.5 & 50.9 & 43.4 & 53.5 & 53.1 & 27.2 \\
Nemotron-Nano-12B-V2-VL & 35.0 & 49.0 & 27.6 & 37.5 & 25.2 & 27.2 & 27.5 & 25.9 \\
ERNIE-4.5-VL-28B-A3B & 53.7 & 67.7 & 37.7 & 56.3 & 41.5 & 53.8 & 43.9 & 23.0 \\
ERNIE-4.5-VL-424B-A47B & 64.6 & 64.4 & 33.5 & 62.4 & 47.9 & 58.4 & 47.3 & 27.9 \\
\bottomrule
\end{tabular}
\end{table}

\begin{table}[t]
\centering
\caption{ID accuracy by action type (\%, all 38 models).}
\label{tab:T06}
\scriptsize
\setlength{\tabcolsep}{4pt}
\begin{tabular}{@{}lrrrrr@{}}
\toprule
Model & Exo & Perc & Insp & Navi & Mani \\
\midrule
GPT-5.4 & 62.4 & 61.2 & 58.3 & 68.8 & 81.2 \\
GPT-5.2 & 58.9 & 52.3 & 50.7 & 63.4 & 67.4 \\
Qwen-VL-Max & 40.8 & 46.2 & 52.1 & 60.8 & 79.4 \\
Qwen2.5-VL-3B & 24.2 & 24.9 & 22.8 & 32.2 & 28.9 \\
Qwen2.5-VL-7B & 33.5 & 32.2 & 34.4 & 46.4 & 64.3 \\
Qwen2.5-VL-32B & 39.6 & 43.2 & 39.6 & 61.0 & 58.3 \\
Qwen2.5-VL-72B & 53.7 & 47.7 & 47.8 & 66.4 & 83.4 \\
Qwen3-VL-8B-Instruct & 28.1 & 29.0 & 29.5 & 42.3 & 38.2 \\
Qwen3-VL-30B-A3B & 33.0 & 42.1 & 38.0 & 51.2 & 51.4 \\
Qwen3-VL-32B-Instruct & 43.7 & 52.3 & 47.8 & 61.9 & 73.2 \\
Qwen3-VL-235B-A22B & 56.2 & 55.5 & 43.7 & 67.3 & 75.8 \\
Intern-S1 & 24.5 & 22.9 & 26.2 & 23.9 & 22.9 \\
Intern-S1-Mini & 25.2 & 23.5 & 24.9 & 23.0 & 24.9 \\
Intern-S1-Pro & 56.2 & 52.8 & 53.7 & 70.1 & 87.1 \\
InternVL3.5-241B-A28B & 26.6 & 24.6 & 26.7 & 22.7 & 24.3 \\
InternVL3.5-Latest & 25.5 & 25.3 & 25.8 & 24.0 & 24.8 \\
Gemma-3-4B & 22.7 & 26.9 & 27.3 & 39.1 & 29.3 \\
Gemma-3-12B & 37.3 & 28.0 & 39.8 & 46.4 & 56.4 \\
Gemma-3-27B & 36.6 & 31.1 & 44.6 & 54.0 & 69.6 \\
Gemma-4-26B-A4B & 40.5 & 32.1 & 42.9 & 47.9 & 56.0 \\
Gemma-4-31B & 50.8 & 49.4 & 53.1 & 61.2 & 74.3 \\
Llama-4-Scout & 23.6 & 34.0 & 32.2 & 42.4 & 30.5 \\
Llama-4-Maverick & 20.9 & 27.7 & 26.8 & 42.5 & 26.3 \\
Mistral-Small-3.2 & 34.1 & 27.5 & 44.4 & 64.6 & 44.1 \\
Mistral-Large-2512 & 36.5 & 31.8 & 40.8 & 51.2 & 41.3 \\
Pixtral-Large-2411 & 33.6 & 35.5 & 55.6 & 80.6 & 48.6 \\
MiniMax-01 & 27.7 & 27.5 & 29.8 & 37.6 & 37.9 \\
Qwen3.5-9B & 22.5 & 28.3 & 27.0 & 40.1 & 33.9 \\
Qwen3.5-27B & 30.3 & 29.5 & 21.5 & 44.2 & 43.1 \\
Qwen3.5-35B-A3B & 10.3 & 29.3 & 15.3 & 27.6 & 30.0 \\
Qwen3.5-122B-A10B & 41.5 & 31.4 & 22.7 & 23.3 & 21.6 \\
Qwen3.5-397B-A17B & 53.4 & 57.7 & 42.4 & 61.5 & 65.2 \\
GLM-4.6V & 32.2 & 16.0 & 20.3 & 30.2 & 17.2 \\
Seed-1.6-Flash & 35.5 & 31.2 & 42.4 & 49.5 & 68.8 \\
Seed-2.0-Mini & 47.1 & 50.1 & 50.5 & 58.4 & 66.7 \\
Nemotron-Nano-12B-V2-VL & 26.0 & 24.4 & 31.9 & 34.6 & 40.7 \\
ERNIE-4.5-VL-28B-A3B & 36.6 & 34.2 & 40.8 & 51.4 & 63.0 \\
ERNIE-4.5-VL-424B-A47B & 40.1 & 40.5 & 48.7 & 66.6 & 63.4 \\
\bottomrule
\end{tabular}
\end{table}

\begin{table}[t]
\centering
\caption{Exogenous sub-principle accuracy (\%, ID; capable models shown, full 38 in repo).}
\label{tab:T07}
\scriptsize
\setlength{\tabcolsep}{3pt}
\begin{tabular}{@{}lrrrrrrrr@{}}
\toprule
Model & permanence & cohesion & solidity & gravity & support & inertia & collision & containment \\
\midrule
GPT-5.4 & 56.2 & 61.8 & 69.4 & 75.0 & 62.5 & 60.4 & 43.8 & 70.1 \\
GPT-5.2 & 57.6 & 60.4 & 62.5 & 75.0 & 50.7 & 55.6 & 43.8 & 66.0 \\
Qwen-VL-Max & 29.9 & 44.4 & 38.9 & 56.9 & 41.7 & 47.2 & 18.8 & 48.6 \\
Qwen2.5-VL-7B & 37.5 & 37.5 & 28.5 & 47.2 & 34.0 & 41.0 & 21.5 & 20.8 \\
Qwen2.5-VL-32B & 41.0 & 42.3 & 36.4 & 44.2 & 37.6 & 41.6 & 32.4 & 41.8 \\
Qwen2.5-VL-72B & 54.2 & 51.4 & 45.8 & 76.4 & 58.3 & 59.0 & 32.6 & 52.1 \\
Qwen3-VL-32B-Instruct & 34.7 & 51.4 & 41.7 & 56.9 & 43.8 & 44.4 & 27.1 & 49.3 \\
Qwen3-VL-235B-A22B & 52.8 & 61.1 & 49.3 & 73.6 & 61.8 & 61.1 & 33.3 & 56.2 \\
Intern-S1-Pro & 45.8 & 41.7 & 61.8 & 77.8 & 68.8 & 61.1 & 36.8 & 56.2 \\
Gemma-3-12B & 37.5 & 44.4 & 35.4 & 61.8 & 38.2 & 33.3 & 25.7 & 22.2 \\
Gemma-3-27B & 34.7 & 38.9 & 31.9 & 61.8 & 34.0 & 47.2 & 27.8 & 16.7 \\
Gemma-4-26B-A4B & 34.0 & 47.1 & 46.5 & 56.4 & 42.6 & 44.2 & 22.9 & 30.4 \\
Gemma-4-31B & 55.5 & 54.2 & 48.5 & 49.6 & 58.6 & 58.1 & 23.7 & 58.9 \\
Pixtral-Large-2411 & 39.3 & 31.2 & 31.2 & 50.7 & 26.3 & 26.8 & 25.4 & 37.3 \\
Qwen3.5-397B-A17B & 51.0 & 58.6 & 46.5 & 60.0 & 54.1 & 62.5 & 30.8 & 66.3 \\
Seed-2.0-Mini & 51.5 & 50.7 & 42.0 & 66.2 & 46.1 & 62.9 & 29.9 & 28.2 \\
ERNIE-4.5-VL-424B-A47B & 41.1 & 37.0 & 37.7 & 60.4 & 37.0 & 36.8 & 25.2 & 45.3 \\
\bottomrule
\end{tabular}
\end{table}

\begin{table}[t]
\centering
\caption{T07b Manipulative subject $\times$ view accuracy (\%, ID; capable models).}
\label{tab:T07b}
\scriptsize
\setlength{\tabcolsep}{4pt}
\begin{tabular}{@{}lrrrr@{}}
\toprule
Model & h/ego & h/allo & hh/ego & hh/allo \\
\midrule
GPT-5.4 & 87.2 & 80.9 & 82.6 & 74.3 \\
GPT-5.2 & 69.8 & 67.7 & 64.9 & 67.4 \\
Qwen-VL-Max & 85.1 & 79.2 & 81.2 & 72.2 \\
Qwen2.5-VL-7B & 68.8 & 65.6 & 66.7 & 56.2 \\
Qwen2.5-VL-32B & 66.7 & 59.0 & 60.8 & 46.9 \\
Qwen2.5-VL-72B & 91.3 & 84.4 & 85.8 & 72.2 \\
Qwen3-VL-32B-Instruct & 83.0 & 70.5 & 75.0 & 64.2 \\
Qwen3-VL-235B-A22B & 83.0 & 77.1 & 79.2 & 63.9 \\
Intern-S1-Pro & 89.6 & 87.5 & 89.6 & 81.6 \\
Gemma-3-27B & 76.7 & 67.4 & 75.0 & 59.4 \\
Gemma-4-31B & 78.1 & 74.7 & 77.8 & 66.7 \\
Pixtral-Large-2411 & 60.8 & 40.6 & 56.6 & 36.5 \\
Qwen3.5-397B-A17B & 68.1 & 61.8 & 65.6 & 65.3 \\
Seed-2.0-Mini & 70.1 & 66.7 & 70.1 & 59.7 \\
ERNIE-4.5-VL-424B-A47B & 75.7 & 63.5 & 69.8 & 44.4 \\
\bottomrule
\end{tabular}
\end{table}

\begin{table}[t]
\centering
\caption{Perturbation accuracy (overall / geo / app, $n{=}1{,}116$, all 38).}
\label{tab:T08}
\small
\setlength{\tabcolsep}{4pt}
\begin{tabular}{@{}lrrr@{}}
\toprule
Model & overall & geo & app \\
\midrule
GPT-5.4 & 79.1 & 62.7 & 95.5 \\
GPT-5.2 & 60.0 & 45.7 & 74.4 \\
Qwen-VL-Max & 48.9 & 33.7 & 64.2 \\
Qwen2.5-VL-3B & 21.1 & 22.6 & 19.7 \\
Qwen2.5-VL-7B & 37.0 & 23.5 & 50.5 \\
Qwen2.5-VL-32B & 37.4 & 30.1 & 44.6 \\
Qwen2.5-VL-72B & 60.8 & 38.5 & 83.0 \\
Qwen3-VL-8B-Instruct & 23.0 & 25.1 & 21.0 \\
Qwen3-VL-30B-A3B & 34.1 & 25.6 & 42.5 \\
Qwen3-VL-32B-Instruct & 50.4 & 32.4 & 68.5 \\
Qwen3-VL-235B-A22B & 55.5 & 32.8 & 78.1 \\
Intern-S1 & 26.7 & 26.5 & 27.0 \\
Intern-S1-Mini & 26.4 & 26.4 & 26.4 \\
Intern-S1-Pro & 56.3 & 33.3 & 79.2 \\
InternVL3.5-241B-A28B & 26.7 & 26.3 & 27.1 \\
InternVL3.5-Latest & 26.8 & 26.0 & 27.6 \\
Gemma-3-4B & 24.4 & 24.6 & 24.2 \\
Gemma-3-12B & 40.3 & 32.6 & 48.0 \\
Gemma-3-27B & 49.5 & 36.4 & 62.5 \\
Gemma-4-26B-A4B & 34.1 & 25.4 & 42.8 \\
Gemma-4-31B & 77.2 & 59.0 & 95.5 \\
Llama-4-Scout & 15.2 & 17.0 & 13.4 \\
Llama-4-Maverick & 11.3 & 9.9 & 12.7 \\
Mistral-Small-3.2 & 34.3 & 27.3 & 41.3 \\
Mistral-Large-2512 & 31.7 & 27.1 & 36.4 \\
Pixtral-Large-2411 & 50.1 & 35.1 & 65.1 \\
MiniMax-01 & 25.0 & 26.3 & 23.7 \\
Qwen3.5-9B & 20.6 & 20.8 & 20.4 \\
Qwen3.5-27B & 36.8 & 31.2 & 42.5 \\
Qwen3.5-35B-A3B & 12.6 & 10.6 & 14.7 \\
Qwen3.5-122B-A10B & 25.8 & 24.4 & 27.2 \\
Qwen3.5-397B-A17B & 50.0 & 32.8 & 67.2 \\
GLM-4.6V & 7.2 & 5.4 & 9.0 \\
Seed-1.6-Flash & 54.9 & 38.4 & 71.5 \\
Seed-2.0-Mini & 48.3 & 27.1 & 69.5 \\
Nemotron-Nano-12B-V2-VL & 25.5 & 27.1 & 24.0 \\
ERNIE-4.5-VL-28B-A3B & 42.5 & 32.5 & 52.5 \\
ERNIE-4.5-VL-424B-A47B & 35.3 & 25.1 & 45.5 \\
\bottomrule
\end{tabular}
\end{table}

\section{Diagnostic results}
\label{sec:assess:findings}

\begin{figure}[!htbp]
\centering
\includegraphics[width=\linewidth]{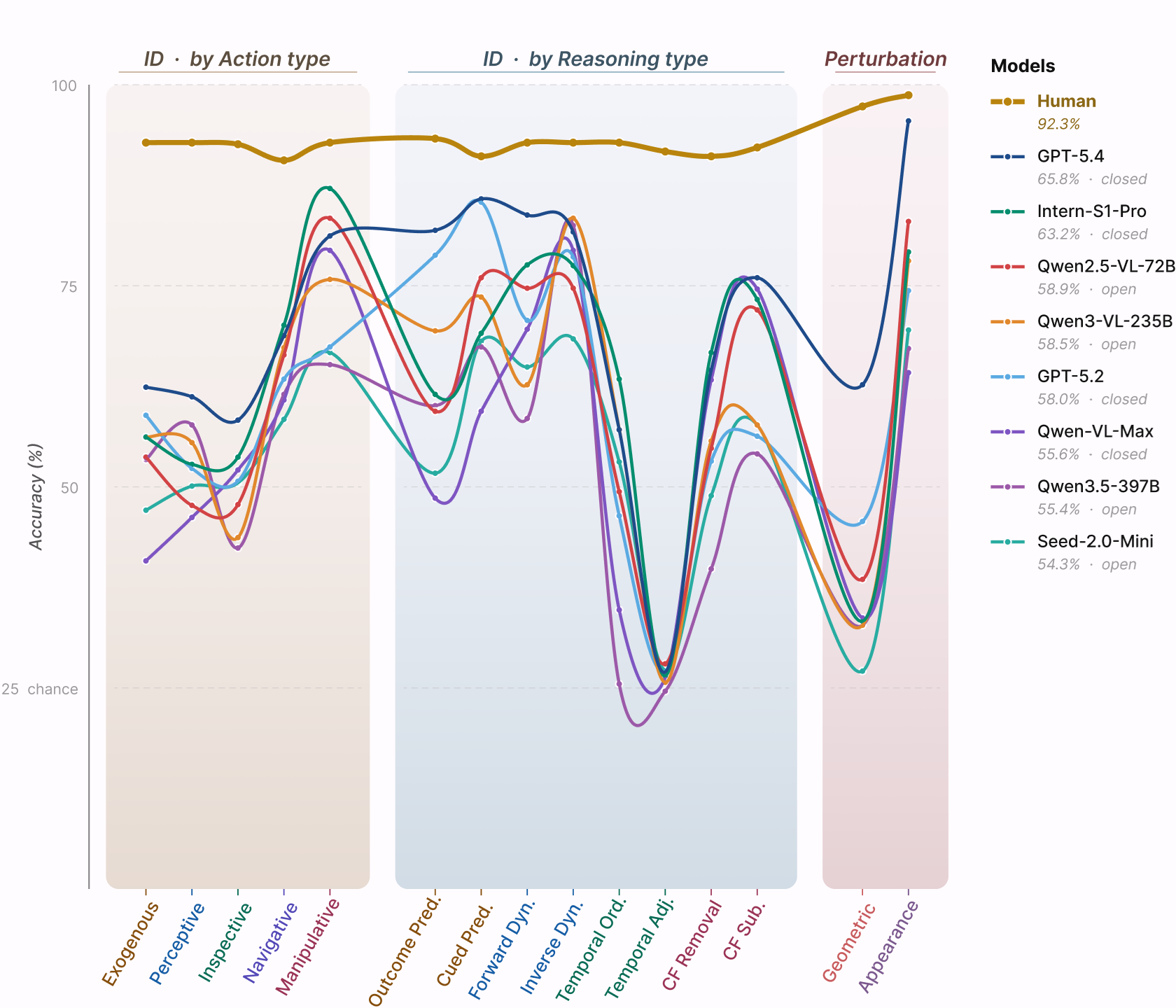}
\caption{\textbf{Diagnostic profiles of eight frontier MLLMs against the human baseline.} Performance is grouped by action type (left), reasoning type (middle), and state-axis perturbation (right). Full $38$-model results are reported in Appendix~\ref{app:eval-tables}.}
\label{fig:assess-glance}
\end{figure}

This appendix details the diagnostic baseline summarized in \S\ref{sec:assess}. Figure~\ref{fig:assess-glance} compares the action, reasoning, and perturbation profiles of eight frontier models; Tables~\ref{tab:T05},~\ref{tab:T06}, and~\ref{tab:T08} report the full cohort. The following analyses examine geometric reasoning, scaling, temporal and counterfactual asymmetries, architecture, and scenario factors.

\subsection{Rotation deficit details}
\label{app:F-A}
\label{sec:assess:rotation}

The geometric reasoning deficit is supported by four converging observations. (A.1) Inspective accuracy falls below manipulative accuracy for $11/11$ frontier models, with manipulative$-$inspective gaps of $+16.2$\,to\,$+35.6$pp (Table~\ref{tab:T21}). (A.2) Within perceptive, translation $\gg$ rotation accuracy by $+8$\,to\,$+55$pp ($28/29$ above-chance models positive; Table~\ref{tab:T24}). (A.3) Errors on rotation cells concentrate on the mirror-direction distractor at frontier scale. (A.4) $\chi^2(\text{df}{=}2) > 100$ for $9/11$ frontier models on perturbation geometric distractors when restricted to orientation-flip items, Cramér's $V > 0.30$.

\begin{table}[!htbp]
\centering
\caption{Manipulative$-$inspective accuracy gap (frontier).}
\label{tab:T21}
\small
\setlength{\tabcolsep}{4pt}
\begin{tabular}{@{}lrrr@{}}
\toprule
Model & manipulative & inspective & gap \\
\midrule
GPT-5.4 & 81.2 & 58.3 & ${+}23.0$ \\
GPT-5.2 & 67.4 & 50.7 & ${+}16.7$ \\
Qwen-VL-Max & 79.4 & 52.1 & ${+}27.3$ \\
Qwen2.5-VL-7B & 64.3 & 34.4 & ${+}29.9$ \\
Qwen2.5-VL-32B & 58.3 & 39.6 & ${+}18.8$ \\
Qwen2.5-VL-72B & 83.4 & 47.8 & ${+}35.6$ \\
Qwen3-VL-32B & 73.2 & 47.8 & ${+}25.4$ \\
Qwen3-VL-235B-A22B & 75.8 & 43.7 & ${+}32.1$ \\
Intern-S1-Pro & 87.1 & 53.7 & ${+}33.4$ \\
Gemma-4-31B & 74.3 & 53.1 & ${+}21.2$ \\
Seed-2.0-Mini & 66.7 & 50.5 & ${+}16.2$ \\
\bottomrule
\end{tabular}
\end{table}

\begin{table}[!htbp]
\centering
\caption{Perceptive translation vs.\ rotation (frontier).}
\label{tab:T24}
\small
\setlength{\tabcolsep}{4pt}
\begin{tabular}{@{}lrrr@{}}
\toprule
Model & translation & rotation & gap \\
\midrule
GPT-5.4 & 92.4 & 74.0 & ${+}18.4$ \\
GPT-5.2 & 88.2 & 56.6 & ${+}31.6$ \\
Qwen-VL-Max & 82.6 & 42.7 & ${+}39.9$ \\
Qwen2.5-VL-7B & 56.6 & 17.0 & ${+}39.6$ \\
Qwen2.5-VL-32B & 58.0 & 41.3 & ${+}16.7$ \\
Qwen2.5-VL-72B & 77.1 & 42.7 & ${+}34.4$ \\
Qwen3-VL-32B & 86.1 & 61.5 & ${+}24.7$ \\
Qwen3-VL-235B-A22B & 88.2 & 60.1 & ${+}28.1$ \\
Intern-S1-Pro & 76.7 & 57.6 & ${+}19.1$ \\
Gemma-4-31B & 91.0 & 47.9 & ${+}43.1$ \\
Pixtral-Large-2411 & 67.4 & 12.2 & ${+}55.2$ \\
Seed-2.0-Mini & 79.9 & 34.4 & ${+}45.5$ \\
\bottomrule
\end{tabular}
\end{table}

\subsection{Geometric vs.\ appearance scaling dissociation}
\label{app:F-B}
\label{sec:assess:scaling}

\begin{figure}[!htbp]
\centering
\includegraphics[width=\linewidth]{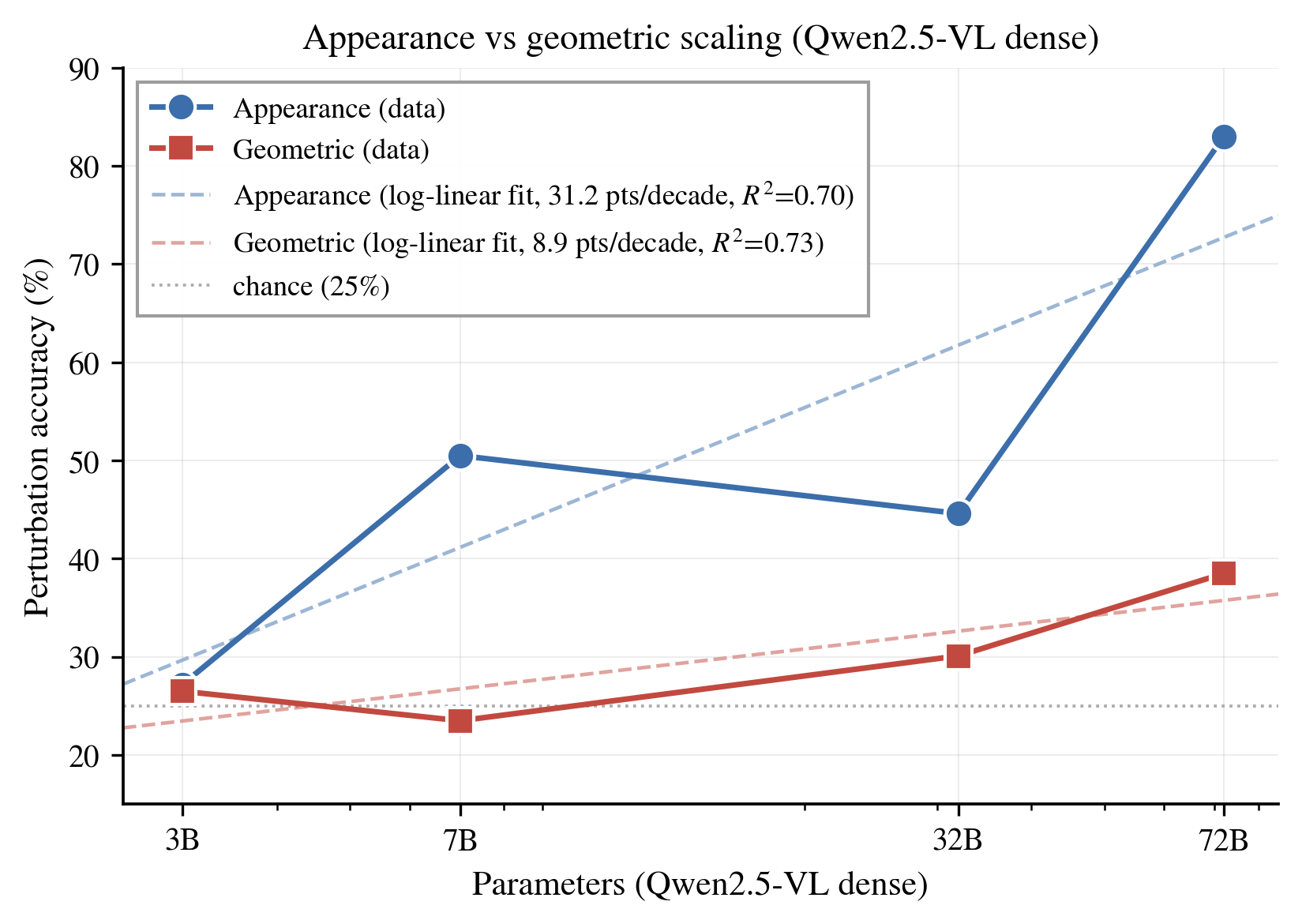}
\caption{\textbf{Appearance and geometric robustness across the Qwen2.5-VL dense model ladder ($\{3,7,32,72\}$B).} Appearance robustness scales $3.5{\times}$ faster than geometric robustness, with fitted slopes of $31.2$ and $8.9$ percentage points per decade of parameters, respectively.}
\label{fig:fb-scaling}
\end{figure}

Within-family scaling fit on Qwen2.5-VL dense ladder is OLS log-linear $y = a + b\log_{10} N$ with $b_{\text{app}} = 31.2$ pts/decade ($R^2{=}0.70$), $b_{\text{geo}} = 8.9$ pts/decade ($R^2{=}0.73$), ratio $3.5{\times}$ (Figure~\ref{fig:fb-scaling}). Pooled over the four dense ladders (Qwen2.5-VL $\{3,7,32,72\}$B, Qwen3-VL $\{8,32\}$B, Gemma-3 $\{4,12,27\}$B, Qwen3.5 $\{9,27\}$B; $11$ models) with a separate intercept per ladder, both slopes are significant (appearance $39.8$ pts/decade, $p{=}0.004$; geometric $11.1$ pts/decade, $p{=}0.003$; $\mathrm{df}{=}6$), and the appearance slope exceeds the geometric slope by $28.7$ pts/decade ($95\%$ CI $[8.1, 49.4]$, $p{=}0.015$). The lower appearance $R^2$ reflects a non-monotonicity at $7$B; we leave the point in the fit. First-order extrapolation places $50\%$ geometric at $N \approx 2.9{\times}10^{12}$ and $60\%$ at $N \approx 3.8{\times}10^{13}$ parameters; `we report this to communicate the gap's severity, with the log-linear fit's $R^2{=}0.73$ holding within the $3$B--$72$B range only and no claim made beyond it. Within-model two-proportion $z$-test on Qwen2.5-VL-7B: $p_{\text{app}} = 0.505$, $p_{\text{geo}} = 0.235$, $n_{\text{app}}=n_{\text{geo}}=558$, $z{=}9.34$, $p{<}10^{-20}$.
\begin{table}[!htbp]
\centering
\caption{Qwen2.5-VL dense scaling (\%).}
\label{tab:T25}
\small
\setlength{\tabcolsep}{4pt}
\begin{tabular}{@{}lcrrrr@{}}
\toprule
Model & Params & ID & overall & geo & app \\
\midrule
Qwen2.5-VL-3B & 3B & 26.4 & 21.1 & 22.6 & 19.7 \\
Qwen2.5-VL-7B & 7B & 41.6 & 37.0 & 23.5 & 50.5 \\
Qwen2.5-VL-32B & 32B & 47.7 & 37.4 & 30.1 & 44.6 \\
Qwen2.5-VL-72B & 72B & 58.9 & 60.8 & 38.5 & 83.0 \\
\bottomrule
\end{tabular}
\end{table}

\begin{table}[!htbp]
\centering
\caption{Qwen3-VL family.}
\label{tab:T26}
\small
\setlength{\tabcolsep}{4pt}
\begin{tabular}{@{}lcrrrr@{}}
\toprule
Model & Params & ID & overall & geo & app \\
\midrule
Qwen3-VL-8B-Instruct & 8B & 33.1 & 23.0 & 25.1 & 21.0 \\
Qwen3-VL-30B-A3B & 30B-A3B (MoE) & 42.8 & 34.1 & 25.6 & 42.5 \\
Qwen3-VL-32B-Instruct & 32B & 55.2 & 50.4 & 32.4 & 68.5 \\
Qwen3-VL-235B-A22B & 235B-A22B (MoE) & 58.5 & 55.5 & 32.8 & 78.1 \\
\bottomrule
\end{tabular}
\end{table}

\begin{table}[!htbp]
\centering
\caption{Gemma-3/Gemma-4.}
\label{tab:T27}
\small
\setlength{\tabcolsep}{4pt}
\begin{tabular}{@{}lcrrrr@{}}
\toprule
Model & Params & ID & overall & geo & app \\
\midrule
Gemma-3-4B & 4B & 28.9 & 24.4 & 24.6 & 24.2 \\
Gemma-3-12B & 12B & 41.5 & 40.3 & 32.6 & 48.0 \\
Gemma-3-27B & 27B & 47.0 & 49.5 & 36.4 & 62.5 \\
Gemma-4-26B-A4B & 26B-A4B (MoE) & 43.9 & 34.1 & 25.4 & 42.8 \\
Gemma-4-31B & 31B & 57.8 & 77.2 & 59.0 & 95.5 \\
\bottomrule
\end{tabular}
\end{table}

\begin{table}[!htbp]
\centering
\caption{Qwen3.5 family (reasoning disabled).}
\label{tab:T28}
\small
\setlength{\tabcolsep}{4pt}
\begin{tabular}{@{}lcrrrr@{}}
\toprule
Model & Params & ID & overall & geo & app \\
\midrule
Qwen3.5-9B & 9B & 30.1 & 20.6 & 20.8 & 20.4 \\
Qwen3.5-27B & 27B & 33.3 & 36.8 & 31.2 & 42.5 \\
Qwen3.5-35B-A3B & 35B-A3B (MoE) & 22.5 & 12.6 & 10.6 & 14.7 \\
Qwen3.5-122B-A10B & 122B-A10B (MoE) & 26.9 & 25.8 & 24.4 & 27.2 \\
Qwen3.5-397B-A17B & 397B-A17B (MoE) & 55.4 & 50.0 & 32.8 & 67.2 \\
\bottomrule
\end{tabular}
\end{table}

\subsection{Capability--representation decoupling}
\label{app:F-C}

ID-overall vs.\ perturbation Spearman correlations across $n{=}38$ models: $\rho_{\text{app}}{=}{+}0.90$ ($R^2{=}0.86$), $\rho_{\text{geo}}{=}{+}0.76$ ($R^2{=}0.52$). Roughly $48\%$ of geometric-robustness variance is independent of overall capability. The rank-shift table makes the decoupling visible: ERNIE-4.5-VL-424B-A47B drops from ID rank $11$ to geometric rank $30$ ($\Delta {=} {+}19$); Seed-1.6-Flash rises from ID rank $15$ to geometric rank $5$ ($\Delta {=} {-}10$).

\begin{table}[!htbp]
\centering
\caption{ID vs.\ perturbation-geometric rank shift (subset; positive $\Delta$ = relatively worse on geometric).}
\label{tab:T31}
\small
\setlength{\tabcolsep}{4pt}
\begin{tabular}{@{}lrrrrr@{}}
\toprule
Model & ID acc & ID rank & Geo acc & Geo rank & $\Delta$ \\
\midrule
ERNIE-4.5-VL-424B-A47B & 51.7 & 11 & 25.1 & 30 & ${+}19$ \\
Qwen2.5-VL-7B & 41.6 & 20 & 23.5 & 33 & ${+}13$ \\
Gemma-4-26B-A4B & 43.9 & 17 & 25.4 & 28 & ${+}11$ \\
Llama-4-Scout & 32.6 & 25 & 17.0 & 35 & ${+}10$ \\
Qwen3-VL-30B-A3B & 42.8 & 18 & 25.6 & 27 & ${+}9$ \\
Seed-2.0-Mini & 54.3 & 10 & 27.1 & 19 & ${+}9$ \\
Intern-S1-Pro & 63.2 & 2 & 33.3 & 9 & ${+}7$ \\
GPT-5.4 & 65.8 & 1 & 62.7 & 1 & $0$ \\
GPT-5.2 & 58.0 & 5 & 45.7 & 3 & ${-}2$ \\
Gemma-4-31B & 57.8 & 6 & 59.0 & 2 & ${-}4$ \\
Seed-1.6-Flash & 45.3 & 15 & 38.4 & 5 & ${-}10$ \\
Intern-S1 & 24.2 & 36 & 26.5 & 22 & ${-}14$ \\
\bottomrule
\end{tabular}
\end{table}

\subsection{Temporal ordering--adjacency dissociation}
\label{app:F-D}

The frontier median ordering--adjacency gap is ${+}20.7$pp (mean ${+}20.9$pp, range ${+}5.5$\,to\,${+}36.9$, $11/11$ frontier models positive; Table~\ref{tab:T22}). For GPT-5.4, temporal adjacency is $27.0\%$, within $2$pp of chance, consistent with temporal adjacency being the most isolated reasoning type family in WOVEN. Both queries operate within a single action's frame sequence, so the dissociation indicates that models recover the coarse order of the transition's intermediate states but not their fine temporal resolution, the two grains of the trajectory representation of \S\ref{app:two-axes}.

\begin{table}[!htbp]
\centering
\caption{Temporal ordering--adjacency gap (frontier).}
\label{tab:T22}
\small
\setlength{\tabcolsep}{4pt}
\begin{tabular}{@{}lrrr@{}}
\toprule
Model & Ord & Adj & Gap \\
\midrule
GPT-5.4 & 57.1 & 27.0 & ${+}30.1$ \\
GPT-5.2 & 46.4 & 27.1 & ${+}19.3$ \\
Qwen-VL-Max & 34.7 & 26.3 & ${+}8.4$ \\
Qwen2.5-VL-32B & 49.3 & 28.6 & ${+}20.7$ \\
Qwen2.5-VL-72B & 49.4 & 28.0 & ${+}21.4$ \\
Qwen3-VL-32B & 37.1 & 26.3 & ${+}10.8$ \\
Qwen3-VL-235B-A22B & 57.2 & 25.7 & ${+}31.5$ \\
Intern-S1-Pro & 63.4 & 26.5 & ${+}36.9$ \\
Gemma-4-31B & 30.8 & 25.3 & ${+}5.5$ \\
Seed-2.0-Mini & 53.1 & 27.2 & ${+}25.9$ \\
ERNIE-4.5-VL-424B-A47B & 47.3 & 27.9 & ${+}19.4$ \\
\bottomrule
\end{tabular}
\end{table}

\subsection{MoE active-parameter tax}
\label{app:F-E}

Three matched-total within-family pairs yield dense premia of ${+}12.4$, ${+}13.9$, ${+}10.9$pp respectively (Table~\ref{tab:T29}). Visual world-modeling capability tracks active parameters rather than total. Scaling plots that compare MoE to dense must use the active-parameter axis under penalty of $\sim\!12$pp overstatement.

\begin{table}[!htbp]
\centering
\caption{Dense vs.\ MoE comparisons (matched total).}
\label{tab:T29}
\small
\setlength{\tabcolsep}{4pt}
\begin{tabular}{@{}lrrr@{}}
\toprule
Comparison & Dense acc & MoE acc & gap \\
\midrule
Qwen3-VL-32B vs.\ 30B-A3B & 55.2 & 42.8 & ${+}12.4$ \\
Gemma-4-31B vs.\ 26B-A4B & 57.8 & 43.9 & ${+}13.9$ \\
Qwen3.5-27B vs.\ 35B-A3B & 33.3 & 22.5 & ${+}10.9$ \\
\bottomrule
\end{tabular}
\end{table}

\subsection{Emergence ordering and gravity violation asymmetry}
\label{app:F-F}
\label{sec:assess:emergence}

\textbf{Emergence ordering.} Eight exogenous principles cross the $40\%$ ID threshold in a cross-ladder order on Qwen2.5-VL: \emph{gravity, inertia} at $7$B; \emph{cohesion, permanence, containment} at $32$B; \emph{support, solidity} at $72$B; \emph{collision} never crosses (frontier mean $31.4\%$). The ordering replicates on Qwen3-VL ladder ($8$B$\to 235$B-A22B) and Gemma-3 ladder ($4$B$\to 27$B), aligning with the developmental sequence in infant cognition~\citep{spelke2007coreknowledge,baillargeon1987objectpermanence}.

\textbf{Gravity violation asymmetry.} The violation-of-expectation distractor breakdown on cued prediction for the gravity principle shows directional asymmetry: across $12$ above-chance models from $7$ architecturally distinct families, $P(\texttt{rise}\mid\text{err}){=}20.5\%$ vs.\ $P(\texttt{float}\mid\text{err}){=}37.0\%$ vs.\ $P(\texttt{lateral}\mid\text{err}){=}38.6\%$ (one-sample $t{=}{-}3.85$, $p{=}0.003$). The pattern replicates on outcome prediction.

\subsection{Counterfactual cost asymmetry}
\label{app:F-G}

Define $\mathrm{cost}(\mathrm{Fwd}\to X) := \mathrm{acc}(\mathrm{Fwd}) - \mathrm{acc}(X)$ for $X \in \{\mathrm{Rmv}, \mathrm{Sub}\}$. Across $24$ above-chance models, $23$ satisfy $\mathrm{cost}(\mathrm{Rmv}) > \mathrm{cost}(\mathrm{Sub})$ with mean gap ${+}11.0$pp (paired Wilcoxon $p{<}10^{-5}$). Mechanism is scale-dependent: frontier models distribute counterfactual-removal errors away from the actually-occurring action's outcome (mean $P_{\text{actual}\mid\text{err}}{=}20.4\%$, all below $1/3$ uniform; GPT-5.4 $24.0\%$); mid-scale models revert to it (Qwen2.5-VL-7B $63.6\%$). Pearson $r$(ID overall, $P_{\text{actual}\mid\text{err}}){=}{-}0.06$ across $38$ models. Appendix~\ref{app:E-F6} provides converse evidence (Appendix~\ref{app:E-F6}).

\subsection{Manipulative pair-mate independence + view-axis dominance}
\label{app:F-H}

\textbf{Cell-hardness ladder}: $\texttt{h\_ego} > \texttt{hh\_ego} > \texttt{h\_allo} > \texttt{hh\_allo}$ (cohort-mean source-acc $54.2 > 52.6 > 49.5 > 44.6$). View axis (ego vs.\ allo) shifts $\sim\!5$pp; combined with subject shift adds another $\sim\!5$pp.

\textbf{Pair-mate independence.} Pearson $r$ between cell pairs across $18$ scenario-groups, averaged over $\sim\!152$ (model $\times$ reasoning type) combinations: all six pairwise $r$ in $[{+}0.02, {+}0.16]$. Diagonal cells (\texttt{h\_ego} $\leftrightarrow$ \texttt{hh\_allo}) essentially uncorrelated; same-axis (\texttt{allo} $\leftrightarrow$ \texttt{allo}) highest. Models do not have a unified scenario representation.

\textbf{View-axis dominates subject-axis in single-axis failures.} Of $14$ (model, reasoning type) combinations with $\geq 4/18$ scenarios exhibiting single-axis failure, $11$ are view-axis (ego/allo) and only $3$ are subject-axis (human/humanoid). Within view-axis failures, ego-only-correct dominates.

\subsection{Per-scene invariance within ID}
\label{app:F-I}

Across $12$ above-chance models on $15$ ID scenes, per-model scene CV ranges $4.04$\,to\,$8.58\%$, median $5.6\%$. Per-model max$-$min scene gap $8$\,to\,$16$pp. Most uniform: GPT-5.4 (CV $4.04\%$, gap $8.4$pp); least uniform: ERNIE-4.5-VL-424B (CV $8.58\%$, gap $16.1$pp). This sets a measurement-validity floor for OOD-gap interpretation: \texttt{beach} + \texttt{restaurant} OOD vs.\ ID gaps larger than $\sim\!5$\,to\,$9$pp can be attributed to scene novelty.

\subsection{Small-scale no-op prior}
\label{app:F-Minor}

Qwen2.5-VL-3B picks the \texttt{static} (``no action'') option on $34.0\%$ of inspective forward-dynamics items, well above the $25\%$ uniform-chance baseline (one-sided binomial, $k{=}102$, $N{=}300$, $p{=}3.1{\times}10^{-4}$). When the smallest model is uncertain about camera motion, it defaults to ``nothing happened.'' This signature disappears at $\geq 7$B and is absent from frontier models.

\subsection{Family fingerprinting via principle vectors}
\label{app:F-Fingerprint}

Each model's $8$-dim per-principle accuracy vector serves as a representational fingerprint. Centred cosine similarity (after subtracting each model's own mean, measures \emph{shape} rather than capability level) shows within-family similarity is high but not uniform.

\begin{table}[!htbp]
\centering
\caption{F-Fingerprint: centered cosine similarity on $8$-dim per-principle accuracy vectors.}
\label{tab:Tfingerprint}
\small
\begin{tabular}{@{}lr@{}}
\toprule
Pair & centered cos \\
\midrule
Qwen2.5-VL-72B $\leftrightarrow$ Qwen3-VL-235B-A22B & ${+}0.951$ \\
GPT-5.4 $\leftrightarrow$ GPT-5.2 & ${+}0.895$ \\
Qwen3-VL-235B $\leftrightarrow$ Qwen3-VL-32B & ${+}0.895$ \\
ERNIE-424B $\leftrightarrow$ GPT-5.4 & ${+}0.825$ \\
Qwen3-VL-235B $\leftrightarrow$ GPT-5.4 & ${+}0.761$ \\
Qwen2.5-VL-72B $\leftrightarrow$ GPT-5.4 & ${+}0.712$ \\
Seed-2.0-Mini $\leftrightarrow$ GPT-5.4 & ${+}0.341$ \\
Gemma-4-31B $\leftrightarrow$ Gemma-3-27B (same provider, gen step) & $\mathbf{{+}0.104}$ \\
\bottomrule
\end{tabular}
\end{table}

Gemma's $3{\to}4$ generation transition produces an \emph{orthogonal} per-principle reasoning type profile (${+}0.104$), unlike Qwen2.5\,$\to$\,Qwen3 which preserves shape (${+}0.951$). Within-provider scaling does not preserve representational shape.

\section{Training recipes and finding details}

\subsection{Shared supervised fine-tuning protocol}
\label{app:E-recipes}

\textbf{Base models and reporting.} The main teachability and transfer comparisons use Qwen2.5-VL-3B-Instruct~\citep{bai2025qwen25vl}. Model-scale experiments use Qwen2.5-VL-7B-Instruct and Qwen2.5-VL-32B-Instruct (Appendix~\ref{app:E-F13}); the matched-backbone video-supervision comparison uses Qwen3.5-4B (Appendix~\ref{app:E-F14}). SFT and subsequent GRPO~\citep{shao2024deepseekmath} results are reported separately where both stages are evaluated.

\textbf{Shared SFT optimization.} All SFT experiments, including full-mix, single-cell, and temporal supervision, use LoRA with $r{=}16$, $\alpha{=}4$, dropout $0.05$, and target modules $\{q,k,v,o\}_{\text{proj}}$. Training uses $3$ epochs, batch size $32$, AdamW with learning rate $1{\times}10^{-4}$, cosine scheduling, $5\%$ warmup, and fp$16$ precision.

\textbf{Option-permutation augmentation.} For WOVEN-only training, each item is presented under $4$ permutations of its MCQ options, with the answer updated consistently, to reduce dependence on answer position. Fixed-budget substitution experiments use the answer-balanced data construction described in Appendix~\ref{app:E-F12}.

\subsection{Full-mix SFT calibration}
\label{app:E-F1}

Full-mix SFT uses the $22{,}728$-item WOVEN training set and the shared optimization protocol in Appendix~\ref{app:E-recipes}. On Qwen2.5-VL-3B-Instruct, it reaches $89.3\%$ on the in-distribution test and $88.2\%$ on held-out scenes (\S\ref{sec:enhance:sft}).

\subsection{Cross-benchmark transfer}
\label{app:E-F2}

Full-mix cross-benchmark transfer under the final locked protocol is reported in the main text (Tables~\ref{tab:enhance-transfer} and~\ref{tab:enhance-bestcell}); per-benchmark, per-category breakdowns for all training configurations are in Appendix~\ref{app:E-F11}. Within WM-ABench, the gains concentrate on the transition dimensions (Prediction/Transitivity ${+}21.7$, Prediction/Mechanistic ${+}18.6$) while the static Visual dimension is flat (${\pm}1$); see the WM-ABench table in Appendix~\ref{app:E-F11}.

\begin{table}[!htbp]
\centering
\caption{\textbf{SFT accuracy on WOVEN's own test sets (ID, scene-OOD, state-axis perturbation probe) (\%).} Full-mix SFT is on Qwen2.5-VL-3B; inverse-text is narrow SFT on inverse dynamics with text options ($4$ endogenous action types); mv-balanced is balanced \texttt{mv\_fwd}+\texttt{mv\_bwd} training on Qwen2.5-VL-7B. Downstream transfer is in Tables~\ref{tab:enhance-transfer} and~\ref{tab:enhance-bestcell}. Per-cell own deltas, the remaining narrow-run families, and full recipe details in the appendix.}
\label{tab:enhance-main-detail}
\scriptsize
\setlength{\tabcolsep}{3pt}
\begin{tabular}{@{}lrrrrr@{}}
\toprule
& \multicolumn{2}{c}{} & \multicolumn{3}{c}{Perturbation probe} \\
\cmidrule(lr){4-6}
Variant & ID & OOD & all & geo & app \\
\midrule
Qwen2.5-VL-3B (base)         & 26.4 & --- & 21.1 & 22.6 & 19.7 \\
\quad +full-mix SFT          & \textbf{89.3} ($\uparrow${62.9}) & \textbf{88.2} & \textbf{38.3} ($\uparrow${17.1}) & \textbf{35.0} ($\uparrow${12.4}) & \textbf{41.6} ($\uparrow${21.9}) \\
\quad +inverse-text SFT & 95.5\textsuperscript{$\dagger$} ($\uparrow${54}) & --- & --- & --- & --- \\
\midrule
Qwen2.5-VL-7B (base)         & 41.6 & --- & 37.0 & 23.5 & 50.5 \\
\quad +mv-balanced SFT & 50.1 ($\uparrow${8.5}) & 57.5 & 52.4 ($\uparrow${15.4}) & 38.4 ($\uparrow${14.9}) & 66.5 ($\uparrow${16.0}) \\
\bottomrule
\end{tabular}

{\footnotesize $\dagger$ inverse-text own-cell mean over 4 endogenous action types: base $41\%$, post-SFT $95.5\%$ ($\uparrow$54).}
\end{table}

\subsection{Targeted training and diagnostic correspondence}
\label{sec:enhance:targeted}

\textbf{Inverse-text is the strongest narrow world-modeling signal.} Narrow SFT on inverse dynamics with text-described action options across the $4$ endogenous action types ($n{=}4{,}132$, shared SFT protocol; Appendix~\ref{app:E-F3}) yields ${+}54$pp own-cell average (inspective ${+}77$, perceptive ${+}70$, navigative ${+}39$, manipulative ${+}32$). Holding option modality constant and switching only direction from forward to inverse gives ${+}51$pp: the gain comes from the inverse direction rather than the text-options modality.

\textbf{Cross-action-type forward transfer is asymmetric.} Single-action-type forward-dynamics training across the four endogenous action types yields three asymmetries (Appendix~\ref{app:E-F5}): (i) navigative$\leftrightarrow$manipulative double-strong (${+}61$pp both directions); (ii) perceptive$\leftrightarrow$inspective weakly symmetric (${+}13$/${+}14$pp); (iii) P/I$\to$N/M strong but N/M$\to$P/I weak (${+}47$\,to\,${+}57$ vs.\ ${+}1$\,to\,${+}11$). One hypothesis is that P/I training teaches both visual grounding and spatial reasoning, while N/M training teaches only grounding.

\textbf{Imagining absence stays hard even when its neighbors are trained.} Within an action type (Appendix~\ref{app:E-F6}), inspective forward dynamics teaches own ${+}65$pp, transfers to counterfactual substitution ${+}48$ and counterfactual removal ${+}25$, but not to temporal templates. Counterfactual-substitution training transfers to forward dynamics ${+}66$ and inverse dynamics ${+}34$ but to removal $\approx 0$pp. The Sub\,$\not\to$\,Rmv non-transfer triangulates the cohort-level cost asymmetry: ``imagining absence'' is intrinsic to the reasoning type rather than a coverage shortage.

\textbf{Temporal reasoning is the most isolated family in WOVEN.} Temporal adjacency is teachable in narrow SFT (own ${+}30$\,to\,${+}35$pp across $4$ action types; Appendix~\ref{app:E-F7}); temporal ordering is not under our setup; the two sister reasoning types show $\approx 0$pp transfer in both directions. Combined with the cohort-level dissociation of \S\ref{sec:assess}, this means temporal skill neither arrives with the other families nor spreads to them: it must be trained directly.

\textbf{Chirality is direction-asymmetric and capacity-dependent.} Balanced rotation training teaches both directions (Appendix~\ref{app:E-F8-7B-scaling},~\ref{app:E-F9}) (${+}55$\,to\,${+}61$pp own); single-direction teaches own-direction but cross-direction $\Delta\approx 0$ at $3$B. At $7$B, perceptive cross-direction becomes \emph{negative} (model specialises harder than at $3$B), and inspective single-direction breaks symmetry: $\texttt{orb\_l}\to\texttt{orb\_r}$ jumps ${+}27$pp but reverse stays at $-4$. Balanced \texttt{mv\_fwd}+\texttt{mv\_bwd} on $7$B saturates own directions to $100\%$ while cross-onto-rotation drops $-25$\,to\,$-27$pp: move and rotation are treated as separate action classes.

\textbf{The developmental emergence ordering works as a training curriculum.} The exogenous emergence ordering of Appendix~\ref{app:F-F} (Appendix~\ref{app:E-F10}) (gravity, inertia\,$\to$\,cohesion, permanence, containment\,$\to$\,support, solidity) operationalizes as a curriculum: training the same data in emergence order vs.\ random shuffle gives ${+}4.8$pp own-bench at $1$ epoch. Developmental ordering can be a useful prior on data scheduling.

\textbf{Five of the seven evaluation findings of Appendix~\ref{sec:assess:findings} replicate train-side.}\label{sec:enhance:triangulation} In detail: (i) the appearance/geometric scaling dissociation (Appendix~\ref{app:F-B}) replicates as ${+}21.9$ vs.\ ${+}12.4$pp on the perturbation probe after full-mix SFT, preserving the stronger improvement under appearance perturbations; (ii) temporal isolation appears in both the cohort and the training transfers (Appendix~\ref{app:E-F7}); (iii) the counterfactual-removal cost asymmetry is intrinsic rather than a coverage shortage (F$6$); (iv) chirality blindness manifests train-side as direction-asymmetric, capacity-dependent transfer (Appendices~\ref{app:E-F8-7B-scaling} and~\ref{app:E-F9}); (v) the exogenous developmental ordering operationalizes as a curriculum prior with ${+}4.8$pp gain (Appendix~\ref{app:E-F10}). The two non-replicated findings, the MoE active-parameter dependence (Appendix~\ref{app:F-E}) and scenario pair-mate independence (F-H), target base-model architecture and item-factor structure, axes outside the post-training intervention surface. The convergence of observation and intervention supports the cohort-side structural claims as representational rather than artifactual.

\subsection{Inverse-text per-cell own results}
\label{app:E-F3}

\begin{table}[!htbp]
\centering
\caption{Inverse-text SFT per-cell own ($\Delta$ vs.\ base 3B). Columns are inverse-dynamics cells by action type; the last column is the mean over the four own cells.}
\label{tab:E-F3}
\small
\begin{tabular}{@{}lrrrrr@{}}
\toprule
 & inspective & perceptive & navigative & manipulative & mean \\
\midrule
base 3B & $16$ & $27$ & $59$ & $62$ & $41$ \\
inverse-text SFT & $\mathbf{93}$ & $\mathbf{97}$ & $\mathbf{98}$ & $\mathbf{94}$ & $\mathbf{95.5}$ \\
$\Delta$ & ${+}77$ & ${+}70$ & ${+}39$ & ${+}32$ & ${+}54$ \\
\bottomrule
\end{tabular}
\end{table}

The matched forward-text run (text options, the same SFT protocol and $n{=}4{,}132$) has own-cell mean $44.3\%$; switching only direction from forward to inverse yields ${+}51$pp, driven by inverse direction.

\subsection{Narrow SFT data-scale margin}
\label{app:E-F4}

A $1{\times}$-data run ($n{=}4132$) vs.\ a $2{\times}$-data run ($n{=}8264$), all other config identical. Own-bench overall: $1{\times}$ $50.7\%$, $2{\times}$ $56.4\%$, $\Delta {=} {+}5.7$pp on $2{\times}$ data. Sample efficiency exhibits diminishing marginal returns under narrow SFT.

\subsection{Cross-action-type forward transfer matrix}
\label{app:E-F5}

\begin{table}[!htbp]
\centering
\caption{Cross-action-type forward transfer ($4{\times}4$). $n{\approx}1{,}020$ each, shared SFT protocol.}
\label{tab:E-F5}
\small
\setlength{\tabcolsep}{6pt}
\begin{tabular}{@{}lrrrr@{}}
\toprule
Source $\downarrow$ \,/\, Target $\rightarrow$ & Perc Fwd & Insp Fwd & Navi Fwd & Mani Fwd \\
\midrule
Base 3B (acc, \%) & $28.1$ & $28.1$ & $30.9$ & $21.9$ \\
\midrule
perceptive ($n{=}1036$) $\Delta$ & $\mathbf{{+}68.8}$ own & ${+}13.8$ & ${+}55.6$ & ${+}50.0$ \\
inspective ($n{=}1020$) $\Delta$ & ${+}12.9$ & $\mathbf{{+}64.9}$ own & ${+}57.3$ & ${+}46.9$ \\
navigative ($n{=}1020$) $\Delta$ & ${+}11.1$ & ${+}10.0$ & $\mathbf{{+}67.0}$ own & ${+}61.4$ \\
manipulative ($n{=}1020$) $\Delta$ & ${+}7.0$ & ${+}1.5$ & ${+}61.5$ & $\mathbf{{+}74.6}$ own \\
\bottomrule
\end{tabular}
\end{table}

\subsection{Within-action-type cross-reasoning-type}
\label{app:E-F6}

\begin{table}[!htbp]
\centering
\caption{(a) Within-action-type cross-comp transfer (inspective forward run).}
\label{tab:E-F6a}
\small
\setlength{\tabcolsep}{4pt}
\begin{tabular}{@{}lrrrrrr@{}}
\toprule
& Fwd & Sub & Rmv & Inv & Adj & Ord \\
\midrule
Base 3B (acc, \%) & $28.1$ & $26.7$ & $25.6$ & $15.2$ & $26.1$ & $27.0$ \\
SFT $\Delta$ & $\mathbf{{+}65}$ own & ${+}48$ & ${+}25$ & ${+}19$ & $0$ & ${-}1$ \\
\bottomrule
\end{tabular}
\end{table}

\begin{table}[!htbp]
\centering
\caption{(b) Sub within-action-type cross-comp, per-action-type split.}
\label{tab:E-F6b}
\small
\setlength{\tabcolsep}{4pt}
\begin{tabular}{@{}lrrrr@{}}
\toprule
& Fwd & Sub & Rmv & Inv \\
\midrule
Base 3B Insp (acc, \%) & $28.1$ & $26.7$ & $25.6$ & $15.2$ \\
Sub $\Delta$ Insp & ${+}60$ & $\mathbf{{+}66}$ own & ${-}1$ & ${+}48$ \\
Base 3B Mani (acc, \%) & $21.9$ & $20.8$ & $24.0$ & $62.5$ \\
Sub $\Delta$ Mani & ${+}71$ & $\mathbf{{+}74}$ own & $0$ & ${+}20$ \\
\bottomrule
\end{tabular}
\end{table}

\subsection{Temporal ordering and adjacency}
\label{app:E-F7}

One run trains temporal ordering across $4$ action types ($n{=}4{,}132$); a second trains temporal adjacency across $4$ action types ($n{=}4{,}132$). Both use the shared SFT optimization protocol (Appendix~\ref{app:E-recipes}). Across $4$ action types (averaged): the adjacency run gains ${+}30$\,to\,${+}35$pp on Adj (own) and $\approx 0$pp on Ord (cross); the ordering run gains $\approx 0$pp on Ord (own) and $\approx 0$pp on Adj (cross). Conversely, the four temporal subsets leave every other reasoning type within $3.3$ points of the base model under SFT (Table~\ref{tab:own-T2}).

\subsection{Rotation chirality at 7B}
\label{app:E-F8-7B-scaling}

\begin{table}[!htbp]
\centering
\caption{Balanced rotation: 3B vs.\ 7B (\texttt{arc\_up} interference reverses). Columns are inspective $\times$ forward cells by camera move.}
\label{tab:E-F8-D1}
\small
\setlength{\tabcolsep}{4pt}
\begin{tabular}{@{}lrrr@{}}
\toprule
 & orb\_l (own) & orb\_r (own) & arc\_up (orth) \\
\midrule
base 3B (\%) & $30.0$ & $20.0$ & $31.1$ \\
3B $\Delta$ & ${+}40.0$ & ${+}45.6$ & $\mathbf{{-}24.4}$ \\
7B $\Delta$ & $\mathbf{{+}50.0}$ & $\mathbf{{+}75.6}$ & $\mathbf{{+}43.3}$ \\
\bottomrule
\end{tabular}
\end{table}

\begin{table}[!htbp]
\centering
\caption{Perceptive single-direction runs: 3B vs.\ 7B. Columns are perceptive cells by camera move.}
\label{tab:E-F8-D2-perc}
\small
\setlength{\tabcolsep}{4pt}
\begin{tabular}{@{}lrr@{}}
\toprule
 & rot\_l & rot\_r \\
\midrule
base 3B (\%) & $27.8$ & $33.3$ \\
rot\_l only (7B) $\Delta$ & $\mathbf{{+}65.3}$ own & $-15.2$ cross \\
rot\_r only (7B) $\Delta$ & $-13.9$ cross & $\mathbf{{+}63.9}$ own \\
\bottomrule
\end{tabular}
\end{table}

\begin{table}[!htbp]
\centering
\caption{Inspective single-direction runs: 3B vs.\ 7B, asymmetric chirality sharing. Columns are inspective cells by camera move.}
\label{tab:E-F8-D2-insp}
\small
\setlength{\tabcolsep}{4pt}
\begin{tabular}{@{}lrrr@{}}
\toprule
 & orb\_l & orb\_r & arc\_up \\
\midrule
base 3B (\%) & $30.0$ & $20.0$ & $31.1$ \\
orb\_l only (7B) $\Delta$ & $\mathbf{{+}58.9}$ own & $\mathbf{{+}26.7}$ cross & ${+}36.7$ cross \\
orb\_r only (7B) $\Delta$ & ${-}4.4$ cross & $\mathbf{{+}72.2}$ own & ${+}37.8$ cross \\
\bottomrule
\end{tabular}
\end{table}

\subsection{Move chirality at 7B}
\label{app:E-F9}

\begin{table}[!htbp]
\centering
\caption{mv-balanced ($7$B) move chirality (perceptive forward). Columns are perceptive cells by camera move.}
\label{tab:E-F9}
\small
\begin{tabular}{@{}lrrrr@{}}
\toprule
 & mv\_fwd (own) & mv\_bwd (own) & rot\_l (cross) & rot\_r (cross) \\
\midrule
base 3B (\%) & $20.8$ & $19.4$ & $27.8$ & $33.3$ \\
mv-balanced 7B (\%) & $\mathbf{100.0}$ & $\mathbf{100.0}$ & $2.8$ & $5.6$ \\
$\Delta$ & ${+}79.2$ & ${+}80.6$ & $\mathbf{{-}25.0}$ & $\mathbf{{-}27.7}$ \\
\bottomrule
\end{tabular}
\end{table}

\subsection{Emergence ordering as curriculum}
\label{app:E-F10}

Train data: 7 exogenous principles (cohesion, solidity, gravity, support, inertia, containment, permanence; collision excluded), $1{,}820$ raw items $\times 4{\times}$ within-item permutation $= 7{,}280$ items. Both curriculum conditions use the shared SFT optimization protocol (Appendix~\ref{app:E-recipes}); Table~\ref{tab:E-F10} reports accuracy after the first epoch.

\begin{table}[!htbp]
\centering
\caption{Curriculum vs.\ shuffled at 1 epoch (own\_exo subset).}
\label{tab:E-F10}
\small
\begin{tabular}{@{}lr@{}}
\toprule
Variant & own\_exo subset acc \\
\midrule
$V_1$ (shuffled) & $33.2\%$ \\
$V_2$ (emergence-ordered) & $\mathbf{38.0\%}$ \\
$\Delta$ ($V_2 - V_1$) & $\mathbf{{+}4.8}$pp \\
\bottomrule
\end{tabular}
\end{table}

Order: gravity\,$\to$\,inertia\,$\to$\,cohesion\,$\to$\,containment\,$\to$\,permanence\,$\to$\,support\,$\to$\,solidity.

\clearpage
\FloatBarrier

\subsection{Downstream transfer: protocol, full results, and analyses}
\label{app:E-F11}

\textbf{Benchmark suite.} The downstream suite comprises $26$ external benchmarks: $22$ transition-relevant targets, SAT~\citep{ray2024sat}, MindCube~\citep{yin2025mindcube}, ViewSpatial~\citep{li2025viewspatial}, SpatialViz~\citep{wang2025spatialviz}, DSI-Bench~\citep{dsibench2025}, WM-ABench~\citep{huang2025wmabench}, WorldPrediction~\citep{worldprediction2025}, TemporalBench~\citep{temporalbench}, TOMATO~\citep{shangguan2025tomato}, TempCompass~\citep{liu2024tempcompass}, TVBench~\citep{cores2024tvbench}, BlackSwan~\citep{chinchure2025blackswan}, CLEVRER~\citep{yi2020clevrer}, InPhyRe~\citep{sreekumar2025inphyre}, MVP~\citep{krojer2025mvp}, ActionEQA~\citep{actioneqa2025}, Robo2VLM~\citep{chen2025robo2vlm}, ERQA~\citep{geminirobotics2025}, Cosmos~\citep{azzolini2025cosmosreason1}, PaiBench~\citep{zhou2025paibench}, CV-Bench~\citep{tong2024cambrian}, and BLINK~\citep{fu2024blink}, plus $4$ adjacent-capability probes (static perceptual grounding: 3DSRBench~\citep{sdrbench2025}, CoreCognition~\citep{li2025corecognition}; holistic video QA: PerceptionTest~\citep{patraucean2023perceptiontest}, EgoTaskQA~\citep{jia2022egotaskqa}). Every benchmark is evaluated under its official full protocol and split: SAT uses circular evaluation ($150$ questions $\times$ $2$ answer orders, both orders must be correct); TemporalBench uses dual-order binary accuracy (BA); MVP uses official paired accuracy; TVBench uses the official per-task macro average; DSI-Bench reports the official sample-wise aggregate over its four spatio-temporal flip variants; CLEVRER and ENACT use stratified samples ($12{,}000$ of $70{,}862$ per-option rows; not used in the main tables). Answers are generated greedily with \texttt{max\_new\_tokens}$=8$; unparseable outputs count as incorrect. A gain is called \emph{clear} above ${+}2$ points, roughly $5\times$ the observed run-to-run standard deviation of the full-mix RL configuration.

\begin{table}[p]
\centering
\rotatebox{-90}{\begin{minipage}{0.92\textheight}
\centering
\captionof{table}{\textbf{Best controlled-subset training for all $22$ improved benchmarks.} Each subset is defined by action type and reasoning family and contains approximately $2{,}000$ unique examples. Each column reports the subset with the largest gain on that benchmark, named in the last row; best SFT/GRPO accuracy is bold. Abbreviations as in Tables~\ref{tab:enhance-transfer} and~\ref{tab:enhance-bestcell}; ViewSp: ViewSpatial; TempComp: TempCompass. Benchmarks: SAT~\citep{ray2024sat}, ActionEQA~\citep{actioneqa2025}, TemporalBench-long~\citep{temporalbench}, InPhyRe~\citep{sreekumar2025inphyre}, Robo2VLM~\citep{chen2025robo2vlm}, WorldPrediction~\citep{worldprediction2025}, DSI-Bench~\citep{dsibench2025}, MindCube~\citep{yin2025mindcube}, CLEVRER~\citep{yi2020clevrer}, WM-ABench~\citep{huang2025wmabench}, MVP~\citep{krojer2025mvp}, ERQA~\citep{geminirobotics2025}, SpatialViz~\citep{wang2025spatialviz}, ViewSpatial~\citep{li2025viewspatial}, PaiBench~\citep{zhou2025paibench}, CV-Bench~\citep{tong2024cambrian}, Cosmos~\citep{azzolini2025cosmosreason1}, BLINK~\citep{fu2024blink}, TOMATO~\citep{shangguan2025tomato}, TVBench~\citep{cores2024tvbench}, BlackSwan~\citep{chinchure2025blackswan}, TempCompass~\citep{liu2024tempcompass}. The four remaining benchmarks and all subset results are reported below.}
\label{tab:enhance-bestcell-full}
\scriptsize
\setlength{\tabcolsep}{2pt}
\begin{adjustbox}{max width=\linewidth}
\begin{tabular}{@{}l*{22}{r}@{}}
\toprule
 & SAT & AEQA & TB-L & InPhyRe & R2VLM & WPred & DSI & MindCube & CLEVRER & WM-AB & MVP & ERQA & SpViz & ViewSp & PaiBench & CV-Bench & Cosmos & BLINK & TOMATO & TVBench & BSwan & TempComp \\
\midrule
Base & 37.3 & 43.4 & 16.6 & 22.0 & 33.5 & 34.8 & 34.9 & 37.2 & 58.2 & 31.0 & 24.4 & 33.8 & 27.7 & 35.3 & 48.3 & 69.3 & 76.7 & 46.2 & 31.6 & 36.5 & 61.8 & 67.7 \\
Subset SFT & 63.3 & 47.5 & 22.0 & 32.3 & 40.5 & 40.1 & 40.3 & 40.0 & 62.7 & \textbf{35.6} & 28.1 & 37.5 & \textbf{32.2} & 37.9 & 51.5 & 72.4 & \textbf{80.0} & 48.9 & 34.2 & 38.6 & 63.8 & 69.9 \\
Subset GRPO & \textbf{64.7} & \textbf{55.9} & \textbf{28.2} & \textbf{32.6} & \textbf{42.4} & \textbf{42.9} & \textbf{40.9} & \textbf{43.1} & \textbf{63.3} & 35.4 & \textbf{29.0} & \textbf{38.3} & 31.8 & \textbf{39.6} & \textbf{51.8} & \textbf{72.9} & 79.5 & \textbf{49.5} & \textbf{34.7} & \textbf{39.4} & \textbf{64.3} & \textbf{70.2} \\
\rowcolor{gray!12} $\Delta$ & +27.3 & +12.5 & +11.7 & +10.6 & +8.9 & +8.1 & +6.0 & +5.9 & +5.1 & +4.6 & +4.5 & +4.5 & +4.5 & +4.3 & +3.5 & +3.5 & +3.3 & +3.3 & +3.1 & +2.8 & +2.5 & +2.4 \\
\addlinespace[2pt]
Subset & \hd{perc}{causal} & \hd{perc}{causal} & \hd{mani}{cf} & \hd{insp}{cf} & \hd{insp}{cf} & \hd{mani}{causal} & \hd{perc}{causal} & \hd{perc}{temporal} & \hd{insp}{cf} & \hd{mani}{cf} & \hd{perc}{causal} & \hd{mani}{causal} & \hd{navi}{temporal} & \hd{mani}{cf} & \hd{insp}{causal} & \hd{insp}{causal} & \hd{insp}{causal} & \hd{exo}{physical} & \hd{insp}{temporal} & \hd{perc}{causal} & \hd{mani}{causal} & \hd{mani}{causal} \\
\bottomrule
\end{tabular}
\end{adjustbox}
\end{minipage}}
\end{table}

\textbf{RL optimization.} The RL stage uses Group Relative Policy Optimization~\citep{shao2024deepseekmath}, which estimates advantages by comparing each sampled response to a group of responses under the same prompt, with no separate value network. We run GRPO via verl~\citep{sheng2024verl} + vLLM on Qwen2.5-VL-3B initialized from a world modeling SFT checkpoint. Shared config: KL-to-reference \texttt{kl\_loss\_coef} $0.005$ (low-variance estimator), lr $1{\times}10^{-6}$, entropy coefficient $0$, train batch $16$ prompts $\times\,8$ rollouts (mini-batch $8$), max prompt $1{,}536$ tokens, max response $8$ tokens (single-letter answers), $300$ optimization steps with checkpoints every $50$; the reported checkpoint is step $150$ (validation accuracy saturates over steps $50$--$150$).

\textbf{RL reward.} To mitigate answer-symbol bias in multiple-choice prediction~\citep{zheng2024mcq,pezeshkpour2023order,xue2024symbolbinding}, we combine answer correctness with an anti-bias penalty:
\begin{equation*}
\mathrm{reward}(\text{rollout}_i) = \mathrm{letter\_match}_i \;-\; \lambda \cdot \mathrm{KL}\!\left(\hat{P}_i \,\|\, \mathrm{uniform}\right),
\end{equation*}
where $\mathrm{letter\_match}_i$ is $1$ if the predicted answer letter matches the gold letter and $0$ otherwise. The empirical distribution $\hat{P}_i$ is computed over the $W{=}64$ most recent predicted letters at the time rollout $i$ is scored, including rollout $i$ itself; $\mathrm{uniform}$ assigns probability $0.25$ to each of A/B/C/D, and $\lambda{=}0.1$. The window is updated for each scored rollout, allowing the penalty to vary within a GRPO group. The uniform target reflects the answer-letter balance of the training data. All reported RL experiments use this reward.

\textbf{Transfer-fingerprint normalizations.} Table~\ref{tab:fingerprint-norms} reports the fingerprint analysis of \S\ref{sec:enhance:org} under four treatments of benchmark difficulty. The ordering (same reasoning family $>$ same action type $\approx$ unrelated) is invariant.

\begin{table}[!htbp]
\centering
\caption{Mean pairwise Pearson correlation of the $11$ cells' $26$-dimensional transfer profiles, grouped by what the cell pair shares, under four normalizations of the per-benchmark gains.}
\label{tab:fingerprint-norms}
\small
\begin{tabular}{@{}lrrr@{}}
\toprule
Normalization & same reasoning family & same action type & neither \\
\midrule
raw $\Delta$ (points) & $+0.71$ & $+0.53$ & $+0.42$ \\
relative $\Delta/\text{base}$ & $+0.65$ & $+0.49$ & $+0.42$ \\
headroom $\Delta/(100-\text{base})$ & $+0.70$ & $+0.46$ & $+0.36$ \\
per-benchmark $z$-score & $+0.32$ & $-0.22$ & $-0.23$ \\
\bottomrule
\end{tabular}
\end{table}

Figure~\ref{fig:fingerprint-grouping} visualizes the analysis as the full $11{\times}11$ correlation matrices. The left panel shows the raw profiles: every cell pair is positively correlated, including pairs sharing neither axis, which is the shared primitive in correlational form. The middle and right panels display the identical residual matrix (after the per-benchmark $z$-score) under the two candidate orderings. Ordered by action type, no coherent structure appears along the diagonal; reordered by reasoning family, the remaining positive correlation concentrates into the diagonal family blocks (causal, counterfactual, temporal). The structure is revealed purely by reordering, so the family organization resides in the data rather than in the display. The exogenous-physical cell is visible as its own singleton: within the causal-adjacent region it correlates weakly with the agentic causal cells, consistent with its negative transfer to endogenous transitions discussed below.

\paragraph{Common component and dose--response across subsets.}\label{app:E-F11-common}
Two analyses connect the transfer of the $11$ controlled subsets to the capability they learn on WOVEN (\S\ref{sec:enhance:substitution}). Both use the $22$ benchmarks with per-configuration tables in this appendix (the $18$ improved benchmarks reported there and the $4$ static-perception and general video QA benchmarks) and the WOVEN in-distribution gains of Table~\ref{tab:own-T1}. \emph{Common component.} A subset's transfer profile is its vector of gains over the base model across the benchmarks. All $55$ pairwise Pearson correlations between the $11$ profiles are positive (mean $0.56$, range $0.11$--$0.92$; Fig.~\ref{fig:fingerprint-grouping}, left). For an entrywise-positive correlation matrix, the Perron--Frobenius theorem guarantees a leading eigenvector whose components all share one sign, that is, a single factor on which every subset loads positively; here its eigenvalue is $6.7$ of $11$. Equivalently, the first principal component of the profile matrix explains $75\%$ of the variance of the raw profiles ($54\%$ after centering each benchmark, $42\%$ after standardizing each benchmark), and its scores correlate with the subsets' WOVEN gains at $|r|{=}0.92$--$0.93$ under all three normalizations. The subsets therefore improve largely the same benchmarks, and the shared part of their transfer tracks how much each improves WOVEN. \emph{Dose--response.} Across the $11$ subsets, the WOVEN in-distribution gain after SFT predicts the mean external gain: Pearson $r{=}0.86$ ($p{=}0.0007$), Spearman $\rho{=}0.86$; $r{=}0.82$ with the post-GRPO WOVEN gain, $0.82$ against the number of benchmarks improved by more than $2$ points, $0.84$--$0.92$ under leave-one-subset-out, and positive within the temporal ($0.68$, $n{=}4$) and non-temporal ($0.79$, $n{=}7$) subsets separately, so the relation is not driven by the difference between these two groups. Specificity complements the dose--response: the benchmarks that improve are exactly the transition-related ones, and none of the $4$ static-perception and general video QA benchmarks gains more than $2$ points under any subset or under the full training set.

\textbf{Complementary reasoning operations.} Two within-WOVEN cases show that supervision of one reasoning operation does not necessarily teach another. (i) Temporal adjacency training does not produce temporal ordering nor vice versa (Appendix~\ref{app:E-F7}). (ii) The exogenous-physical cell, whose queries never condition on an agent's action, attains the largest state-perturbation robustness gain of any cell (${+}29.9$) yet transfers negatively to the two benchmarks built on endogenous, agent-driven transitions (WorldPrediction ${-}10.2$, ActionEQA ${-}1.7$): exogenous supervision does not teach endogenous transitions.

\textbf{InPhyRe subset anomaly.} A subset of InPhyRe (the ``irregular'' scenes) deliberately constructs fictional, real-world-violating physics (e.g.\ objects swap colors after collision) and marks the real-world-correct answer wrong; base models therefore score far below chance on these subsets ($0.0\%$ on \texttt{color\_constancy}). Post-training moves several subsets violently in both directions (\texttt{color\_constancy} $0.0 \to 84.2$; \texttt{off\_the\_wall\_regular} $91.0 \to 0.3$), so the aggregate InPhyRe gain of ${+}13.0$ nets over large opposing subset swings; the per-category table below gives the full picture, and we flag this benchmark's aggregate as protocol-sensitive.

\textbf{Per-benchmark, per-category results.} The tables below give, for each suite benchmark, accuracy by official category for the base model and all $12$ training configurations (columns: \texttt{base}; \texttt{full-mix}, the full training set; then the $11$ controlled subsets). Each configuration column reports its selected checkpoint (\S\ref{sec:enhance:setup}); category rows come from the same checkpoint as that configuration's overall score.

\begin{table}[!htbp]
\centering
\caption{SAT (circular): accuracy (\%) by official category. Columns: training configurations; rows: categories.}
\scriptsize
\setlength{\tabcolsep}{2.5pt}
\begin{adjustbox}{max width=\textwidth}
\begin{tabular}{@{}l*{13}{r}@{}}
\toprule
Category & \hd{}{base} & \hd{}{full-mix} & \hd{perc}{causal} & \hd{perc}{temporal} & \hd{insp}{causal} & \hd{insp}{cf} & \hd{insp}{temporal} & \hd{navi}{causal} & \hd{navi}{temporal} & \hd{mani}{causal} & \hd{mani}{cf} & \hd{exo}{physical} & \hd{exo}{temporal} \\
\midrule
Overall & \textbf{37.3} & \textbf{61.3} & \textbf{64.7} & \textbf{46.0} & \textbf{63.3} & \textbf{56.7} & \textbf{42.7} & \textbf{54.0} & \textbf{40.7} & \textbf{51.3} & \textbf{53.3} & \textbf{45.3} & \textbf{42.7} \\
\midrule
action\_conseq & 62.2 & 73.0 & 73.0 & 64.9 & 81.1 & 73.0 & 59.5 & 54.1 & 64.9 & 54.1 & 67.6 & 59.5 & 56.8 \\
ego\_movement & 34.8 & 87.0 & 100.0 & 43.5 & 78.3 & 52.2 & 60.9 & 82.6 & 52.2 & 78.3 & 73.9 & 73.9 & 30.4 \\
goal\_aim & 35.3 & 79.4 & 73.5 & 50.0 & 64.7 & 67.6 & 44.1 & 55.9 & 38.2 & 61.8 & 61.8 & 41.2 & 64.7 \\
obj\_movement & 8.7 & 26.1 & 30.4 & 17.4 & 39.1 & 34.8 & 8.7 & 21.7 & 4.3 & 13.0 & 17.4 & 13.0 & 4.3 \\
perspective & 33.3 & 36.4 & 45.5 & 42.4 & 48.5 & 45.5 & 33.3 & 54.5 & 33.3 & 45.5 & 39.4 & 36.4 & 39.4 \\
\bottomrule
\end{tabular}
\end{adjustbox}
\end{table}

\begin{table}[!htbp]
\centering
\caption{ActionEQA: accuracy (\%) by official category. Columns: training configurations; rows: categories.}
\scriptsize
\setlength{\tabcolsep}{2.5pt}
\begin{adjustbox}{max width=\textwidth}
\begin{tabular}{@{}l*{13}{r}@{}}
\toprule
Category & \hd{}{base} & \hd{}{full-mix} & \hd{perc}{causal} & \hd{perc}{temporal} & \hd{insp}{causal} & \hd{insp}{cf} & \hd{insp}{temporal} & \hd{navi}{causal} & \hd{navi}{temporal} & \hd{mani}{causal} & \hd{mani}{cf} & \hd{exo}{physical} & \hd{exo}{temporal} \\
\midrule
Overall & \textbf{43.4} & \textbf{54.4} & \textbf{55.9} & \textbf{51.4} & \textbf{54.3} & \textbf{55.9} & \textbf{45.2} & \textbf{47.8} & \textbf{51.4} & \textbf{55.0} & \textbf{53.2} & \textbf{41.7} & \textbf{47.4} \\
\midrule
bridge\_high\_forward & 25.0 & 49.4 & 56.4 & 49.6 & 52.3 & 52.5 & 37.1 & 33.3 & 50.5 & 40.5 & 40.4 & 31.2 & 40.1 \\
bridge\_high\_inverse & 37.0 & 48.7 & 45.9 & 46.2 & 39.7 & 45.0 & 44.4 & 40.2 & 50.3 & 57.8 & 50.5 & 41.5 & 42.2 \\
droid\_high\_forward & 36.8 & 47.0 & 40.2 & 40.8 & 53.5 & 52.4 & 34.0 & 43.6 & 36.5 & 36.5 & 43.1 & 26.1 & 41.9 \\
droid\_high\_inverse & 64.9 & 68.0 & 61.2 & 66.9 & 62.9 & 66.0 & 62.0 & 62.9 & 62.3 & 64.0 & 65.7 & 61.8 & 60.3 \\
rt1\_high\_forward & 62.0 & 59.1 & 76.4 & 57.5 & 74.4 & 67.4 & 53.5 & 64.5 & 55.1 & 73.9 & 67.2 & 42.9 & 57.5 \\
rt1\_high\_inverse & 59.3 & 64.7 & 58.9 & 54.4 & 58.6 & 65.4 & 48.8 & 63.1 & 53.9 & 62.7 & 65.0 & 56.6 & 54.4 \\
\bottomrule
\end{tabular}
\end{adjustbox}
\end{table}

\begin{table}[!htbp]
\centering
\caption{TemporalBench-long (BA): accuracy (\%) by official category. Columns: training configurations; rows: categories.}
\scriptsize
\setlength{\tabcolsep}{2.5pt}
\begin{adjustbox}{max width=\textwidth}
\begin{tabular}{@{}l*{13}{r}@{}}
\toprule
Category & \hd{}{base} & \hd{}{full-mix} & \hd{perc}{causal} & \hd{perc}{temporal} & \hd{insp}{causal} & \hd{insp}{cf} & \hd{insp}{temporal} & \hd{navi}{causal} & \hd{navi}{temporal} & \hd{mani}{causal} & \hd{mani}{cf} & \hd{exo}{physical} & \hd{exo}{temporal} \\
\midrule
Overall & \textbf{16.6} & \textbf{26.8} & \textbf{17.7} & \textbf{25.7} & \textbf{15.5} & \textbf{20.3} & \textbf{11.8} & \textbf{14.9} & \textbf{24.9} & \textbf{19.5} & \textbf{28.2} & \textbf{22.2} & \textbf{22.3} \\
\midrule
ActivityNet & 17.5 & 34.1 & 18.9 & 30.4 & 17.5 & 21.0 & 12.6 & 15.8 & 27.3 & 22.2 & 29.6 & 24.3 & 24.2 \\
COIN & 16.4 & 26.5 & 18.2 & 25.3 & 16.6 & 21.2 & 10.9 & 13.8 & 23.7 & 20.4 & 29.0 & 23.3 & 21.5 \\
Charades & 15.8 & 15.8 & 19.0 & 20.2 & 13.7 & 16.0 & 10.3 & 13.7 & 21.6 & 19.9 & 26.2 & 19.9 & 20.5 \\
EgoExo4D & 16.7 & 28.2 & 16.6 & 27.4 & 12.8 & 20.3 & 13.7 & 14.3 & 27.0 & 16.0 & 28.8 & 21.3 & 23.3 \\
FineGym & 14.9 & 19.3 & 14.9 & 14.6 & 19.3 & 21.8 & 8.0 & 21.8 & 19.3 & 21.8 & 21.8 & 17.9 & 19.3 \\
\bottomrule
\end{tabular}
\end{adjustbox}
\end{table}

\begin{table}[!htbp]
\centering
\caption{InPhyRe: accuracy (\%) by official category. Columns: training configurations; rows: categories.}
\scriptsize
\setlength{\tabcolsep}{2.5pt}
\begin{adjustbox}{max width=\textwidth}
\begin{tabular}{@{}l*{13}{r}@{}}
\toprule
Category & \hd{}{base} & \hd{}{full-mix} & \hd{perc}{causal} & \hd{perc}{temporal} & \hd{insp}{causal} & \hd{insp}{cf} & \hd{insp}{temporal} & \hd{navi}{causal} & \hd{navi}{temporal} & \hd{mani}{causal} & \hd{mani}{cf} & \hd{exo}{physical} & \hd{exo}{temporal} \\
\midrule
Overall & \textbf{22.0} & \textbf{34.9} & \textbf{29.1} & \textbf{29.7} & \textbf{29.9} & \textbf{32.6} & \textbf{26.8} & \textbf{25.9} & \textbf{29.3} & \textbf{28.3} & \textbf{29.9} & \textbf{30.6} & \textbf{31.6} \\
\midrule
ang\_mom\_cons & 29.2 & 73.2 & 37.4 & 45.3 & 51.7 & 57.8 & 39.2 & 32.0 & 46.5 & 15.2 & 40.5 & 62.1 & 49.3 \\
color\_constancy & 0.0 & 84.2 & 3.7 & 5.3 & 30.1 & 3.0 & 5.5 & 3.9 & 18.4 & 0.2 & 10.7 & 16.1 & 20.2 \\
color\_lin\_mom\_cons & 2.9 & 0.6 & 5.1 & 3.0 & 3.3 & 8.4 & 4.5 & 4.3 & 4.0 & 14.6 & 3.4 & 5.0 & 4.8 \\
color\_passthrough & 2.9 & 0.8 & 4.8 & 2.9 & 3.2 & 9.7 & 4.0 & 4.0 & 4.2 & 14.6 & 3.5 & 5.3 & 5.0 \\
lin\_mom\_cons & 0.1 & 0.0 & 2.4 & 0.5 & 0.1 & 9.9 & 1.1 & 1.0 & 0.8 & 14.6 & 0.3 & 3.3 & 2.5 \\
lin\_mom\_cons\_regular & 5.7 & 99.8 & 64.7 & 69.7 & 79.8 & 69.7 & 66.8 & 52.6 & 78.2 & 46.1 & 71.0 & 25.4 & 63.8 \\
off\_the\_wall & 8.9 & 23.2 & 1.6 & 34.7 & 45.9 & 41.6 & 45.5 & 20.2 & 41.2 & 2.4 & 26.7 & 25.9 & 52.2 \\
off\_the\_wall\_regular & 91.0 & 0.2 & 96.8 & 63.0 & 25.1 & 38.0 & 45.0 & 68.7 & 42.1 & 97.5 & 63.9 & 72.8 & 43.9 \\
size\_bias & 0.3 & 14.6 & 0.2 & 2.2 & 0.2 & 0.8 & 4.6 & 5.2 & 0.5 & 12.8 & 0.8 & 0.0 & 2.6 \\
size\_bias\_regular & 79.2 & 52.8 & 75.3 & 71.1 & 59.5 & 87.4 & 52.2 & 67.3 & 57.4 & 65.8 & 79.1 & 90.5 & 72.0 \\
\bottomrule
\end{tabular}
\end{adjustbox}
\end{table}

\begin{table}[!htbp]
\centering
\caption{Robo2VLM: accuracy (\%) by official category. Columns: training configurations; rows: categories.}
\scriptsize
\setlength{\tabcolsep}{2.5pt}
\begin{adjustbox}{max width=\textwidth}
\begin{tabular}{@{}l*{13}{r}@{}}
\toprule
Category & \hd{}{base} & \hd{}{full-mix} & \hd{perc}{causal} & \hd{perc}{temporal} & \hd{insp}{causal} & \hd{insp}{cf} & \hd{insp}{temporal} & \hd{navi}{causal} & \hd{navi}{temporal} & \hd{mani}{causal} & \hd{mani}{cf} & \hd{exo}{physical} & \hd{exo}{temporal} \\
\midrule
Overall & \textbf{33.5} & \textbf{31.8} & \textbf{39.1} & \textbf{34.2} & \textbf{39.1} & \textbf{42.4} & \textbf{37.7} & \textbf{38.4} & \textbf{34.2} & \textbf{37.2} & \textbf{41.6} & \textbf{35.0} & \textbf{33.9} \\
\midrule
robo2vlm & 33.5 & 31.8 & 39.1 & 34.2 & 39.1 & 42.4 & 37.7 & 38.4 & 34.2 & 37.2 & 41.6 & 35.0 & 33.9 \\
\bottomrule
\end{tabular}
\end{adjustbox}
\end{table}

\begin{table}[!htbp]
\centering
\caption{WorldPrediction-WM: accuracy (\%) by official category. Columns: training configurations; rows: categories.}
\scriptsize
\setlength{\tabcolsep}{2.5pt}
\begin{adjustbox}{max width=\textwidth}
\begin{tabular}{@{}l*{13}{r}@{}}
\toprule
Category & \hd{}{base} & \hd{}{full-mix} & \hd{perc}{causal} & \hd{perc}{temporal} & \hd{insp}{causal} & \hd{insp}{cf} & \hd{insp}{temporal} & \hd{navi}{causal} & \hd{navi}{temporal} & \hd{mani}{causal} & \hd{mani}{cf} & \hd{exo}{physical} & \hd{exo}{temporal} \\
\midrule
Overall & \textbf{34.8} & \textbf{41.5} & \textbf{37.2} & \textbf{35.4} & \textbf{39.9} & \textbf{29.7} & \textbf{31.4} & \textbf{38.9} & \textbf{29.7} & \textbf{42.9} & \textbf{42.5} & \textbf{24.6} & \textbf{32.7} \\
\midrule
COIN & 29.7 & 35.2 & 30.5 & 33.1 & 34.3 & 28.0 & 31.4 & 29.2 & 28.4 & 31.8 & 29.7 & 16.9 & 28.8 \\
CrossTask & 30.3 & 32.1 & 28.4 & 27.5 & 33.0 & 24.8 & 25.7 & 35.8 & 25.7 & 33.0 & 34.9 & 21.1 & 24.8 \\
EPIC-KITCHENS-100 & 48.2 & 57.0 & 54.4 & 49.7 & 55.4 & 37.3 & 39.9 & 55.4 & 37.8 & 59.6 & 65.3 & 33.2 & 44.6 \\
EgoExo4D & 36.7 & 43.8 & 32.0 & 34.4 & 39.8 & 30.5 & 32.0 & 40.6 & 31.2 & 52.3 & 40.6 & 23.4 & 33.6 \\
IKEAASM & 27.7 & 36.5 & 36.5 & 27.7 & 34.0 & 25.8 & 24.5 & 34.0 & 23.3 & 38.4 & 40.9 & 28.9 & 28.9 \\
\bottomrule
\end{tabular}
\end{adjustbox}
\end{table}

\begin{table}[!htbp]
\centering
\caption{DSI-Bench: accuracy (\%) by official category. Columns: training configurations; rows: categories.}
\scriptsize
\setlength{\tabcolsep}{2.5pt}
\begin{adjustbox}{max width=\textwidth}
\begin{tabular}{@{}l*{13}{r}@{}}
\toprule
Category & \hd{}{base} & \hd{}{full-mix} & \hd{perc}{causal} & \hd{perc}{temporal} & \hd{insp}{causal} & \hd{insp}{cf} & \hd{insp}{temporal} & \hd{navi}{causal} & \hd{navi}{temporal} & \hd{mani}{causal} & \hd{mani}{cf} & \hd{exo}{physical} & \hd{exo}{temporal} \\
\midrule
Overall & \textbf{34.9} & \textbf{35.2} & \textbf{40.9} & \textbf{36.7} & \textbf{34.0} & \textbf{38.9} & \textbf{36.5} & \textbf{39.5} & \textbf{32.2} & \textbf{38.4} & \textbf{39.0} & \textbf{37.8} & \textbf{33.9} \\
\midrule
cate0 & 34.7 & 35.5 & 47.3 & 40.0 & 36.9 & 45.9 & 41.4 & 48.8 & 35.7 & 45.9 & 47.0 & 44.1 & 39.1 \\
cate1 & 39.3 & 31.7 & 44.3 & 43.3 & 33.0 & 46.3 & 43.3 & 46.0 & 35.5 & 44.7 & 45.9 & 42.1 & 37.4 \\
cate2 & 26.3 & 40.9 & 33.5 & 23.8 & 32.2 & 30.1 & 20.2 & 25.2 & 25.3 & 25.4 & 27.1 & 24.9 & 26.4 \\
cate3 & 32.4 & 29.4 & 36.6 & 32.4 & 28.1 & 35.7 & 33.3 & 34.8 & 26.7 & 32.8 & 33.8 & 34.1 & 28.6 \\
cate4 & 48.0 & 65.9 & 53.8 & 51.6 & 62.1 & 34.8 & 47.5 & 50.7 & 47.5 & 52.2 & 48.2 & 53.6 & 50.9 \\
cate5 & 22.9 & 31.2 & 30.0 & 23.8 & 32.4 & 25.6 & 26.5 & 25.9 & 30.9 & 27.9 & 25.0 & 28.2 & 25.9 \\
\bottomrule
\end{tabular}
\end{adjustbox}
\end{table}

\begin{table}[!htbp]
\centering
\caption{DSI-Bench: accuracy (\%) by spatio-temporal variant. Columns: training configurations; rows: variants.}
\scriptsize
\setlength{\tabcolsep}{2.5pt}
\begin{adjustbox}{max width=\textwidth}
\begin{tabular}{@{}l*{13}{r}@{}}
\toprule
Variant & \hd{}{base} & \hd{}{full-mix} & \hd{perc}{causal} & \hd{perc}{temporal} & \hd{insp}{causal} & \hd{insp}{cf} & \hd{insp}{temporal} & \hd{navi}{causal} & \hd{navi}{temporal} & \hd{mani}{causal} & \hd{mani}{cf} & \hd{exo}{physical} & \hd{exo}{temporal} \\
\midrule
Overall & \textbf{34.9} & \textbf{35.2} & \textbf{40.9} & \textbf{36.7} & \textbf{34.0} & \textbf{38.9} & \textbf{36.5} & \textbf{39.5} & \textbf{32.2} & \textbf{38.4} & \textbf{39.0} & \textbf{37.8} & \textbf{33.9} \\
\midrule
std & 42.3 & 31.1 & 42.0 & 48.0 & 42.6 & 49.6 & 42.4 & 46.5 & 39.1 & 48.3 & 49.1 & 50.0 & 41.0 \\
hflip & 41.8 & 30.9 & 41.6 & 46.9 & 42.3 & 49.9 & 41.5 & 46.5 & 38.3 & 48.3 & 49.0 & 48.7 & 40.3 \\
reverse & 28.0 & 39.2 & 39.5 & 25.6 & 25.7 & 27.9 & 31.0 & 32.3 & 25.6 & 28.3 & 28.7 & 26.5 & 27.4 \\
reverse\_hflip & 27.4 & 39.6 & 40.4 & 26.5 & 25.5 & 28.4 & 31.1 & 32.9 & 25.8 & 28.7 & 29.2 & 26.2 & 27.0 \\
\bottomrule
\end{tabular}
\end{adjustbox}
\end{table}

\begin{table}[!htbp]
\centering
\caption{MindCube: accuracy (\%) by official category. Columns: training configurations; rows: categories.}
\scriptsize
\setlength{\tabcolsep}{2.5pt}
\begin{adjustbox}{max width=\textwidth}
\begin{tabular}{@{}l*{13}{r}@{}}
\toprule
Category & \hd{}{base} & \hd{}{full-mix} & \hd{perc}{causal} & \hd{perc}{temporal} & \hd{insp}{causal} & \hd{insp}{cf} & \hd{insp}{temporal} & \hd{navi}{causal} & \hd{navi}{temporal} & \hd{mani}{causal} & \hd{mani}{cf} & \hd{exo}{physical} & \hd{exo}{temporal} \\
\midrule
Overall & \textbf{37.2} & \textbf{48.6} & \textbf{31.4} & \textbf{43.1} & \textbf{38.3} & \textbf{41.4} & \textbf{35.7} & \textbf{32.2} & \textbf{40.4} & \textbf{33.8} & \textbf{39.0} & \textbf{41.4} & \textbf{41.0} \\
\midrule
linear/O-O/mw/self & 51.5 & 75.7 & 40.4 & 65.4 & 42.6 & 53.7 & 52.9 & 41.2 & 55.1 & 50.0 & 47.1 & 60.3 & 52.9 \\
linear/P-O/mw/level2 & 55.6 & 44.4 & 55.6 & 55.6 & 55.6 & 66.7 & 44.4 & 11.1 & 77.8 & 11.1 & 55.6 & 55.6 & 55.6 \\
linear/P-P/mw/self & 41.0 & 58.1 & 33.3 & 41.9 & 49.5 & 45.7 & 33.3 & 32.4 & 49.5 & 33.3 & 48.6 & 46.7 & 48.6 \\
perp/O-O/mw/level2 & 33.3 & 33.3 & 33.3 & 27.8 & 38.9 & 27.8 & 38.9 & 27.8 & 33.3 & 33.3 & 38.9 & 27.8 & 44.4 \\
perp/O-O/mw/self & 21.7 & 27.4 & 21.7 & 21.7 & 29.7 & 25.1 & 22.3 & 24.6 & 20.0 & 20.6 & 21.7 & 21.7 & 24.0 \\
perp/P-O/mw/level1 & 50.0 & 75.0 & 75.0 & 75.0 & 50.0 & 75.0 & 50.0 & 50.0 & 50.0 & 50.0 & 100.0 & 50.0 & 50.0 \\
perp/P-O/mw/level2 & 25.0 & 75.0 & 25.0 & 75.0 & 25.0 & 50.0 & 25.0 & 75.0 & 50.0 & 50.0 & 75.0 & 50.0 & 25.0 \\
perp/P-O/mw/self & 18.2 & 18.2 & 19.5 & 18.2 & 19.5 & 22.1 & 26.0 & 16.9 & 20.8 & 15.6 & 23.4 & 24.7 & 27.3 \\
perp/P-O/rot/self & 33.0 & 29.5 & 32.5 & 34.0 & 34.0 & 30.5 & 33.5 & 35.0 & 34.5 & 32.5 & 29.5 & 35.5 & 35.5 \\
perp/P-O/seq/self & 50.3 & 87.6 & 41.8 & 71.2 & 48.4 & 58.8 & 50.3 & 32.7 & 55.6 & 48.4 & 54.9 & 55.6 & 50.3 \\
perp/P-P/mw/self & 40.8 & 44.4 & 25.4 & 44.4 & 40.2 & 50.9 & 30.2 & 36.1 & 44.4 & 32.0 & 45.6 & 45.6 & 47.3 \\
\bottomrule
\end{tabular}
\end{adjustbox}
\end{table}

\begin{table}[!htbp]
\centering
\caption{CLEVRER: accuracy (\%) by official category. Columns: training configurations; rows: categories.}
\scriptsize
\setlength{\tabcolsep}{2.5pt}
\begin{adjustbox}{max width=\textwidth}
\begin{tabular}{@{}l*{13}{r}@{}}
\toprule
Category & \hd{}{base} & \hd{}{full-mix} & \hd{perc}{causal} & \hd{perc}{temporal} & \hd{insp}{causal} & \hd{insp}{cf} & \hd{insp}{temporal} & \hd{navi}{causal} & \hd{navi}{temporal} & \hd{mani}{causal} & \hd{mani}{cf} & \hd{exo}{physical} & \hd{exo}{temporal} \\
\midrule
Overall & \textbf{58.2} & \textbf{56.8} & \textbf{62.0} & \textbf{61.7} & \textbf{58.9} & \textbf{63.3} & \textbf{58.8} & \textbf{57.8} & \textbf{58.7} & \textbf{55.2} & \textbf{58.7} & \textbf{56.7} & \textbf{62.6} \\
\midrule
counterfactual & 52.3 & 51.1 & 53.6 & 52.8 & 52.3 & 56.0 & 52.4 & 51.0 & 51.4 & 50.1 & 54.9 & 50.6 & 54.1 \\
explanatory & 49.5 & 51.2 & 50.6 & 54.7 & 51.5 & 56.1 & 52.3 & 47.9 & 51.4 & 45.7 & 51.8 & 49.8 & 56.8 \\
predictive & 72.8 & 68.2 & 81.7 & 77.5 & 73.0 & 77.7 & 71.6 & 74.5 & 73.3 & 69.8 & 69.3 & 69.7 & 76.8 \\
\bottomrule
\end{tabular}
\end{adjustbox}
\end{table}

\begin{table}[!htbp]
\centering
\caption{WM-ABench: accuracy (\%) by official category. Columns: training configurations; rows: categories.}
\scriptsize
\setlength{\tabcolsep}{2.5pt}
\begin{adjustbox}{max width=\textwidth}
\begin{tabular}{@{}l*{13}{r}@{}}
\toprule
Category & \hd{}{base} & \hd{}{full-mix} & \hd{perc}{causal} & \hd{perc}{temporal} & \hd{insp}{causal} & \hd{insp}{cf} & \hd{insp}{temporal} & \hd{navi}{causal} & \hd{navi}{temporal} & \hd{mani}{causal} & \hd{mani}{cf} & \hd{exo}{physical} & \hd{exo}{temporal} \\
\midrule
Overall & \textbf{31.0} & \textbf{39.8} & \textbf{34.2} & \textbf{33.3} & \textbf{33.7} & \textbf{34.2} & \textbf{30.9} & \textbf{34.0} & \textbf{30.3} & \textbf{34.6} & \textbf{35.6} & \textbf{32.3} & \textbf{30.2} \\
\midrule
Perc/Motion/MotDir\_maniskill\_subtask & 22.0 & 46.0 & 24.0 & 20.0 & 37.0 & 24.0 & 18.0 & 21.0 & 21.0 & 20.0 & 26.0 & 28.0 & 22.0 \\
Perc/Motion/MotDir\_maniskill\_towards & 23.0 & 11.0 & 13.0 & 16.0 & 30.0 & 16.0 & 8.0 & 26.0 & 13.0 & 32.0 & 24.0 & 19.0 & 11.0 \\
Perc/Motion/MotDir\_tdw & 28.0 & 42.0 & 30.0 & 25.0 & 35.0 & 30.0 & 19.0 & 38.0 & 25.0 & 34.0 & 37.0 & 24.0 & 22.0 \\
Perc/Motion/MotDir\_tdw\_subtask & 12.0 & 31.0 & 21.0 & 16.0 & 23.0 & 19.0 & 10.0 & 23.0 & 12.0 & 23.0 & 25.0 & 22.0 & 11.0 \\
Perc/Motion/MotSpd\_maniskill & 26.0 & 29.0 & 20.0 & 25.0 & 25.0 & 25.0 & 24.0 & 27.0 & 24.0 & 26.0 & 25.0 & 23.0 & 27.0 \\
Perc/Motion/MotSpd\_tdw & 35.0 & 36.0 & 37.0 & 43.0 & 42.0 & 34.0 & 33.0 & 35.0 & 39.0 & 49.0 & 36.0 & 41.0 & 36.0 \\
\bottomrule
\end{tabular}
\end{adjustbox}
\end{table}

\begin{table}[!htbp]
\centering
\caption{MVP: accuracy (\%) by official category. Columns: training configurations; rows: categories.}
\scriptsize
\setlength{\tabcolsep}{2.5pt}
\begin{adjustbox}{max width=\textwidth}
\begin{tabular}{@{}l*{13}{r}@{}}
\toprule
Category & \hd{}{base} & \hd{}{full-mix} & \hd{perc}{causal} & \hd{perc}{temporal} & \hd{insp}{causal} & \hd{insp}{cf} & \hd{insp}{temporal} & \hd{navi}{causal} & \hd{navi}{temporal} & \hd{mani}{causal} & \hd{mani}{cf} & \hd{exo}{physical} & \hd{exo}{temporal} \\
\midrule
Overall & \textbf{24.4} & \textbf{26.2} & \textbf{28.9} & \textbf{21.6} & \textbf{27.1} & \textbf{24.8} & \textbf{22.8} & \textbf{25.7} & \textbf{22.1} & \textbf{28.4} & \textbf{26.0} & \textbf{25.7} & \textbf{20.4} \\
\midrule
human\_object\_interactions & 22.6 & 25.6 & 25.7 & 19.1 & 24.5 & 22.2 & 21.2 & 24.1 & 19.9 & 26.2 & 25.3 & 23.3 & 20.9 \\
intuitive\_physics & 17.0 & 14.7 & 18.4 & 16.4 & 16.3 & 18.1 & 16.5 & 17.5 & 14.3 & 16.8 & 16.4 & 15.1 & 16.5 \\
robot\_object\_interactions & 25.8 & 29.6 & 32.4 & 19.8 & 33.3 & 28.7 & 22.2 & 28.2 & 23.4 & 34.4 & 27.3 & 28.4 & 13.7 \\
temporal\_reasoning & 32.2 & 35.1 & 39.3 & 30.9 & 34.2 & 30.2 & 31.4 & 33.0 & 30.8 & 36.1 & 35.1 & 35.9 & 30.7 \\
\bottomrule
\end{tabular}
\end{adjustbox}
\end{table}

\begin{table}[!htbp]
\centering
\caption{ERQA: accuracy (\%) by official category. Columns: training configurations; rows: categories.}
\scriptsize
\setlength{\tabcolsep}{2.5pt}
\begin{adjustbox}{max width=\textwidth}
\begin{tabular}{@{}l*{13}{r}@{}}
\toprule
Category & \hd{}{base} & \hd{}{full-mix} & \hd{perc}{causal} & \hd{perc}{temporal} & \hd{insp}{causal} & \hd{insp}{cf} & \hd{insp}{temporal} & \hd{navi}{causal} & \hd{navi}{temporal} & \hd{mani}{causal} & \hd{mani}{cf} & \hd{exo}{physical} & \hd{exo}{temporal} \\
\midrule
Overall & \textbf{33.8} & \textbf{37.5} & \textbf{35.8} & \textbf{35.2} & \textbf{37.0} & \textbf{34.8} & \textbf{35.5} & \textbf{34.5} & \textbf{37.5} & \textbf{38.2} & \textbf{34.8} & \textbf{36.2} & \textbf{32.8} \\
\midrule
Action Reasoning & 40.3 & 38.9 & 37.5 & 41.7 & 37.5 & 34.7 & 40.3 & 34.7 & 34.7 & 45.8 & 40.3 & 40.3 & 34.7 \\
Multi-view Reasoning & 27.0 & 13.5 & 32.4 & 29.7 & 24.3 & 21.6 & 18.9 & 24.3 & 29.7 & 18.9 & 18.9 & 35.1 & 16.2 \\
Other & 14.3 & 35.7 & 28.6 & 14.3 & 35.7 & 28.6 & 28.6 & 28.6 & 28.6 & 28.6 & 35.7 & 14.3 & 7.1 \\
Pointing & 35.3 & 38.2 & 35.3 & 38.2 & 38.2 & 29.4 & 41.2 & 47.1 & 44.1 & 44.1 & 29.4 & 35.3 & 38.2 \\
Spatial Reasoning & 29.8 & 36.9 & 33.3 & 38.1 & 44.0 & 35.7 & 32.1 & 28.6 & 34.5 & 35.7 & 25.0 & 34.5 & 32.1 \\
State Estimation & 40.0 & 50.9 & 36.4 & 40.0 & 40.0 & 40.0 & 43.6 & 45.5 & 49.1 & 34.5 & 43.6 & 47.3 & 49.1 \\
Task Reasoning & 42.1 & 47.4 & 50.0 & 26.3 & 42.1 & 47.4 & 42.1 & 44.7 & 42.1 & 52.6 & 52.6 & 39.5 & 31.6 \\
Trajectory Reasoning & 28.8 & 33.3 & 31.8 & 31.8 & 28.8 & 33.3 & 31.8 & 27.3 & 34.8 & 37.9 & 34.8 & 28.8 & 30.3 \\
\bottomrule
\end{tabular}
\end{adjustbox}
\end{table}

\begin{table}[!htbp]
\centering
\caption{SpatialViz: accuracy (\%) by official category. Columns: training configurations; rows: categories.}
\scriptsize
\setlength{\tabcolsep}{2.5pt}
\begin{adjustbox}{max width=\textwidth}
\begin{tabular}{@{}l*{13}{r}@{}}
\toprule
Category & \hd{}{base} & \hd{}{full-mix} & \hd{perc}{causal} & \hd{perc}{temporal} & \hd{insp}{causal} & \hd{insp}{cf} & \hd{insp}{temporal} & \hd{navi}{causal} & \hd{navi}{temporal} & \hd{mani}{causal} & \hd{mani}{cf} & \hd{exo}{physical} & \hd{exo}{temporal} \\
\midrule
Overall & \textbf{27.7} & \textbf{31.1} & \textbf{29.0} & \textbf{30.0} & \textbf{30.3} & \textbf{28.6} & \textbf{30.8} & \textbf{30.7} & \textbf{32.2} & \textbf{31.6} & \textbf{28.6} & \textbf{30.8} & \textbf{31.6} \\
\midrule
MentalAnimation/ArrowMoving & 21.2 & 33.8 & 30.0 & 23.8 & 21.2 & 23.8 & 36.2 & 27.5 & 30.0 & 33.8 & 33.8 & 22.5 & 28.8 \\
MentalAnimation/BlockMoving & 43.8 & 31.2 & 45.0 & 40.0 & 38.8 & 30.0 & 38.8 & 41.2 & 40.0 & 38.8 & 32.5 & 36.2 & 38.8 \\
MentalAnimation/MechanicalSystem & 52.5 & 32.5 & 61.2 & 45.0 & 50.0 & 53.8 & 46.2 & 56.2 & 51.2 & 48.8 & 38.8 & 50.0 & 53.8 \\
MentalFolding/CubeReconstruction & 32.5 & 30.8 & 23.3 & 21.7 & 32.5 & 24.2 & 31.7 & 25.0 & 30.8 & 30.0 & 25.8 & 26.7 & 23.3 \\
MentalFolding/CubeUnfolding & 28.3 & 25.0 & 24.2 & 31.7 & 25.0 & 25.0 & 27.5 & 30.0 & 30.8 & 29.2 & 28.3 & 30.8 & 32.5 \\
MentalFolding/PaperFolding & 20.8 & 29.2 & 22.5 & 27.5 & 22.5 & 26.7 & 19.2 & 36.7 & 30.0 & 35.8 & 30.0 & 30.8 & 23.3 \\
\bottomrule
\end{tabular}
\end{adjustbox}
\end{table}

\begin{table}[!htbp]
\centering
\caption{ViewSpatial: accuracy (\%) by official category. Columns: training configurations; rows: categories.}
\scriptsize
\setlength{\tabcolsep}{2.5pt}
\begin{adjustbox}{max width=\textwidth}
\begin{tabular}{@{}l*{13}{r}@{}}
\toprule
Category & \hd{}{base} & \hd{}{full-mix} & \hd{perc}{causal} & \hd{perc}{temporal} & \hd{insp}{causal} & \hd{insp}{cf} & \hd{insp}{temporal} & \hd{navi}{causal} & \hd{navi}{temporal} & \hd{mani}{causal} & \hd{mani}{cf} & \hd{exo}{physical} & \hd{exo}{temporal} \\
\midrule
Overall & \textbf{35.3} & \textbf{34.1} & \textbf{38.1} & \textbf{38.3} & \textbf{37.2} & \textbf{37.2} & \textbf{32.7} & \textbf{36.6} & \textbf{31.6} & \textbf{37.0} & \textbf{39.6} & \textbf{37.4} & \textbf{35.8} \\
\midrule
Camera perspective - Object View Orientation & 29.7 & 28.1 & 34.6 & 30.5 & 29.4 & 27.9 & 28.2 & 28.0 & 30.2 & 28.0 & 26.2 & 31.2 & 25.8 \\
Camera perspective - Relative Direction & 42.9 & 43.3 & 43.8 & 43.2 & 44.1 & 41.4 & 40.7 & 44.2 & 41.3 & 43.1 & 44.9 & 44.6 & 40.7 \\
Person perspective - Object View Orientation & 42.4 & 35.4 & 50.6 & 49.2 & 47.5 & 54.1 & 36.8 & 48.6 & 29.8 & 50.9 & 59.1 & 45.8 & 47.0 \\
Person perspective - Relative Direction & 28.6 & 28.3 & 30.3 & 34.2 & 28.7 & 34.0 & 27.2 & 28.3 & 25.4 & 32.1 & 34.6 & 30.3 & 33.8 \\
Person perspective - Scene Simulation Relative Direction & 26.8 & 28.1 & 26.7 & 30.5 & 30.1 & 26.1 & 24.4 & 27.7 & 23.6 & 26.6 & 29.4 & 29.5 & 28.5 \\
\bottomrule
\end{tabular}
\end{adjustbox}
\end{table}

\begin{table}[!htbp]
\centering
\caption{PaiBench: accuracy (\%) by official category. Columns: training configurations; rows: categories.}
\scriptsize
\setlength{\tabcolsep}{2.5pt}
\begin{adjustbox}{max width=\textwidth}
\begin{tabular}{@{}l*{13}{r}@{}}
\toprule
Category & \hd{}{base} & \hd{}{full-mix} & \hd{perc}{causal} & \hd{perc}{temporal} & \hd{insp}{causal} & \hd{insp}{cf} & \hd{insp}{temporal} & \hd{navi}{causal} & \hd{navi}{temporal} & \hd{mani}{causal} & \hd{mani}{cf} & \hd{exo}{physical} & \hd{exo}{temporal} \\
\midrule
Overall & \textbf{48.3} & \textbf{50.2} & \textbf{50.3} & \textbf{50.2} & \textbf{51.8} & \textbf{49.6} & \textbf{50.2} & \textbf{48.7} & \textbf{49.8} & \textbf{49.5} & \textbf{48.5} & \textbf{48.0} & \textbf{50.1} \\
\midrule
Common Sense/Physical world & 43.4 & 47.3 & 42.0 & 46.0 & 46.0 & 41.6 & 49.6 & 43.4 & 48.2 & 43.4 & 39.4 & 42.5 & 50.0 \\
Common Sense/Space: Affordance & 18.8 & 43.8 & 37.5 & 31.2 & 25.0 & 25.0 & 31.2 & 25.0 & 18.8 & 31.2 & 25.0 & 31.2 & 25.0 \\
Common Sense/Space: Environment & 68.2 & 59.1 & 77.3 & 72.7 & 63.6 & 59.1 & 72.7 & 77.3 & 81.8 & 68.2 & 72.7 & 77.3 & 77.3 \\
Common Sense/Space: Plausibility & 23.5 & 41.2 & 17.6 & 70.6 & 17.6 & 29.4 & 52.9 & 23.5 & 41.2 & 29.4 & 17.6 & 23.5 & 47.1 \\
Common Sense/Space: Relationship & 72.0 & 68.0 & 60.0 & 60.0 & 72.0 & 56.0 & 76.0 & 60.0 & 72.0 & 68.0 & 60.0 & 68.0 & 68.0 \\
Common Sense/Time: Actions & 51.0 & 47.6 & 52.4 & 50.3 & 55.8 & 51.0 & 49.7 & 47.6 & 50.3 & 50.3 & 46.9 & 44.2 & 51.0 \\
\bottomrule
\end{tabular}
\end{adjustbox}
\end{table}

\begin{table}[!htbp]
\centering
\caption{CV-Bench: accuracy (\%) by official category. Columns: training configurations; rows: categories.}
\scriptsize
\setlength{\tabcolsep}{2.5pt}
\begin{adjustbox}{max width=\textwidth}
\begin{tabular}{@{}l*{13}{r}@{}}
\toprule
Category & \hd{}{base} & \hd{}{full-mix} & \hd{perc}{causal} & \hd{perc}{temporal} & \hd{insp}{causal} & \hd{insp}{cf} & \hd{insp}{temporal} & \hd{navi}{causal} & \hd{navi}{temporal} & \hd{mani}{causal} & \hd{mani}{cf} & \hd{exo}{physical} & \hd{exo}{temporal} \\
\midrule
Overall & \textbf{69.3} & \textbf{72.4} & \textbf{72.3} & \textbf{70.4} & \textbf{72.9} & \textbf{70.5} & \textbf{68.0} & \textbf{72.1} & \textbf{68.2} & \textbf{69.5} & \textbf{71.4} & \textbf{68.7} & \textbf{69.1} \\
\midrule
2D/Count/ADE20K & 50.9 & 55.8 & 54.4 & 56.4 & 53.5 & 53.2 & 52.6 & 50.3 & 52.0 & 51.5 & 51.5 & 49.7 & 54.1 \\
2D/Count/COCO & 71.7 & 76.0 & 71.3 & 72.0 & 72.4 & 72.2 & 71.1 & 70.6 & 70.6 & 70.6 & 71.3 & 69.7 & 70.4 \\
2D/Relation/ADE20K & 69.4 & 72.9 & 75.9 & 71.1 & 79.7 & 72.5 & 70.8 & 75.6 & 71.8 & 72.2 & 77.0 & 68.4 & 72.9 \\
2D/Relation/COCO & 76.0 & 77.4 & 79.4 & 77.2 & 83.0 & 76.3 & 74.7 & 80.5 & 78.0 & 74.9 & 80.5 & 75.8 & 77.7 \\
3D/Depth/Omni3D & 75.5 & 77.5 & 78.2 & 78.3 & 78.0 & 75.7 & 75.5 & 80.3 & 76.5 & 78.2 & 80.2 & 72.8 & 79.2 \\
3D/Distance/Omni3D & 67.8 & 70.7 & 71.3 & 64.7 & 69.7 & 69.7 & 61.8 & 70.5 & 59.5 & 65.7 & 66.0 & 70.7 & 59.7 \\
\bottomrule
\end{tabular}
\end{adjustbox}
\end{table}

\begin{table}[!htbp]
\centering
\caption{Cosmos: accuracy (\%) by official category. Columns: training configurations; rows: categories.}
\scriptsize
\setlength{\tabcolsep}{2.5pt}
\begin{adjustbox}{max width=\textwidth}
\begin{tabular}{@{}l*{13}{r}@{}}
\toprule
Category & \hd{}{base} & \hd{}{full-mix} & \hd{perc}{causal} & \hd{perc}{temporal} & \hd{insp}{causal} & \hd{insp}{cf} & \hd{insp}{temporal} & \hd{navi}{causal} & \hd{navi}{temporal} & \hd{mani}{causal} & \hd{mani}{cf} & \hd{exo}{physical} & \hd{exo}{temporal} \\
\midrule
Overall & \textbf{76.7} & \textbf{75.2} & \textbf{80.0} & \textbf{75.2} & \textbf{80.0} & \textbf{79.0} & \textbf{75.7} & \textbf{79.0} & \textbf{76.7} & \textbf{78.6} & \textbf{79.5} & \textbf{76.7} & \textbf{75.7} \\
\midrule
robofail & 67.0 & 63.0 & 72.0 & 63.0 & 71.0 & 69.0 & 65.0 & 72.0 & 67.0 & 70.0 & 71.0 & 70.0 & 64.0 \\
robovqa & 85.5 & 86.4 & 87.3 & 86.4 & 88.2 & 88.2 & 85.5 & 85.5 & 85.5 & 86.4 & 87.3 & 82.7 & 86.4 \\
\bottomrule
\end{tabular}
\end{adjustbox}
\end{table}

\begin{table}[!htbp]
\centering
\caption{BLINK: accuracy (\%) by official category. Columns: training configurations; rows: categories.}
\scriptsize
\setlength{\tabcolsep}{2.5pt}
\begin{adjustbox}{max width=\textwidth}
\begin{tabular}{@{}l*{13}{r}@{}}
\toprule
Category & \hd{}{base} & \hd{}{full-mix} & \hd{perc}{causal} & \hd{perc}{temporal} & \hd{insp}{causal} & \hd{insp}{cf} & \hd{insp}{temporal} & \hd{navi}{causal} & \hd{navi}{temporal} & \hd{mani}{causal} & \hd{mani}{cf} & \hd{exo}{physical} & \hd{exo}{temporal} \\
\midrule
Overall & \textbf{46.2} & \textbf{45.4} & \textbf{49.0} & \textbf{46.4} & \textbf{47.1} & \textbf{48.4} & \textbf{46.5} & \textbf{46.8} & \textbf{45.9} & \textbf{49.0} & \textbf{47.6} & \textbf{49.4} & \textbf{45.0} \\
\midrule
Art\_Style & 51.3 & 64.1 & 53.0 & 52.1 & 65.8 & 55.6 & 54.7 & 59.0 & 59.0 & 61.5 & 53.0 & 58.1 & 56.4 \\
Counting & 64.2 & 69.2 & 65.8 & 59.2 & 65.8 & 65.8 & 67.5 & 67.5 & 61.7 & 65.0 & 65.0 & 66.7 & 66.7 \\
Forensic\_Detection & 26.5 & 25.8 & 45.5 & 39.4 & 25.8 & 33.3 & 29.5 & 18.2 & 25.8 & 40.2 & 29.5 & 32.6 & 28.0 \\
Functional\_Correspondence & 24.6 & 23.1 & 29.2 & 24.6 & 26.9 & 26.9 & 31.5 & 29.2 & 25.4 & 27.7 & 27.7 & 26.2 & 23.8 \\
IQ\_Test & 28.0 & 29.3 & 32.0 & 30.0 & 28.7 & 28.0 & 29.3 & 27.3 & 28.0 & 33.3 & 25.3 & 29.3 & 27.3 \\
Jigsaw & 50.7 & 47.3 & 46.7 & 47.3 & 46.0 & 42.0 & 46.0 & 42.0 & 48.7 & 45.3 & 50.0 & 46.7 & 51.3 \\
\bottomrule
\end{tabular}
\end{adjustbox}
\end{table}

\begin{table}[!htbp]
\centering
\caption{TOMATO: accuracy (\%) by official category. Columns: training configurations; rows: categories.}
\scriptsize
\setlength{\tabcolsep}{2.5pt}
\begin{adjustbox}{max width=\textwidth}
\begin{tabular}{@{}l*{13}{r}@{}}
\toprule
Category & \hd{}{base} & \hd{}{full-mix} & \hd{perc}{causal} & \hd{perc}{temporal} & \hd{insp}{causal} & \hd{insp}{cf} & \hd{insp}{temporal} & \hd{navi}{causal} & \hd{navi}{temporal} & \hd{mani}{causal} & \hd{mani}{cf} & \hd{exo}{physical} & \hd{exo}{temporal} \\
\midrule
Overall & \textbf{31.6} & \textbf{36.1} & \textbf{29.2} & \textbf{33.2} & \textbf{34.3} & \textbf{30.4} & \textbf{34.7} & \textbf{30.7} & \textbf{34.7} & \textbf{30.1} & \textbf{31.7} & \textbf{32.3} & \textbf{34.7} \\
\midrule
count/human & 19.3 & 36.8 & 10.5 & 15.8 & 17.5 & 10.5 & 17.5 & 21.1 & 17.5 & 19.3 & 15.8 & 21.1 & 21.1 \\
count/object & 47.6 & 43.9 & 46.3 & 48.2 & 45.7 & 40.2 & 47.0 & 46.3 & 53.7 & 42.7 & 45.1 & 45.7 & 51.8 \\
count/simulated & 63.4 & 59.2 & 57.7 & 56.3 & 62.0 & 67.6 & 62.0 & 62.0 & 60.6 & 49.3 & 57.7 & 56.3 & 53.5 \\
direction/human & 21.8 & 26.3 & 20.2 & 24.8 & 24.0 & 23.7 & 26.7 & 24.4 & 26.7 & 22.5 & 27.9 & 23.3 & 26.0 \\
direction/object & 42.4 & 40.2 & 52.2 & 45.7 & 51.1 & 45.7 & 37.0 & 42.4 & 37.0 & 40.2 & 41.3 & 45.7 & 39.1 \\
direction/simulated & 28.6 & 24.5 & 26.5 & 36.7 & 28.6 & 38.8 & 28.6 & 24.5 & 32.7 & 24.5 & 30.6 & 32.7 & 24.5 \\
\bottomrule
\end{tabular}
\end{adjustbox}
\end{table}

\begin{table}[!htbp]
\centering
\caption{TVBench: accuracy (\%) by official category. Columns: training configurations; rows: categories.}
\scriptsize
\setlength{\tabcolsep}{2.5pt}
\begin{adjustbox}{max width=\textwidth}
\begin{tabular}{@{}l*{13}{r}@{}}
\toprule
Category & \hd{}{base} & \hd{}{full-mix} & \hd{perc}{causal} & \hd{perc}{temporal} & \hd{insp}{causal} & \hd{insp}{cf} & \hd{insp}{temporal} & \hd{navi}{causal} & \hd{navi}{temporal} & \hd{mani}{causal} & \hd{mani}{cf} & \hd{exo}{physical} & \hd{exo}{temporal} \\
\midrule
Overall & \textbf{36.5} & \textbf{37.4} & \textbf{39.4} & \textbf{35.8} & \textbf{36.4} & \textbf{38.4} & \textbf{36.7} & \textbf{37.1} & \textbf{35.0} & \textbf{36.3} & \textbf{38.7} & \textbf{35.7} & \textbf{35.4} \\
\midrule
tvbench\_action\_count & 0.0 & 0.6 & 1.9 & 1.7 & 0.2 & 1.5 & 20.7 & 1.3 & 0.6 & 0.2 & 2.4 & 0.7 & 7.6 \\
tvbench\_action\_localization & 36.2 & 38.1 & 35.6 & 36.2 & 35.0 & 35.6 & 34.4 & 34.4 & 35.6 & 32.5 & 35.0 & 31.9 & 35.6 \\
tvbench\_action\_seq & 62.0 & 62.9 & 59.0 & 63.4 & 62.7 & 63.6 & 60.4 & 59.3 & 61.6 & 60.9 & 62.2 & 61.3 & 62.2 \\
tvbench\_egocentric\_seq & 31.5 & 32.5 & 41.0 & 37.5 & 33.5 & 37.5 & 34.0 & 35.5 & 27.0 & 34.0 & 39.5 & 25.0 & 30.0 \\
tvbench\_moving\_direction & 24.6 & 34.1 & 36.6 & 22.8 & 29.7 & 31.0 & 22.4 & 26.3 & 29.3 & 25.4 & 25.9 & 28.4 & 25.9 \\
tvbench\_object\_count & 24.3 & 21.6 & 23.6 & 25.0 & 25.7 & 25.7 & 23.6 & 28.4 & 25.0 & 23.0 & 24.3 & 23.6 & 26.4 \\
tvbench\_object\_shuffle & 39.6 & 37.8 & 37.3 & 36.0 & 36.9 & 38.7 & 40.0 & 41.8 & 35.1 & 42.2 & 43.6 & 39.1 & 34.2 \\
tvbench\_scene\_transition & 78.9 & 80.5 & 80.0 & 73.5 & 73.5 & 78.9 & 71.9 & 78.9 & 69.2 & 79.5 & 78.9 & 81.1 & 71.4 \\
tvbench\_unexpected\_action & 31.7 & 28.0 & 39.0 & 25.6 & 30.5 & 32.9 & 23.2 & 28.0 & 31.7 & 29.3 & 36.6 & 30.5 & 25.6 \\
\bottomrule
\end{tabular}
\end{adjustbox}
\end{table}

\begin{table}[!htbp]
\centering
\caption{BlackSwan: accuracy (\%) by official category. Columns: training configurations; rows: categories.}
\scriptsize
\setlength{\tabcolsep}{2.5pt}
\begin{adjustbox}{max width=\textwidth}
\begin{tabular}{@{}l*{13}{r}@{}}
\toprule
Category & \hd{}{base} & \hd{}{full-mix} & \hd{perc}{causal} & \hd{perc}{temporal} & \hd{insp}{causal} & \hd{insp}{cf} & \hd{insp}{temporal} & \hd{navi}{causal} & \hd{navi}{temporal} & \hd{mani}{causal} & \hd{mani}{cf} & \hd{exo}{physical} & \hd{exo}{temporal} \\
\midrule
Overall & \textbf{61.8} & \textbf{65.6} & \textbf{60.6} & \textbf{60.5} & \textbf{62.7} & \textbf{60.8} & \textbf{62.7} & \textbf{62.5} & \textbf{62.4} & \textbf{64.2} & \textbf{59.6} & \textbf{59.5} & \textbf{62.0} \\
\midrule
Detective/easy & 64.0 & 66.9 & 63.8 & 63.4 & 66.3 & 63.0 & 65.3 & 67.6 & 64.8 & 65.1 & 63.4 & 62.8 & 59.4 \\
Detective/hard & 45.6 & 50.4 & 51.2 & 44.0 & 47.2 & 43.2 & 47.2 & 46.4 & 44.0 & 44.8 & 44.8 & 44.8 & 47.2 \\
Detective/medium & 55.6 & 59.5 & 56.6 & 54.1 & 58.9 & 57.0 & 55.3 & 53.9 & 55.3 & 59.1 & 54.9 & 54.5 & 56.4 \\
Reporter/easy & 77.9 & 80.5 & 68.7 & 75.4 & 77.4 & 75.9 & 79.5 & 77.9 & 80.5 & 79.5 & 71.3 & 70.8 & 81.0 \\
Reporter/hard & 62.4 & 66.7 & 59.8 & 64.6 & 55.0 & 57.7 & 64.6 & 61.9 & 64.6 & 67.2 & 56.1 & 61.4 & 69.3 \\
Reporter/medium & 64.9 & 70.6 & 61.3 & 61.3 & 64.9 & 63.7 & 66.1 & 65.7 & 65.3 & 68.5 & 62.5 & 60.1 & 66.1 \\
\bottomrule
\end{tabular}
\end{adjustbox}
\end{table}

\begin{table}[!htbp]
\centering
\caption{TempCompass: accuracy (\%) by official category. Columns: training configurations; rows: categories.}
\scriptsize
\setlength{\tabcolsep}{2.5pt}
\begin{adjustbox}{max width=\textwidth}
\begin{tabular}{@{}l*{13}{r}@{}}
\toprule
Category & \hd{}{base} & \hd{}{full-mix} & \hd{perc}{causal} & \hd{perc}{temporal} & \hd{insp}{causal} & \hd{insp}{cf} & \hd{insp}{temporal} & \hd{navi}{causal} & \hd{navi}{temporal} & \hd{mani}{causal} & \hd{mani}{cf} & \hd{exo}{physical} & \hd{exo}{temporal} \\
\midrule
Overall & \textbf{67.7} & \textbf{65.9} & \textbf{70.0} & \textbf{68.2} & \textbf{68.7} & \textbf{68.7} & \textbf{67.8} & \textbf{69.5} & \textbf{67.8} & \textbf{70.2} & \textbf{68.8} & \textbf{69.0} & \textbf{67.4} \\
\midrule
capmatch/action & 96.6 & 96.6 & 97.0 & 96.6 & 96.6 & 97.3 & 97.3 & 97.6 & 97.3 & 98.0 & 97.3 & 97.3 & 97.6 \\
capmatch/attribute\_change & 74.0 & 79.9 & 78.1 & 75.3 & 80.6 & 76.7 & 76.7 & 81.6 & 75.3 & 82.6 & 77.4 & 81.6 & 73.3 \\
capmatch/direction & 60.9 & 62.7 & 56.9 & 60.6 & 63.9 & 59.0 & 61.5 & 60.9 & 61.2 & 63.0 & 57.8 & 59.3 & 62.7 \\
capmatch/order & 82.7 & 81.0 & 86.3 & 82.7 & 84.7 & 84.3 & 79.0 & 85.0 & 77.7 & 86.0 & 83.7 & 86.3 & 75.0 \\
capmatch/speed & 62.2 & 61.9 & 68.7 & 61.5 & 59.1 & 67.0 & 61.5 & 64.6 & 62.9 & 68.0 & 67.4 & 63.2 & 57.7 \\
mc/action & 95.9 & 95.3 & 95.6 & 95.6 & 93.8 & 96.2 & 95.3 & 94.7 & 95.0 & 95.3 & 97.0 & 97.9 & 95.6 \\
\bottomrule
\end{tabular}
\end{adjustbox}
\end{table}

\begin{table}[!htbp]
\centering
\caption{PerceptionTest (val): accuracy (\%) by official category. Columns: training configurations; rows: categories.}
\scriptsize
\setlength{\tabcolsep}{2.5pt}
\begin{adjustbox}{max width=\textwidth}
\begin{tabular}{@{}l*{13}{r}@{}}
\toprule
Category & \hd{}{base} & \hd{}{full-mix} & \hd{perc}{causal} & \hd{perc}{temporal} & \hd{insp}{causal} & \hd{insp}{cf} & \hd{insp}{temporal} & \hd{navi}{causal} & \hd{navi}{temporal} & \hd{mani}{causal} & \hd{mani}{cf} & \hd{exo}{physical} & \hd{exo}{temporal} \\
\midrule
Overall & \textbf{62.9} & \textbf{58.1} & \textbf{63.7} & \textbf{61.8} & \textbf{60.6} & \textbf{63.2} & \textbf{60.4} & \textbf{62.8} & \textbf{60.4} & \textbf{62.6} & \textbf{62.4} & \textbf{63.7} & \textbf{60.5} \\
\midrule
abstraction & 54.6 & 54.6 & 55.1 & 53.7 & 52.5 & 54.5 & 53.4 & 54.1 & 53.6 & 53.7 & 53.6 & 55.0 & 53.7 \\
memory & 47.6 & 46.8 & 46.4 & 46.2 & 48.8 & 44.9 & 44.4 & 46.0 & 44.6 & 45.3 & 45.8 & 46.4 & 46.0 \\
physics & 58.4 & 50.7 & 58.9 & 57.8 & 53.7 & 58.8 & 55.8 & 58.0 & 56.4 & 58.6 & 56.6 & 59.5 & 55.6 \\
semantics & 78.0 & 71.8 & 79.7 & 76.2 & 78.0 & 79.0 & 74.7 & 79.1 & 73.7 & 78.1 & 79.8 & 79.2 & 74.5 \\
\bottomrule
\end{tabular}
\end{adjustbox}
\end{table}

\begin{table}[!htbp]
\centering
\caption{EgoTaskQA: accuracy (\%) by official category. Columns: training configurations; rows: categories.}
\scriptsize
\setlength{\tabcolsep}{2.5pt}
\begin{adjustbox}{max width=\textwidth}
\begin{tabular}{@{}l*{13}{r}@{}}
\toprule
Category & \hd{}{base} & \hd{}{full-mix} & \hd{perc}{causal} & \hd{perc}{temporal} & \hd{insp}{causal} & \hd{insp}{cf} & \hd{insp}{temporal} & \hd{navi}{causal} & \hd{navi}{temporal} & \hd{mani}{causal} & \hd{mani}{cf} & \hd{exo}{physical} & \hd{exo}{temporal} \\
\midrule
Overall & \textbf{81.5} & \textbf{74.1} & \textbf{77.8} & \textbf{81.3} & \textbf{79.8} & \textbf{81.6} & \textbf{80.2} & \textbf{78.3} & \textbf{80.7} & \textbf{74.2} & \textbf{81.6} & \textbf{82.6} & \textbf{80.9} \\
\midrule
egotaskqa & 81.5 & 74.1 & 77.8 & 81.3 & 79.8 & 81.6 & 80.2 & 78.3 & 80.7 & 74.2 & 81.6 & 82.6 & 80.9 \\
\bottomrule
\end{tabular}
\end{adjustbox}
\end{table}

\begin{table}[!htbp]
\centering
\caption{3DSRBench (circular): accuracy (\%) by official category. Columns: training configurations; rows: categories.}
\scriptsize
\setlength{\tabcolsep}{2.5pt}
\begin{adjustbox}{max width=\textwidth}
\begin{tabular}{@{}l*{13}{r}@{}}
\toprule
Category & \hd{}{base} & \hd{}{full-mix} & \hd{perc}{causal} & \hd{perc}{temporal} & \hd{insp}{causal} & \hd{insp}{cf} & \hd{insp}{temporal} & \hd{navi}{causal} & \hd{navi}{temporal} & \hd{mani}{causal} & \hd{mani}{cf} & \hd{exo}{physical} & \hd{exo}{temporal} \\
\midrule
Overall & \textbf{51.2} & \textbf{50.6} & \textbf{51.1} & \textbf{50.7} & \textbf{51.6} & \textbf{51.1} & \textbf{51.2} & \textbf{52.6} & \textbf{50.5} & \textbf{52.3} & \textbf{52.6} & \textbf{50.7} & \textbf{50.4} \\
\midrule
height\_higher & 53.6 & 50.7 & 53.6 & 54.7 & 52.2 & 53.3 & 52.6 & 49.4 & 50.9 & 51.7 & 54.6 & 55.0 & 50.7 \\
location\_above & 61.8 & 60.8 & 59.4 & 59.2 & 64.0 & 61.6 & 61.1 & 67.3 & 59.9 & 64.8 & 64.2 & 62.9 & 59.1 \\
location\_closer\_to\_camera & 71.8 & 71.5 & 73.5 & 72.9 & 74.8 & 70.0 & 71.3 & 71.6 & 74.0 & 70.6 & 70.9 & 70.4 & 72.6 \\
location\_next\_to & 51.3 & 56.6 & 53.7 & 48.4 & 54.0 & 51.3 & 55.5 & 62.5 & 49.7 & 57.2 & 57.5 & 51.3 & 52.4 \\
multi\_object\_closer\_to & 49.9 & 47.3 & 49.6 & 49.1 & 48.1 & 51.3 & 53.6 & 52.7 & 51.6 & 53.0 & 50.7 & 47.7 & 49.1 \\
multi\_object\_facing & 59.4 & 59.7 & 56.5 & 59.1 & 63.6 & 60.0 & 60.0 & 63.4 & 60.5 & 70.1 & 67.6 & 58.1 & 58.5 \\
multi\_object\_parallel & 52.1 & 51.2 & 52.2 & 52.2 & 50.6 & 53.1 & 52.5 & 53.7 & 51.5 & 48.8 & 52.7 & 52.2 & 50.6 \\
multi\_object\_same\_direction & 53.1 & 49.9 & 53.6 & 51.7 & 55.1 & 52.3 & 52.6 & 53.5 & 51.7 & 50.9 & 51.3 & 48.3 & 52.0 \\
multi\_object\_viewpoint\_towards\_object & 28.1 & 29.7 & 27.4 & 26.7 & 26.2 & 28.9 & 27.5 & 27.0 & 26.5 & 27.1 & 28.1 & 27.3 & 27.6 \\
orientation\_in\_front\_of & 56.5 & 50.9 & 54.7 & 58.6 & 50.0 & 57.4 & 53.8 & 53.3 & 56.4 & 56.4 & 58.6 & 61.0 & 56.5 \\
orientation\_on\_the\_left & 49.9 & 50.1 & 49.9 & 50.1 & 50.0 & 50.0 & 49.9 & 50.0 & 50.3 & 50.1 & 50.0 & 50.1 & 49.9 \\
orientation\_viewpoint & 33.8 & 34.8 & 35.8 & 33.0 & 35.9 & 32.8 & 34.1 & 37.5 & 31.9 & 37.1 & 34.8 & 31.0 & 34.3 \\
\bottomrule
\end{tabular}
\end{adjustbox}
\end{table}

\begin{table}[!htbp]
\centering
\caption{CoreCognition: accuracy (\%) by official category. Columns: training configurations; rows: categories.}
\scriptsize
\setlength{\tabcolsep}{2.5pt}
\begin{adjustbox}{max width=\textwidth}
\begin{tabular}{@{}l*{13}{r}@{}}
\toprule
Category & \hd{}{base} & \hd{}{full-mix} & \hd{perc}{causal} & \hd{perc}{temporal} & \hd{insp}{causal} & \hd{insp}{cf} & \hd{insp}{temporal} & \hd{navi}{causal} & \hd{navi}{temporal} & \hd{mani}{causal} & \hd{mani}{cf} & \hd{exo}{physical} & \hd{exo}{temporal} \\
\midrule
Overall & \textbf{59.8} & \textbf{53.2} & \textbf{57.4} & \textbf{57.8} & \textbf{52.6} & \textbf{57.7} & \textbf{57.5} & \textbf{56.0} & \textbf{58.3} & \textbf{57.9} & \textbf{56.5} & \textbf{58.8} & \textbf{57.4} \\
\midrule
boundary/Stage Sensorimotor & 76.4 & 74.2 & 76.4 & 76.4 & 79.8 & 78.7 & 76.4 & 78.7 & 73.0 & 73.0 & 71.9 & 78.7 & 71.9 \\
conservation/Stage Concrete Operational & 69.5 & 63.8 & 58.1 & 59.0 & 21.9 & 65.7 & 65.7 & 49.5 & 65.7 & 61.0 & 46.7 & 68.6 & 65.7 \\
continuity/Stage Sensorimotor & 67.2 & 60.9 & 64.1 & 67.2 & 67.2 & 64.1 & 60.9 & 62.5 & 62.5 & 64.1 & 64.1 & 67.2 & 62.5 \\
hierarchy/Stage Concrete Operational & 67.0 & 57.3 & 60.2 & 62.1 & 66.0 & 67.0 & 59.2 & 61.2 & 63.1 & 57.3 & 65.0 & 66.0 & 65.0 \\
intentionality/Stage Formal Operational & 76.1 & 60.6 & 76.1 & 75.2 & 71.6 & 72.5 & 77.1 & 74.3 & 76.1 & 74.3 & 70.6 & 70.6 & 76.1 \\
intuitivephysics/Stage Concrete Operational & 58.6 & 50.0 & 53.4 & 56.9 & 56.9 & 53.4 & 51.7 & 63.8 & 56.9 & 58.6 & 58.6 & 58.6 & 53.4 \\
\bottomrule
\end{tabular}
\end{adjustbox}
\end{table}

\subsection{In-domain substitution sweep: protocol and full tables}
\label{app:E-F12}

\textbf{Protocol.} Three external benchmarks with public training splits are paired with their best-matching WOVEN cell: SAT with \texttt{insp\_causal}, ActionEQA with \texttt{mani\_causal}, and CLEVRER with \texttt{insp\_cf}. For each share $p \in \{0,10,20,30,40,50,70,100\}\%$, the training set holds a fixed $2{,}000$-item budget: $2{,}000\cdot(1-p)$ items drawn from the benchmark's own training split in its native format, and $2{,}000\cdot p$ items from the WOVEN cell. Items are packed into $16$-row blocks that are answer-letter balanced within each source type, and the answer-letter sequence is identical across arms, so arms differ only in the substituted content. Training uses the shared SFT optimization protocol of Appendix~\ref{app:E-recipes} for $3$ epochs, with seeds $42/43/44$, followed by $300$ GRPO steps with the anti-bias reward. The fixed-budget datasets use the answer-balanced blocks described above without additional option-permutation augmentation. Evaluation uses each benchmark's own protocol (SAT scored circularly).

\textbf{Trend statistics.} An OLS fit of in-domain accuracy on the share over $p \le 50\%$ ($n{=}18$ runs per benchmark) gives slopes, per ${+}10\%$ share, of $-0.03 \pm 1.00$\,pp for SAT ($t(16){=}{-}0.03$, $p{=}0.98$), $-1.54 \pm 0.30$\,pp for ActionEQA ($t(16){=}{-}5.08$, $p{=}1.1\times10^{-4}$), and $-0.01 \pm 0.30$\,pp for CLEVRER ($t(16){=}{-}0.04$, $p{=}0.97$). Per-arm seed ranges are visible in Table~\ref{tab:g2-mix-full} (SAT carries the widest), so the sweep is read at the level of the fitted trend rather than of individual arms.

\begin{table}[h]
\centering
\caption{\textbf{In-domain substitution sweep, full per-seed results} (in-domain accuracy, \%). $p$ is the WOVEN share of the fixed $2{,}000$-item training budget; the $p{=}0$ row is the pure in-domain baseline.}
\label{tab:g2-mix-full}
\begin{adjustbox}{max width=\linewidth}
\begin{tabular}{r rrrr rrrr rrrr}
\toprule
& \multicolumn{4}{c}{SAT (circular)} & \multicolumn{4}{c}{ActionEQA} & \multicolumn{4}{c}{CLEVRER} \\
\cmidrule(lr){2-5}\cmidrule(lr){6-9}\cmidrule(lr){10-13}
$p$ (\%) & s42 & s43 & s44 & mean & s42 & s43 & s44 & mean & s42 & s43 & s44 & mean \\
\midrule
0 & 45.33 & 43.33 & 58.00 & \textbf{48.89} & 79.30 & 80.01 & 78.69 & \textbf{79.33} & 83.20 & 81.73 & 83.00 & \textbf{82.64} \\
10 & 45.33 & 50.67 & 49.33 & \textbf{48.44} & 77.60 & 77.55 & 78.82 & \textbf{77.99} & 82.47 & 81.53 & 82.60 & \textbf{82.20} \\
20 & 54.67 & 42.67 & 35.33 & \textbf{44.22} & 77.50 & 77.61 & 75.46 & \textbf{76.86} & 82.53 & 84.47 & 82.93 & \textbf{83.31} \\
30 & 37.33 & 54.00 & 38.00 & \textbf{43.11} & 79.51 & 79.98 & 78.42 & \textbf{79.30} & 83.60 & 78.67 & 82.33 & \textbf{81.53} \\
40 & 58.67 & 40.00 & 53.33 & \textbf{50.67} & 74.30 & 74.19 & 73.88 & \textbf{74.12} & 76.27 & 81.47 & 83.93 & \textbf{80.56} \\
50 & 44.67 & 46.67 & 51.33 & \textbf{47.56} & 67.66 & 70.84 & 72.55 & \textbf{70.35} & 85.00 & 84.80 & 81.93 & \textbf{83.91} \\
70 & 54.67 & 54.00 & 46.67 & \textbf{51.78} & 64.15 & 66.68 & 67.77 & \textbf{66.20} & 75.73 & 81.80 & 80.47 & \textbf{79.33} \\
100 & 58.00 & 60.00 & 56.67 & \textbf{58.22} & 46.54 & 46.51 & 46.38 & \textbf{46.47} & 56.80 & 54.40 & 54.40 & \textbf{55.20} \\
\bottomrule
\end{tabular}
\end{adjustbox}
\end{table}

\textbf{Deletion control.} To test whether preserved accuracy under substitution merely reflects saturation of the task-specific training budget, we ablate the WOVEN contribution at $p \in \{20,40,50\}\%$. Each deletion arm retains exactly the same task-specific subset as its substitution counterpart but omits the WOVEN examples, leaving $1{,}600$, $1{,}200$, or $1{,}000$ training items under the same optimization recipe. At $40\%$ and $50\%$, substitution outperforms deletion on all three benchmarks by $3.60$--$11.00$ percentage points (Table~\ref{tab:g2-deletion}); at $20\%$, it improves ActionEQA and CLEVRER but is $1.99$ points lower on SAT. This pattern supports an additional contribution from WOVEN supervision beyond simply discarding redundant task-specific data. In Fig.~\ref{fig:g2-mix}, the deletion curves connect these three ablation settings to the shared $p{=}0$ reference.

\begin{table}[h]
\centering
\caption{\textbf{Deletion ablation for in-domain substitution.} Accuracy (\%) after deleting the WOVEN portion (Del.) or retaining it (Sub.). $\Delta$ is substitution minus deletion in percentage points. Both arms retain the same task-specific subset.}
\label{tab:g2-deletion}
\begin{adjustbox}{max width=\linewidth}
\begin{tabular}{r rrr rrr rrr}
\toprule
& \multicolumn{3}{c}{SAT (circular)} & \multicolumn{3}{c}{ActionEQA} & \multicolumn{3}{c}{CLEVRER} \\
\cmidrule(lr){2-4}\cmidrule(lr){5-7}\cmidrule(lr){8-10}
$p$ (\%) & Del. & Sub. & $\Delta$ & Del. & Sub. & $\Delta$ & Del. & Sub. & $\Delta$ \\
\midrule
20 & 46.21 & 44.22 & $-1.99$ & 74.21 & 76.86 & $+2.65$ & 80.31 & 83.31 & $+3.00$ \\
40 & 39.67 & 50.67 & $+11.00$ & 68.41 & 74.12 & $+5.71$ & 76.96 & 80.56 & $+3.60$ \\
50 & 39.72 & 47.56 & $+7.84$ & 66.20 & 70.35 & $+4.15$ & 77.12 & 83.91 & $+6.79$ \\
\bottomrule
\end{tabular}
\end{adjustbox}
\end{table}

\subsection{Transfer at scale: 7B and 32B}
\label{app:E-F13}

The transfer of \S\ref{sec:enhance} persists when the same full-mix recipe is applied at larger scale. Table~\ref{tab:scale-transfer} reports zero-shot transfer gains over the same-scale base for Qwen2.5-VL-7B (SFT and GRPO stages, following the 3B recipe) and Qwen2.5-VL-32B (SFT stage). Gains of the same character as at 3B appear at both scales: egocentric spatial reasoning (SAT ${+}20.0$ at 7B, ${+}22.7$ at 32B), paired video perception (MVP), world-model probes (WM-ABench), and dynamic-scene inference (DSI), alongside single-scale gains on temporal, embodied, and intuitive-physics surfaces.

\begin{table}[h]
\centering
\footnotesize
\setlength{\tabcolsep}{2.5pt}
\caption{\textbf{Transfer at scale.} Zero-shot transfer gains (pp over the same-scale base) after WOVEN full-mix post-training of Qwen2.5-VL-7B and Qwen2.5-VL-32B. The recipe follows the 3B protocol; the 32B run is SFT-only; SAT is under circular evaluation. Abbreviations as in Tables~\ref{tab:enhance-transfer} and~\ref{tab:enhance-bestcell}; ViewSp: ViewSpatial. `---': not reported.}
\label{tab:scale-transfer}
\begin{tabular}{@{}l*{12}{c}@{}}
\toprule
 & SAT & DSI & MVP & WM-AB & TB-L & ViewSp & R2VLM & InPhyRe & ERQA & BSwan & CLEVRER & TVBench \\
\midrule
7B SFT & ${+}20.00$ & ${+}6.32$ & ${+}5.08$ & ${+}4.39$ & ${+}4.30$ & --- & ${+}3.24$ & --- & --- & --- & --- & --- \\
7B GRPO & ${+}20.00$ & ${+}5.96$ & ${+}4.49$ & ${+}4.83$ & ${+}5.34$ & ${+}2.10$ & ${+}3.54$ & --- & ${+}2.75$ & --- & ${+}2.19$ & ${+}2.04$ \\
32B SFT & ${+}22.67$ & --- & ${+}6.28$ & ${+}5.91$ & --- & ${+}4.94$ & --- & ${+}3.54$ & --- & ${+}2.29$ & --- & --- \\
\bottomrule
\end{tabular}
\end{table}

\subsection{Same-backbone comparison with video-SSL mid-training}
\label{app:E-F14}

Orca-4B~\citep{wang2026orca} mid-trains the Qwen3.5-4B backbone on latent next-state prediction over $125$K hours of video with $160$M event annotations. We post-train the \emph{same} Qwen3.5-4B backbone on WOVEN supervision alone: SFT on ${\approx}107$K QA pairs drawn from ${\approx}7$ hours of generated video, four orders of magnitude less video and a post-training-only budget. The comparison uses 3DSRBench, a benchmark on which Orca-4B reports results. On this backbone, WOVEN post-training with about four orders of magnitude less video yields a larger gain over the shared base than Orca-4B mid-training ($+6.7$ vs.\ $+4.0$ points; Table~\ref{tab:orca}).

\begin{table}[h]
\centering
\caption{\textbf{Same-backbone comparison with video-SSL mid-training.} Both models start from Qwen3.5-4B; gains in parentheses are over the shared base. Orca-4B: latent next-state mid-training ($125$K hours of video, $160$M event annotations). Ours: WOVEN post-training (SFT, ${\approx}107$K QA pairs, ${\approx}7$ hours of generated video).}
\label{tab:orca}
\begin{tabular}{lc}
\toprule
Model & 3DSRBench \\
\midrule
Qwen3.5-4B base & $48.1$ \\
Orca-4B & $52.1$ (${+}4.0$) \\
${+}$WOVEN (ours) & $\mathbf{54.8}$ (${+}6.7$) \\
\bottomrule
\end{tabular}
\end{table}

\FloatBarrier

\subsection{WOVEN results by supervision source and training stage}
\label{app:E-F15}

We report the complete WOVEN test results for the base Qwen2.5-VL-3B model, full-mix training, and all $11$ individual supervision subsets. Accuracies and error breakdowns were recomputed from model generations. The three test sets are in-distribution (ID; $n=6{,}228$), held-out scenes (OOD; $n=3{,}496$), and state perturbations (Pert; $n=1{,}116$); chance accuracy is $25\%$. Tables~\ref{tab:own-T1}--\ref{tab:own-E7c} have separate panels for supervised fine-tuning (SFT) and reinforcement learning (RL). Full-mix RL results are not available in this set of evaluations and are marked with dashes; the base model is repeated in both panels for comparison.

Configuration prefixes denote camera motion (\texttt{perc}), object inspection (\texttt{insp}), navigation (\texttt{navi}), object manipulation (\texttt{mani}), and passive physical events (\texttt{exo}). Suffixes identify causal, counterfactual (\texttt{cf}), temporal, or physical supervision, following the training configurations in Appendix~\ref{app:E-F11}. Unless a caption specifies otherwise, accuracies are percentages over all items in the named category, and overall accuracy is computed over the complete named test split. Error shares describe the distractors selected on incorrect answers; captions specify the relevant subset and any categories omitted from the display.

\textbf{Overall and cross-action gains.} All $11$ individual subsets improve overall ID accuracy over the $26.38\%$ base score: SFT accuracies range from $31.87\%$ to $50.75\%$, and RL accuracies from $32.45\%$ to $51.65\%$ (Table~\ref{tab:own-T1}). To distinguish these gains from improvement restricted to the trained action category, we also compare each source on the other four action categories. At both stages, $43$ of these $44$ comparisons improve over the base model: ten sources improve all four other action categories, and the remaining source improves three. Table~\ref{tab:own-T2} further identifies which reasoning skills improve under each source.

\FloatBarrier

\subsubsection{Accuracy by action, reasoning type, and test split}

\begin{table}[!htbp]
\centering
\caption{ID accuracy (\%) by action type ($n=6{,}228$). Overall covers the complete ID split. Panels report SFT and RL separately; --- denotes unavailable full-mix RL results.}
\label{tab:own-T1}
\begingroup
\footnotesize
\setlength{\tabcolsep}{3pt}
\renewcommand{\arraystretch}{1.04}
\begin{adjustbox}{max width=\linewidth}
\begin{tabular}{@{}lrrrrrr@{}}
\toprule
\multicolumn{7}{l}{\textbf{SFT}} \\
\addlinespace[2pt]
Config. & Perceptive & Inspective & Navigative & Manipulative & Exogenous & Overall \\
\midrule
base & 27.00 & 23.46 & 30.03 & 28.82 & 23.78 & 26.38 \\
\addlinespace[1pt]
\texttt{full-mix} & 90.71 & 90.49 & 90.02 & 95.31 & 79.43 & 89.29 \\
\texttt{perc\_causal} & 60.76 & 34.75 & 56.94 & 64.67 & 29.77 & 48.28 \\
\texttt{perc\_temporal} & 38.98 & 27.28 & 36.81 & 29.25 & 31.16 & 32.29 \\
\texttt{insp\_causal} & 37.07 & 62.28 & 56.34 & 58.85 & 30.30 & 49.97 \\
\texttt{insp\_cf} & 33.42 & 58.89 & 53.21 & 71.96 & 32.99 & 50.75 \\
\texttt{insp\_temporal} & 33.94 & 27.96 & 37.15 & 30.47 & 32.29 & 32.03 \\
\texttt{navi\_causal} & 34.11 & 34.14 & 60.33 & 78.04 & 31.25 & 46.56 \\
\texttt{navi\_temporal} & 35.42 & 27.78 & 37.59 & 29.08 & 31.16 & 31.87 \\
\texttt{mani\_causal} & 31.42 & 25.86 & 55.56 & 90.36 & 31.77 & 45.41 \\
\texttt{mani\_cf} & 30.99 & 26.54 & 56.68 & 92.53 & 34.55 & 46.63 \\
\texttt{exo\_physical} & 26.74 & 28.83 & 35.16 & 33.51 & 51.30 & 34.63 \\
\texttt{exo\_temporal} & 35.16 & 28.15 & 37.59 & 29.43 & 32.29 & 32.19 \\
\midrule
\addlinespace[3pt]
\multicolumn{7}{l}{\textbf{RL}} \\
\addlinespace[2pt]
Config. & Perceptive & Inspective & Navigative & Manipulative & Exogenous & Overall \\
\midrule
base & 27.00 & 23.46 & 30.03 & 28.82 & 23.78 & 26.38 \\
\addlinespace[1pt]
\texttt{full-mix} & --- & --- & --- & --- & --- & --- \\
\texttt{perc\_causal} & 59.98 & 36.91 & 56.94 & 68.58 & 28.91 & 49.26 \\
\texttt{perc\_temporal} & 37.93 & 28.40 & 38.02 & 30.12 & 32.20 & 32.96 \\
\texttt{insp\_causal} & 35.94 & 63.33 & 57.20 & 66.49 & 30.56 & 51.65 \\
\texttt{insp\_cf} & 34.29 & 61.98 & 54.86 & 71.53 & 30.38 & 51.46 \\
\texttt{insp\_temporal} & 33.42 & 26.85 & 41.23 & 30.03 & 32.99 & 32.45 \\
\texttt{navi\_causal} & 37.07 & 36.17 & 60.07 & 79.77 & 31.42 & 47.94 \\
\texttt{navi\_temporal} & 39.24 & 27.47 & 38.63 & 31.68 & 31.86 & 33.30 \\
\texttt{mani\_causal} & 30.90 & 29.20 & 56.25 & 90.19 & 32.03 & 46.32 \\
\texttt{mani\_cf} & 32.20 & 28.95 & 57.20 & 93.06 & 34.38 & 47.64 \\
\texttt{exo\_physical} & 25.78 & 29.20 & 36.72 & 38.45 & 51.22 & 35.74 \\
\texttt{exo\_temporal} & 34.11 & 28.95 & 40.10 & 31.16 & 32.20 & 32.98 \\
\bottomrule
\end{tabular}
\end{adjustbox}
\endgroup
\end{table}

\begin{table}[!htbp]
\centering
\caption{ID accuracy (\%) by reasoning type ($n=6{,}228$). Overall covers all eight reasoning types. Panels report SFT and RL separately; --- denotes unavailable full-mix RL results.}
\label{tab:own-T2}
\begingroup
\footnotesize
\setlength{\tabcolsep}{3pt}
\renewcommand{\arraystretch}{1.04}
\begin{adjustbox}{max width=\linewidth}
\begin{tabular}{@{}lrrrrrrrrr@{}}
\toprule
\multicolumn{10}{l}{\textbf{SFT}} \\
\addlinespace[2pt]
Config. & \shortstack{Forward\\dynamics} & \shortstack{Inverse\\dynamics} & \shortstack{CF\\substitution} & \shortstack{CF\\removal} & \shortstack{Outcome\\prediction} & \shortstack{Cued\\prediction} & \shortstack{Temporal\\ordering} & \shortstack{Temporal\\adjacency} & Overall \\
\midrule
base & 21.34 & 40.39 & 19.35 & 18.64 & 21.18 & 23.96 & 27.34 & 25.66 & 26.38 \\
\addlinespace[1pt]
\texttt{full-mix} & 97.09 & 94.80 & 94.80 & 96.95 & 80.90 & 83.33 & 92.77 & 69.66 & 89.29 \\
\texttt{perc\_causal} & 75.31 & 76.63 & 62.19 & 28.85 & 33.33 & 33.68 & 25.13 & 26.28 & 48.28 \\
\texttt{perc\_temporal} & 20.81 & 41.71 & 18.82 & 20.61 & 23.26 & 25.35 & 30.34 & 52.73 & 32.29 \\
\texttt{insp\_causal} & 73.72 & 75.84 & 74.19 & 41.22 & 32.64 & 34.03 & 25.40 & 25.75 & 49.97 \\
\texttt{insp\_cf} & 73.72 & 50.53 & 83.15 & 86.92 & 39.93 & 40.62 & 26.63 & 23.72 & 50.75 \\
\texttt{insp\_temporal} & 20.72 & 41.36 & 19.00 & 21.86 & 22.57 & 25.00 & 24.96 & 56.70 & 32.03 \\
\texttt{navi\_causal} & 64.90 & 64.90 & 59.68 & 49.64 & 29.86 & 39.93 & 26.46 & 27.95 & 46.56 \\
\texttt{navi\_temporal} & 21.25 & 42.50 & 17.20 & 20.43 & 22.57 & 20.83 & 25.93 & 55.82 & 31.87 \\
\texttt{mani\_causal} & 63.32 & 62.70 & 60.93 & 48.39 & 29.86 & 38.89 & 25.84 & 26.28 & 45.41 \\
\texttt{mani\_cf} & 63.23 & 54.76 & 68.10 & 57.35 & 38.54 & 42.71 & 27.25 & 28.48 & 46.63 \\
\texttt{exo\_physical} & 25.93 & 47.62 & 15.23 & 38.17 & 76.74 & 77.43 & 28.13 & 23.10 & 34.63 \\
\texttt{exo\_temporal} & 20.55 & 42.77 & 18.10 & 21.51 & 22.92 & 24.65 & 25.31 & 56.61 & 32.19 \\
\midrule
\addlinespace[3pt]
\multicolumn{10}{l}{\textbf{RL}} \\
\addlinespace[2pt]
Config. & \shortstack{Forward\\dynamics} & \shortstack{Inverse\\dynamics} & \shortstack{CF\\substitution} & \shortstack{CF\\removal} & \shortstack{Outcome\\prediction} & \shortstack{Cued\\prediction} & \shortstack{Temporal\\ordering} & \shortstack{Temporal\\adjacency} & Overall \\
\midrule
base & 21.34 & 40.39 & 19.35 & 18.64 & 21.18 & 23.96 & 27.34 & 25.66 & 26.38 \\
\addlinespace[1pt]
\texttt{full-mix} & --- & --- & --- & --- & --- & --- & --- & --- & --- \\
\texttt{perc\_causal} & 76.01 & 79.28 & 67.56 & 31.00 & 29.86 & 34.72 & 24.51 & 25.84 & 49.26 \\
\texttt{perc\_temporal} & 19.93 & 43.92 & 17.74 & 20.97 & 21.88 & 25.35 & 30.60 & 55.56 & 32.96 \\
\texttt{insp\_causal} & 75.22 & 76.10 & 83.51 & 44.80 & 33.33 & 32.64 & 25.93 & 26.54 & 51.65 \\
\texttt{insp\_cf} & 72.13 & 54.94 & 84.23 & 87.28 & 33.33 & 36.11 & 26.54 & 26.98 & 51.46 \\
\texttt{insp\_temporal} & 20.55 & 43.12 & 14.87 & 23.12 & 23.26 & 24.65 & 24.96 & 58.73 & 32.45 \\
\texttt{navi\_causal} & 66.31 & 70.55 & 60.39 & 47.49 & 29.51 & 39.24 & 25.40 & 30.51 & 47.94 \\
\texttt{navi\_temporal} & 20.99 & 48.85 & 17.38 & 22.94 & 23.61 & 22.57 & 23.81 & 57.67 & 33.30 \\
\texttt{mani\_causal} & 64.90 & 62.87 & 64.70 & 48.39 & 32.29 & 39.93 & 24.69 & 27.95 & 46.32 \\
\texttt{mani\_cf} & 66.14 & 55.03 & 70.25 & 60.57 & 38.89 & 43.40 & 26.63 & 28.57 & 47.64 \\
\texttt{exo\_physical} & 27.60 & 50.26 & 15.95 & 42.65 & 73.96 & 77.43 & 26.98 & 24.16 & 35.74 \\
\texttt{exo\_temporal} & 21.34 & 44.36 & 17.56 & 23.66 & 25.35 & 23.26 & 26.54 & 56.26 & 32.98 \\
\bottomrule
\end{tabular}
\end{adjustbox}
\endgroup
\end{table}

\FloatBarrier

\begin{table}[!htbp]
\centering
\caption{Held-out-scene (OOD) accuracy (\%) by action type ($n=3{,}496$). Overall covers the complete OOD split. Panels report SFT and RL separately; --- denotes unavailable full-mix RL results.}
\label{tab:own-T3}
\begingroup
\footnotesize
\setlength{\tabcolsep}{3pt}
\renewcommand{\arraystretch}{1.04}
\begin{adjustbox}{max width=\linewidth}
\begin{tabular}{@{}lrrrrrr@{}}
\toprule
\multicolumn{7}{l}{\textbf{SFT}} \\
\addlinespace[2pt]
Config. & Perceptive & Inspective & Navigative & Manipulative & Exogenous & Overall \\
\midrule
base & 27.19 & 24.04 & 33.75 & 25.16 & 30.47 & 27.77 \\
\addlinespace[1pt]
\texttt{full-mix} & 84.69 & 87.18 & 90.78 & 95.47 & 83.12 & 88.16 \\
\texttt{perc\_causal} & 59.69 & 37.93 & 59.84 & 58.12 & 25.94 & 47.43 \\
\texttt{perc\_temporal} & 33.28 & 23.50 & 33.28 & 31.41 & 28.91 & 29.52 \\
\texttt{insp\_causal} & 31.88 & 59.62 & 59.22 & 53.44 & 32.81 & 48.43 \\
\texttt{insp\_cf} & 36.25 & 62.61 & 59.69 & 63.75 & 35.31 & 52.46 \\
\texttt{insp\_temporal} & 29.38 & 25.11 & 32.50 & 25.47 & 30.94 & 28.38 \\
\texttt{navi\_causal} & 35.16 & 34.08 & 66.88 & 70.47 & 34.38 & 47.00 \\
\texttt{navi\_temporal} & 30.16 & 25.64 & 35.94 & 28.12 & 35.00 & 30.52 \\
\texttt{mani\_causal} & 33.75 & 29.70 & 66.72 & 87.50 & 44.06 & 50.43 \\
\texttt{mani\_cf} & 31.72 & 29.81 & 61.56 & 87.03 & 45.62 & 49.34 \\
\texttt{exo\_physical} & 26.09 & 29.49 & 38.12 & 32.03 & 62.19 & 36.90 \\
\texttt{exo\_temporal} & 32.66 & 29.49 & 34.38 & 25.94 & 36.56 & 31.61 \\
\midrule
\addlinespace[3pt]
\multicolumn{7}{l}{\textbf{RL}} \\
\addlinespace[2pt]
Config. & Perceptive & Inspective & Navigative & Manipulative & Exogenous & Overall \\
\midrule
base & 27.19 & 24.04 & 33.75 & 25.16 & 30.47 & 27.77 \\
\addlinespace[1pt]
\texttt{full-mix} & --- & --- & --- & --- & --- & --- \\
\texttt{perc\_causal} & 58.59 & 39.74 & 59.22 & 60.47 & 24.69 & 47.80 \\
\texttt{perc\_temporal} & 33.44 & 25.75 & 37.66 & 28.75 & 30.31 & 30.72 \\
\texttt{insp\_causal} & 32.19 & 63.68 & 58.91 & 54.69 & 26.41 & 48.57 \\
\texttt{insp\_cf} & 35.31 & 62.71 & 58.75 & 65.94 & 34.84 & 52.46 \\
\texttt{insp\_temporal} & 25.16 & 22.44 & 30.00 & 24.38 & 28.59 & 25.80 \\
\texttt{navi\_causal} & 37.66 & 36.86 & 66.72 & 72.81 & 33.75 & 48.48 \\
\texttt{navi\_temporal} & 34.69 & 30.24 & 40.78 & 27.19 & 35.78 & 33.44 \\
\texttt{mani\_causal} & 36.72 & 34.19 & 67.81 & 87.66 & 47.03 & 52.95 \\
\texttt{mani\_cf} & 36.09 & 33.87 & 62.19 & 88.75 & 43.91 & 51.34 \\
\texttt{exo\_physical} & 29.53 & 30.56 & 41.56 & 35.62 & 61.09 & 38.90 \\
\texttt{exo\_temporal} & 33.44 & 32.37 & 37.03 & 25.94 & 37.97 & 33.27 \\
\bottomrule
\end{tabular}
\end{adjustbox}
\endgroup
\end{table}

\begin{table}[!htbp]
\centering
\caption{Held-out-scene (OOD) accuracy (\%) by reasoning type ($n=3{,}496$). Overall covers all eight reasoning types. Panels report SFT and RL separately; --- denotes unavailable full-mix RL results.}
\label{tab:own-T4}
\begingroup
\footnotesize
\setlength{\tabcolsep}{3pt}
\renewcommand{\arraystretch}{1.04}
\begin{adjustbox}{max width=\linewidth}
\begin{tabular}{@{}lrrrrrrrrr@{}}
\toprule
\multicolumn{10}{l}{\textbf{SFT}} \\
\addlinespace[2pt]
Config. & \shortstack{Forward\\dynamics} & \shortstack{Inverse\\dynamics} & \shortstack{CF\\substitution} & \shortstack{CF\\removal} & \shortstack{Outcome\\prediction} & \shortstack{Cued\\prediction} & \shortstack{Temporal\\ordering} & \shortstack{Temporal\\adjacency} & Overall \\
\midrule
base & 19.18 & 35.69 & 14.56 & 24.37 & 24.38 & 31.88 & 37.26 & 27.04 & 27.77 \\
\addlinespace[1pt]
\texttt{full-mix} & 97.33 & 95.44 & 94.94 & 97.78 & 92.50 & 92.50 & 85.22 & 64.31 & 88.16 \\
\texttt{perc\_causal} & 75.47 & 74.06 & 63.61 & 30.70 & 30.00 & 25.62 & 28.93 & 21.38 & 47.43 \\
\texttt{perc\_temporal} & 20.44 & 37.11 & 23.42 & 24.05 & 27.50 & 30.62 & 25.31 & 41.19 & 29.52 \\
\texttt{insp\_causal} & 70.60 & 76.10 & 74.05 & 38.29 & 41.25 & 42.50 & 22.01 & 20.60 & 48.43 \\
\texttt{insp\_cf} & 72.96 & 47.48 & 79.75 & 85.13 & 45.62 & 42.50 & 33.18 & 30.66 & 52.46 \\
\texttt{insp\_temporal} & 21.38 & 36.32 & 19.62 & 23.10 & 29.38 & 31.25 & 17.61 & 44.18 & 28.38 \\
\texttt{navi\_causal} & 60.69 & 64.15 & 55.38 & 47.15 & 35.00 & 36.88 & 32.86 & 31.60 & 47.00 \\
\texttt{navi\_temporal} & 21.70 & 37.89 & 21.52 & 22.15 & 31.25 & 35.62 & 26.10 & 43.55 & 30.52 \\
\texttt{mani\_causal} & 61.95 & 62.89 & 61.08 & 45.25 & 53.75 & 53.75 & 40.09 & 32.39 & 50.43 \\
\texttt{mani\_cf} & 63.52 & 50.47 & 65.82 & 56.96 & 53.75 & 57.50 & 31.13 & 37.11 & 49.34 \\
\texttt{exo\_physical} & 24.21 & 44.50 & 15.82 & 42.09 & 94.38 & 93.75 & 30.82 & 27.20 & 36.90 \\
\texttt{exo\_temporal} & 21.07 & 36.48 & 19.62 & 25.32 & 28.75 & 36.25 & 24.53 & 52.99 & 31.61 \\
\midrule
\addlinespace[3pt]
\multicolumn{10}{l}{\textbf{RL}} \\
\addlinespace[2pt]
Config. & \shortstack{Forward\\dynamics} & \shortstack{Inverse\\dynamics} & \shortstack{CF\\substitution} & \shortstack{CF\\removal} & \shortstack{Outcome\\prediction} & \shortstack{Cued\\prediction} & \shortstack{Temporal\\ordering} & \shortstack{Temporal\\adjacency} & Overall \\
\midrule
base & 19.18 & 35.69 & 14.56 & 24.37 & 24.38 & 31.88 & 37.26 & 27.04 & 27.77 \\
\addlinespace[1pt]
\texttt{full-mix} & --- & --- & --- & --- & --- & --- & --- & --- & --- \\
\texttt{perc\_causal} & 77.83 & 78.30 & 67.09 & 31.01 & 30.00 & 22.50 & 25.16 & 19.50 & 47.80 \\
\texttt{perc\_temporal} & 22.17 & 38.84 & 22.15 & 23.10 & 30.00 & 30.00 & 25.00 & 45.28 & 30.72 \\
\texttt{insp\_causal} & 71.23 & 78.93 & 74.68 & 42.41 & 30.00 & 30.62 & 23.58 & 19.81 & 48.57 \\
\texttt{insp\_cf} & 72.48 & 52.20 & 80.70 & 87.34 & 44.38 & 40.62 & 31.76 & 27.04 & 52.46 \\
\texttt{insp\_temporal} & 20.28 & 38.36 & 16.77 & 20.57 & 29.38 & 31.88 & 12.74 & 36.48 & 25.80 \\
\texttt{navi\_causal} & 63.05 & 66.98 & 55.70 & 47.78 & 28.75 & 31.25 & 33.33 & 36.64 & 48.48 \\
\texttt{navi\_temporal} & 21.23 & 49.37 & 21.20 & 21.52 & 31.25 & 30.62 & 30.82 & 45.60 & 33.44 \\
\texttt{mani\_causal} & 62.74 & 64.31 & 64.24 & 45.89 & 55.62 & 57.50 & 41.19 & 39.62 & 52.95 \\
\texttt{mani\_cf} & 66.35 & 52.67 & 68.35 & 60.13 & 48.12 & 56.88 & 33.18 & 39.78 & 51.34 \\
\texttt{exo\_physical} & 25.79 & 48.74 & 17.09 & 45.25 & 91.88 & 92.50 & 33.02 & 28.93 & 38.90 \\
\texttt{exo\_temporal} & 21.70 & 38.36 & 19.62 & 23.73 & 28.75 & 33.75 & 28.93 & 56.60 & 33.27 \\
\bottomrule
\end{tabular}
\end{adjustbox}
\endgroup
\end{table}

\FloatBarrier

\begin{table}[!htbp]
\centering
\caption{State-perturbation (Pert) accuracy (\%) by perturbation group ($n=1{,}116$). Geometric changes affect size, orientation, or layout; appearance changes affect color, lighting, or background. Overall covers the complete Pert split. Panels report SFT and RL separately; --- denotes unavailable full-mix RL results.}
\label{tab:own-T5}
\label{tab:app-geo}
\begingroup
\footnotesize
\setlength{\tabcolsep}{3pt}
\renewcommand{\arraystretch}{1.04}
\begin{adjustbox}{max width=\linewidth}
\begin{tabular}{@{}lrrr@{}}
\toprule
\multicolumn{4}{l}{\textbf{SFT}} \\
\addlinespace[2pt]
Config. & Geometric & Appearance & Overall \\
\midrule
base & 22.58 & 19.71 & 21.15 \\
\addlinespace[1pt]
\texttt{full-mix} & 34.95 & 41.58 & 38.26 \\
\texttt{perc\_causal} & 23.12 & 19.53 & 21.33 \\
\texttt{perc\_temporal} & 24.19 & 20.43 & 22.31 \\
\texttt{insp\_causal} & 25.09 & 24.37 & 24.73 \\
\texttt{insp\_cf} & 26.34 & 24.37 & 25.36 \\
\texttt{insp\_temporal} & 23.66 & 23.66 & 23.66 \\
\texttt{navi\_causal} & 37.63 & 30.29 & 33.96 \\
\texttt{navi\_temporal} & 25.27 & 24.19 & 24.73 \\
\texttt{mani\_causal} & 34.95 & 50.00 & 42.47 \\
\texttt{mani\_cf} & 35.13 & 44.62 & 39.87 \\
\texttt{exo\_physical} & 37.10 & 62.37 & 49.73 \\
\texttt{exo\_temporal} & 24.91 & 26.16 & 25.54 \\
\midrule
\addlinespace[3pt]
\multicolumn{4}{l}{\textbf{RL}} \\
\addlinespace[2pt]
Config. & Geometric & Appearance & Overall \\
\midrule
base & 22.58 & 19.71 & 21.15 \\
\addlinespace[1pt]
\texttt{full-mix} & --- & --- & --- \\
\texttt{perc\_causal} & 23.48 & 22.22 & 22.85 \\
\texttt{perc\_temporal} & 28.14 & 22.40 & 25.27 \\
\texttt{insp\_causal} & 23.66 & 24.55 & 24.10 \\
\texttt{insp\_cf} & 25.81 & 22.94 & 24.37 \\
\texttt{insp\_temporal} & 25.09 & 27.60 & 26.34 \\
\texttt{navi\_causal} & 38.53 & 31.00 & 34.77 \\
\texttt{navi\_temporal} & 29.39 & 26.52 & 27.96 \\
\texttt{mani\_causal} & 35.66 & 53.05 & 44.35 \\
\texttt{mani\_cf} & 36.56 & 46.06 & 41.31 \\
\texttt{exo\_physical} & 37.99 & 64.16 & 51.08 \\
\texttt{exo\_temporal} & 28.14 & 27.96 & 28.05 \\
\bottomrule
\end{tabular}
\end{adjustbox}
\endgroup
\end{table}

\FloatBarrier

\begin{table}[!htbp]
\centering
\caption{ID accuracy (\%) by camera or viewpoint action. Direction-specific columns include only perceptive and inspective items. Overall ID is accuracy on the complete ID split ($n=6{,}228$), not the camera/viewpoint subset. Panels report SFT and RL separately; --- denotes unavailable full-mix RL results.}
\label{tab:own-T6}
\begingroup
\footnotesize
\setlength{\tabcolsep}{3pt}
\renewcommand{\arraystretch}{1.04}
\begin{adjustbox}{max width=\linewidth}
\begin{tabular}{@{}lrrrrrrrr@{}}
\toprule
\multicolumn{9}{l}{\textbf{SFT}} \\
\addlinespace[2pt]
Config. & \shortstack{Rotate\\left} & \shortstack{Rotate\\right} & \shortstack{Move\\forward} & \shortstack{Move\\backward} & \shortstack{Orbit\\left} & \shortstack{Orbit\\right} & \shortstack{Arc\\up} & \shortstack{Overall\\ID} \\
\midrule
base & 19.79 & 19.79 & 39.58 & 28.82 & 22.04 & 19.63 & 28.70 & 26.38 \\
\addlinespace[1pt]
\texttt{full-mix} & 90.97 & 87.50 & 93.40 & 90.97 & 89.44 & 89.44 & 92.59 & 89.29 \\
\texttt{perc\_causal} & 58.33 & 60.07 & 60.76 & 63.89 & 29.81 & 39.44 & 35.00 & 48.28 \\
\texttt{perc\_temporal} & 34.38 & 32.64 & 42.71 & 46.18 & 28.89 & 21.48 & 31.48 & 32.29 \\
\texttt{insp\_causal} & 34.38 & 30.21 & 39.58 & 44.10 & 62.41 & 64.81 & 59.63 & 49.97 \\
\texttt{insp\_cf} & 26.74 & 27.08 & 42.01 & 37.85 & 53.70 & 56.48 & 66.48 & 50.75 \\
\texttt{insp\_temporal} & 29.17 & 27.43 & 41.67 & 37.50 & 28.15 & 23.89 & 31.85 & 32.03 \\
\texttt{navi\_causal} & 22.57 & 17.71 & 53.47 & 42.71 & 29.63 & 34.07 & 38.70 & 46.56 \\
\texttt{navi\_temporal} & 34.38 & 28.82 & 40.62 & 37.85 & 28.70 & 22.41 & 32.22 & 31.87 \\
\texttt{mani\_causal} & 16.67 & 16.32 & 37.15 & 55.56 & 24.81 & 21.48 & 31.30 & 45.41 \\
\texttt{mani\_cf} & 14.58 & 14.93 & 48.26 & 46.18 & 25.19 & 26.67 & 27.78 & 46.63 \\
\texttt{exo\_physical} & 11.46 & 14.24 & 38.89 & 42.36 & 28.15 & 30.37 & 27.96 & 34.63 \\
\texttt{exo\_temporal} & 30.90 & 32.99 & 39.58 & 37.15 & 28.52 & 22.96 & 32.96 & 32.19 \\
\midrule
\addlinespace[3pt]
\multicolumn{9}{l}{\textbf{RL}} \\
\addlinespace[2pt]
Config. & \shortstack{Rotate\\left} & \shortstack{Rotate\\right} & \shortstack{Move\\forward} & \shortstack{Move\\backward} & \shortstack{Orbit\\left} & \shortstack{Orbit\\right} & \shortstack{Arc\\up} & \shortstack{Overall\\ID} \\
\midrule
base & 19.79 & 19.79 & 39.58 & 28.82 & 22.04 & 19.63 & 28.70 & 26.38 \\
\addlinespace[1pt]
\texttt{full-mix} & --- & --- & --- & --- & --- & --- & --- & --- \\
\texttt{perc\_causal} & 58.33 & 59.38 & 61.11 & 61.11 & 34.63 & 37.04 & 39.07 & 49.26 \\
\texttt{perc\_temporal} & 35.07 & 32.64 & 39.93 & 44.10 & 27.04 & 24.44 & 33.70 & 32.96 \\
\texttt{insp\_causal} & 33.33 & 32.64 & 41.67 & 36.11 & 65.00 & 66.30 & 58.70 & 51.65 \\
\texttt{insp\_cf} & 28.12 & 26.39 & 47.92 & 34.72 & 55.56 & 60.00 & 70.37 & 51.46 \\
\texttt{insp\_temporal} & 29.86 & 26.04 & 42.36 & 35.42 & 27.78 & 20.93 & 31.85 & 32.45 \\
\texttt{navi\_causal} & 23.96 & 20.83 & 55.21 & 48.26 & 32.59 & 35.19 & 40.74 & 47.94 \\
\texttt{navi\_temporal} & 40.28 & 34.72 & 43.06 & 38.89 & 27.04 & 20.93 & 34.44 & 33.30 \\
\texttt{mani\_causal} & 17.71 & 17.71 & 33.33 & 54.86 & 27.41 & 25.37 & 34.81 & 46.32 \\
\texttt{mani\_cf} & 18.06 & 17.36 & 50.35 & 43.06 & 26.11 & 29.81 & 30.93 & 47.64 \\
\texttt{exo\_physical} & 8.68 & 14.93 & 39.24 & 40.28 & 29.63 & 29.63 & 28.33 & 35.74 \\
\texttt{exo\_temporal} & 27.43 & 29.17 & 43.06 & 36.81 & 28.52 & 22.59 & 35.74 & 32.98 \\
\bottomrule
\end{tabular}
\end{adjustbox}
\endgroup
\end{table}

\begin{table}[!htbp]
\centering
\caption{ID accuracy (\%) by answer-option modality ($n=6{,}228$). Image and Text refer to the options presented to the model; Overall covers the complete ID split. Panels report SFT and RL separately; --- denotes unavailable full-mix RL results.}
\label{tab:own-T7}
\begingroup
\footnotesize
\setlength{\tabcolsep}{3pt}
\renewcommand{\arraystretch}{1.04}
\begin{adjustbox}{max width=\linewidth}
\begin{tabular}{@{}lrrr@{}}
\toprule
\multicolumn{4}{l}{\textbf{SFT}} \\
\addlinespace[2pt]
Config. & Image & Text & Overall \\
\midrule
base & 20.67 & 31.13 & 26.38 \\
\addlinespace[1pt]
\texttt{full-mix} & 93.56 & 85.74 & 89.29 \\
\texttt{perc\_causal} & 55.02 & 42.68 & 48.28 \\
\texttt{perc\_temporal} & 21.09 & 41.59 & 32.29 \\
\texttt{insp\_causal} & 59.16 & 42.33 & 49.97 \\
\texttt{insp\_cf} & 71.37 & 33.63 & 50.75 \\
\texttt{insp\_temporal} & 21.23 & 41.01 & 32.03 \\
\texttt{navi\_causal} & 54.74 & 39.77 & 46.56 \\
\texttt{navi\_temporal} & 20.38 & 41.42 & 31.87 \\
\texttt{mani\_causal} & 54.00 & 38.27 & 45.41 \\
\texttt{mani\_cf} & 58.42 & 36.83 & 46.63 \\
\texttt{exo\_physical} & 36.66 & 32.95 & 34.63 \\
\texttt{exo\_temporal} & 20.91 & 41.56 & 32.19 \\
\midrule
\addlinespace[3pt]
\multicolumn{4}{l}{\textbf{RL}} \\
\addlinespace[2pt]
Config. & Image & Text & Overall \\
\midrule
base & 20.67 & 31.13 & 26.38 \\
\addlinespace[1pt]
\texttt{full-mix} & --- & --- & --- \\
\texttt{perc\_causal} & 56.55 & 43.21 & 49.26 \\
\texttt{perc\_temporal} & 20.45 & 43.36 & 32.96 \\
\texttt{insp\_causal} & 62.24 & 42.86 & 51.65 \\
\texttt{insp\_cf} & 69.89 & 36.16 & 51.46 \\
\texttt{insp\_temporal} & 20.63 & 42.27 & 32.45 \\
\texttt{navi\_causal} & 54.92 & 42.15 & 47.94 \\
\texttt{navi\_temporal} & 21.09 & 43.45 & 33.30 \\
\texttt{mani\_causal} & 55.73 & 38.51 & 46.32 \\
\texttt{mani\_cf} & 60.76 & 36.74 & 47.64 \\
\texttt{exo\_physical} & 38.08 & 33.80 & 35.74 \\
\texttt{exo\_temporal} & 21.66 & 42.39 & 32.98 \\
\bottomrule
\end{tabular}
\end{adjustbox}
\endgroup
\end{table}

\FloatBarrier

\FloatBarrier

\subsubsection{Errors by question type}

\begin{table}[!htbp]
\centering
\caption{Temporal-adjacency errors on ID ($n=1{,}134$). Accuracy is in percent; Errors is the number of incorrect answers. Error shares (\%) distinguish gross localization errors, off-by-one errors, and reversed before/after direction. Shares sum to approximately 100 within each run, allowing for rounding. Panels report SFT and RL separately; --- denotes unavailable full-mix RL results.}
\label{tab:own-E1}
\label{tab:tadj-comp}
\begingroup
\footnotesize
\setlength{\tabcolsep}{3pt}
\renewcommand{\arraystretch}{1.04}
\begin{adjustbox}{max width=\linewidth}
\begin{tabular}{@{}lrrrrr@{}}
\toprule
\multicolumn{6}{l}{\textbf{SFT}} \\
\addlinespace[2pt]
Config. & \shortstack{Accuracy\\(\%)} & Errors & Gross & \shortstack{Off-by-\\one} & Direction \\
\midrule
base & 25.66 & 843 & 17.2 & 34.4 & 48.4 \\
\addlinespace[1pt]
\texttt{full-mix} & 69.66 & 344 & 1.7 & 38.4 & 59.9 \\
\texttt{perc\_causal} & 26.28 & 836 & 16.0 & 35.9 & 48.1 \\
\texttt{perc\_temporal} & 52.73 & 536 & 1.5 & 25.6 & 72.9 \\
\texttt{insp\_causal} & 25.75 & 842 & 17.6 & 31.4 & 51.1 \\
\texttt{insp\_cf} & 23.72 & 865 & 16.5 & 35.1 & 48.3 \\
\texttt{insp\_temporal} & 56.70 & 491 & 1.8 & 26.7 & 71.5 \\
\texttt{navi\_causal} & 27.95 & 817 & 14.7 & 33.3 & 52.0 \\
\texttt{navi\_temporal} & 55.82 & 501 & 3.0 & 27.5 & 69.5 \\
\texttt{mani\_causal} & 26.28 & 836 & 15.3 & 34.0 & 50.7 \\
\texttt{mani\_cf} & 28.48 & 811 & 14.7 & 35.0 & 50.3 \\
\texttt{exo\_physical} & 23.10 & 872 & 15.9 & 34.6 & 49.4 \\
\texttt{exo\_temporal} & 56.61 & 492 & 2.4 & 29.3 & 68.3 \\
\midrule
\addlinespace[3pt]
\multicolumn{6}{l}{\textbf{RL}} \\
\addlinespace[2pt]
Config. & \shortstack{Accuracy\\(\%)} & Errors & Gross & \shortstack{Off-by-\\one} & Direction \\
\midrule
base & 25.66 & 843 & 17.2 & 34.4 & 48.4 \\
\addlinespace[1pt]
\texttt{full-mix} & --- & --- & --- & --- & --- \\
\texttt{perc\_causal} & 25.84 & 841 & 18.7 & 34.4 & 47.0 \\
\texttt{perc\_temporal} & 55.56 & 504 & 3.6 & 28.0 & 68.5 \\
\texttt{insp\_causal} & 26.54 & 833 & 18.0 & 28.2 & 53.8 \\
\texttt{insp\_cf} & 26.98 & 828 & 15.6 & 33.0 & 51.4 \\
\texttt{insp\_temporal} & 58.73 & 468 & 2.6 & 26.3 & 71.2 \\
\texttt{navi\_causal} & 30.51 & 788 & 12.4 & 33.1 & 54.4 \\
\texttt{navi\_temporal} & 57.67 & 480 & 3.3 & 26.0 & 70.6 \\
\texttt{mani\_causal} & 27.95 & 817 & 15.1 & 34.5 & 50.4 \\
\texttt{mani\_cf} & 28.57 & 810 & 15.1 & 33.5 & 51.5 \\
\texttt{exo\_physical} & 24.16 & 860 & 17.4 & 34.0 & 48.6 \\
\texttt{exo\_temporal} & 56.26 & 496 & 2.2 & 31.5 & 66.3 \\
\bottomrule
\end{tabular}
\end{adjustbox}
\endgroup
\end{table}

\begin{table}[!htbp]
\centering
\caption{Temporal-ordering errors on ID ($n=1{,}134$). Accuracy is in percent; Errors counts incorrect answers. Random 0--2 are the three independently sampled permutation distractors. Their shares (\%) describe the distribution of errors and sum to approximately 100, allowing for rounding. Panels report SFT and RL separately; --- denotes unavailable full-mix RL results.}
\label{tab:own-E2}
\begingroup
\footnotesize
\setlength{\tabcolsep}{3pt}
\renewcommand{\arraystretch}{1.04}
\begin{adjustbox}{max width=\linewidth}
\begin{tabular}{@{}lrrrrr@{}}
\toprule
\multicolumn{6}{l}{\textbf{SFT}} \\
\addlinespace[2pt]
Config. & \shortstack{Accuracy\\(\%)} & Errors & \shortstack{Random\\0} & \shortstack{Random\\1} & \shortstack{Random\\2} \\
\midrule
base & 27.34 & 824 & 32.8 & 37.1 & 30.1 \\
\addlinespace[1pt]
\texttt{full-mix} & 92.77 & 82 & 36.6 & 35.4 & 28.0 \\
\texttt{perc\_causal} & 25.13 & 849 & 33.8 & 34.5 & 31.7 \\
\texttt{perc\_temporal} & 30.34 & 790 & 33.4 & 31.8 & 34.8 \\
\texttt{insp\_causal} & 25.40 & 846 & 32.6 & 35.5 & 31.9 \\
\texttt{insp\_cf} & 26.63 & 832 & 34.3 & 35.0 & 30.8 \\
\texttt{insp\_temporal} & 24.96 & 851 & 33.0 & 34.0 & 33.0 \\
\texttt{navi\_causal} & 26.46 & 834 & 32.0 & 35.4 & 32.6 \\
\texttt{navi\_temporal} & 25.93 & 840 & 32.6 & 34.9 & 32.5 \\
\texttt{mani\_causal} & 25.84 & 841 & 31.6 & 33.8 & 34.6 \\
\texttt{mani\_cf} & 27.25 & 825 & 33.6 & 35.5 & 30.9 \\
\texttt{exo\_physical} & 28.13 & 815 & 32.8 & 36.8 & 30.4 \\
\texttt{exo\_temporal} & 25.31 & 847 & 35.3 & 33.5 & 31.2 \\
\midrule
\addlinespace[3pt]
\multicolumn{6}{l}{\textbf{RL}} \\
\addlinespace[2pt]
Config. & \shortstack{Accuracy\\(\%)} & Errors & \shortstack{Random\\0} & \shortstack{Random\\1} & \shortstack{Random\\2} \\
\midrule
base & 27.34 & 824 & 32.8 & 37.1 & 30.1 \\
\addlinespace[1pt]
\texttt{full-mix} & --- & --- & --- & --- & --- \\
\texttt{perc\_causal} & 24.51 & 856 & 31.9 & 36.0 & 32.1 \\
\texttt{perc\_temporal} & 30.60 & 787 & 32.0 & 34.8 & 33.2 \\
\texttt{insp\_causal} & 25.93 & 840 & 35.1 & 32.5 & 32.4 \\
\texttt{insp\_cf} & 26.54 & 833 & 32.7 & 34.6 & 32.8 \\
\texttt{insp\_temporal} & 24.96 & 851 & 35.0 & 33.4 & 31.6 \\
\texttt{navi\_causal} & 25.40 & 846 & 32.9 & 35.1 & 32.0 \\
\texttt{navi\_temporal} & 23.81 & 864 & 34.3 & 31.6 & 34.1 \\
\texttt{mani\_causal} & 24.69 & 854 & 31.7 & 35.1 & 33.1 \\
\texttt{mani\_cf} & 26.63 & 832 & 33.5 & 32.9 & 33.5 \\
\texttt{exo\_physical} & 26.98 & 828 & 36.0 & 34.3 & 29.7 \\
\texttt{exo\_temporal} & 26.54 & 833 & 32.7 & 34.8 & 32.5 \\
\bottomrule
\end{tabular}
\end{adjustbox}
\endgroup
\end{table}

\FloatBarrier

\begin{table}[!htbp]
\centering
\caption{Physical-modeling errors on ID: exogenous outcome and cued prediction ($n=576$). Accuracy is in percent; Errors counts all incorrect answers. The ten displayed violation types are the most frequent in the base model's errors. Each share (\%) uses all errors in that run as its denominator, including unlisted types; displayed shares are not renormalized and need not sum to 100. Panels report SFT and RL separately; --- denotes unavailable full-mix RL results.}
\label{tab:own-E3}
\begingroup
\footnotesize
\setlength{\tabcolsep}{2.3pt}
\renewcommand{\arraystretch}{1.04}
\begin{adjustbox}{max width=\linewidth}
\begin{tabular}{@{}lrrrrrrrrrrrr@{}}
\toprule
\multicolumn{13}{l}{\textbf{SFT}} \\
\addlinespace[2pt]
Config. & \shortstack{Accuracy\\(\%)} & Errors & \shortstack{Pass\\through} & Transform & \shortstack{No\\effect} & Split & Lateral & Float & \shortstack{Instant\\stop} & \shortstack{Embed\\surface} & Escape & \shortstack{Separate\\midair} \\
\midrule
base & 22.57 & 446 & 8.1 & 5.6 & 5.6 & 5.2 & 5.2 & 5.2 & 5.2 & 4.7 & 4.7 & 4.5 \\
\addlinespace[1pt]
\texttt{full-mix} & 82.12 & 103 & 12.6 & 0.0 & 4.9 & 1.9 & 2.9 & 0.0 & 0.0 & 4.9 & 5.8 & 3.9 \\
\texttt{perc\_causal} & 33.51 & 383 & 9.4 & 2.1 & 2.3 & 10.7 & 2.1 & 0.5 & 4.2 & 1.0 & 6.0 & 6.8 \\
\texttt{perc\_temporal} & 24.31 & 436 & 7.3 & 5.3 & 4.8 & 4.6 & 5.3 & 6.2 & 6.0 & 5.0 & 3.0 & 3.9 \\
\texttt{insp\_causal} & 33.33 & 384 & 14.1 & 1.0 & 1.0 & 1.3 & 0.5 & 1.6 & 0.0 & 1.3 & 7.3 & 4.2 \\
\texttt{insp\_cf} & 40.28 & 344 & 12.8 & 0.9 & 2.9 & 2.6 & 2.3 & 1.2 & 0.0 & 0.6 & 4.9 & 4.1 \\
\texttt{insp\_temporal} & 23.78 & 439 & 7.7 & 5.5 & 4.6 & 3.9 & 5.5 & 5.7 & 6.2 & 4.6 & 3.2 & 4.6 \\
\texttt{navi\_causal} & 34.90 & 375 & 7.7 & 1.6 & 4.0 & 0.8 & 2.9 & 2.1 & 4.0 & 2.7 & 2.7 & 4.8 \\
\texttt{navi\_temporal} & 21.70 & 451 & 7.8 & 4.9 & 3.8 & 4.0 & 5.3 & 5.8 & 6.7 & 4.4 & 3.8 & 4.9 \\
\texttt{mani\_causal} & 34.38 & 378 & 8.2 & 0.0 & 4.2 & 0.0 & 4.0 & 2.9 & 5.6 & 1.3 & 4.2 & 2.9 \\
\texttt{mani\_cf} & 40.62 & 342 & 7.6 & 0.0 & 6.7 & 2.0 & 5.3 & 1.8 & 1.5 & 1.5 & 5.8 & 2.9 \\
\texttt{exo\_physical} & 77.08 & 132 & 4.5 & 0.0 & 16.7 & 0.0 & 0.8 & 3.8 & 0.0 & 10.6 & 0.0 & 3.0 \\
\texttt{exo\_temporal} & 23.78 & 439 & 8.9 & 5.5 & 4.6 & 3.2 & 4.8 & 6.2 & 5.7 & 4.6 & 3.6 & 3.6 \\
\midrule
\addlinespace[3pt]
\multicolumn{13}{l}{\textbf{RL}} \\
\addlinespace[2pt]
Config. & \shortstack{Accuracy\\(\%)} & Errors & \shortstack{Pass\\through} & Transform & \shortstack{No\\effect} & Split & Lateral & Float & \shortstack{Instant\\stop} & \shortstack{Embed\\surface} & Escape & \shortstack{Separate\\midair} \\
\midrule
base & 22.57 & 446 & 8.1 & 5.6 & 5.6 & 5.2 & 5.2 & 5.2 & 5.2 & 4.7 & 4.7 & 4.5 \\
\addlinespace[1pt]
\texttt{full-mix} & --- & --- & --- & --- & --- & --- & --- & --- & --- & --- & --- & --- \\
\texttt{perc\_causal} & 32.29 & 390 & 5.1 & 0.8 & 1.5 & 8.2 & 1.5 & 1.3 & 2.8 & 1.3 & 6.4 & 8.2 \\
\texttt{perc\_temporal} & 23.61 & 440 & 8.2 & 5.2 & 5.2 & 5.7 & 4.1 & 6.4 & 4.3 & 4.5 & 3.2 & 4.8 \\
\texttt{insp\_causal} & 32.99 & 386 & 12.4 & 1.0 & 1.6 & 1.8 & 0.0 & 0.0 & 0.0 & 2.1 & 7.0 & 2.3 \\
\texttt{insp\_cf} & 34.72 & 376 & 13.8 & 0.5 & 4.3 & 1.6 & 4.3 & 3.5 & 0.0 & 3.2 & 4.3 & 4.3 \\
\texttt{insp\_temporal} & 23.96 & 438 & 7.1 & 5.0 & 4.3 & 4.3 & 5.5 & 5.0 & 7.1 & 4.3 & 2.5 & 4.1 \\
\texttt{navi\_causal} & 34.38 & 378 & 6.3 & 0.8 & 5.3 & 0.5 & 2.9 & 2.1 & 3.4 & 2.4 & 2.4 & 4.8 \\
\texttt{navi\_temporal} & 23.09 & 443 & 7.2 & 5.2 & 5.0 & 6.8 & 3.8 & 6.3 & 5.0 & 5.0 & 3.8 & 4.1 \\
\texttt{mani\_causal} & 36.11 & 368 & 9.5 & 0.0 & 3.5 & 0.3 & 4.1 & 1.9 & 5.2 & 0.8 & 4.6 & 1.9 \\
\texttt{mani\_cf} & 41.15 & 339 & 8.0 & 0.0 & 6.5 & 3.2 & 2.4 & 1.5 & 0.9 & 0.9 & 5.0 & 1.8 \\
\texttt{exo\_physical} & 75.69 & 140 & 3.6 & 0.0 & 19.3 & 0.0 & 1.4 & 5.0 & 0.0 & 8.6 & 0.0 & 6.4 \\
\texttt{exo\_temporal} & 24.31 & 436 & 8.7 & 3.4 & 5.0 & 5.0 & 4.4 & 4.6 & 3.9 & 4.8 & 4.6 & 5.5 \\
\bottomrule
\end{tabular}
\end{adjustbox}
\endgroup
\end{table}

\FloatBarrier

\begin{table}[!htbp]
\centering
\caption{Incorrect option choices on the state-perturbation probe (Pert, $n=1{,}116$). Accuracy is in percent; Errors counts incorrect answers. Shares (\%) identify the perturbation dimension of the selected incorrect option; None denotes an unperturbed option. Shares sum to approximately 100, allowing for rounding. Panels report SFT and RL separately; --- denotes unavailable full-mix RL results.}
\label{tab:own-E4}
\begingroup
\footnotesize
\setlength{\tabcolsep}{3pt}
\renewcommand{\arraystretch}{1.04}
\begin{adjustbox}{max width=\linewidth}
\begin{tabular}{@{}lrrrrrrrrr@{}}
\toprule
\multicolumn{10}{l}{\textbf{SFT}} \\
\addlinespace[2pt]
Config. & \shortstack{Accuracy\\(\%)} & Errors & None & Size & \shortstack{Orien-\\tation} & Layout & Color & Lighting & \shortstack{Back-\\ground} \\
\midrule
base & 21.15 & 880 & 0.0 & 16.2 & 17.6 & 15.2 & 17.0 & 16.2 & 17.6 \\
\addlinespace[1pt]
\texttt{full-mix} & 38.26 & 689 & 0.0 & 20.8 & 22.6 & 9.3 & 14.8 & 20.2 & 12.3 \\
\texttt{perc\_causal} & 21.33 & 878 & 0.0 & 12.2 & 19.2 & 17.4 & 20.0 & 16.1 & 15.0 \\
\texttt{perc\_temporal} & 22.31 & 867 & 0.0 & 15.5 & 18.2 & 15.1 & 16.3 & 15.6 & 19.4 \\
\texttt{insp\_causal} & 24.73 & 840 & 0.0 & 16.8 & 19.0 & 13.9 & 13.8 & 17.6 & 18.8 \\
\texttt{insp\_cf} & 25.36 & 833 & 0.0 & 19.3 & 16.8 & 13.2 & 20.0 & 17.3 & 13.3 \\
\texttt{insp\_temporal} & 23.66 & 852 & 0.0 & 18.0 & 18.1 & 14.0 & 17.4 & 15.5 & 17.1 \\
\texttt{navi\_causal} & 33.96 & 737 & 0.0 & 11.1 & 27.8 & 8.3 & 17.9 & 22.8 & 12.1 \\
\texttt{navi\_temporal} & 24.73 & 840 & 0.0 & 16.7 & 19.0 & 13.9 & 16.2 & 17.1 & 17.0 \\
\texttt{mani\_causal} & 42.47 & 642 & 0.0 & 22.9 & 25.4 & 8.3 & 17.4 & 15.9 & 10.1 \\
\texttt{mani\_cf} & 39.87 & 671 & 0.0 & 21.5 & 23.8 & 8.6 & 17.1 & 16.4 & 12.5 \\
\texttt{exo\_physical} & 49.73 & 561 & 0.0 & 26.4 & 28.2 & 8.0 & 13.2 & 20.0 & 4.3 \\
\texttt{exo\_temporal} & 25.54 & 831 & 0.0 & 17.7 & 18.9 & 13.8 & 17.4 & 16.6 & 15.5 \\
\midrule
\addlinespace[3pt]
\multicolumn{10}{l}{\textbf{RL}} \\
\addlinespace[2pt]
Config. & \shortstack{Accuracy\\(\%)} & Errors & None & Size & \shortstack{Orien-\\tation} & Layout & Color & Lighting & \shortstack{Back-\\ground} \\
\midrule
base & 21.15 & 880 & 0.0 & 16.2 & 17.6 & 15.2 & 17.0 & 16.2 & 17.6 \\
\addlinespace[1pt]
\texttt{full-mix} & --- & --- & --- & --- & --- & --- & --- & --- & --- \\
\texttt{perc\_causal} & 22.85 & 861 & 0.0 & 12.1 & 18.4 & 19.2 & 21.6 & 15.7 & 13.1 \\
\texttt{perc\_temporal} & 25.27 & 834 & 0.0 & 17.0 & 16.8 & 14.3 & 16.4 & 15.6 & 19.9 \\
\texttt{insp\_causal} & 24.10 & 847 & 0.0 & 18.1 & 19.2 & 13.0 & 13.5 & 15.9 & 20.3 \\
\texttt{insp\_cf} & 24.37 & 844 & 0.0 & 18.6 & 17.9 & 12.6 & 15.4 & 19.0 & 16.6 \\
\texttt{insp\_temporal} & 26.34 & 822 & 0.0 & 16.8 & 19.2 & 14.8 & 17.8 & 16.1 & 15.3 \\
\texttt{navi\_causal} & 34.77 & 728 & 0.0 & 11.4 & 27.2 & 8.5 & 18.4 & 23.8 & 10.7 \\
\texttt{navi\_temporal} & 27.96 & 804 & 0.0 & 19.7 & 18.3 & 11.1 & 17.8 & 18.9 & 14.3 \\
\texttt{mani\_causal} & 44.35 & 621 & 0.0 & 20.8 & 28.0 & 9.0 & 15.5 & 17.9 & 8.9 \\
\texttt{mani\_cf} & 41.31 & 655 & 0.0 & 19.8 & 25.2 & 9.0 & 13.6 & 15.9 & 16.5 \\
\texttt{exo\_physical} & 51.08 & 546 & 0.0 & 30.8 & 28.2 & 4.4 & 11.2 & 23.1 & 2.4 \\
\texttt{exo\_temporal} & 28.05 & 803 & 0.0 & 18.8 & 16.3 & 14.8 & 20.0 & 15.4 & 14.6 \\
\bottomrule
\end{tabular}
\end{adjustbox}
\endgroup
\end{table}

\FloatBarrier

\begin{table}[!htbp]
\centering
\caption{Default-state choices on ID. Forward/inverse dynamics uses the subset with a no-change option ($n=540$); counterfactual substitution and removal each use $n=558$. Accuracy and $P(\mathrm{static}\mid\mathrm{error})$ are percentages. The latter is reported separately for forward/inverse dynamics and substitution, where the default-state choice is wrong. In removal, the unchanged state is correct, so accuracy measures selecting it appropriately. Panels report SFT and RL separately; --- denotes unavailable full-mix RL results.}
\label{tab:own-E5}
\label{tab:static-pull}
\begingroup
\footnotesize
\setlength{\tabcolsep}{3pt}
\renewcommand{\arraystretch}{1.04}
\begin{adjustbox}{max width=\linewidth}
\begin{tabular}{@{}lrrrrr@{}}
\toprule
\multicolumn{6}{l}{\textbf{SFT}} \\
\addlinespace[2pt]
& \multicolumn{2}{c}{Forward/inverse dynamics} & \multicolumn{2}{c}{CF substitution} & CF removal \\
\cmidrule(lr){2-3}\cmidrule(lr){4-5}\cmidrule(l){6-6}
Config. & \shortstack{Accuracy\\(\%)} & \shortstack{$P(\mathrm{static}\mid\mathrm{error})$\\(\%)} & \shortstack{Accuracy\\(\%)} & \shortstack{$P(\mathrm{static}\mid\mathrm{error})$\\(\%)} & \shortstack{Accuracy\\(\%)} \\
\midrule
base & 22.22 & 63.8 & 19.35 & 27.8 & 18.64 \\
\addlinespace[1pt]
\texttt{full-mix} & 93.52 & 0.0 & 94.80 & 51.7 & 96.95 \\
\texttt{perc\_causal} & 43.52 & 7.2 & 62.19 & 36.5 & 28.85 \\
\texttt{perc\_temporal} & 23.70 & 56.3 & 18.82 & 29.1 & 20.61 \\
\texttt{insp\_causal} & 93.70 & 2.9 & 74.19 & 51.4 & 41.22 \\
\texttt{insp\_cf} & 55.19 & 89.3 & 83.15 & 19.1 & 86.92 \\
\texttt{insp\_temporal} & 20.93 & 57.6 & 19.00 & 26.8 & 21.86 \\
\texttt{navi\_causal} & 39.07 & 35.9 & 59.68 & 47.6 & 49.64 \\
\texttt{navi\_temporal} & 22.22 & 57.4 & 17.20 & 28.8 & 20.43 \\
\texttt{mani\_causal} & 30.00 & 58.5 & 60.93 & 33.5 & 48.39 \\
\texttt{mani\_cf} & 22.78 & 75.1 & 68.10 & 54.5 & 57.35 \\
\texttt{exo\_physical} & 29.44 & 37.5 & 15.23 & 48.6 & 38.17 \\
\texttt{exo\_temporal} & 24.81 & 43.1 & 18.10 & 28.0 & 21.51 \\
\midrule
\addlinespace[3pt]
\multicolumn{6}{l}{\textbf{RL}} \\
\addlinespace[2pt]
& \multicolumn{2}{c}{Forward/inverse dynamics} & \multicolumn{2}{c}{CF substitution} & CF removal \\
\cmidrule(lr){2-3}\cmidrule(lr){4-5}\cmidrule(l){6-6}
Config. & \shortstack{Accuracy\\(\%)} & \shortstack{$P(\mathrm{static}\mid\mathrm{error})$\\(\%)} & \shortstack{Accuracy\\(\%)} & \shortstack{$P(\mathrm{static}\mid\mathrm{error})$\\(\%)} & \shortstack{Accuracy\\(\%)} \\
\midrule
base & 22.22 & 63.8 & 19.35 & 27.8 & 18.64 \\
\addlinespace[1pt]
\texttt{full-mix} & --- & --- & --- & --- & --- \\
\texttt{perc\_causal} & 42.96 & 8.1 & 67.56 & 32.0 & 31.00 \\
\texttt{perc\_temporal} & 24.44 & 56.9 & 17.74 & 31.4 & 20.97 \\
\texttt{insp\_causal} & 89.63 & 0.0 & 83.51 & 51.1 & 44.80 \\
\texttt{insp\_cf} & 60.00 & 90.3 & 84.23 & 37.5 & 87.28 \\
\texttt{insp\_temporal} & 22.59 & 57.7 & 14.87 & 28.4 & 23.12 \\
\texttt{navi\_causal} & 46.11 & 27.1 & 60.39 & 46.2 & 47.49 \\
\texttt{navi\_temporal} & 28.52 & 40.4 & 17.38 & 32.1 & 22.94 \\
\texttt{mani\_causal} & 34.81 & 51.1 & 64.70 & 32.5 & 48.39 \\
\texttt{mani\_cf} & 26.48 & 72.5 & 70.25 & 45.2 & 60.57 \\
\texttt{exo\_physical} & 31.30 & 38.0 & 15.95 & 53.5 & 42.65 \\
\texttt{exo\_temporal} & 23.52 & 51.3 & 17.56 & 27.4 & 23.66 \\
\bottomrule
\end{tabular}
\end{adjustbox}
\endgroup
\end{table}

\FloatBarrier

\begin{table}[!htbp]
\centering
\caption{Temporal-ordering errors on held-out scenes (OOD, $n=636$). Accuracy is in percent; Errors counts incorrect answers. Error shares (\%) distinguish adjacent swaps in the middle, adjacent swaps at an end, and exchanges of the first and last states. Shares sum to approximately 100, allowing for rounding. Panels report SFT and RL separately; --- denotes unavailable full-mix RL results.}
\label{tab:own-E6a}
\begingroup
\footnotesize
\setlength{\tabcolsep}{3pt}
\renewcommand{\arraystretch}{1.04}
\begin{adjustbox}{max width=\linewidth}
\begin{tabular}{@{}lrrrrr@{}}
\toprule
\multicolumn{6}{l}{\textbf{SFT}} \\
\addlinespace[2pt]
Config. & \shortstack{Accuracy\\(\%)} & Errors & \shortstack{Center\\swap} & \shortstack{Edge\\swap} & \shortstack{Endpoint\\swap} \\
\midrule
base & 37.26 & 399 & 44.9 & 39.8 & 15.3 \\
\addlinespace[1pt]
\texttt{full-mix} & 85.22 & 94 & 26.6 & 71.3 & 2.1 \\
\texttt{perc\_causal} & 28.93 & 452 & 31.0 & 41.2 & 27.9 \\
\texttt{perc\_temporal} & 25.31 & 475 & 36.4 & 27.2 & 36.4 \\
\texttt{insp\_causal} & 22.01 & 496 & 35.3 & 36.3 & 28.4 \\
\texttt{insp\_cf} & 33.18 & 425 & 39.8 & 34.1 & 26.1 \\
\texttt{insp\_temporal} & 17.61 & 524 & 27.3 & 32.1 & 40.6 \\
\texttt{navi\_causal} & 32.86 & 427 & 40.3 & 37.7 & 22.0 \\
\texttt{navi\_temporal} & 26.10 & 470 & 40.4 & 41.5 & 18.1 \\
\texttt{mani\_causal} & 40.09 & 381 & 41.7 & 34.6 & 23.6 \\
\texttt{mani\_cf} & 31.13 & 438 & 38.8 & 34.5 & 26.7 \\
\texttt{exo\_physical} & 30.82 & 440 & 38.0 & 35.9 & 26.1 \\
\texttt{exo\_temporal} & 24.53 & 480 & 39.6 & 42.7 & 17.7 \\
\midrule
\addlinespace[3pt]
\multicolumn{6}{l}{\textbf{RL}} \\
\addlinespace[2pt]
Config. & \shortstack{Accuracy\\(\%)} & Errors & \shortstack{Center\\swap} & \shortstack{Edge\\swap} & \shortstack{Endpoint\\swap} \\
\midrule
base & 37.26 & 399 & 44.9 & 39.8 & 15.3 \\
\addlinespace[1pt]
\texttt{full-mix} & --- & --- & --- & --- & --- \\
\texttt{perc\_causal} & 25.16 & 476 & 31.5 & 39.1 & 29.4 \\
\texttt{perc\_temporal} & 25.00 & 477 & 38.6 & 27.3 & 34.2 \\
\texttt{insp\_causal} & 23.58 & 486 & 32.3 & 39.5 & 28.2 \\
\texttt{insp\_cf} & 31.76 & 434 & 38.7 & 37.1 & 24.2 \\
\texttt{insp\_temporal} & 12.74 & 555 & 27.4 & 29.5 & 43.1 \\
\texttt{navi\_causal} & 33.33 & 424 & 43.6 & 37.0 & 19.3 \\
\texttt{navi\_temporal} & 30.82 & 440 & 32.5 & 42.5 & 25.0 \\
\texttt{mani\_causal} & 41.19 & 374 & 38.2 & 38.2 & 23.5 \\
\texttt{mani\_cf} & 33.18 & 425 & 40.2 & 34.6 & 25.2 \\
\texttt{exo\_physical} & 33.02 & 426 & 37.3 & 37.6 & 25.1 \\
\texttt{exo\_temporal} & 28.93 & 452 & 36.5 & 51.3 & 12.2 \\
\bottomrule
\end{tabular}
\end{adjustbox}
\endgroup
\end{table}

\begin{table}[!htbp]
\centering
\caption{Temporal-adjacency errors on held-out scenes (OOD, $n=636$). Accuracy is in percent; Errors counts incorrect answers. Successor and predecessor off-by-one errors extend one step beyond the correct neighbor; Swapped reverses before and after. Error shares (\%) sum to approximately 100, allowing for rounding. Panels report SFT and RL separately; --- denotes unavailable full-mix RL results.}
\label{tab:own-E6b}
\begingroup
\footnotesize
\setlength{\tabcolsep}{3pt}
\renewcommand{\arraystretch}{1.04}
\begin{adjustbox}{max width=\linewidth}
\begin{tabular}{@{}lrrrrr@{}}
\toprule
\multicolumn{6}{l}{\textbf{SFT}} \\
\addlinespace[2pt]
Config. & \shortstack{Accuracy\\(\%)} & Errors & \shortstack{Successor\\off-by-one} & \shortstack{Predecessor\\off-by-one} & Swapped \\
\midrule
base & 27.04 & 464 & 29.3 & 30.8 & 39.9 \\
\addlinespace[1pt]
\texttt{full-mix} & 64.31 & 227 & 0.0 & 4.8 & 95.2 \\
\texttt{perc\_causal} & 21.38 & 500 & 35.4 & 32.4 & 32.2 \\
\texttt{perc\_temporal} & 41.19 & 374 & 1.6 & 7.2 & 91.2 \\
\texttt{insp\_causal} & 20.60 & 505 & 23.8 & 39.4 & 36.8 \\
\texttt{insp\_cf} & 30.66 & 441 & 24.9 & 32.9 & 42.2 \\
\texttt{insp\_temporal} & 44.18 & 355 & 1.1 & 3.9 & 94.9 \\
\texttt{navi\_causal} & 31.60 & 435 & 23.4 & 25.7 & 50.8 \\
\texttt{navi\_temporal} & 43.55 & 359 & 0.0 & 1.4 & 98.6 \\
\texttt{mani\_causal} & 32.39 & 430 & 22.3 & 23.0 & 54.7 \\
\texttt{mani\_cf} & 37.11 & 400 & 23.2 & 23.2 & 53.5 \\
\texttt{exo\_physical} & 27.20 & 463 & 37.4 & 28.9 & 33.7 \\
\texttt{exo\_temporal} & 52.99 & 299 & 0.0 & 2.3 & 97.7 \\
\midrule
\addlinespace[3pt]
\multicolumn{6}{l}{\textbf{RL}} \\
\addlinespace[2pt]
Config. & \shortstack{Accuracy\\(\%)} & Errors & \shortstack{Successor\\off-by-one} & \shortstack{Predecessor\\off-by-one} & Swapped \\
\midrule
base & 27.04 & 464 & 29.3 & 30.8 & 39.9 \\
\addlinespace[1pt]
\texttt{full-mix} & --- & --- & --- & --- & --- \\
\texttt{perc\_causal} & 19.50 & 512 & 36.9 & 31.1 & 32.0 \\
\texttt{perc\_temporal} & 45.28 & 348 & 1.7 & 4.0 & 94.3 \\
\texttt{insp\_causal} & 19.81 & 510 & 25.9 & 36.5 & 37.6 \\
\texttt{insp\_cf} & 27.04 & 464 & 24.1 & 31.7 & 44.2 \\
\texttt{insp\_temporal} & 36.48 & 404 & 0.7 & 3.7 & 95.5 \\
\texttt{navi\_causal} & 36.64 & 403 & 19.9 & 18.4 & 61.8 \\
\texttt{navi\_temporal} & 45.60 & 346 & 0.6 & 0.0 & 99.4 \\
\texttt{mani\_causal} & 39.62 & 384 & 20.6 & 18.2 & 61.2 \\
\texttt{mani\_cf} & 39.78 & 383 & 19.1 & 17.5 & 63.4 \\
\texttt{exo\_physical} & 28.93 & 452 & 36.1 & 29.4 & 34.5 \\
\texttt{exo\_temporal} & 56.60 & 276 & 0.4 & 0.4 & 99.3 \\
\bottomrule
\end{tabular}
\end{adjustbox}
\endgroup
\end{table}

\FloatBarrier

\begin{table}[!htbp]
\centering
\caption{Action-confusion errors in forward dynamics on perceptive and inspective ID items ($n=558$). Options are outcome images. Mirror and Reverse select outcomes of the left/right or forward/backward counterpart; Other selects another action's outcome; Static selects the unchanged scene. Accuracy and error shares are percentages; Errors counts incorrect answers. Panels report SFT and RL separately; --- denotes unavailable full-mix RL results.}
\label{tab:own-E7a}
\begingroup
\footnotesize
\setlength{\tabcolsep}{3pt}
\renewcommand{\arraystretch}{1.04}
\begin{adjustbox}{max width=\linewidth}
\begin{tabular}{@{}lrrrrrr@{}}
\toprule
\multicolumn{7}{l}{\textbf{SFT}} \\
\addlinespace[2pt]
Config. & \shortstack{Accuracy\\(\%)} & Errors & \shortstack{Mirror\\(L/R)} & \shortstack{Reverse\\(F/B)} & \shortstack{Other\\action} & \shortstack{Static\\option} \\
\midrule
base & 24.37 & 422 & 15.4 & 10.7 & 58.3 & 15.6 \\
\addlinespace[1pt]
\texttt{full-mix} & 97.31 & 15 & 73.3 & 0.0 & 26.7 & 0.0 \\
\texttt{perc\_causal} & 71.33 & 160 & 40.6 & 0.0 & 51.9 & 7.5 \\
\texttt{perc\_temporal} & 24.19 & 423 & 15.1 & 8.5 & 64.3 & 12.1 \\
\texttt{insp\_causal} & 70.43 & 165 & 22.4 & 7.3 & 69.7 & 0.6 \\
\texttt{insp\_cf} & 71.51 & 159 & 40.9 & 11.3 & 40.9 & 6.9 \\
\texttt{insp\_temporal} & 24.73 & 420 & 16.0 & 8.6 & 60.0 & 15.5 \\
\texttt{navi\_causal} & 37.63 & 348 & 13.5 & 8.3 & 64.7 & 13.5 \\
\texttt{navi\_temporal} & 25.09 & 418 & 16.3 & 12.2 & 55.7 & 15.8 \\
\texttt{mani\_causal} & 30.11 & 390 & 9.2 & 11.5 & 48.7 & 30.5 \\
\texttt{mani\_cf} & 33.87 & 369 & 14.9 & 12.7 & 50.9 & 21.4 \\
\texttt{exo\_physical} & 29.75 & 392 & 20.4 & 18.9 & 59.7 & 1.0 \\
\texttt{exo\_temporal} & 24.19 & 423 & 16.8 & 11.8 & 53.9 & 17.5 \\
\midrule
\addlinespace[3pt]
\multicolumn{7}{l}{\textbf{RL}} \\
\addlinespace[2pt]
Config. & \shortstack{Accuracy\\(\%)} & Errors & \shortstack{Mirror\\(L/R)} & \shortstack{Reverse\\(F/B)} & \shortstack{Other\\action} & \shortstack{Static\\option} \\
\midrule
base & 24.37 & 422 & 15.4 & 10.7 & 58.3 & 15.6 \\
\addlinespace[1pt]
\texttt{full-mix} & --- & --- & --- & --- & --- & --- \\
\texttt{perc\_causal} & 70.61 & 164 & 52.4 & 0.0 & 39.0 & 8.5 \\
\texttt{perc\_temporal} & 22.04 & 435 & 14.9 & 7.8 & 60.9 & 16.3 \\
\texttt{insp\_causal} & 68.82 & 174 & 18.4 & 5.7 & 75.9 & 0.0 \\
\texttt{insp\_cf} & 71.68 & 158 & 31.6 & 20.9 & 38.0 & 9.5 \\
\texttt{insp\_temporal} & 24.01 & 424 & 17.2 & 10.6 & 55.9 & 16.3 \\
\texttt{navi\_causal} & 39.25 & 339 & 13.3 & 6.8 & 64.9 & 15.0 \\
\texttt{navi\_temporal} & 23.12 & 429 & 17.5 & 9.3 & 55.0 & 18.2 \\
\texttt{mani\_causal} & 33.15 & 373 & 10.7 & 11.0 & 51.7 & 26.5 \\
\texttt{mani\_cf} & 38.35 & 344 & 16.6 & 12.2 & 53.5 & 17.7 \\
\texttt{exo\_physical} & 30.65 & 387 & 20.7 & 18.1 & 60.2 & 1.0 \\
\texttt{exo\_temporal} & 24.19 & 423 & 16.1 & 10.2 & 55.8 & 18.0 \\
\bottomrule
\end{tabular}
\end{adjustbox}
\endgroup
\end{table}

\begin{table}[!htbp]
\centering
\caption{Action-confusion errors in inverse dynamics on perceptive and inspective ID items ($n=558$). Options are action descriptions, not outcome images. Mirror and Reverse select the left/right or forward/backward counterpart; Other selects another action; Static denotes the no-action choice. Accuracy and error shares are percentages; Errors counts incorrect answers. Panels report SFT and RL separately; --- denotes unavailable full-mix RL results.}
\label{tab:own-E7b}
\begingroup
\footnotesize
\setlength{\tabcolsep}{3pt}
\renewcommand{\arraystretch}{1.04}
\begin{adjustbox}{max width=\linewidth}
\begin{tabular}{@{}lrrrrrr@{}}
\toprule
\multicolumn{7}{l}{\textbf{SFT}} \\
\addlinespace[2pt]
Config. & \shortstack{Accuracy\\(\%)} & Errors & \shortstack{Mirror\\(L/R)} & \shortstack{Reverse\\(F/B)} & \shortstack{Other\\action} & \shortstack{Static\\option} \\
\midrule
base & 23.30 & 428 & 1.6 & 15.9 & 35.3 & 47.2 \\
\addlinespace[1pt]
\texttt{full-mix} & 94.62 & 30 & 73.3 & 0.0 & 26.7 & 0.0 \\
\texttt{perc\_causal} & 72.04 & 156 & 44.2 & 1.3 & 48.1 & 6.4 \\
\texttt{perc\_temporal} & 23.48 & 427 & 3.0 & 18.0 & 36.5 & 42.4 \\
\texttt{insp\_causal} & 68.46 & 176 & 23.3 & 14.2 & 62.5 & 0.0 \\
\texttt{insp\_cf} & 24.01 & 424 & 0.0 & 17.9 & 33.7 & 48.3 \\
\texttt{insp\_temporal} & 23.12 & 429 & 6.8 & 14.0 & 37.1 & 42.2 \\
\texttt{navi\_causal} & 39.43 & 338 & 12.1 & 13.6 & 53.3 & 21.0 \\
\texttt{navi\_temporal} & 24.91 & 419 & 9.1 & 14.3 & 34.8 & 41.8 \\
\texttt{mani\_causal} & 35.13 & 362 & 10.8 & 7.5 & 53.6 & 28.2 \\
\texttt{mani\_cf} & 22.76 & 431 & 0.5 & 12.5 & 32.7 & 54.3 \\
\texttt{exo\_physical} & 27.96 & 402 & 2.5 & 13.7 & 49.3 & 34.6 \\
\texttt{exo\_temporal} & 29.03 & 396 & 18.7 & 15.7 & 40.2 & 25.5 \\
\midrule
\addlinespace[3pt]
\multicolumn{7}{l}{\textbf{RL}} \\
\addlinespace[2pt]
Config. & \shortstack{Accuracy\\(\%)} & Errors & \shortstack{Mirror\\(L/R)} & \shortstack{Reverse\\(F/B)} & \shortstack{Other\\action} & \shortstack{Static\\option} \\
\midrule
base & 23.30 & 428 & 1.6 & 15.9 & 35.3 & 47.2 \\
\addlinespace[1pt]
\texttt{full-mix} & --- & --- & --- & --- & --- & --- \\
\texttt{perc\_causal} & 71.51 & 159 & 50.9 & 3.1 & 39.0 & 6.9 \\
\texttt{perc\_temporal} & 25.27 & 417 & 4.6 & 18.2 & 38.6 & 38.6 \\
\texttt{insp\_causal} & 65.23 & 194 & 20.1 & 17.5 & 62.4 & 0.0 \\
\texttt{insp\_cf} & 29.39 & 394 & 0.3 & 18.0 & 36.0 & 45.7 \\
\texttt{insp\_temporal} & 25.27 & 417 & 7.9 & 16.1 & 34.8 & 41.2 \\
\texttt{navi\_causal} & 49.64 & 281 & 18.1 & 12.8 & 59.1 & 10.0 \\
\texttt{navi\_temporal} & 36.92 & 352 & 23.0 & 13.4 & 41.5 & 22.2 \\
\texttt{mani\_causal} & 36.02 & 357 & 14.3 & 10.1 & 52.9 & 22.7 \\
\texttt{mani\_cf} & 21.86 & 436 & 0.2 & 14.4 & 33.3 & 52.1 \\
\texttt{exo\_physical} & 27.96 & 402 & 1.2 & 15.9 & 48.8 & 34.1 \\
\texttt{exo\_temporal} & 26.88 & 408 & 4.9 & 13.5 & 48.3 & 33.3 \\
\bottomrule
\end{tabular}
\end{adjustbox}
\endgroup
\end{table}

\begin{table}[!htbp]
\centering
\caption{Action-confusion errors in counterfactual substitution on inspective ID items only ($n=270$). Options are outcome images. Mirror selects the opposite left/right orbit, Other another action's outcome, and Static the unchanged scene. The forward/backward Reverse column is retained as supplied. Accuracy and error shares are percentages; Errors counts incorrect answers. Panels report SFT and RL separately; --- denotes unavailable full-mix RL results.}
\label{tab:own-E7c}
\begingroup
\footnotesize
\setlength{\tabcolsep}{3pt}
\renewcommand{\arraystretch}{1.04}
\begin{adjustbox}{max width=\linewidth}
\begin{tabular}{@{}lrrrrrr@{}}
\toprule
\multicolumn{7}{l}{\textbf{SFT}} \\
\addlinespace[2pt]
Config. & \shortstack{Accuracy\\(\%)} & Errors & \shortstack{Mirror\\(L/R)} & \shortstack{Reverse\\(F/B)} & \shortstack{Other\\action} & \shortstack{Static\\option} \\
\midrule
base & 18.52 & 220 & 8.2 & 0.0 & 61.4 & 30.5 \\
\addlinespace[1pt]
\texttt{full-mix} & 94.81 & 14 & 14.3 & 0.0 & 35.7 & 50.0 \\
\texttt{perc\_causal} & 47.78 & 141 & 7.1 & 0.0 & 53.9 & 39.0 \\
\texttt{perc\_temporal} & 20.74 & 214 & 8.9 & 0.0 & 59.3 & 31.8 \\
\texttt{insp\_causal} & 80.00 & 54 & 0.0 & 0.0 & 13.0 & 87.0 \\
\texttt{insp\_cf} & 93.70 & 17 & 0.0 & 0.0 & 82.4 & 17.6 \\
\texttt{insp\_temporal} & 19.26 & 218 & 8.3 & 0.0 & 62.8 & 28.9 \\
\texttt{navi\_causal} & 34.81 & 176 & 4.0 & 0.0 & 48.9 & 47.2 \\
\texttt{navi\_temporal} & 18.52 & 220 & 8.6 & 0.0 & 59.5 & 31.8 \\
\texttt{mani\_causal} & 28.52 & 193 & 3.6 & 0.0 & 70.5 & 25.9 \\
\texttt{mani\_cf} & 38.89 & 165 & 4.2 & 0.0 & 40.0 & 55.8 \\
\texttt{exo\_physical} & 11.85 & 238 & 11.8 & 0.0 & 38.2 & 50.0 \\
\texttt{exo\_temporal} & 18.52 & 220 & 9.1 & 0.0 & 59.5 & 31.4 \\
\midrule
\addlinespace[3pt]
\multicolumn{7}{l}{\textbf{RL}} \\
\addlinespace[2pt]
Config. & \shortstack{Accuracy\\(\%)} & Errors & \shortstack{Mirror\\(L/R)} & \shortstack{Reverse\\(F/B)} & \shortstack{Other\\action} & \shortstack{Static\\option} \\
\midrule
base & 18.52 & 220 & 8.2 & 0.0 & 61.4 & 30.5 \\
\addlinespace[1pt]
\texttt{full-mix} & --- & --- & --- & --- & --- & --- \\
\texttt{perc\_causal} & 54.44 & 123 & 7.3 & 0.0 & 59.3 & 33.3 \\
\texttt{perc\_temporal} & 17.04 & 224 & 8.5 & 0.0 & 58.9 & 32.6 \\
\texttt{insp\_causal} & 86.30 & 37 & 0.0 & 0.0 & 16.2 & 83.8 \\
\texttt{insp\_cf} & 95.19 & 13 & 0.0 & 0.0 & 53.8 & 46.2 \\
\texttt{insp\_temporal} & 14.81 & 230 & 13.0 & 0.0 & 56.1 & 30.9 \\
\texttt{navi\_causal} & 36.30 & 172 & 4.1 & 0.0 & 49.4 & 46.5 \\
\texttt{navi\_temporal} & 16.30 & 226 & 8.0 & 0.0 & 59.3 & 32.7 \\
\texttt{mani\_causal} & 35.19 & 175 & 3.4 & 0.0 & 71.4 & 25.1 \\
\texttt{mani\_cf} & 43.33 & 153 & 5.9 & 0.0 & 45.8 & 48.4 \\
\texttt{exo\_physical} & 12.59 & 236 & 11.0 & 0.0 & 37.3 & 51.7 \\
\texttt{exo\_temporal} & 15.19 & 229 & 14.4 & 0.0 & 58.1 & 27.5 \\
\bottomrule
\end{tabular}
\end{adjustbox}
\endgroup
\end{table}

\FloatBarrier

\FloatBarrier
\subsubsection{Question-dependent state judgments and the scene generalization gap}

\textbf{Shortcut check for the default-state result.} A deflationary reading of Table~\ref{tab:static-pull} is that causal supervision, whose correct options are always changed frames, merely teaches a surface rule, \emph{prefer the changed frame}, which would lower the static pull as a side effect. The counterfactual-removal probes separate the two readings: their correct option is exactly the unchanged scene, so a blanket changed-frame rule would suppress it and push accuracy \emph{below} base. At the SFT stage, the opposite happens: every causally-trained cell improves counterfactual-removal accuracy over base ($18.6\% \to 28.9/41.2/49.6/48.4$ for \texttt{perc/insp/navi/mani\_causal}) while simultaneously cutting the static pull on forward probes. The same weights move the unchanged-scene option in opposite directions depending on what the question asks, the signature of a removed prior with question-conditioned state selection.

Table~\ref{tab:static-pull} reports both stages and includes counterfactual substitution and removal alongside the forward/inverse probes. Table~\ref{tab:tadj-comp} gives the corresponding temporal-adjacency error analysis. Table~\ref{tab:app-geo} reports absolute accuracies under geometric and appearance perturbations, with the base row providing the reference for gains.

\textbf{Competency-matched scene gap.} Table~\ref{tab:ood-gap} reports the SFT comparison that averages OOD-minus-ID accuracy differences equally across reasoning types. This statistic differs from subtracting the overall accuracies in Tables~\ref{tab:own-T1} and~\ref{tab:own-T3}, which weight individual test items equally. Temporal-ordering distractors also differ between the two tests: ID uses random permutations, whereas OOD uses structured swaps. Their error distributions are therefore reported separately in Tables~\ref{tab:own-E2} and~\ref{tab:own-E6a}.

\begin{table}[h]
\centering
\caption{\textbf{Scene-OOD minus ID accuracy gap by supervision} (competency-matched mean, pp).}
\label{tab:ood-gap}
\begin{adjustbox}{max width=0.42\linewidth}
\begin{tabular}{lc}
\toprule
Run & OOD $-$ ID (pp) \\
\midrule
base & ${+}2.06$ \\
full-mix & ${+}1.21$ \\
perc\_causal & ${-}1.45$ \\
perc\_temporal & ${-}0.50$ \\
insp\_causal & ${+}0.33$ \\
insp\_cf & ${+}1.51$ \\
insp\_temporal & ${-}1.17$ \\
navi\_causal & ${+}0.05$ \\
navi\_temporal & ${+}1.66$ \\
mani\_causal & ${+}6.87$ \\
mani\_cf & ${+}4.48$ \\
exo\_physical & ${+}5.05$ \\
exo\_temporal & ${+}1.57$ \\
\bottomrule
\end{tabular}
\end{adjustbox}
\end{table}

\FloatBarrier

\section{Controls for the transfer results}
\label{app:controls}

The three controls below each vary one aspect of training or evaluation: the transition content of the training data with its format kept fixed (Appendix~\ref{app:content-control}), the scoring of answer letters (Appendix~\ref{app:letter-control}), and the unchanged-scene option in cross-operation transfer (Appendix~\ref{app:static-option}). The experiments in Appendices~\ref{app:controls}, \ref{app:gemma}, and~\ref{app:recipe-prospective} re-evaluate every model, including the base model, with one common script (greedy decoding with vLLM, with the model answering with the option letter only), so their base scores differ from those in the main text.

\subsection{Content control: transition content versus training format}
\label{app:content-control}

The downstream gains of WOVEN training come from learned transition reasoning: they appear only when the model acquires the correct action--outcome relations, not from the question format, the answer-letter balance, or the amount of training. Training on WOVEN teaches more than transitions: answering four-option questions, spreading answers over the letters, reading several images with images as options, and simply more training. We therefore train on four data sets that match a WOVEN subset in everything except that content and compare them with the real data (Table~\ref{tab:cc-arms}). The real data are the perceptive causal-dynamics subset. Two controls keep its items, images, options, and answer-letter balance and change only the action--outcome mapping: a per-source permutation maps the four camera actions to the outcomes by a random permutation without fixed points within each source, so that no consistent rule can be learned (the answer changes in $95.1\%$ of the examples), and a global mirror reverses the direction of every action under one mapping for all sources (the answer changes in $95.0\%$), a consistent rule with the wrong semantics. A third control trains on A-OKVQA multiple-choice questions~\citep{schwenk2022aokvqa} with a balanced answer-letter distribution and no WOVEN content. If the gains came from the format or from the amount of training, the controls would gain as much as the real data; if they require the transition content, only the real data gains, which is what we observe.

\begin{table}[!htbp]
\centering
\footnotesize
\setlength{\tabcolsep}{4pt}
\caption{Content control: the four training sets ($9{,}750$ training examples each, same number of optimizer steps).}
\label{tab:cc-arms}
\begin{tabular}{@{}p{2.0cm}p{6.6cm}p{3.9cm}@{}}
\toprule
Arm & Training data & Relation to the real data \\
\midrule
Real & WOVEN perceptive $\times$ causal dynamics (forward and inverse dynamics questions, four option orderings). & --- \\
Per-source permutation & Same items, images, options, and answer-letter balance. Within each source, the four camera actions are mapped to the outcomes by a random permutation with no fixed point: the forward question for action $a_i$ has the outcome of $a_{\pi(i)}$ as its answer, and the inverse question showing outcome $s'_k$ has $a_{\pi^{-1}(k)}$ as its answer. The answer changes in $95.1\%$ of the examples. & Identical format; the action--outcome mapping differs across sources, so no consistent rule can be learned. \\
Global mirror & Same items, images, options, and answer-letter balance. One mapping for all sources: move forward $\leftrightarrow$ move backward, rotate left $45^\circ$ $\leftrightarrow$ rotate right $45^\circ$. The answer changes in $95.0\%$ of the examples. & Identical format; a consistent, learnable rule with the direction of every action reversed. \\
A-OKVQA & $2{,}072$ four-option questions from the A-OKVQA training set~\citep{schwenk2022aokvqa} (COCO images), each in its four option orderings, plus $1{,}462$ additional random orderings. & Multiple-choice training with a balanced answer-letter distribution, without WOVEN content. \\
\bottomrule
\end{tabular}
\end{table}

All four training sets have $9{,}750$ examples and the same number of optimizer steps. Each is trained on Qwen2.5-VL-3B-Instruct with LoRA (rank $16$, $\alpha=32$, dropout $0.05$) on the language model's q/k/v/o projections, vision encoder frozen, vision--language merger trained, learning rate $10^{-4}$, $5\%$ warmup, cosine schedule, effective batch size $64$, $3$ epochs, final checkpoint, with $3$ seeds ($12$ models). Evaluation uses greedy decoding (vLLM) with the model asked to output only the option letter, on SAT~\citep{ray2024sat} ($150$ two-option questions, scored circularly: each question is asked once with each option order and counts as correct only if both answers are right), SPAR-Bench view change~\citep{zhang2025spar} on ScanNet and ScanNet++ ($261$ questions with two real images and ground-truth camera poses, asking how the camera moved between them, the format of WOVEN's perceptive inverse-dynamics questions on real images; the numeric questions are converted to four options as in Appendix~\ref{app:recipe-prospective}), ActionEQA~\citep{actioneqa2025} ($800$ questions) and WorldPrediction~\citep{worldprediction2025} ($825$ questions), both scored circularly (each question is asked four times with the options rotated so that the correct option appears at every letter once, and counts as correct only if all four answers are right, so a preference for any answer letter earns no credit), and WOVEN OOD (the $3{,}496$ held-out-scene questions). For each question, the score of an arm is averaged over its $3$ seeds; two arms are compared by a paired sign-flip permutation test over questions, with a bootstrap $95\%$ confidence interval of the difference.

\begin{table}[!htbp]
\centering
\footnotesize
\setlength{\tabcolsep}{4pt}
\caption{Content control: accuracy (\%), mean over $3$ seeds.}
\label{tab:cc-acc}
\begin{tabular}{@{}lrrrrr@{}}
\toprule
Benchmark & Base & Real & Permutation & Mirror & A-OKVQA \\
\midrule
SAT (circular) & 39.3 & \textbf{63.3} & 35.1 & 37.8 & 44.0 \\
SPAR-Bench view change (real images) & 26.4 & \textbf{54.9} & 27.5 & 35.8 & 21.8 \\
ActionEQA (circular) & 25.6 & \textbf{31.8} & 22.2 & 26.5 & 22.6 \\
WorldPrediction (circular) & 14.7 & \textbf{16.6} & 12.4 & 11.3 & 10.3 \\
WOVEN OOD & 36.2 & \textbf{45.9} & 36.8 & 32.9 & 32.6 \\
\bottomrule
\end{tabular}
\end{table}

\begin{table}[!htbp]
\centering
\footnotesize
\caption{Content control: Real minus control, percentage points with bootstrap $95\%$ confidence intervals; every difference has $p<0.001$ under the paired permutation test.}
\label{tab:cc-diff}
\begin{tabular}{@{}lrrr@{}}
\toprule
Benchmark & Real $-$ Permutation & Real $-$ Mirror & Real $-$ A-OKVQA \\
\midrule
SAT (circular) & $+28.2$ $[+21.1, +35.8]$ & $+25.6$ $[+18.2, +33.3]$ & $+19.3$ $[+11.8, +26.9]$ \\
SPAR-Bench view change & $+27.5$ $[+20.7, +34.5]$ & $+19.2$ $[+12.4, +25.5]$ & $+33.1$ $[+26.4, +39.3]$ \\
ActionEQA (circular) & $+9.6$ $[+7.0, +12.3]$ & $+5.3$ $[+3.5, +7.2]$ & $+9.2$ $[+6.7, +11.8]$ \\
WorldPrediction (circular) & $+4.2$ $[+1.9, +6.3]$ & $+5.3$ $[+3.6, +7.1]$ & $+6.3$ $[+4.0, +8.6]$ \\
WOVEN OOD & $+9.1$ $[+7.2, +11.0]$ & $+13.0$ $[+11.4, +14.5]$ & $+13.3$ $[+11.4, +15.3]$ \\
\bottomrule
\end{tabular}
\end{table}

\begin{table}[!htbp]
\centering
\footnotesize
\caption{Content control: per-seed accuracy (\%).}
\label{tab:cc-seeds}
\begin{tabular}{@{}llrrrr@{}}
\toprule
Benchmark & Arm & s1 & s2 & s3 & Mean \\
\midrule
SAT (circular) & Real & 65.3 & 63.3 & 61.3 & 63.3 \\
 & Per-source permutation & 35.3 & 36.7 & 33.3 & 35.1 \\
 & Global mirror & 37.3 & 40.0 & 36.0 & 37.8 \\
 & A-OKVQA & 42.7 & 44.7 & 44.7 & 44.0 \\
SPAR-Bench view change & Real & 56.3 & 55.2 & 53.3 & 54.9 \\
 & Per-source permutation & 29.9 & 25.7 & 26.8 & 27.5 \\
 & Global mirror & 42.9 & 39.5 & 24.9 & 35.8 \\
 & A-OKVQA & 25.3 & 19.2 & 21.1 & 21.8 \\
ActionEQA (circular) & Real & 32.2 & 31.8 & 31.5 & 31.8 \\
 & Per-source permutation & 22.2 & 22.2 & 22.1 & 22.2 \\
 & Global mirror & 27.8 & 26.6 & 25.2 & 26.5 \\
 & A-OKVQA & 22.1 & 22.6 & 23.0 & 22.6 \\
WorldPrediction (circular) & Real & 15.4 & 19.5 & 14.9 & 16.6 \\
 & Per-source permutation & 12.2 & 12.1 & 13.0 & 12.4 \\
 & Global mirror & 9.0 & 8.1 & 16.7 & 11.3 \\
 & A-OKVQA & 9.0 & 11.2 & 10.7 & 10.3 \\
WOVEN OOD & Real & 45.5 & 46.3 & 45.9 & 45.9 \\
 & Per-source permutation & 37.3 & 37.4 & 35.8 & 36.8 \\
 & Global mirror & 34.8 & 35.2 & 28.7 & 32.9 \\
 & A-OKVQA & 32.6 & 33.3 & 31.8 & 32.6 \\
\bottomrule
\end{tabular}
\end{table}

Only the real data gains (Tables~\ref{tab:cc-acc}--\ref{tab:cc-seeds}). The two matched controls fall $4.2$ to $28.2$ points below the real arm on every benchmark (all $p<0.001$; Table~\ref{tab:cc-diff}), so the multiple-choice format, the answer-letter balance, multi-image input with image options, and the amount of training do not by themselves produce the gains; under circular scoring, where a letter preference earns no credit, the real arm stays above all controls. The A-OKVQA arm stays below the real arm on every benchmark and below the base model on three, so general multiple-choice training does not produce the gains either. The mirror arm learns its reversed rule: on the perceptive forward and inverse dynamics questions of the WOVEN ID test set ($288$ each, from sources not seen in training), it answers $85$--$92\%$ of the inverse questions by the reversed rule, whereas the permutation arm stays near chance because its mapping does not carry over to a new source (Table~\ref{tab:cc-manip}). The mirror arm nevertheless obtains at most a fraction of the gains, $9.4$ points above the base model on SPAR-Bench and $0.9$ on ActionEQA and below it on the other three benchmarks, so the gains require the correct action semantics and not only a consistent rule. On SPAR-Bench, real images in WOVEN's inverse-dynamics format, the real arm doubles the base accuracy ($26.4 \to 54.9$) while the permutation arm ($27.5$) and the A-OKVQA arm ($21.8$) stay at or below the base model and the mirror arm reaches $35.8$, so what is learned from the generated rollouts carries over to real images. Several controls fall below the base model (all three on WorldPrediction, the mirror and A-OKVQA arms on WOVEN OOD, the A-OKVQA arm on SPAR-Bench), so part of each difference in Table~\ref{tab:cc-diff} comes from the control falling below the base model rather than from the real arm rising above it. For the perceptive causal-dynamics subset, the external gains therefore come from learned transition reasoning: they require the correct action--outcome relations and are not produced by the format or the amount of training.

\begin{table}[!htbp]
\centering
\scriptsize
\setlength{\tabcolsep}{4pt}
\caption{Content control, manipulation check: answers on WOVEN ID perceptive dynamics questions (\%), per seed.}
\label{tab:cc-manip}
\begin{tabular}{@{}lllll@{}}
\toprule
Arm & Forward: real answer & Forward: mirrored answer & Inverse: real answer & Inverse: mirrored answer \\
\midrule
Base & 24.7 & 27.1 & 25.3 & 28.8 \\
Real & 97.2 / 98.6 / 95.8 & 1.0 / 0.3 / 1.7 & 89.2 / 93.1 / 85.8 & 6.6 / 3.8 / 6.2 \\
Per-source permutation & 21.5 / 24.3 / 21.9 & 18.4 / 20.5 / 21.9 & 22.9 / 24.3 / 22.2 & 31.2 / 29.2 / 32.3 \\
Global mirror & 37.2 / 42.0 / 20.5 & 26.7 / 32.3 / 71.2 & 5.9 / 8.3 / 5.6 & 87.2 / 84.7 / 92.4 \\
A-OKVQA & 24.7 / 25.0 / 22.6 & 25.0 / 25.7 / 26.7 & 31.9 / 33.3 / 25.3 & 29.5 / 30.2 / 31.9 \\
\bottomrule
\end{tabular}
\end{table}

\subsection{Letter-balance control: circular scoring}
\label{app:letter-control}

The external gains of WOVEN training remain when a letter preference earns no credit. SFT uses all four option orderings of every question and the GRPO reward penalizes an uneven answer-letter distribution (Appendix~\ref{app:E-recipes}), so WOVEN-trained models spread their answers over the letters more evenly than the base model, which prefers some letters (Table~\ref{tab:lb-letters}); on a benchmark whose base score is lowered by a letter preference, this alone raises accuracy. We therefore re-score with circular evaluation: each question is asked once per option rotation, so that the correct option appears at every letter once, and it counts as correct only if every rotation is answered correctly, so a model that answers without knowing the answer scores at most $(1/n)^n$ per question ($0.4\%$ for four options, $25\%$ for two) whatever its letter distribution. A gain that remains under circular scoring comes from the content of the options.

We re-score the base model and four WOVEN-trained models from the main experiments, the full mixture after SFT and after SFT + GRPO and the perceptive causal-dynamics subset after SFT and after SFT + GRPO, on one benchmark per domain: SAT~\citep{ray2024sat} (a $150$-question two-option subset), ActionEQA~\citep{actioneqa2025} ($800$ four-option questions), and SPAR-Bench view change~\citep{zhang2025spar} on ScanNet and ScanNet++ ($261$ four-option questions on real images, converted from the numeric questions as in Appendix~\ref{app:recipe-prospective}). Decoding is greedy (vLLM) with the model asked to output only the option letter; standard scoring asks each question once with its original option order (for SAT, the accuracy over the two orders of each question). Because SAT and ActionEQA are evaluated on these subsets, the standard-scoring base scores differ from those in Table~\ref{tab:enhance-transfer}. Each trained model is compared with the base model by a paired sign-flip permutation test over questions with a bootstrap $95\%$ confidence interval of the difference.

\begin{table}[!htbp]
\centering
\footnotesize
\setlength{\tabcolsep}{4pt}
\caption{Letter-balance control: answer-letter distribution (\% of answers), standard scoring.}
\label{tab:lb-letters}
\begin{tabular}{@{}llrrrr@{}}
\toprule
Benchmark & Model & A & B & C & D \\
\midrule
SAT & Correct answers & 50.0 & 50.0 & & \\
 & Base & 62.0 & 38.0 & & \\
 & Full mixture, SFT & 51.0 & 49.0 & & \\
 & Full mixture, SFT + GRPO & 49.0 & 51.0 & & \\
 & Perceptive $\times$ causal dynamics, SFT & 52.3 & 47.7 & & \\
 & Perceptive $\times$ causal dynamics, SFT + GRPO & 55.0 & 45.0 & & \\
\midrule
ActionEQA & Correct answers & 24.2 & 24.0 & 26.5 & 25.2 \\
 & Base & 19.5 & 32.0 & 30.0 & 18.5 \\
 & Full mixture, SFT & 24.8 & 21.1 & 28.0 & 26.1 \\
 & Full mixture, SFT + GRPO & 22.8 & 21.8 & 31.6 & 23.9 \\
 & Perceptive $\times$ causal dynamics, SFT & 24.1 & 23.5 & 28.5 & 23.9 \\
 & Perceptive $\times$ causal dynamics, SFT + GRPO & 25.2 & 19.5 & 28.0 & 27.2 \\
\midrule
SPAR-Bench view change & Correct answers & 28.7 & 23.0 & 23.0 & 25.3 \\
 & Base & 64.0 & 14.2 & 17.6 & 4.2 \\
 & Full mixture, SFT & 25.7 & 24.1 & 27.2 & 23.0 \\
 & Full mixture, SFT + GRPO & 19.9 & 24.9 & 31.4 & 23.8 \\
 & Perceptive $\times$ causal dynamics, SFT & 35.2 & 23.4 & 21.5 & 19.9 \\
 & Perceptive $\times$ causal dynamics, SFT + GRPO & 31.8 & 27.6 & 21.5 & 19.2 \\
\bottomrule
\end{tabular}
\end{table}

\begin{table}[!htbp]
\centering
\footnotesize
\setlength{\tabcolsep}{4pt}
\caption{Letter-balance control: accuracy (\%) under standard and circular scoring.}
\label{tab:lb-acc}
\begin{tabular}{@{}llrrrrr@{}}
\toprule
Benchmark & Scoring & Base & \shortstack[r]{Full mixture,\\SFT} & \shortstack[r]{Full mixture,\\SFT + GRPO} & \shortstack[r]{Perc.\ $\times$ causal,\\SFT} & \shortstack[r]{Perc.\ $\times$ causal,\\SFT + GRPO} \\
\midrule
SAT & Standard & 57.3 & 68.3 & 67.7 & 69.7 & 66.3 \\
 & Circular & 39.3 & 62.7 & 61.3 & 64.7 & 58.7 \\
ActionEQA & Standard & 43.1 & 50.2 & 49.0 & 49.5 & 52.0 \\
 & Circular & 25.6 & 35.4 & 34.2 & 32.8 & 35.5 \\
SPAR-Bench view change & Standard & 26.4 & 41.8 & 39.5 & 54.8 & 55.9 \\
 & Circular & 0.4 & 20.3 & 19.5 & 30.7 & 36.0 \\
\bottomrule
\end{tabular}
\end{table}

\begin{table}[!htbp]
\centering
\scriptsize
\setlength{\tabcolsep}{3pt}
\caption{Letter-balance control: trained minus base, percentage points with bootstrap $95\%$ confidence intervals. Every difference is significant under the paired permutation test ($p \le 0.006$ under standard scoring; $p<0.001$ under circular scoring).}
\label{tab:lb-diff}
\begin{tabular}{@{}llrrrr@{}}
\toprule
Benchmark & Scoring & \shortstack[r]{Full mixture,\\SFT} & \shortstack[r]{Full mixture,\\SFT + GRPO} & \shortstack[r]{Perc.\ $\times$ causal,\\SFT} & \shortstack[r]{Perc.\ $\times$ causal,\\SFT + GRPO} \\
\midrule
SAT & Standard & $+11.0$ $[+4.7, +17.7]$ & $+10.3$ $[+3.7, +16.7]$ & $+12.3$ $[+6.7, +18.0]$ & $+9.0$ $[+3.0, +15.0]$ \\
 & Circular & $+23.3$ $[+14.0, +32.7]$ & $+22.0$ $[+12.7, +31.3]$ & $+25.3$ $[+16.7, +34.0]$ & $+19.3$ $[+11.3, +27.3]$ \\
ActionEQA & Standard & $+7.1$ $[+3.0, +11.2]$ & $+5.9$ $[+1.9, +10.1]$ & $+6.4$ $[+3.0, +9.8]$ & $+8.9$ $[+5.5, +12.4]$ \\
 & Circular & $+9.8$ $[+6.1, +13.4]$ & $+8.6$ $[+5.1, +12.2]$ & $+7.1$ $[+4.4, +10.0]$ & $+9.9$ $[+6.9, +13.0]$ \\
SPAR view change & Standard & $+15.3$ $[+7.3, +23.4]$ & $+13.0$ $[+4.6, +21.1]$ & $+28.4$ $[+21.1, +35.6]$ & $+29.5$ $[+22.2, +36.8]$ \\
 & Circular & $+19.9$ $[+15.3, +24.9]$ & $+19.2$ $[+14.6, +24.1]$ & $+30.3$ $[+24.9, +35.6]$ & $+35.6$ $[+29.9, +41.4]$ \\
\bottomrule
\end{tabular}
\end{table}

Training does even out the answer letters: after training they follow the distribution of the correct answers closely (Table~\ref{tab:lb-letters}). Under circular scoring, all four trained models remain significantly above the base model on all three benchmarks (Tables~\ref{tab:lb-acc} and~\ref{tab:lb-diff}), with larger gains than under standard scoring because the base model's letter preference costs it more; the SFT models, trained without the balancing reward, gain as much as the SFT + GRPO models. The base model's circular score on SPAR-Bench ($0.4\%$) equals the expected score of answering without the content: its answers are determined by the option position. The external gains of WOVEN training thus remain significant when a letter preference earns no credit, after SFT and after GRPO; they cannot be attributed to the model having learned to balance its answer letters or to the multiple-choice format.

\subsection{Shortcut control for cross-operation transfer: the unchanged-scene option}
\label{app:static-option}

The transfer from forward dynamics to inverse dynamics and to counterfactual reasoning within an action type does not depend on the unchanged-scene option. Training only on inspective forward dynamics questions (initial view and camera action, pick the resulting view) transfers to inverse dynamics, counterfactual substitution, and counterfactual removal of the same action type (Appendix~\ref{app:E-F6}). Inspective questions contain an option showing the unchanged scene (the initial view, or the text ``the camera remains stationary''), which is always wrong in forward dynamics, inverse dynamics, and counterfactual substitution and always correct in counterfactual removal, so part of the transfer to substitution could be a rule about that option rather than structure shared across reasoning types. We remove the option from the forward dynamics, inverse dynamics, and substitution questions, which become three-option questions (chance $33.3\%$) on which avoiding it no longer helps, and evaluate removal in its original form, where the unchanged scene is the correct answer and a model that avoids it would fall below the base model.

The model is Qwen2.5-VL-3B-Instruct trained only on inspective forward dynamics ($1{,}020$ questions, each in its four option orderings, $4{,}080$ training examples) in one run, with LoRA (rank $16$, $\alpha=32$, dropout $0.05$), learning rate $10^{-4}$, $3$ epochs, final checkpoint, compared with the untrained base model; this configuration differs from that of Appendix~\ref{app:E-F6}, so the gains differ in size. Evaluation uses the WOVEN ID inspective questions, $270$ per reasoning type, with greedy decoding (vLLM) and the model asked to output only the option letter; in the static-free versions the remaining options are relabeled A--C in their original order. Differences (forward-only $-$ base) are tested by a paired sign-flip permutation test over questions with a bootstrap $95\%$ confidence interval.

\begin{table}[!htbp]
\centering
\footnotesize
\caption{Static-free questions (unchanged-scene option removed, three options), accuracy (\%).}
\label{tab:so-free}
\begin{tabular}{@{}lrrrr@{}}
\toprule
Reasoning type & Base & Forward-only & $\Delta$ $[95\%$ CI$]$ & $p$ \\
\midrule
Counterfactual substitution & 28.5 & 93.3 & $+64.8$ $[+58.5, +71.1]$ & $<0.001$ \\
Inverse dynamics & 43.3 & 60.7 & $+17.4$ $[+12.2, +22.6]$ & $<0.001$ \\
Forward dynamics (trained) & 37.8 & 87.4 & $+49.6$ $[+42.2, +57.0]$ & $<0.001$ \\
\bottomrule
\end{tabular}
\end{table}

\begin{table}[!htbp]
\centering
\footnotesize
\setlength{\tabcolsep}{4pt}
\caption{Original questions (four options, unchanged-scene option included), accuracy (\%).}
\label{tab:so-orig}
\begin{tabular}{@{}lrrrr@{}}
\toprule
Reasoning type & Base & Forward-only & $\Delta$ $[95\%$ CI$]$ & $p$ \\
\midrule
Counterfactual substitution & 24.8 & 78.1 & $+53.3$ $[+45.9, +60.4]$ & $<0.001$ \\
Counterfactual removal (answer: unchanged scene) & 27.8 & 55.2 & $+27.4$ $[+19.3, +35.6]$ & $<0.001$ \\
Inverse dynamics & 17.4 & 44.1 & $+26.7$ $[+21.5, +32.2]$ & $<0.001$ \\
Forward dynamics (trained) & 28.9 & 86.3 & $+57.4$ $[+50.7, +64.1]$ & $<0.001$ \\
\bottomrule
\end{tabular}
\end{table}

With the unchanged-scene option removed, the forward-only model still gains $64.8$ points on counterfactual substitution and $17.4$ points on inverse dynamics (Table~\ref{tab:so-free}), and on counterfactual removal, where the unchanged scene is the correct answer, it gains $27.4$ points (Table~\ref{tab:so-orig}). The transfer from forward dynamics to inverse dynamics and to counterfactual reasoning within an action type therefore does not depend on that option.

\section{Robustness and consistency of the results}
\label{app:robustness}

This section checks the best-subset gains against selection and seed variation (Appendix~\ref{app:best-subset-robustness}), repeats the training on a different model family (Appendix~\ref{app:gemma}), and tests whether forward and inverse queries about the same transition succeed together after training (Appendix~\ref{app:fwd-inv}).

\subsection{Selection and seed robustness of the best-subset gains}
\label{app:best-subset-robustness}

The best-subset gains on representative benchmarks of every capability domain are neither an artifact of selecting the best of $22$ candidates nor of a single training seed. The controlled-subset result reports, for each benchmark, the best of the $11$ subsets after SFT and after GRPO ($22$ candidates); taking the maximum of many candidates can produce a gain of a few points by selection alone, and a single training run can produce a gain by seed variation alone. To check both, we ran two additional GRPO seeds for every subset ($3$ in total) and selected two representative benchmarks for each capability domain of \S\ref{sec:enhance}: SAT and ViewSpatial (spatial), ActionEQA and Robo2VLM (embodied), InPhyRe and CLEVRER (physical), WorldPrediction and WM-ABench (procedural and world-model prediction), and TempCompass and TemporalBench (long) (temporal). Scores are recomputed from the model outputs under each benchmark's official protocol (Table~\ref{tab:bsr-units}); GRPO scores are per-unit means over the $3$ GRPO seeds, so the best candidate and its gain can differ from the single reported checkpoint in Table~\ref{tab:enhance-bestcell}. The selection check compares the best gain among the $22$ candidates with the distribution of the best gain that selection alone produces: per evaluation unit, the $22$ differences from the base model are sign-flipped jointly, the $22$ gains are recomputed, and their maximum is taken ($10{,}000$ flips); Holm correction over the $22$ candidates is reported as well. The seed check evaluates the best subset after GRPO separately for each of its $3$ seeds against the base model (paired sign-flip test, $10{,}000$ flips).

\begin{table}[!htbp]
\centering
\footnotesize
\setlength{\tabcolsep}{4pt}
\caption{Representative benchmarks, evaluation units, and scoring.}
\label{tab:bsr-units}
\begin{tabular}{@{}llrp{5.9cm}@{}}
\toprule
Domain & Benchmark & Units & Scoring \\
\midrule
Spatial & SAT & 150 & circular: a question counts only if both answer orders are correct \\
Spatial & ViewSpatial & 5,712 & question accuracy \\
Embodied & ActionEQA & 3,782 & question accuracy \\
Embodied & Robo2VLM & 6,676 & question accuracy \\
Physical & InPhyRe & 19,981 & question accuracy \\
Physical & CLEVRER & 12,000 & per-option accuracy (official stratified sample) \\
Procedural / world-model & WorldPrediction & 825 & question accuracy \\
Procedural / world-model & WM-ABench & 4,807 & question accuracy, resampled by question group \\
Temporal & TempCompass & 5,536 & question accuracy \\
Temporal & TemporalBench (long) & 5,485 & dual-order binary accuracy: a caption counts only if both answer orders are correct \\
\bottomrule
\end{tabular}
\end{table}

\begin{table}[!htbp]
\centering
\scriptsize
\setlength{\tabcolsep}{3pt}
\caption{Best of $22$ candidates versus the max-selection null (accuracy \%, gains in points). Perc., Insp., Mani.: perceptive, inspective, manipulative; cf: counterfactual. Null: 95th percentile of the best gain under selection alone; $p_{\max}$: $p$ under the max-selection null; $p_{\mathrm{Holm}}$: Holm-corrected over the 22 candidates.}
\label{tab:bsr-selection}
\begin{tabular}{@{}lrlrrrr@{}}
\toprule
Benchmark & Base & Best candidate & Best gain $[95\%$ CI$]$ & Null & $p_{\max}$ & $p_{\mathrm{Holm}}$ \\
\midrule
SAT & 37.3 & Perc.\ $\times$ causal, SFT & $+26.0$ $[+17.3, +34.7]$ & $+10.4$ & $<0.001$ & 0.002 \\
ViewSpatial & 35.3 & Mani.\ $\times$ cf, SFT + GRPO & $+3.6$ $[+2.7, +4.5]$ & $+1.3$ & $<0.001$ & 0.002 \\
ActionEQA & 43.4 & Mani.\ $\times$ causal, SFT & $+11.5$ $[+9.9, +13.1]$ & $+2.0$ & $<0.001$ & 0.002 \\
Robo2VLM & 33.5 & Mani.\ $\times$ cf, SFT & $+7.6$ $[+6.6, +8.6]$ & $+1.2$ & $<0.001$ & 0.002 \\
InPhyRe & 22.0 & Insp.\ $\times$ cf, SFT & $+10.3$ $[+9.7, +10.9]$ & $+0.8$ & $<0.001$ & 0.002 \\
CLEVRER & 58.2 & Insp.\ $\times$ cf, SFT + GRPO & $+4.6$ $[+3.8, +5.4]$ & $+1.0$ & $<0.001$ & 0.002 \\
WorldPrediction & 34.8 & Mani.\ $\times$ causal, SFT + GRPO & $+6.9$ $[+3.0, +10.8]$ & $+4.9$ & 0.003 & 0.009 \\
WM-ABench & 31.0 & Mani.\ $\times$ cf, SFT & $+4.6$ $[+3.4, +6.0]$ & $+1.6$ & $<0.001$ & 0.002 \\
TempCompass & 67.7 & Mani.\ $\times$ causal, SFT & $+2.2$ $[+1.3, +3.1]$ & $+1.2$ & $<0.001$ & 0.002 \\
TemporalBench (long) & 16.6 & Mani.\ $\times$ cf, SFT + GRPO & $+8.1$ $[+7.2, +9.1]$ & $+1.2$ & $<0.001$ & 0.002 \\
\bottomrule
\end{tabular}
\end{table}

\begin{table}[!htbp]
\centering
\scriptsize
\setlength{\tabcolsep}{3pt}
\caption{Best subset after GRPO: gain over the base model per GRPO seed (points). Every per-seed gain has $p<0.001$ under the paired sign-flip test except WorldPrediction seeds 1 and 2 ($p=0.011$ and $p=0.001$).}
\label{tab:bsr-seeds}
\begin{tabular}{@{}llrrrr@{}}
\toprule
Benchmark & Subset & Seed 1 & Seed 2 & Seed 3 & Mean \\
\midrule
SAT & Perc.\ $\times$ causal & $+20.7$ & $+24.0$ & $+27.3$ & $+24.0$ \\
ViewSpatial & Mani.\ $\times$ cf & $+3.3$ & $+3.1$ & $+4.3$ & $+3.6$ \\
ActionEQA & Mani.\ $\times$ causal & $+11.6$ & $+11.4$ & $+9.5$ & $+10.8$ \\
Robo2VLM & Mani.\ $\times$ cf & $+8.1$ & $+7.0$ & $+6.8$ & $+7.3$ \\
InPhyRe & Insp.\ $\times$ cf & $+10.2$ & $+9.8$ & $+10.6$ & $+10.2$ \\
CLEVRER & Insp.\ $\times$ cf & $+5.1$ & $+4.6$ & $+4.2$ & $+4.6$ \\
WorldPrediction & Mani.\ $\times$ causal & $+5.5$ & $+7.2$ & $+8.1$ & $+6.9$ \\
WM-ABench & Mani.\ $\times$ cf & $+3.9$ & $+4.5$ & $+2.6$ & $+3.7$ \\
TempCompass & Mani.\ $\times$ causal & $+1.9$ & $+2.4$ & $+2.3$ & $+2.2$ \\
TemporalBench (long) & Mani.\ $\times$ cf & $+5.7$ & $+7.0$ & $+11.7$ & $+8.1$ \\
\bottomrule
\end{tabular}
\end{table}

On all ten benchmarks, the best-subset gain exceeds what selecting the best of $22$ candidates produces by chance and remains significant after Holm correction (Table~\ref{tab:bsr-selection}), and the best subset improves significantly over the base model in each of its three GRPO seeds (Table~\ref{tab:bsr-seeds}).

\subsection{Generalization across model families}
\label{app:gemma}

The effect of WOVEN training holds across model families. The main experiments train Qwen2.5-VL; we repeat the training on Gemma-3-4B-it~\citep{gemma2025gemma3}, which has a different vision encoder and a different way of handling multiple images, and compare with the untrained Gemma-3-4B-it. Training uses the WOVEN perceptive causal-dynamics subset (forward and inverse dynamics questions, four option orderings, $9{,}750$ training examples) with LoRA (rank $16$, $\alpha=32$, dropout $0.05$) on the language model's q/k/v/o projections, vision encoder and multimodal projector frozen, learning rate $10^{-4}$, $5\%$ warmup, cosine schedule, effective batch size $64$, $3$ epochs, final checkpoint, with $3$ seeds. Evaluation uses greedy decoding (vLLM) with the model asked to output only the option letter, on WOVEN ID ($6{,}228$ questions), WOVEN OOD ($3{,}496$ held-out-scene questions), SPAR-Bench view change~\citep{zhang2025spar} on ScanNet and ScanNet++ ($261$ questions with two real images, asking how the camera moved between them, converted to four options as in Appendix~\ref{app:recipe-prospective}), and SAT~\citep{ray2024sat} ($150$ two-option questions, scored circularly: each question is asked once with each option order and counts as correct only if both answers are right). For each question, the score of the trained model is averaged over the $3$ seeds and compared with the base model by a paired sign-flip permutation test over questions, with a bootstrap $95\%$ confidence interval of the difference. Table~\ref{tab:gemma-qwen} reports the same training and evaluation on Qwen2.5-VL-3B-Instruct for reference; because the training configuration and the decoding protocol differ from those of the main experiments, these Qwen scores differ from the corresponding scores in the main text.

\begin{table}[!htbp]
\centering
\footnotesize
\setlength{\tabcolsep}{4pt}
\caption{Gemma-3-4B-it: accuracy (\%). $\Delta$ = trained (mean of $3$ seeds) $-$ base, percentage points with bootstrap $95\%$ confidence intervals.}
\label{tab:gemma}
\scriptsize
\setlength{\tabcolsep}{3pt}
\begin{tabular}{@{}lrrrrrrr@{}}
\toprule
Benchmark & Base & Seed 1 & Seed 2 & Seed 3 & Mean & $\Delta$ $[95\%$ CI$]$ & $p$ \\
\midrule
WOVEN ID & 27.9 & 44.4 & 44.5 & 44.3 & \textbf{44.4} & $+16.5$ $[+15.1, +18.0]$ & $<0.001$ \\
WOVEN OOD & 33.0 & 45.2 & 44.3 & 43.9 & \textbf{44.5} & $+11.5$ $[+9.4, +13.5]$ & $<0.001$ \\
SPAR-Bench view change (real images) & 28.0 & 46.4 & 45.6 & 51.0 & \textbf{47.6} & $+19.7$ $[+12.5, +26.6]$ & $<0.001$ \\
SAT (circular) & 30.0 & 40.0 & 46.0 & 55.3 & \textbf{47.1} & $+17.1$ $[+6.9, +27.1]$ & $0.002$ \\
\bottomrule
\end{tabular}
\end{table}

\begin{table}[!htbp]
\centering
\footnotesize
\caption{The same training on Qwen2.5-VL-3B-Instruct, accuracy (\%), for reference.}
\label{tab:gemma-qwen}
\begin{tabular}{@{}lrrr@{}}
\toprule
Benchmark & Base & Mean of $3$ seeds & $\Delta$ \\
\midrule
WOVEN ID & 29.8 & 47.7 & $+17.9$ \\
WOVEN OOD & 36.2 & 45.9 & $+9.7$ \\
SPAR-Bench view change (real images) & 26.4 & 54.9 & $+28.5$ \\
SAT (circular) & 39.3 & 63.3 & $+24.0$ \\
\bottomrule
\end{tabular}
\end{table}

WOVEN training raises Gemma-3-4B-it by $16.5$ points on WOVEN ID, $11.5$ points on held-out WOVEN scenes, $19.7$ points on real-image camera-motion questions, and $17.1$ points on SAT (all $p\le0.002$), with every seed above the base model on every benchmark (Table~\ref{tab:gemma}). The effect of WOVEN training holds across model families.

\subsection{Consistency between forward and inverse dynamics}
\label{app:fwd-inv}

After WOVEN training, the forward and inverse queries about the same transition succeed and fail together, whereas for the base model they are independent, so the two reasoning operations draw on one shared mapping between states, actions, and outcomes rather than on two separate skills. Training raises accuracy on both forward dynamics (initial state and action, predict the outcome) and inverse dynamics (initial state and outcome, identify the action); the two could improve independently. We therefore query every transition $(s, a_i, s'_i)$ of the WOVEN ID test set twice, by the forward question for $a_i$ and by the inverse question for $s'_i$ ($1{,}134$ pairs from $306$ sources, all ID transitions that have both questions), and compare the rate at which both are answered correctly (joint) with the rate expected if the two were independent, $P(\text{forward correct}) \times P(\text{inverse correct})$, computed within each action type and weighted by the number of pairs so that differences in difficulty between action types cannot produce a correlation. We also report the cycle rate, the share of transitions for which the inverse question about the outcome the model itself picked in the forward question returns the original action $a_i$. GRPO values are averaged over $3$ GRPO seeds; $95\%$ confidence intervals are by bootstrap over sources ($2{,}000$ resamples).

\begin{table}[!htbp]
\centering
\scriptsize
\setlength{\tabcolsep}{3pt}
\caption{Forward and inverse consistency on WOVEN ID (\%). Joint: both questions of a pair correct; Independent: expected joint rate under independence within action type; Excess: joint $-$ independent; Cycle: the inverse query recovers the action behind the outcome the model predicted; $\Delta$ joint: joint $-$ joint of the base model.}
\label{tab:fwd-inv}
\begin{tabular}{@{}lrrrrrrr@{}}
\toprule
Model & Forward & Inverse & Joint & Indep. & Excess $[95\%$ CI$]$ & Cycle & $\Delta$ joint $[95\%$ CI$]$ \\
\midrule
Base & 21.3 & 40.4 & 8.0 & 8.1 & $-0.1$ $[-1.2, +0.9]$ & 19.0 & --- \\
Full mixture, SFT & 97.1 & 94.8 & 93.5 & 92.1 & $+1.4$ $[+0.8, +2.1]$ & 94.3 & $+85.4$ $[+83.2, +87.7]$ \\
Full mixture, SFT + GRPO & 96.6 & 96.9 & 94.1 & 93.7 & $+0.4$ $[+0.1, +0.9]$ & 94.4 & $+86.1$ $[+83.7, +88.2]$ \\
Perceptive $\times$ causal, SFT & 75.3 & 76.6 & 65.5 & 61.8 & $+3.7$ $[+2.6, +4.8]$ & 71.3 & $+57.5$ $[+53.6, +61.5]$ \\
Perceptive $\times$ causal, SFT + GRPO & 75.7 & 77.4 & 66.9 & 62.8 & $+4.1$ $[+3.2, +4.9]$ & 72.8 & $+58.8$ $[+55.1, +62.6]$ \\
Inspective $\times$ causal, SFT & 73.7 & 75.8 & 61.9 & 59.3 & $+2.7$ $[+1.6, +3.7]$ & 69.0 & $+53.9$ $[+49.6, +58.1]$ \\
Inspective $\times$ causal, SFT + GRPO & 73.9 & 77.5 & 64.1 & 60.8 & $+3.3$ $[+2.4, +4.1]$ & 70.5 & $+56.1$ $[+51.9, +60.0]$ \\
Navigative $\times$ causal, SFT & 64.9 & 64.9 & 52.9 & 49.0 & $+3.9$ $[+2.7, +5.0]$ & 58.6 & $+44.9$ $[+40.3, +49.3]$ \\
Navigative $\times$ causal, SFT + GRPO & 63.1 & 64.3 & 51.5 & 47.3 & $+4.2$ $[+3.3, +5.1]$ & 57.0 & $+43.5$ $[+39.3, +47.9]$ \\
Manipulative $\times$ causal, SFT & 63.3 & 62.7 & 53.2 & 48.6 & $+4.5$ $[+3.5, +5.6]$ & 57.1 & $+45.1$ $[+40.7, +49.6]$ \\
Manipulative $\times$ causal, SFT + GRPO & 63.5 & 61.8 & 51.8 & 48.4 & $+3.4$ $[+2.7, +4.2]$ & 56.3 & $+43.8$ $[+39.4, +48.3]$ \\
Inspective $\times$ counterfactual, SFT & 73.7 & 50.5 & 40.0 & 38.0 & $+2.0$ $[+1.0, +3.1]$ & 45.0 & $+32.0$ $[+28.3, +35.6]$ \\
Inspective $\times$ counterfactual, SFT + GRPO & 72.5 & 53.4 & 41.7 & 39.6 & $+2.1$ $[+1.2, +3.0]$ & 46.8 & $+33.7$ $[+30.2, +37.2]$ \\
Manipulative $\times$ counterfactual, SFT & 63.2 & 54.8 & 46.9 & 43.7 & $+3.2$ $[+2.0, +4.3]$ & 50.9 & $+38.9$ $[+34.2, +43.3]$ \\
Manipulative $\times$ counterfactual, SFT + GRPO & 64.9 & 54.3 & 47.8 & 44.8 & $+3.0$ $[+2.1, +3.8]$ & 51.3 & $+39.8$ $[+35.2, +44.2]$ \\
\bottomrule
\end{tabular}
\end{table}

For the base model, success on the forward and the inverse question of a transition is independent (excess $-0.1$, confidence interval including $0$). After training on causal or counterfactual transitions, the two queries of the same transition succeed together more often than independence predicts, for every trained model (excess $+0.4$ to $+4.5$, every confidence interval above $0$); the full-mixture models are close to ceiling on both queries (at least $94.8\%$), which leaves little room for excess. The cycle rate, where the inverse query recovers the action behind the outcome the model itself predicted, rises from $19.0\%$ to $45$--$94\%$ (Table~\ref{tab:fwd-inv}).

\section{Prospective validation of the training recipe}
\label{app:recipe-prospective}

\textbf{Goal.} The recipe in \S\ref{sec:enhance:org} is distilled from transfer results on the $26$ benchmarks. The prospective test fixes training mixtures by the recipe before training and compares them with other mixtures of the same size on questions from external benchmarks that were not used to derive the recipe. Every comparison asks how a fixed training budget should be allocated: all mixtures start from the same base model and are trained with the same protocol, so the base model enters none of the comparisons; its scores are reported for reference.

\textbf{Training.} Qwen2.5-VL-3B-Instruct, single-stage SFT with LoRA (rank $16$, $\alpha=32$, dropout $0.05$), learning rate $10^{-4}$, $5\%$ warmup, $3$ epochs. The WOVEN training set is partitioned into the $11$ controlled subsets of \S\ref{sec:enhance}, one per combination of action type and reasoning family. Each model is trained on $1{,}500$ WOVEN items drawn from these subsets, each item with its four option orderings, about $7{,}000$ training examples in total. The five mixtures differ only in how the $1{,}500$ items are allocated across the $11$ subsets (Table~\ref{tab:prosp-mix}). Each mixture is trained with $8$ seeds, where the seed determines which items are drawn and the training order, $40$ models in total.

\begin{table}[!htbp]
\centering
\footnotesize
\setlength{\tabcolsep}{3.5pt}
\caption{Prospective test: number of training items per subset in each mixture (of $1{,}500$).}
\label{tab:prosp-mix}
\begin{tabular}{@{}lrrrrr@{}}
\toprule
Subset & Recipe & Uniform & Temporal-heavy & Camera-heavy & Exogenous-heavy \\
\midrule
Perceptive $\times$ causal dynamics & 197 & 137 & 45 & 245 & 45 \\
Inspective $\times$ causal dynamics & 197 & 137 & 45 & 245 & 45 \\
Navigative $\times$ causal dynamics & 196 & 137 & 45 & 46 & 45 \\
Manipulative $\times$ causal dynamics & 196 & 137 & 45 & 46 & 45 \\
Inspective $\times$ counterfactual & 226 & 136 & 44 & 245 & 45 \\
Manipulative $\times$ counterfactual & 226 & 136 & 44 & 46 & 45 \\
Exogenous $\times$ physical modeling & 54 & 136 & 44 & 46 & 548 \\
Perceptive $\times$ temporal & 52 & 136 & 297 & 245 & 45 \\
Inspective $\times$ temporal & 52 & 136 & 297 & 245 & 45 \\
Navigative $\times$ temporal & 52 & 136 & 297 & 46 & 45 \\
Exogenous $\times$ temporal & 52 & 136 & 297 & 45 & 547 \\
\bottomrule
\end{tabular}
\end{table}

\textbf{Mixtures.} The target questions of the main comparisons (Set~A below) ask for the outcome of an agent's action or for the action behind an observed change, that is, they call for causal and counterfactual reasoning about agent-driven transitions. By item (1) of the recipe, the Recipe mixture allocates $82\%$ of the budget to the six agent-action subsets of the causal and counterfactual families; by item (3), temporal and passive-physical supervision does not transfer to such questions, so the remaining five subsets keep $52$--$54$ items each. Each comparison mixture represents another way of selecting supervision: Camera-heavy selects by matching the action (the camera-motion and object-inspection subsets, across reasoning families); Temporal-heavy selects the wrong reasoning operation (the temporal subsets); Uniform makes no selection (equal shares over the $11$ subsets); Exogenous-heavy is dominated by passive physical events.

\textbf{Evaluation questions.} All questions are drawn from external benchmarks, MMSI-Bench~\citep{yang2025mmsi}, VLM4D~\citep{vlm4d2025}, PhysBench~\citep{chow2025physbench}, SPAR-Bench~\citep{zhang2025spar}, MVBench~\citep{mvbench2023}, IntPhys~2~\citep{intphys2}, and CameraBench~\citep{lin2025camerabench}, none of which is among the $26$ benchmarks used to derive the recipe. Decoding is greedy (vLLM), and the model is asked to output only the option letter. The questions are grouped by what each comparison requires of them. Recipe versus Temporal-heavy, Uniform, and Exogenous-heavy require only that the questions call for causal or counterfactual reasoning about an agent's action (Set~A). The other direction of item (3) requires questions about passive physical events (Set~B). Recipe versus Camera-heavy asks whether matching the reasoning operation or matching the action matters more, so it is run on questions whose action Camera-heavy matches, with the action fixed to camera motion (Set~C). Item (4) requires static questions that involve no change (Set~D).

\emph{Set A, agent-action questions ($1{,}673$).} All require the outcome of an action or the motion behind a change; the action type is not restricted (Table~\ref{tab:prosp-setA}). The score is the accuracy on each subset, averaged with equal weight over the five subsets. The PhysBench questions come from a sample of $1{,}001$ PhysBench questions stratified by category.

\begin{table}[!htbp]
\centering
\footnotesize
\setlength{\tabcolsep}{4pt}
\caption{Prospective test: Set A.}
\label{tab:prosp-setA}
\begin{tabular}{@{}lp{3.2cm}rp{5.0cm}@{}}
\toprule
Source & Subset & Questions & Example \\
\midrule
MMSI-Bench & Motion (Cam.) & 74 & ``The images are taken continuously from a first-person perspective. In which direction are you moving?'' \\
VLM4D (real videos) & object motion direction & 1,111 & ``Which way does the person turn?'' \\
VLM4D (real videos) & motion direction from a stated viewpoint & 260 & ``From the camera perspective, which direction are the cyclists moving toward?'' \\
PhysBench & relationships $\cdot$ motion & 145 & ``Moving in the direction indicated by the arrow in the picture, which of the following options are you most likely to see?'' \\
PhysBench & dynamics $\cdot$ manipulation & 83 & ``Which of the following options presents the correct sequence to use the spoon to pick the objects in the green bowl and put them in the green pot?'' \\
\bottomrule
\end{tabular}
\end{table}

\emph{Set B, passive-physics questions ($1{,}174$).} Every subset shows objects that move, collide, or persist without any agent acting on them (Table~\ref{tab:prosp-setB}). MVBench items use $8$ frames sampled uniformly from each video; IntPhys~2 items ask ``Is what happens in this video physically possible?'' with the answers Yes / No. The score is the accuracy over all questions.

\begin{table}[!htbp]
\centering
\footnotesize
\setlength{\tabcolsep}{4pt}
\caption{Prospective test: Set B.}
\label{tab:prosp-setB}
\begin{tabular}{@{}lp{3.6cm}rp{5.2cm}@{}}
\toprule
Source & Subset & Questions & What is asked \\
\midrule
MVBench & counterfactual inference & 200 & which collision happens if an object is removed \\
MVBench & moving attribute & 200 & attributes of the objects that move or stay still \\
MVBench & moving count & 200 & how many objects of a kind are moving \\
MVBench & object existence & 198 & whether moving objects are still present at the end \\
IntPhys~2 & solidity, fixed camera & 104 & whether the event is physically possible \\
IntPhys~2 & immutability, moving camera & 136 & whether the event is physically possible \\
IntPhys~2 & continuity, moving camera & 136 & whether the event is physically possible \\
\bottomrule
\end{tabular}
\end{table}

\emph{Set C, camera-motion questions ($7{,}592$).} In every subset the action is the camera's own motion, which corresponds to WOVEN's perceptive actions, and the reasoning is to infer how the camera moved from the observations (inverse dynamics; Table~\ref{tab:prosp-setC}). The $74$ MMSI-Bench questions also belong to Set~A. For SPAR view change, the original questions ask for the camera's translation and rotation as numbers and are converted to four-option questions: the correct option is the motion the camera actually made (a turn of at least $30^\circ$, or, with a turn of at most $10^\circ$, a translation of at least $0.6$\,m in a single direction), and the three wrong options are motions that clearly did not happen, including the opposite of the correct one. The CameraBench questions are the yes/no questions of the CameraBench VQA release, grouped by the camera motion they ask about, with $8$ frames sampled uniformly from each video. The score is the accuracy over all questions.

\begin{table}[!htbp]
\centering
\footnotesize
\setlength{\tabcolsep}{4pt}
\caption{Prospective test: Set C.}
\label{tab:prosp-setC}
\begin{tabular}{@{}lp{3.6cm}rp{5.2cm}@{}}
\toprule
Source & Subset & Questions & What is asked \\
\midrule
MMSI-Bench & Motion (Cam.) & 74 & how the camera moved or rotated across consecutive images \\
VLM4D & camera motion & 6 & which way the camera turned or moved in the video \\
SPAR-Bench & view change, ScanNet / ScanNet++ & 125 / 136 & which motion the camera made between two images \\
CameraBench & pan & 1,200 & whether the camera pans left or right \\
CameraBench & tilt & 703 & whether the camera tilts up or down \\
CameraBench & roll & 524 & whether the camera rolls clockwise or counterclockwise \\
CameraBench & dolly & 753 & whether the camera moves forward or backward \\
CameraBench & truck & 863 & whether the camera moves laterally left or right \\
CameraBench & pedestal & 220 & whether the camera moves up or down \\
CameraBench & static & 628 & whether the camera stays still \\
CameraBench & tracking & 1,344 & whether the camera follows a moving subject \\
CameraBench & stability & 828 & whether the camera motion is smooth or shaky \\
CameraBench & other camera motion & 188 & pedestal-only and single-direction movements \\
\bottomrule
\end{tabular}
\end{table}

\emph{Set D, static questions ($886$).} Questions about a single state with no change, from the same benchmarks as Set~A: PhysBench property (attribute, color, mass, number) and relationships (depth, distance, location, size), $327$ questions; MMSI-Bench Attribute (Appr.), Attribute (Meas.), and the five positional-relationship categories, $559$ questions. The two benchmarks are scored separately.

\textbf{Statistics.} Each of the $8$ models of a mixture gives one score. Two mixtures are compared by an exact two-sided permutation test over all $12{,}870$ splits of their $16$ model scores; $p<0.05$ is significant. The $95\%$ confidence interval of the difference in means uses Welch's $t$. One comparison is made per recipe item and direction; no family-wise correction is applied.

\textbf{Results.} Table~\ref{tab:prosp-results} reports the five comparisons; in every comparison both mixtures score above the untrained base model. Selecting the wrong reasoning operation costs $3.5$ points (Temporal-heavy) and selecting passive physical events costs $4.7$ points (Exogenous-heavy); making no selection costs $1.5$ points (Uniform). On camera-motion questions, Camera-heavy holds more exact-match supervision than Recipe ($735$ versus $620$ items in the camera-motion and object-inspection subsets of the causal and counterfactual families) and still trails by $0.6$ points ($p=0.04$); the two mixtures differ in the remaining budget, which Camera-heavy spends on same-action temporal subsets ($490$ items) and Recipe on same-operation navigation and manipulation subsets ($618$ items). On passive-physics questions, Exogenous-heavy scores $0.5$ points above Recipe ($p=0.04$), consistent with the direction of item (3). In the other direction, Exogenous-heavy gains least on Set~A (Table~\ref{tab:prosp-static}), consistent with \S\ref{sec:recipe:composition}: passive-physical supervision contributes least to agent-action questions. Its small gain is attributable to the $405$ items ($27\%$) it draws from agent-action subsets, whereas \S\ref{sec:recipe:composition} uses the passive-physical subset alone.

\textbf{Item (4).} This item concerns whether training helps at all on tasks that involve no transition, so the comparison is with the untrained base model (Table~\ref{tab:prosp-static}). On static questions the five mixtures differ from the base model by $-0.2$ to $+1.2$ points, and Recipe does not differ significantly from any other mixture ($p\ge0.37$), whereas the same models gain $2.1$ to $6.8$ points over the base model on Set~A. The PhysBench static questions are the main evidence; the MMSI-Bench static questions are close to chance for the base model and are reported for completeness.

\begin{table}[!htbp]
\centering
\footnotesize
\setlength{\tabcolsep}{3.5pt}
\caption{Prospective test: static questions versus agent-action questions (accuracy \%, mean over $8$ models).}
\label{tab:prosp-static}
\scriptsize
\begin{tabular}{@{}lrrrrrr@{}}
\toprule
Questions & Base & Recipe & Uniform & Temporal-heavy & Camera-heavy & Exogenous-heavy \\
\midrule
Set D, PhysBench static (327) & 55.7 & 56.1 & 55.4 & 56.5 & 56.5 & 56.5 \\
Set D, MMSI-Bench static (559) & 28.4 & 29.0 & 28.6 & 29.0 & 29.0 & 29.6 \\
Set A (1,673) & 31.1 & 37.9 & 36.4 & 34.4 & 36.3 & 33.2 \\
\bottomrule
\end{tabular}
\end{table}

\textbf{Item (2)} (supervision with larger state changes gives more robustness) is outside this test: the external benchmarks have no perturbation questions graded by the extent of the state change.

\begin{table}[!htbp]
\centering
\footnotesize
\caption{Prospective test: per-seed scores (accuracy \%). Base model: Set A $31.1$, Set B $70.3$, Set C $54.1$.}
\label{tab:prosp-seeds}
\begin{tabular}{@{}llrrrrrrrrr@{}}
\toprule
Questions & Mixture & s1 & s2 & s3 & s4 & s5 & s6 & s7 & s8 & Mean \\
\midrule
Set A & Recipe & 38.7 & 36.8 & 38.3 & 37.0 & 38.4 & 38.3 & 36.3 & 39.6 & 37.9 \\
Set A & Uniform & 38.4 & 36.9 & 35.2 & 34.5 & 38.0 & 36.1 & 36.2 & 36.1 & 36.4 \\
Set A & Temporal-heavy & 35.8 & 35.1 & 33.0 & 35.8 & 35.6 & 35.3 & 32.5 & 32.4 & 34.4 \\
Set A & Exogenous-heavy & 30.6 & 33.5 & 34.5 & 32.2 & 35.2 & 33.9 & 34.4 & 31.6 & 33.2 \\
Set B & Recipe & 70.3 & 71.1 & 71.5 & 70.7 & 70.6 & 71.0 & 71.3 & 71.1 & 71.0 \\
Set B & Exogenous-heavy & 71.4 & 71.2 & 71.2 & 70.7 & 71.8 & 72.6 & 71.5 & 71.6 & 71.5 \\
Set C & Recipe & 58.2 & 58.3 & 58.3 & 58.8 & 58.5 & 58.8 & 58.2 & 57.7 & 58.4 \\
Set C & Camera-heavy & 57.5 & 56.8 & 57.4 & 58.8 & 58.0 & 57.4 & 58.0 & 58.3 & 57.8 \\
\bottomrule
\end{tabular}
\end{table}

\section{Limitations}
\label{app:limitations}

WOVEN deliberately scopes to world modeling, single-step, vision-grounded, action-conditioned reasoning over $(s_t,a,s_{t+1})$ triples, and therefore does not directly evaluate the broader world-model lineages: long-horizon procedural planning over abstract action sequences~\citep{worldprediction2025}, symbolic-environment dynamics inference~\citep{warrier2025worldtest,chen2025spa}, and purely latent-space predictive representations evaluated by downstream control rather than per-cell attribution~\citep{hafner2023dreamerv3,assran2025vjepa2}. These programmes operate over distinct state spaces and admit different yardsticks; extending the cell-level statistical-attribution methodology to them is a natural follow-up but outside the present scope.

\end{document}